\documentclass[a4paper,twoside,11pt]{dissertation_single} 

\usepackage[ansinew]{inputenc} 
\usepackage[T1]{fontenc} 
\usepackage{supertabular}

\usepackage{fancyhdr}
\usepackage{times} 
\usepackage{tabularx} 
\usepackage{longtable} 
\usepackage{verbatim} 

\usepackage{graphicx} 
\usepackage{placeins} 
\usepackage[hang]{caption} 
\setcaptionmargin{0.5\parindent} 
\usepackage{bibentry} 
\usepackage{tocbibind} 

\usepackage{time} 

\date{\today~~\now} 

\usepackage{listings}
\usepackage{makeidx} 

\usepackage{graphicx}

\usepackage{amssymb}  
\usepackage[centertags]{amsmath}

\makeindex
\newcommand{\textindex} [1]{#1\index{#1}}

\begin{document}
\newcommand{\textmainindex} [1]{#1\index{#1|hyperbf}}
\newcommand{\hyperbf} [1]{\textbf{\hyperpage{#1}}}
\setcounter{page}{1} \pagenumbering{roman}

\begin{titlepage}
    \begin{center}
        \thispagestyle{empty}
        \renewcommand{\baselinestretch}{1.5}
        \vspace*{\fill}
        \vspace{-2.5cm}
        \huge \textbf{The MATLAB Toolbox SciXMiner: \\ User's Manual and Programmer's Guide}\\
        \vspace{10mm} \large
        Ralf Mikut, Andreas Bartschat, Wolfgang Doneit, Jorge Ángel González Ordiano, Benjamin Schott, Johannes Stegmaier, Simon Waczowicz, Markus Reischl \\
        Karlsruhe Institute of Technology (KIT), Institute for Applied Computer Science \\
        P.O. Box 3640, 76021 Karlsruhe, Germany\\
        Phone: ++49/721/608-25731, Fax: ++49/721/608-25702\\
        Email: ralf.mikut@kit.edu\\
        \vspace{5mm}
        Version 2017a (12.04.2017)  \\
        \vspace{5mm}

        \centering{\includegraphics[width = 0.7\textwidth]{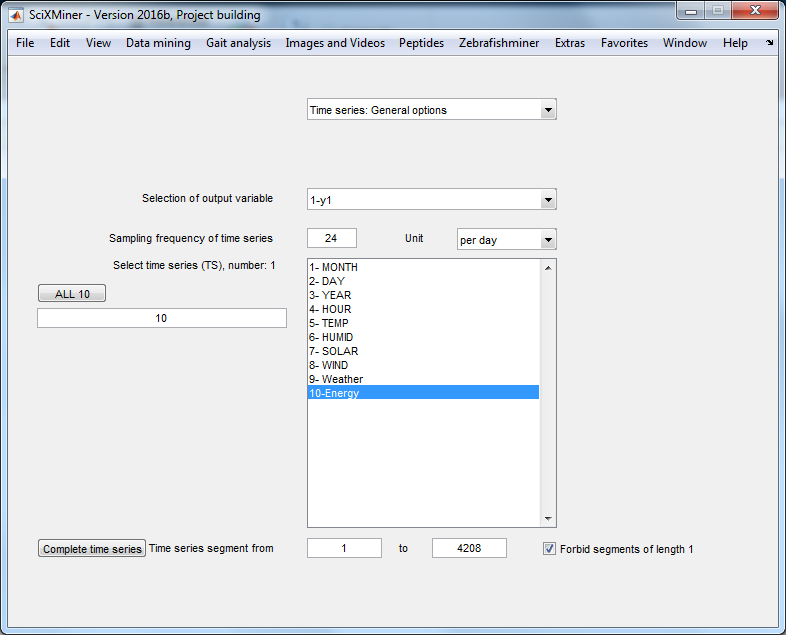}}
       \end{center}
\end{titlepage}

\newpage
\mbox{ }
\newpage

\setlength{\captionwidth}{\textwidth}
\bibliographystyle{diss}
\setcounter{secnumdepth}{5} \setcounter{tocdepth}{2}

\tableofcontents

\setlength{\parindent}{0em} \setlength{\parskip}{1.5ex plus0.5ex minus0.5ex}

\newpage
\setcounter{page}{1} \pagenumbering{arabic}

\lstset{basicstyle=\ttfamily\footnotesize,breaklines=true}

\chapter{Motivation}\thispagestyle{empty}
The Matlab toolbox SciXMiner is designed for the visualization and analysis of time series and
features with a special focus to classification problems. It was developed at the Institute of
Applied Computer Science of the Karlsruhe Institute of Technology (KIT), a member of the Helmholtz
Association of German Research Centres in Germany. The aim was to provide an open platform for the
development and improvement of data mining methods and its applications to various medical and
technical problems.

SciXMiner bases on Matlab (tested for the version 2017a). Many functions do not require
additional standard toolboxes but some parts of Signal, Statistics and Wavelet toolboxes are used
for special cases. The decision to a Matlab-based solution was made to use the wide mathematical
functionality of this package provided by The Mathworks Inc. MATLAB\textregistered and Simulink\textregistered are registered trademarks of The MathWorks, Inc.

SciXMiner is controlled by a graphical user interface (GUI) with menu items and control elements
like popup lists, checkboxes and edit elements. This makes it easier to work with SciXMiner for
inexperienced users. Furthermore, an automatization and batch standardization of analyzes is
possible using macros. The standard Matlab style using the command line is also available.
\index{control elements}\index{menu items}

SciXMiner is an open source software. The download page is

\href{http://sourceforge.net/projects/SciXMiner/}{http://sourceforge.net/projects/SciXMiner/}.

It is licensed under the conditions of the GNU General Public License (GNU-GPL) of The Free Software Foundation (see
\href{http://www.fsf.org/}{http://www.fsf.org/}).

The toolbox bases on earlier internal versions of Gait-CAD \cite{Loose04,Mikut08Biosig}. Please refer to \cite{Mikut17SciXMiner} if you
use SciXMiner for your scientific work.

SciXMiner contains a base toolbox and various application-specific extension packages. Up to know, six extension packages were published:
\begin{itemize}
\item analysis of images and videos \cite{Stegmaier12,Stegmaier16Diss} including a link to the ITK-based pipeline generation tool XPIWIT \cite{Bartschat16},
\item object tracking \cite{Stegmaier12,Stegmaier16Diss},
\item tissue detection and fluorescence quantification for zebrafish larvae (algorithms from \cite{Gehrig09}),
\item feature extraction and peptide optimization based on amino-acid distributions and chemical descriptors \cite{Mikut10Humana},
\item image processing and feature extraction for segmentation of grains in asphalt samples \cite{Reischl14AsphaltEngl}, and
\item extended measures and visualization to evaluate the quality of data and regression models \cite{Doneit17Eng}.
\end{itemize}

This manual is organized as follows: Chapter~\ref{sec:installation} explains the installation
procedure. Chapter~\ref{sec:leistungsumfang} outlines the implemented functionality of SciXMiner
followed by some recommendations for working with SciXMiner. The analysis of two benchmark data sets
is discussed in Chapter~\ref{sec:demo}. Detailed information for the use of menu items
(Chapter~\ref{sec:menu}) and control elements (Chapter~\ref{sec:optionen}) follow.
Chapter~\ref{sec:plugins_und_einzuege} present possible application-specific extensions for the
feature extraction from time series.

The appendix provide information about file formats (Appendix~\ref{sec:dateien}), internal data
structures (Appendix~\ref{sec:variablen}), necessary standard toolboxes
(Appendix~\ref{sec:toolbox_Matlab}) and integrated GNU-GPL-Matlab toolboxes of different groups
(Appendix~\ref{sec:toolbox_gnu}), a list of symbols and abbreviations (Appendix~\ref{sec:symbol})
and a list of known bugs and problems (Appendix~\ref{sec:anhang_fehler}).

\chapter{Installation}\label{sec:installation}\thispagestyle{empty}
The file \texttt{SciXMiner-Installer.exe} contains all necessary files as self-extracting
executable. After starting the destination folder for SciXMiner has to be selected, e.g.
\texttt{d:$\backslash$matlab$\backslash$scixminer$\backslash$}.

In a next step, the recent user path of the installed MATLAB version must be defined. An example is
\texttt{C:$\backslash$Dokumente und Einstellungen$\backslash$firstname.surname$\backslash$My Documents$\backslash$MATLAB}.
SciXMiner uses a subdirectory called \texttt{scixminer} of this directory to save options files. The user needs writing permissions for this directory.

A starting file called \texttt{scixminer.m} is individually created for this computer and stored in the user path. For a different user on the same computer, \texttt{scixminer.m} must be copied to his/her user path. Alternatively, \texttt{scixminer.m} can be copied to any other path of the MATLAB search path (\texttt{matlabpath}).

Alternatively, a zipped version scixminer.zip can be used. For the installation, the following steps are necessary:
\begin{enumerate}
\item extract in a directory,
\item copy start file \texttt{admin$\backslash$scixminer.m} in a directory of the matlab work directory (check using command \texttt{matlabpath}),
\item modify the directories in this file with the directory from the first step.
\end{enumerate}

SciXMiner is now ready installed and can be started with \texttt{scixminer}. \index{installation}

For developers, a SVN repository is available at:
$https://svn.code.sf.net/p/scixminer/scixminer$.

Please check SciXMiner out in a directory on your computer, e.g. $d:\\scixminer$.

\chapter{Methods}\label{sec:leistungsumfang}\thispagestyle{empty}
\section{Handling and functionality}
Data mining methods are useful in searching for unknown or partially known relations in large data
sets (\textindex{KDD}: Knowledge Discovery from Databases). A well known definition is given
by~\cite{Fayyad96}:

\emph{Data mining is a step in the KDD process that consists of applying data analysis and
discovery algorithms that produce a particular enumeration of patterns (or models) over the data.}

\textindex{Pattern} describes typical significant characteristics in features of the data set.
Hereby, a feature is an input variable for the data mining algorithm, which is relevant with
respect to the data mining problem. In this manual every input variable is regarded as possible
feature, as it may be helpful solving the problem. \index{features}

\begin{figure}[hbt]
\begin{center}
\includegraphics[width=\columnwidth]{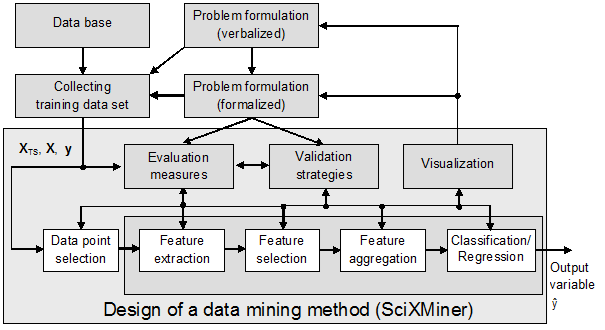}
\end{center}
\caption{\label{fig:probleme} Link between SciXMiner and the design of data mining
methods}
\end{figure}

Starting with a verbalized data mining problem, an adequate formalization has to be found. This
formalization influences as well the collection of the training data set from an (external) data
base (i.\,e. by special import routines, like HeiDATAProViT in gait analysis~\cite{Schablowski04})
as the collective of possible evaluation measures in SciXMiner (Figure~\ref{fig:probleme}).

SciXMiner permits a comfortable handling of numerous algorithms for the
\begin{itemize}
 \item \textindex{selection of data points} (i.\,e. detection of outliers, discarding of incomplete data points and features, selection of
        parts of data sets),
 \item \textindex{extraction of features} (i.\,e. spectrograms, FFT analysis, correlation analysis, linear filtering, calculation of
        extrema, mean values, fuzzification etc.),
 \item \textindex{evaluation of features} and their selection (i.\,e. multivariate analysis of variances, t-test, information measures,
        regression analysis),
 \item \textindex{aggregation of features}
       (synonym: \textindex{feature transformation}, e.g. Discriminant Analysis, Principal Component Analysis (\textindex{PCA}),
       Independent Component Analysis (\textindex{ICA})),
 \item supervised and unsupervised \textindex{classification} (i.\,e. decision trees, cluster algorithms,
       Bayes classifier, Artificial Neural Networks (ANN), Nearest Neighbor algorithms,
       Support Vector Machines (SVM), fuzzy systems) and
 \item \textindex{validation strategies} (i.\,e. cross-validation, bootstrap).
\end{itemize}
Additionally, there are various possibilities to visualize results, import and export data sets,
automatically log results and process steps in text and \LaTeX files, rename variables etc.

Depending on the availability, functions from different toolboxes are called, thereunder standard
Matlab functions, functions from internal Matlab toolboxes (see Appendix~\ref{sec:toolbox_Matlab}),
free available Matlab toolboxes (FastICA\footnote{http://www.cis.hut.fi/projects/ica/fastica/}, SVM
and Kernel Methods Matlab Toolbox
\cite{Canu03}\footnote{http://asi.insa-rouen.fr/$\sim$arakotom/toolbox/index.html}, SOM
Toolbox~\cite{Vesanto00}\footnote{http://www.cis.hut.fi/projects/somtoolbox/},
lp\_solve\footnote{http://lpsolve.sourceforge.net/5.5/}, see Appendix~\ref{sec:toolbox_gnu}) and
many SciXMiner internal functions.

\section{Problem description}\label{ssec:problemstellung}
A mandatory item to start a calculation is a \textmainindex{training data set} with $n=1,\ldots, N$
data points, each containing\index{data point}
\begin{itemize}
 \item $s_z$ \textindex{time series} (matrices $\mathbf{X}_{TS}[n]$ with $x_{TS,r}[k,n] \in \mathbb{R}$, $r=1,\ldots,s_z$, $k=1,\ldots,K$
sample points),
 \item $s$ features (vectors $\mathbf{x} [n] \in \mathbb{R}$ with $x_l[n], \, l=1,\ldots,s$) \index{features} and
 \item $s_y$ discrete \textindex{output variables} (vector  $\mathbf{y}[n] \in \mathbb{N^+}$ with $y_j[n], j=1,\ldots,s_y$).
\end{itemize}
Here, $\mathbb{R}$ is the set of real numbers and $\mathbb{N^+}$ the set of natural numbers.
Ordinal, interval-scaled, and rational-scaled \textmainindex{features} may be processed. Ordinal
features are discrete in value with respect to order scale (i.\,e. quantities with values like very
small, small, middle, and big). The values do not contain any information about the semantics of
their distances. If distances between all values are the same the scale is called interval-scaled
(i.\,e. temperature in~[°C] or~[°F]). Rational-scaled values additionally contain a natural
zero-point (i.\,e. length in~[m], temperature in [K]).

Besides further information like a priori preferences for features may be processed.

The purpose is mainly the generation of static or quasi-static estimations
\begin{align}
 \hat y_j [N+1] = & f(\mathbf{x}[N+1]) \text{ resp.} \\
 \hat y_j [N+1] = & f(\mathbf{x}[N+1] (\mathbf{X}_{TS}[N+1]))
\end{align}
for a data point $N+1$ with an unknown output variable, as well es the generation of intermediate results like tables
or catalogues of relevant features for certain problems.

The management of multiple output variables (i.\,e. diagnoses with respect to diseases in medical
applications, decisions for therapies, qualitative evaluations of therapy successes, gender,
age-groups etc.) for each data point allows a flexible selection of multiple classification
problems. Additionally, input and output variables may be changed in dependence of the problem.

The training data set is given by a binary Matlab project file, containing matrices with
standardized names (i.\,e. \texttt{d\_orgs} for time series, \texttt{d\_org} for features and
\texttt{code\_alle} for output variables). Additionally, matrices and structures are possible (not
mandatory) to denote textual identifiers and further information. Missing information is
compensated by standard values and identifiers. Using 1~GB of RAM a \textindex{SciXMiner project
file} may have a size up to 500~MB without causing any problems with memory. Larger project files
are problematic, due to the chosen data structures (arrays instead of cell arrays).

\section{Further reading}
A comprehensive illustration of all algorithms mentioned in this section can not be done within
this manual. Therefore, some examples for further reading are given:
\begin{itemize}
 \item basic knowledge about multivariate statistics and classification~\cite{Tatsuoka88,Michie94,Jain00}
 and specialties for time series \cite{Mikut08IEEE},
 \item basics in \textindex{data mining} \cite{Fayyad96,Borgelt00,Cios01,Mikut06BMT,Runkler16} including tuning concepts \cite{Konen11},
 \item \textindex{cluster algorithms} and \textindex{Fuzzy C-Means} (FCM): basics: \cite{Hoeppner99,Jain10}, specialties for time series
 \cite{Loose04},
 \item decision trees (basics: \cite{Breiman84,Quinlan93}, implemented design algorithms~\cite{JaekelGM99,Mikut05}),\index{decision tree}
 \item fuzzy systems (basics: \cite{Zadeh65,Kiendl97,Kruse94,Kroll16}, implemented design algorithms~\cite{JaekelGM99,Mikut05,Beck04Mathware,Beck05}), \index{fuzzy system}, specialities to improve interpretability \cite{Gacto11,Mikut05}
 \item \textindex{a priori relevances} \cite{Wolf06,Mikut05},
 \item \textindex{Independent Component Analysis} \cite{Hyvarinen99,Stone02},
 \item \textindex{Support Vector Machines} \cite{Burges98,Canu03,Schoelkopf97},
 \item $k$-Nearest Neighbor Methods \cite{Cover67},
 \item Artificial Neural Networks \cite{Haykin94},\index{Artificial Neural Network}
 \item feature aggregation and selection using wrapper approaches \cite{Reischl04a},
 \item \textindex{validation strategies} \cite{Michie94}, and
 \item alternative data mining software as Weka \cite{Hall09}, Knime \cite{Berthold08}, Apache Spark's machine learning library \cite{Meng16}, Keel \cite{Alcala11}, Rattle/R \cite{Williams11}, see surveys in \cite{Mikut11Wiley,Landset15,Gupta16}
\end{itemize}

Selected SciXMiner applications are listed in Table~\ref{tab:SciXMinerApplications}.
\begin{table}[htb]
\begin{tabular}{|p{4cm}|p{3cm}|p{3cm}|p{4cm}|}
\hline
\textbf{Application} & Data & Domain & References \\
\hline
Asphalt grain analysis & Images & Engineering & \cite{Reischl14AsphaltEngl} \\ \hline
Brain-Machine-Interfaces & Time series & Medicine & \cite{Mikut08} \\ \hline
Cell classification & Images and single features & Biology & \cite{Khan13Krusebook} \\ \hline
Ceramic actuator optimization & Time series and single features & Engineering & \cite{Brueckner12} \\ \hline
Circadian parameter estimation & Time series & Medical technology & \cite{Schott16} \\ \hline
Climate change events on streets & Time series & Engineering & \cite{Keller15} \\ \hline
Control of prosthesis with muscle signals & Time series & Medical technology & \cite{Pylatiuk09,Reischl06,Reischl11ICAE,Reischl16,Schill09,Schill14} \\ \hline
Heartbeat detection of zebrafish & Videos & Biology & \cite{Alshut16,Pylatiuk14,Pfriem16} \\ \hline
Light-sheet microscopy & 3D-Images, 3D-Videos & Biology & \cite{Stegmaier16Diss,Stegmaier16,Stegmaier16AT} \\ \hline
Mobile working machines & Time series & Engineering & \cite{Gerland10,Gerland09} \\ \hline
Modeling of labyrinth seals & Single features & Engineering & \cite{Pychynski10} \\ \hline
Morphology analysis of zebrafish & Images & Biology & \cite{Alshut16,Alshut10,Shahid16,Stegmaier14a} \\ \hline
Movement analysis & Time series & Medicine & \cite{Burmeister09,Schablowski06,Wolf11,Wolf06,Wolf14} \\ \hline
Object segmentation of hardware items & Images & Engineering & \cite{Khan15,Khan16a} \\ \hline
Optimization of antimicrobial peptides & Sequences & Chemistry & \cite{vonGundlach16,Knappe16,Mikut10Humana,Mikut09,Ramon-Garcia13,Mikut16,Yu15a} \\ \hline
Peripheral nerve signals of rats & Time series & Biology & \cite{Hiebl10} \\ \hline
pH sensor diagnosis & Time series & Engineering & \cite{Grube11} \\ \hline
Photo-Motor Response & Images, Time series & Biology & \cite{Kokel13,Marcato15} \\ \hline
Photovoltaic systems & Time series & Engineering & \cite{GonzalezOrdiano16a,GonzalezOrdiano16} \\ \hline
Robotics & Time series & Engineering & \cite{Bauer14,Bauer10,Bauer12,Bauer12at} \\ \hline
Smart energy grids & Time series & Engineering & \cite{Burmeister08,Hagenmeyer16,Maass13,Maass15,Waczowicz14_ENG,Waczowicz16} \\ \hline
Toxicity tests & Images & Biology & \cite{Alshut10,Alshut11AT,Ho13} \\ \hline
Wind power systems & Time series & Engineering & \cite{Arnoldt10a} \\ \hline
\end{tabular}
\caption{Selected SciXMiner applications}
\label{tab:SciXMinerApplications}
\end{table}

Further details about parameterization are given in the description of menu items
(Chapter~\ref{sec:menu}) and control elements (Chapter~\ref{sec:optionen}).

\section{Further toolboxes for classification (selection)}
\begin{itemize}
\item PRTools (http://www.prtools.org/) University of Delft (The Netherlands):
      Matlab toolbox, e.g. Principal Component Analysis, Subspace classifiers, Artificial Neural Networks, without GUI,
      free for academic use
\item NEFCLASS (http://fuzzy.cs.uni-magdeburg.de/nefclass/) University Magdeburg (Germany):
      JAVA, e.g. Neuro-Fuzzy classifiers, free for academic use
\end{itemize}

\chapter{Working with SciXMiner}\label{sec:tutorial}\thispagestyle{empty}
\section{Handling}
The graphical user interface (\textindex{GUI}\index{Graphical User Interface}) of SciXMiner contains
menu items and \textindex{control elements} like listboxes, checkboxes and text fields. They are
implemented using Matlab functions (\texttt{uicontrol}, \texttt{uimenu}), which are partially
called by encapsulated SciXMiner functions with additional functionality. These elements are using
\textindex{callback functions} to exchange data with variables from the workspace and call
functions which are independent from the GUI. Thus, the Matlab-typical way of programming using a
\textindex{command prompt} and variables is possible, too.

The elements of the graphical user interface are described in the following sections.

\section{Import and export of projects}\label{sec:imundexport}
SciXMiner offers the opportunity for an \textindex{import} and \textindex{export} of projects from
and into ASCII files. The possible options will be explained in the next two sections with a focus
to the import and export of time series. Some differences for the export of single features will be
also explained.

\subsection{Import of data}
Data can be imported from one file or from many files in a directory. The latter option is also
possible for existing subdirectories, but requires special naming conventions.

For the import of time series, each data point is read from a separate file. It contains the time
series in the columns and the sample points in the rows. For single features, all data points are
normally read from one file containing the single features in the columns and the data points in
the rows. An import from multiple files is also possible to characterize different output classes
by the file and directory names.

\begin{figure}[htb]
    \begin{center}
        \includegraphics[width=.7\textwidth]{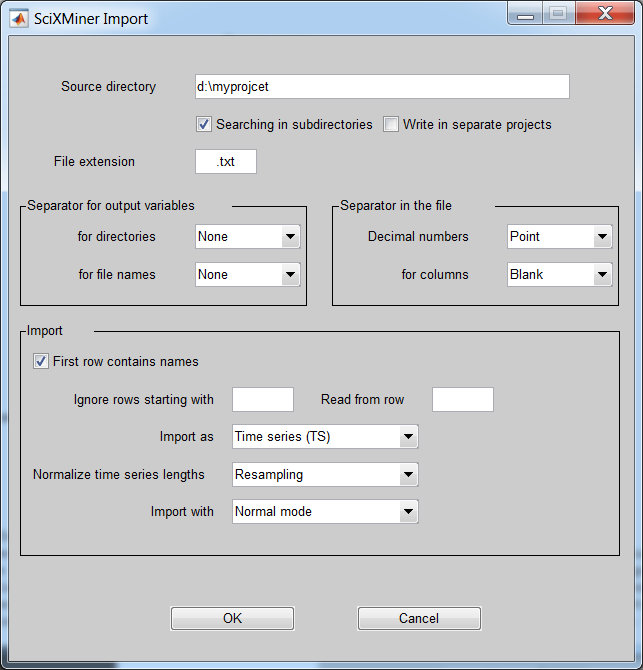}
        \caption{\label{img:import}Dialog window for the data import}
    \end{center}
\end{figure}
The window in Fig.~\ref{img:import} will be shown for an \textindex{import} using
\hyperlink{MI_Import_Ordner}{\emph{File - Import data - From a directory...}}.

The option ''\textindex{Searching in subdirectories}'' defines if subdirectories are scanned for matching files or not. The option ''\textindex{Write in separate projects}'' defines that separate SciXMiner projects are generated for each import file. These separate projects can later fused by \hyperlink{MI_Project_Fusion}{\emph{File - Fusion of projects}}.

''File extension'' specifies all files for the import. Wildcards are not
possible (e.g. .t*)! The ''Separator for output variables'' controls the assignment of output
variables. As a consequence, it is extremely important for a comfortable assignment of output
variables to data points - especially for time series. It can be specified for directory names and
for file names.

Example:

The import of three files

 \texttt{d:$\backslash$prj$\backslash$names$\backslash$Anna$\backslash$Post-therapeutic.txt}

 \texttt{d:$\backslash$prj$\backslash$names$\backslash$Anna$\backslash$Pre-therapeutic.txt}

 \texttt{d:$\backslash$prj$\backslash$names$\backslash$Peter$\backslash$Post-therapeutic.txt}

from a folder \texttt{d:$\backslash$prj$\backslash$names} and the file extension '*.txt' generates
two terms for the first output variable variables (''Anna'', ''Peter'') and two terms for the
second output variable (''Pre-therapeutic'', ''Post-therapeutic'').


For \hyperlink{MI_Import_Datei}{\emph{File - Import data - From a file...}}, the single file to be imported will be selected by a standard window.
After this, a slightly different window is shown to specify the import, but the lower part of the
windows agrees with Fig.~\ref{img:import}.

Inside a file, two different separators are expected for the separation of columns (e.g. tabulators
for different time series or single features) and decimal numbers (e.g. ''Point'': 17.3, ''Comma:
17,3''). Such different separators for decimal numbers are necessary for compatibility reason for
some foreign language versions as e.g. German software.

The option ''First row contains names'' reads the names of the single features and time series from
the first row of the import file. Furthermore, a fix number of the rows at the beginning of the
file (''Read from row'') resp. all rows starting with a specified sign (''Ignore rows starting
with'') can be ignored. This is especially useful for the import of files with a header. Empty rows
will be ignored too.

The option ''Import as'' switches between an import of single features and an import of time
series. A simultaneous import is not possible at the moment.

The option ''Normalization of time series length'' is only necessary if time series with different length will be
imported. SciXMiner assumes always identical lengths. Different options exist: a resampling by the standard Matlab
command (from the signal toolbox) to get the length of the longest time series, a filling of shorter time series with
zeros or the last valid value.

The last option ''Import with'' switches between different import techniques. The ''Normal mode'' generates a temporary
file after a possible separator correction. It is normally the best version especially for large files. ''Write copy
and read again'' make the same line by line. ''Line-Option'' is a standard line-wise reading by Matlab.

''\textindex{Standard-ASCII}'' uses the Matlab command load -ascii and works only, if the file contains only numbers (without variable names etc.). This option has a reduced functionality but it can be very fast.

''\textindex{Structured (with Strings)}'' tries to import numbers and strings for files with single features. The strings are converted into linguistic terms of output variables. This option should be used together with the option ''\textindex{Write in separate projects}''.

''\textindex{Importdata (MATLAB)}'' uses the MATLAB function ''importdata''. It tries to import numbers and strings for files with single features. The strings are converted into linguistic terms of output variables. All columns with a string element in the first data row are interpreted as string columns. Such data is read from ''importdata'' as ''textdata''. This option is faster than ''Structured (with Strings)''.

\subsection{Export of data}
\begin{figure}[hbt]
    \begin{center}
        \includegraphics[width=.7\textwidth]{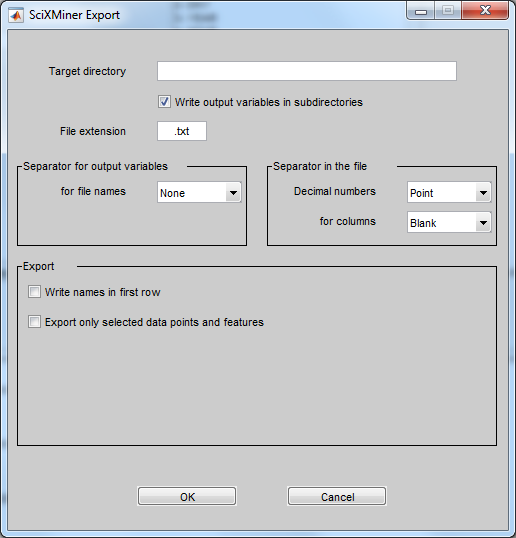}
        \caption{\label{img:export}Dialog window for the export of data}
    \end{center}
\end{figure}

For the export of time series, each data point is written into a separate file. It contains the
time series in the columns and the sample points in the rows. The possible options are controlled
be the window in Fig.~\ref{img:export}.

The target directory must be defined manually for compatibility reasons with Matlab Version 5.3 (due to a missing
window for the selection of directories). A copy from the clipboard is possible. The target directory must exist.

The directory and file structure names are mainly determined by the following options.

''Write output variables in subdirectories'' generates subdirectories for the first $(n-1)$ output
variables and codes the $n$-th output variable in the file name. The file for each data point is
saved in the directories and file with the corresponding name.

Otherwise, the file name is generated by all terms. The parts are separated by the separator
defined by ''Separator for the output variables for file names''. The option ''None'' works without
separators.

If more than one data point for a term combination exists, a further directory splitting will be
done. In this case, the file name is only the number of the data point.

''File extension'' control the extension for the generated files.

Example:

An export of time series into the destination folder \texttt{d:$\backslash$prj$\backslash$names} in
a project with the output variables ''Names'' (terms: ''Anna'', ''Maria'', ''Thomas'', ''Peter'')
and ''Examination'' (terms: ''Pre-therapeutic'', ''Post-therapeutic'') give the following file
structure using the option ''Write output variables in subdirectories'':

 \texttt{d:$\backslash$prj$\backslash$names$\backslash$Anna$\backslash$Post-therapeutic.txt}

 \texttt{d:$\backslash$prj$\backslash$names$\backslash$Anna$\backslash$Pre-therapeutic.txt}

 \texttt{d:$\backslash$prj$\backslash$names$\backslash$Peter$\backslash$Post-therapeutic.txt}

 \texttt{d:$\backslash$prj$\backslash$names$\backslash$Peter$\backslash$Pre-therapeutic.txt}

 \texttt{d:$\backslash$prj$\backslash$names$\backslash$Maria$\backslash$Post-therapeutic.txt}

 \texttt{d:$\backslash$prj$\backslash$names$\backslash$Maria$\backslash$Pre-therapeutic.txt}

 \texttt{d:$\backslash$prj$\backslash$names$\backslash$Thomas$\backslash$Post-therapeutic.txt}

 \texttt{d:$\backslash$prj$\backslash$names$\backslash$Thomas$\backslash$Pre-therapeutic.txt}

If more than one data point for a term combination exist, the directory and file names are slightly
different, e.g.:

 \texttt{d:$\backslash$prj$\backslash$names$\backslash$Anna$\backslash$Post-therapeutic$\backslash$1.txt}

 \texttt{d:$\backslash$prj$\backslash$names$\backslash$Anna$\backslash$Post-therapeutic$\backslash$5.txt}

The separators for columns and decimal numbers are the same as for the import.

The option ''Write names in first row'' writes the names of the single features and time series in
the first row of the export file.

''Export only selected data points and features'' uses the recent selection in the project to
reduce the data for the export.

For the export of single features, all data points are written in one file. It contains the single
features in the columns and the data points in the rows.

Two differences exist for the \textindex{export} of single features:
\begin{itemize}
    \item A dialog box is shown for a file selection. Single features will be always exported into one file.
    \item An additional option ''Export output variables as single features'' add all output
          variables to the single features for the export. It simplifies the matching by data
          points and output classes. For time series, this task is solved by the file and directory names.
\end{itemize}

\subsection{Manual creation of SciXMiner projects}
\label{ssec:manuelles_projekt} In this section the creation of a \textmainindex{SciXMiner project
file} is described, either manual or semi-automatic, if a fully automatic \textindex{import} is not
possible. An overview of identifiers which are used by SciXMiner is given in
Appendix~\ref{sec:variablen}.

SciXMiner projects are binary Matlab files. These files contain variables and at least one of the
following identifiers has to be defined within the project file:
\begin{itemize}
    \item{\texttt{d\_orgs}: contains time series of the project. 3D-array with dimension $N \times K \times s_z$}
    \item{\texttt{d\_org}: contains features of the project. 2D-array with dimension $N \times s$}.
\end{itemize}
$N$ denotes the number of data points, $K$ denotes the number of sample points of the time series,
$s_z$ is the number of time series and $s$ is the number of features (see
Section~\ref{ssec:problemstellung}).

The most comfortable way is to use the function \texttt{generate\_new\_scixminer\_project.m}. This function supports the conversion of MATLAB variables into SciXMiner projects. All possible elements are explained in the following text and in the help text of the function, including some examples for the conversion.

If output variables are known, they may be saved in an identifier called \texttt{code\_alle}.

This identifier contains an $N \times s_y$ matrix with the $s_y$ different types of output
variables in the columns and the class assignments  of $N$ data points in the rows. Thereby, each
class is represented by a (positive) integer number. If this identifier is not defined, the class
assignment of each data point is set to 1.

Optionally, the following identifier may be defined. In order to simplify the handling of SciXMiner, the definition is recommended:
\begin{itemize}
    \item{Names for output variables may be given in the identifier \texttt{bez\_code}.
    This identifier contains a string-matrix, where the number of rows has to match the number of output variables. The number of columns depends
    on the length of the strings. An easy method to create \texttt{bez\_code} is the integrated Matlab function \texttt{strvcat}.
    If \texttt{bez\_code} is not defined, the output variables are named ''y1'', ''y2'', ...}
    \item{The classes of the output variable may also be named by an identifier. Therefore, the struct \texttt{zgf\_y\_bez} with the dimension $s_y \times
    \max(m_{y,i})$ is used. $m_{y,i}$ denotes the number of linguistic terms of the $i$-th output variable. The identifier is contained in the
    field \texttt{name}.\\

        Example: There are two output variables, containing two respectively four classes. The identification of these classes can be done using\\
        \texttt{zgf\_y\_bez(1,1).name = 'Anna'};\\
        \texttt{zgf\_y\_bez(1,2).name = 'Maria'};\\
        \texttt{zgf\_y\_bez(1,3).name = 'Thomas'};\\
        \texttt{zgf\_y\_bez(1,4).name = 'Peter'};\\
        \texttt{zgf\_y\_bez(2,1).name = 'pre-therapeutic'};\\
        \texttt{zgf\_y\_bez(2,2).name = 'post-therapeutic'};\\
    }
    \item{The identifiers of time series are arranged in a matrix called \texttt{var\_bez}. The number of rows of the matrix is equal to
    the number of time series within the project. The number of columns depends on the length of the identifiers (equal to \texttt{bez\_code}).}
    \item{The identifiers of features are arranged in a matrix called \texttt{dorgbez}. The number of rows of the matrix is equal to
    the number of features within the project. The number of columns depends on the length of the identifiers (equal to \texttt{bez\_code}).}
\end{itemize}

Another two variables may be inserted optionally, although the use is rather untypical:
\begin{itemize}
    \item{Project-specific struct \texttt{projekt}: This struct might contain additional information for the project. It will read out during
    loading the project and are written to \texttt{parameter.projekt}.}
    \item{To weight features so called \textindex{a priori relevances} may be used. They are multiplied by the calculated feature relevances.
    Thus, features with doubtful values based on difficult environments for sensor measurements may be downgraded, to force the feature selection
    to use reliable features for classification. These a priori relevances have to be stored in a variable called
    \texttt{interpret\_merk}. \texttt{interpret\_merk} is a $s \times 1$ vector.}
\end{itemize}

The project has to be saved using the suffix ''.prjz''. An easy way is the use of the Matlab
command \texttt{save}. As additional parameter \texttt{-mat} has to be used.

\section{Automation of analysis}\label{sec:automatisierungen}
The main strategy elements for the automation of analysis processes are
\begin{itemize}
 \item macros,
 \item SciXMiner batch files and
 \item generated PDF files with project results.
\end{itemize}
Macros are files containing sequences of clicked menu items and control elements. A manual modification of this file is possible due to its textual Matlab syntax. They can be recorded, executed, and modified (see remarks at  \hyperlink{MI_Makro_Simultan}{\emph{Extras - Record macro...}} and example in Section~\ref{sec:building}).

To \textindex{Save a SciXMiner project} automatically in a \textmainindex{macro}, the name of the project must be manually defined inside a macro to avoid the input from a GUI. This can be done by adding the following lines in a macro:

\begin{lstlisting}
%adds '_new' to the name of the current SciXMiner project,
%e.g. myproject.prjz will be saved as myproject_new.prjz
next_function_parameter = [parameter.projekt.datei '_new'];

%% save the new project
eval(gaitfindobj_callback('MI_Speichern'));
\end{lstlisting}

To \textindex{execute a macro inside a other macro}, the following lines must be added in the calling macro:

\begin{lstlisting}
%defines the name of the macro
%here: computing the exponential function for all selected single features
next_function_parameter = 'feature_plugin_exp.makrog';

%execute the macro
execute_macro_inside_macro;
\end{lstlisting}

A \textindex{SciXMiner batch file} is a file for the automatic execution of macros for one or more projects. The batch file might contain one project or a complete directory with all SciXMiner projects in all subdirectories. Optionally, a file with a list of control elements with the extension *.uihdg is loaded by \hyperlink{MI_Einstellungen_laden}{\emph{File - Options - Load options}}. For
each of these projects, the following list of macros or *.m-files or variable assignments into the fields of the variable \texttt{gaitcad\_extern.user} is executed. These variable assignments have to be complete matlab commands. They are valid for all following macros. A recursive definition with other batch files is possible.

During execution, the related projects are loaded and the macros are executed. All files must be exist. Errors does not caused a stop of the execution. Errors are only written in a file called  error.log.

The files must hold the following scheme:

\begin{lstlisting}
1-Inf x following block:
[ 1    x *.batch file or
[ 1    x *.prjz file or directory
 0-1   x *.uihdg file
 1-Inf x *.makrog file or *.m file or variable assignments in gaitcad_extern.user]]
\end{lstlisting}

Example for a SciXMiner batch file:
\begin{lstlisting}
building.prjz
building_regression.makrog
show_report.makrog
building_trigger.makrog
building_teilen.makrog
building_day.prjz
building_cluster.makrog
gaitcad_extern.user.test = 1;
show_report.makrog
\end{lstlisting}

All project, batch, macro, option and m files can be written with absolute directory names. Relative directory names should be predefined in the variable \texttt{gaitcad\_extern.user}. This strategy simplifies the modification of SciXMiner batch files with complex function calls, e.g., for the transfer to a computer with a different folder structure. Here, a project and a macro directory as well as the recent working directory (pwd) are supported.

As an example, the directories can be defined in a SciXMiner batch file called \texttt{local\_directories.batch}:
\begin{lstlisting}
gaitcad_extern.user.project_directory = 'd:\projects'
gaitcad_extern.user.macro_directory = 'd:\macros'
\end{lstlisting}
The main SciXMiner batch file can be organized as follows:
\begin{lstlisting}
local_directories.batch
project_directory\building.prjz
macro_directory\building_regression.makrog
macro_directory\show_report.m
macro_directory\building_trigger.makrog
macro_directory\building_teilen.makrog
project_directory\building_day.prjz
macro_directory\building_cluster.makrog
gaitcad_extern.user.test = 1;
macro_directory\show_report.m
\end{lstlisting}

SciXMiner batch files can be started from the GUI or using the Matlab command line:

\begin{lstlisting}
gaitbatchfile = 'd:\matlab\scixminer\prj\building_demo.batch';scixminerbatch;
\end{lstlisting}

A special option is the use of an empty project called \textindex{noproject.prjz}. After loading, the menu items can be enabled. Consequently, otherwise disabled menu items as e.g. the import of text files can be executed.

PDF files can be generated for the documentation of project containing figures and latex files. This option requires a Latex installation (Open-Source). Further remarks can be found under \hyperlink{MI_Extras_Autoreport}{\emph{Extras - Generate PDF project report (needs  Latex)}}.

Using these strategy elements, even larger projects with a high computational effort or complete directories with many projects can be processed automatically.

\section{Errors and warnings}\label{sec:fehler_und_warnungen}
\begin{figure}[hbt]
\begin{center}
 \includegraphics[width=.7\columnwidth]{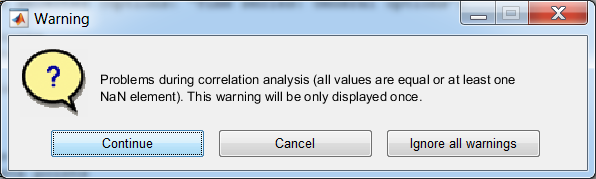}
\end{center}
\caption{\label{fig:warnmeldung} Warning message in SciXMiner}
\end{figure}

Many errors and warnings in functions are displayed in a separated message box (see example for a
warning message in Fig.~\ref{fig:warnmeldung}).\index{error message}

Three different options are available for a \textindex{warning message}:
\begin{itemize}
 \item 'Continue': The function will be continued.
 \item 'Cancel': The function will be canceled. An error message
       'Canceled by user!' will be shown in the Matlab command window.
 \item 'Ignore all warnings': It is similar to 'Continue', but all future warning methods
       will be ignored for the rest of the SciXMiner session.
\end{itemize}
Error and warning messages stop SciXMiner until the message was confirmed or the message window was
closed. The possible debugger options ''Stop if Error'' or ''Stop if Warning'' in the MATLAB editor
redirect to the source code. It makes the confirming of the messages more complicated. It is
recommended to avoid these options.

Tip: Please change the waiting options in myerror.m resp. mywarning.m if you prefer a different
behavior.

\section{Generation of application-specific extension packages}\index{Erweiterungen}
\textmainindex{Application-specific extension packages} can be easily implemented by the function
\texttt{newpackage}. It integrates the new package into the graphical user interface (GUI) of
SciXMiner. The function is located in the directory
\begin{verbatim}
\...\application_specials\template
\end{verbatim}
and can be started after the start of SciXMiner.

The name of the new package is defined by a call of \texttt{newpackage}. Here, blanks and symbols
should be avoided. To guarantee valid function names, very long names will be cut to 12 characters.

The function generates templates for new menu items and control elements in a new directory. These
templates are a starting point for a manual modification. Details are explained as comments in the
files.

The new package is available after the next SciXMiner start. Packages can be switched off by
deactivating of the package in \hyperlink{MI_Extras_Erweiterungen}{\emph{Extras - Choose application-specific extension packages...}} and finally removed by deleting the
directory of the package.

Example:

The command
\begin{verbatim}
 newpackage('Diagnosis',parameter)
\end{verbatim}
generates a new directory
\begin{verbatim}
 \...\application_specials\diagnosis
\end{verbatim}
with functions as e.g. \texttt{menu\_elements\_diagnosis.m}  (Menu items),
\texttt{control\_elements\_diagnosis.m}  (Control elements), \texttt{optionen\_felder\_diagnosis.m}
(location of control elements). In addition, the templates for callback functions are can be used
for the integration of own functions using any parameter from the control elements or available
variables.

\section{Known errors and problems}\index{error handling}
The toolbox was tested using Matlab 2017a. However, we may have overlooked one or
another bug. To some extent, there arise problems based on Matlab itself. Especially, many problems
can be traced back to a missing downward compatibility from Matlab which leads
to differences in the syntax of some commands and demands for long-winded compromises.

Regarding the activation and deactivation of menu items, we decided to activate some menu items although the recently
chosen parameters do not permit the execution of the underlying function. However, in this way the user is enabled to
recognize and to correct his mistake. If the menu items remained deactivated until correct parameters are chosen, the
user would have to try various combinations without getting an output.

Errors in functions are normally caught by separate warning and error messages. Possible Matlab
editor settings ''Stop if Error'' or ''Stop if Warning'' lead the user in the case of warnings or
errors to the corresponding source code. For safe use of Matlab, these options are better
deactivated.

Known bugs and problems can be looked up in detail in Appendix~\ref{sec:anhang_fehler}.

\chapter{Sample projects}\label{sec:demo}\thispagestyle{empty}
\section{Building data set} \label{sec:building}
\subsection{Introduction}
The goal of the Building-data set was to examine the effect of different parameters on the power
consumption of a known building \cite{Kreider93}. It contains a single example with ten time series
(e.g. year, month, day, hour, energy consumption, temperature) and no further features. The time
series are recorded for $175$ days, $24$ measures a day (every hour).\index{Building data set}

With this data set the following topics are addressed:
\begin{itemize}
 \item{splitting of time series and manipulation of data sets}
 \item{usage of macros and plugins}
 \item{application of cluster algorithms}
\end{itemize}
The given problem is the examination and comparison of the power consumption over a day.

The subdirectory "prj" of the SciXMiner installation contains the Building data set as a SciXMiner
project file (building.prjz).

\subsection{Preparation of the data set}
For the analysis of the power consumption over a day the time series have to be split in several
data points, so that only the data of a single day is included in the time series. This is done via
\hyperlink{MI_Extraktion_Teile_ZR}{\emph{Edit - Convert - Separate time series with trigger events...}}. The time series is split with respect to a given trigger signal.

A typical application is the extraction of segments of time series out of long recordings. Examples
for trigger events are the occurrence of errors, start of a new measurement or external events. The
trigger signal contains only zeros, except the sample points where a \textindex{trigger event}
occurs. At these sample points, the amplitude of the trigger signal equals the linguistic term of
the output variable.

First, since this data set contains no defined trigger events, a trigger signal has to be created.
Three different options exist
\begin{itemize}
    \item using a graphical editor for a manual definition by \hyperlink{MI_Erstelle_Trigger}{\emph{Edit - Generating trigger time series}} resp. \hyperlink{MI_Bearbeite_Trigger}{\emph{Edit - Edit trigger time series}},
    \item using a macro, or
    \item using a plugin for extraction of features or time series.
\end{itemize}
The result is the same for all approaches: a time series containing values unequal to $0$ at
trigger events. This time series is called trigger signal.

In this project, the trigger events are extracted out of a time series containing the hours of a
day. The trigger events are defined as the sample points at which the hour declines from $23$ to
$0$.

For the sake of completeness, we consider all three approaches for the extraction of the trigger
events:

\subsubsection{Graphical generation and editing of a trigger time series}\label{sec:edit_trigger}
\begin{figure}[hbt]
\begin{center}
\includegraphics[width=\columnwidth]{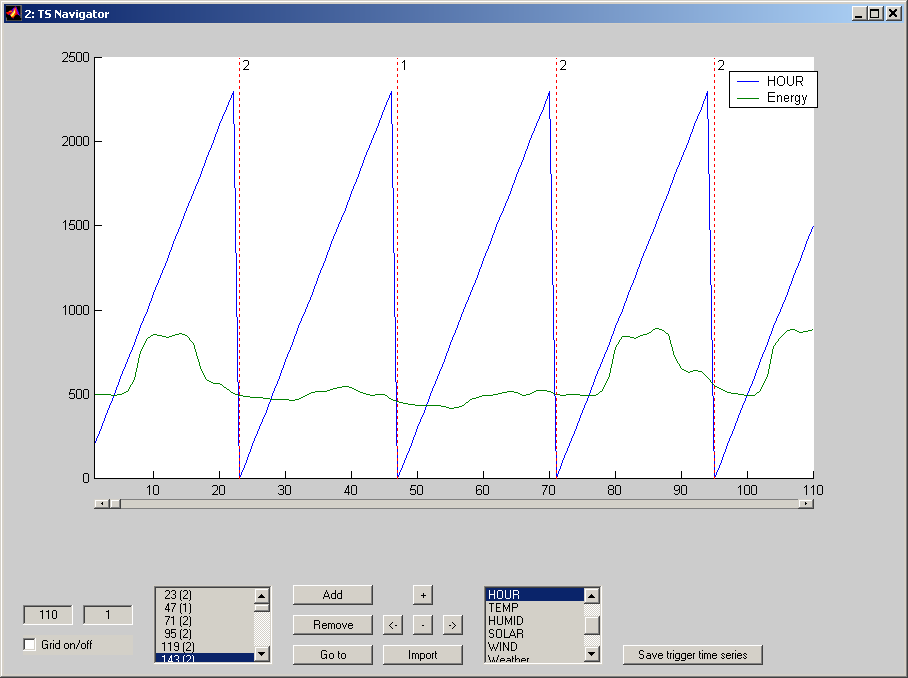}
\caption{\label{fig:trigger_grafik} Graphical generation and editing of a trigger time series}
\end{center}
\end{figure}

A trigger time series can be generated or edited with the window in Fig~\ref{fig:trigger_grafik}.
The details are explained in \hyperlink{MI_Erstelle_Trigger}{\emph{Edit - Generating trigger time series}}. Fig.~\ref{fig:trigger_grafik} shows four
trigger events for the sample points 23, 47, 71 and 95 with two different classes for the output
variable (weekend:~1 and working day:~2). The displayed time series show the ''Hour'' and the
''Energy''. The temporal resolution is tuned by the two fields in the lower left corner (left:
length of the shown time series segment, here:~110, right: each n-th element is shown, here:~1).

The disadvantage of this methods is very high manual effort for long time series. Consequently, the
two following methods are better:

\subsubsection{Extraction of trigger signal by use of macros}
The following code have been created automatically by recording of a macro. The used time series
and interval for the extraction are specific for every project. Thus, a macro is generally not
usable for different projects.

The recording contains the following commands (applied directly in the GUI):
\begin{itemize}
    \item {\hyperlink{MI_Makro_Simultan}{\emph{Extras - Record macro...}}}
    \item {\hyperlink{MI_Extraktion_ZRZR}{\emph{Edit - Extract - Time series -> Time series, Time series -> Single features...}}}
    \item{Selection of time series "Day", interval "Whole time series" and plugin "Velocity (V)".
    Finish the selection by a click on "OK"} \index{segments}
    \item {\hyperlink{MI_Makro_Beenden}{\emph{Extras - Stop macro record}}}
\end{itemize}
The content of the macro file is as follows: \lstset{language={}, numbers=left, numberstyle=\tiny,
stepnumber=3, numbersep=5pt, basicstyle=\small, commentstyle=\textit}
\begin{lstlisting}
 % MAKRO AUSWAHLFENSTER Zeitreihe -> Zeitreihe, Zeitreihe -> Einzelmerkmal
 auswahl.gen=[];
 auswahl.gen{1}={'DAY'};
 auswahl.gen{2}={'Whole time series (0...100%)'};
 auswahl.gen{3}={'Velocity (V)'};
 eval(gaitfindobj_callback('MI_Extraktion_ZRZR'));
 eval(get(figure_handle(size(figure_handle,1),1),'callback'));
\end{lstlisting}

Due to the computation of the velocity in SciXMiner by the formula $$f'(t) =
\frac{f(t+1)-f(t-1)}{2}$$ the trigger signal contains two sample points with values unequal to zero
at the trigger events. Thus, the macro has to be expanded by the following code.
\begin{lstlisting}
% The new time series is the last one
day_abl = squeeze(d_orgs(1, :, end));
% Search for all samples with value unequal to zero
indx = find(day_abl ~= 0);
% Two subsequent samples are zero. The second
% is the wanted sample. Set the first to zero.
d_orgs(1, indx(1:2:end), end) = 0;
% The first day is not identified by this approach.
% Set it manually:
d_orgs(1, 1, end) = 1;
% The amplitude of the trigger signal corresponds to the
% class of this trial. Add temporarily a new class
% containing the subsequent number of the day:
indx = find(squeeze(d_orgs(1, :, end)) ~= 0);
d_orgs(1, indx, end) = [1:length(indx)];
% Clear used variables
clear day_abl indx;
\end{lstlisting}

Applying this macro by \hyperlink{MI_Makro_Ausfuehren}{\emph{Extras - Play macro...}} leads to a new time series which is called "Day
V".

The subdirectory "prj" of the SciXMiner installation contains the file of the macro (comments in
German) (building\_trigger.makrog).

\subsubsection{Extraction of trigger signal by plugins}
A further approach of extracting a trigger signal is the usage of plugins. A plugin is a function
which is automatically included in SciXMiner. The related M file has to be in the subdirectory
"$\backslash$plugins$\backslash$mgenerierung". The code of a plugin for the extraction of the
trigger signal is as follows (for a detailed description of plugins refer to
Chapter~\ref{sec:plugins_und_einzuege}):
\begin{lstlisting}
  function [datenOut, ret, info] = plugin_nsprung(paras, datenIn)

anz_zr = 1;
info = struct('beschreibung', 'Jump to zero', 'bezeichner', 'NSprung', 'anz_zr', anz_zr, 'anz_em', 0, 'laenge_zr', paras.par.laenge_zeitreihe, 'typ', 'TS');
info.einzug_OK = 0;
info.richtung_entfernen = 0;
info.anz_benoetigt_zr = 1;

info.explanation      = strcat('set value  = 1 for a jump to zeros, and value = 0 otherwise.');

if (nargin < 2 | isempty(datenIn))
   datenOut = [];
   ret = [];
   return;
end;

%detect zero values
datenOut.dat_zr = (datenIn.dat == 0);

%detect a jump form a non-zero value to a zero value
datenOut.dat_zr(:,2:end,:) = datenOut.dat_zr(:,2:end,:) & (datenIn.dat(:,1:size(datenIn.dat,2)-1,:)~=0);

ret.ungueltig = 0;
ret.bezeichner = 'NSprung';
\end{lstlisting}
A sample of the new time series is zero if the sample of the original time series is zero and the
preceding sample of the original time series is unequal to zero. For the extraction of the trigger
signal we have to apply the plugin by \hyperlink{MI_Extraktion_ZRZR}{\emph{Edit - Extract - Time series -> Time series, Time series -> Single features...}} with the selections "Original data
(time series)": "Hour", "interval": "Whole time series" and plugin "Jump to zero (NSprung)". This
plugin is unable to detect the first day, since there is no "jump" to zero. Thus, the splitting of
the time series by usage of the macro leads to one data point more than with extraction by plugin.

The subdirectory "prj" of the SciXMiner installation contains the file of the plugin
(plugin\_nsprung.m). It is automatically detected when a project is loaded.

\subsubsection{Splitting of the time series}
After extracting the trigger signal, the project is ready to be converted. This is done by
\hyperlink{MI_Extraktion_Teile_ZR}{\emph{Edit - Convert - Separate time series with trigger events...}}. The options are set as shown in the following figure (use "Hour
NSprung" as time series if the trigger signal is extracted by the plugin).\index{splitting of time
series}
\begin{figure}[hbt]
\begin{center}
\includegraphics[width=.8\textwidth]{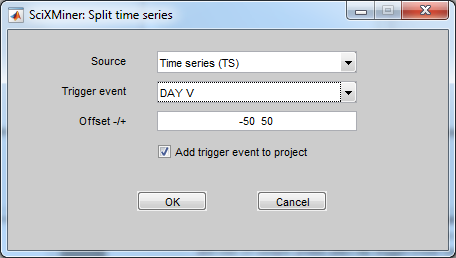}
\caption{Options for the splitting of time series}
\end{center}
\end{figure}

If the first option is set to "Workspace" instead of "time series", variables from the workspace
can be used as trigger signals. Since the extraction of the single trials should be start exactly
at a trigger event and end $23$ sample points after the trigger event (resulting in $24$ hours
after the trigger event), the offset is set to $[0,23]$.

By clicking on "OK" the project is saved with a new name and loaded automatically. The new project
contains the same amount of time series ($10$), but $175$ examples instead of $1$ (resp. $174$
after extracting the trigger signal by the plugin).

\subsection{Application of cluster algorithms}
In this section the application of clustering is shown. It is applied to the time series "Energy" in the project
containing the split time series. If you did not apply the conversion of the time series described in the previous
sections please use the project "building\_day.prjz", included in the subdirectory "prj" of the SciXMiner installation.

\begin{figure}[hbt]
\begin{center}
\includegraphics[width=.8\textwidth]{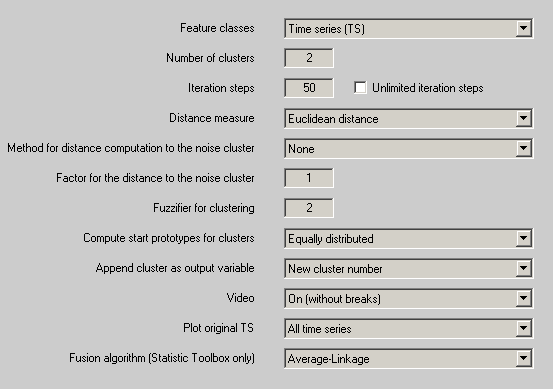}
\caption{\label{img:cluster_ein}Options for the cluster algorithm}
\end{center}
\end{figure}

In the options window \hyperlink{CEO_ZRAllg}{\emph{Control elements:  Time series: General options}} the time series "Energy" is selected. The settings for the cluster algorithm
are shown in Fig.~\ref{img:cluster_ein}.

The clustering is started by a click at \hyperlink{MI_Cluster_Ber}{\emph{Data mining - Clustering - Design and apply}}. If the computation is finished the result can be
visualized by \hyperlink{MI_Ansicht_ClusterZGH_unsortiert}{\emph{View - Clustering - Cluster memberships (sorted by data points)}}.

\begin{figure}[hbt]
\begin{center}
\includegraphics[width=.7\textwidth]{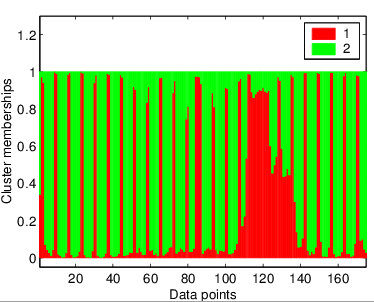}
\caption{Results of the cluster algorithm for the time series ''Energy''}
\end{center}
\end{figure}
\begin{figure}[hbt]
\begin{center}
\includegraphics[width=.7\textwidth]{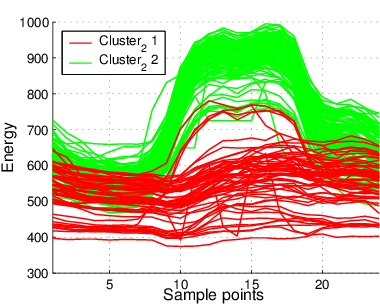}
\caption{\label{img:cluster_zr}Visualization of the original data of the time series ''Energy'' vs.
the cluster membership values}
\end{center}
\end{figure}

The clusters correspond to weekday (Cluster~2, green), resp. weekend or vacations (Cluster~1, red). The rather long red
interval belongs to Christmas time. To visualize the single time series with respect to the cluster membership, the
following steps have to be accomplished (as a prerequisite, \hyperlink{CE_Cluster_Anhaengen}{\emph{Control element: Data mining: Clustering - Append cluster as output variable}} has to be set to "new cluster
number"): Set the output variable to the new one created by application of the cluster algorithm (option window
\hyperlink{CEO_ZRAllg}{\emph{Control elements:  Time series: General options}}) and the time series to "Energy". By clicking at \hyperlink{MI_Anzeige_ZR_Orig}{\emph{View - Time series (TS) - Original time series}} every example of time
series "Energy" is plotted as single curve (cf. Fig.~\ref{img:cluster_zr}). The class means of the time series are
plotted by clicking at \hyperlink{MI_Anzeige_ZR_MW}{\emph{View - Time series (TS) - Mean time series}}.

All commands are included in the macro building\_cluster.makrog in subdirectory "prj" of the
SciXMiner installation.

\subsection{Time series prediction}
\begin{figure}[hbt]
    \begin{center}
        \includegraphics[width=.8\textwidth]{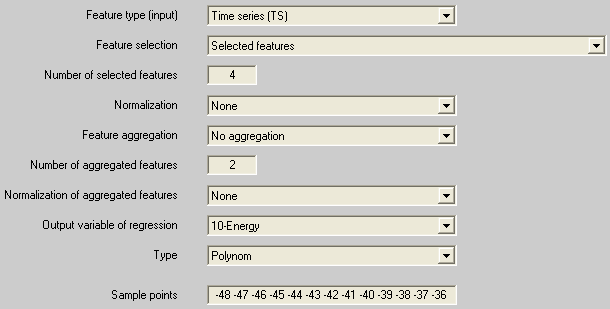}
        \caption{\label{img:building_regression_ein}Settings for the regression.}
    \end{center}
\end{figure}

In this section the prediction of the energy time series of the building dataset is shown. The used settings for the
regression are included in Fig.~\ref{img:building_regression_ein}. All time series must be selected, including the
energy time series!

In this example, all selected time series are used for the regression. An example using linear regression coefficients
for a feature preselection is shown later using the iris dataset. The samples of the previous day are used as features
for the regression (samples $k-24, k-25, \hdots, k-48$). To edit the samples normal Matlab syntax is allowed (e.g.
-48:-24 or -72:-24:-168). The sample $0$ is automatically removed from the vector to avoid trivial prediction.

Necessary settings for the polynomial are the degree (here: 2, \hyperlink{CE_Regression_GradPolynom}{\emph{Control element: Data mining: Special methods - Polynomial degree}}) and the maximal number of
internal features (here: 4, \hyperlink{CE_Regression_AnzahlPolyMerkmale}{\emph{Control element: Data mining: Special methods - Maximal number of internal features}}). The maximal number of internal features specifies
how many coefficients should be computed by the regression algorithm. A selection of $10$ time series and $24$ samples
leads to a total amount of possible coefficients of $240$. In this case, $4$ are selected by the regression algorithm.

\begin{figure}[hbt]
    \begin{center}
        \includegraphics[width=.5\textwidth]{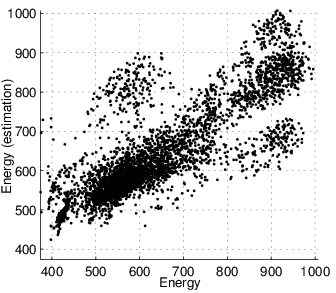}
        \caption{\label{img:building_regression_scatter}Result of regression.}
    \end{center}
\end{figure}

By clicking \hyperlink{MI_Regression_EnAn}{\emph{Data mining - Regression - Design and apply}} the regression algorithm computes the coefficients and a prediction of the time
series. A graphical examination of the result can be obtained by the functions included in \hyperlink{MI_Anzeige_Regression}{\emph{View - Regression}}
(see Fig.~\ref{img:building_regression_scatter} for a scatter plot of the output variable and the estimation).

\begin{figure}[hbt]
    \begin{center}
        \includegraphics[width=.5\textwidth]{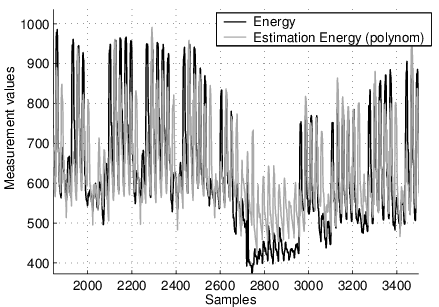}
        \caption{\label{img:zeitreihen_vorhersage2}Part of the original time series and predicted time series.}
    \end{center}
\end{figure}

The scatter plot includes all values of the output variable versus the estimation of the output variable for all
samples. A disadvantage of this plot is the lost time information. But since the prediction of the output variable is
added as a new time series to the project all build-in visualizations can be used, e.g. the visualization of the
original time series versus the predicted time series. To force SciXMiner to plot both time series in the same subplot,
the option \hyperlink{CE_Zeitreihen_Subplots}{\emph{Control element: View: Time series - Show time series as subplots}} must be turned off. A section of such a plot is included in Fig.~\ref{img:zeitreihen_vorhersage2}.

By \hyperlink{MI_ZRAusw_RegrUni}{\emph{Data mining - Evaluation of time series - Linear regression coefficients (univariate)}} univariate regression coefficients (multivariate is also possible) can be computed and
visualized by \hyperlink{MI_Anzeige_ZRRelevanz}{\emph{View - Time series (TS) - Show all feature relevances (sorted table)}} (sorted by importance). The used sample is enclosed by squared brackets.

\section{Iris data set}
\subsection{Introduction}
The \textindex{Iris data set} (downloadable from UCI-Repository \cite{Hettich99}) is one of the
most-famous benchmark data sets for the comparison of classifiers. It describes three different
irises by means of their petal width and length and sepal width and length. The data set contains
$150$ examples of three different irises, every single iris is represented by $50$ examples
\cite{Anderson35}. It is used to show the application of classifiers to a data set containing
single features.

\subsection{Classification}
Since the data set contains no different examples for design and validation of the classifiers all
examples are used for both phases of classification.

First, all examples are chosen by clicking on \hyperlink{MI_Datenauswahl_Alle}{\emph{Edit - Select - All data points}}. Single classes are selectable by clicking on
\hyperlink{MI_Datenauswahl_Klassen}{\emph{Edit - Select - Data points using classes ...}}, e.g. to select examples for the design of the classifier.

For this example, the parameters have been set as shown in figure~\ref{img:iris_ein} (\hyperlink{CEO_DM_Grund}{\emph{Control elements:  Data mining: Classification of single features}}):
\begin{figure}[hbt]
    \begin{center}
        \includegraphics[width=.8\textwidth]{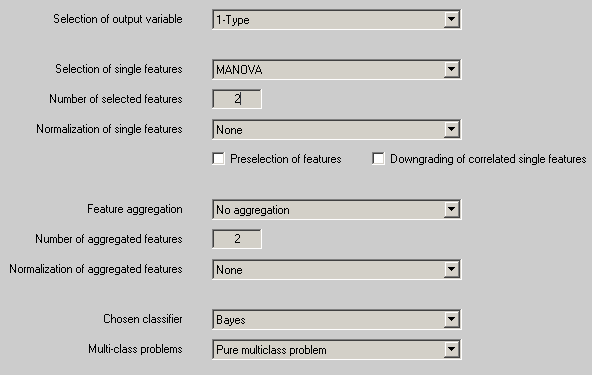}
        \caption{\label{img:iris_ein}Options of the classifier for the Iris data set}
    \end{center}
\end{figure}

The selected parameters of the classifier are the default values for the Bayes classifier (Metric:
Tatsuoka distance, no use of \textindex{a-priori probabilities}). To change these settings select
the options window \hyperlink{CEO_DM_Klass}{\emph{Control elements:  Data mining: Special methods}}. The \textindex{Tatsuoka distance} is an internal notion of
class-specific distances comparable to the Mahalanobis distance. The means and covariance matrices
are approximated separately for every single class.\index{Bayes classifier}

By a click on \hyperlink{MI_EMKlassi_EnAn}{\emph{Data mining - Classification - Design and apply}} the design of the selected classifier with respect to the currently selected
examples is started and afterwards applied to the selected examples. The separate design and application of the
classifier is done by \hyperlink{MI_EMKlassi_En}{\emph{Data mining - Classification - Design}} and \hyperlink{MI_EMKlassi_An}{\emph{Data mining - Classification - Apply}} respectively.

The classification result is displayed on the Matlab command window, independently from the
dimension of the feature space:
\begin{small}
\begin{verbatim}
Apply classifier ...
150 data points 2 features ...
Complete ...

Analyze classification result ...
Number of misclassifications: 7 of 150 examples ( 4.67 %)
Confusion matrix (Rows: True class assignments, columns: Result of classification):
 50    0    0
  0   47    3
  0    4   46
Complete ...
\end{verbatim}
\end{small}
A graphical visualization of the classification result is only possible for features spaces with dimensions less or
equal to three.

\begin{figure}[hbt]
    \begin{center}
        \includegraphics[width=.8\textwidth]{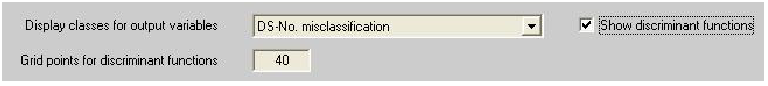}
        \caption{\label{img:iris_visuein}Options for the visualization of the classification results}
    \end{center}
\end{figure}

\begin{figure}[hbt]
    \begin{center}
        \includegraphics[width=.6\textwidth]{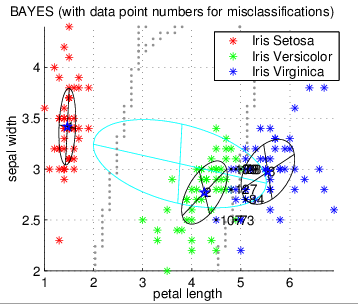}
        \caption{\label{img:iris_erg}Classification result for the Iris data set with two features and a Bayes classifier}
    \end{center}
\end{figure}

The settings for the graphical visualization of the classification result are made in the options
window \hyperlink{CEO_Ans_Kl}{\emph{Control elements:  View: Classification and regression}}. For the following visualization, the settings from
Figure~\ref{img:iris_visuein} are selected. ''DS-No. misclassification'' means, that the examples
are colored with respect to their real class membership and misclassified examples are marked by
their number (see Figure~\ref{img:iris_erg}).

For the visualization in Figure~\ref{img:iris_erg} the menu item \hyperlink{MI_Anzeige_Klassi_Kova}{\emph{View - Classification - 2D classification with covariance matrixes}} was used. This
visualization is only available after application of a Bayes classifier, due to the need of covariance matrices. For
Support Vector Machines, the used support vectors can be displayed by clicking on \hyperlink{MI_Anzeige_Klassi_SVM}{\emph{View - Classification - 2D plot classifier with support vectors}}.
Generally, the visualization of the classification result is done by \hyperlink{MI_Anzeige_Klassi_Erg}{\emph{View - Classification - Result}}.

\subsection{Regression}
This section contains a brief description of the application of regression algorithms to projects containing single
features.

\begin{figure}[hbt]
    \begin{center}
        \includegraphics[width=.8\textwidth]{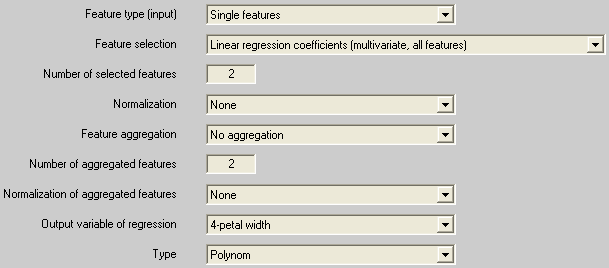}
        \caption{\label{img:iris_regression_ein}Settings for the regression.}
    \end{center}
\end{figure}

To select all data tupels, use \hyperlink{MI_Datenauswahl_Alle}{\emph{Edit - Select - All data points}}. Then, set the settings for the regression as shown in
picture~\ref{img:iris_regression_ein}.

The iris dataset contains only single features (no time series), thus time series can not be selected as input feature
type. The feature selection should be done by multivariate regression coefficients, the number of features to be
selected must be set in \hyperlink{CE_Regression_AnzahlMerkmale}{\emph{Control element: Data mining: Regression - Number of selected features}}. As an alternative, the feature selection could be done
manually by setting the number of features (in \hyperlink{CE_Anzahl_Merkmale}{\emph{Control element: Data mining: Classification of single features - Number of selected features}}) and a feature selection e.g. by using
\hyperlink{MI_EMAusw_MANOVA}{\emph{Data mining - Selection and evaluation of single features - MANOVA, multivariate}}. In this case, the features selection must be set to "selected features" for the regression.

The output variable of the regression is set to "petal width" the type of the regression is set to "polynom". The
parameter for the polynomial are to be set in \hyperlink{CEO_DM_Klass}{\emph{Control elements:  Data mining: Special methods}}. In this example the degree of the polynomial is set to $1$
and the maximal number of internal features to $4$ (see the example using the building dataset for a description of
this option).

The design and application of the regression is started by \hyperlink{MI_Regression_EnAn}{\emph{Data mining - Regression - Design and apply}}. After exiting the algorithm the
command window contains:
\begin{verbatim}
Fitness of regression:
Mean absolute value: 0.154808
Correlation coefficient between true value and estimation: 0.964228
\end{verbatim}
to give a first impression of the regression's quality. The estimation of the feature "petal width" has been added as a
new feature. Make sure that you exclude the estimated feature from the list of used features for further regressions.
Select a feature selection method including "selected features" in the description. (e.g.. "linear regression
coefficients (multivariat, selected features)"). The output variable is automatically removed from the list of used
features, but not estimations of the output variable.

\begin{figure}[hbt]
    \begin{center}
        \includegraphics[width=.5\textwidth]{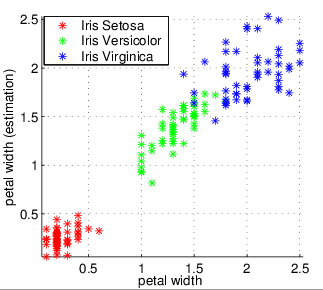}
        \caption{\label{img:regression_iris}Result of the regression of ''petal width''.}
    \end{center}
\end{figure}

More detailed information about the regression's quality can be obtained by the functions in
\hyperlink{MI_Anzeige_Regression}{\emph{View - Regression}}, e.g. the real value versus the estimated value, see picture ~\ref{img:regression_iris}
and the coefficients of the polynomial model.

The used polynomial is:
\begin{align}
\text{petal width} &= -0.722 + 0.427 \cdot \text{petal length}\\
                   &+ 0.103 \cdot \text{sepal width}\nonumber
\end{align}
An examination of single regression coefficients is possible via \hyperlink{MI_EMAusw_RegrUni}{\emph{Data mining - Selection and evaluation of single features - Linear regression coefficients (univariate)}} (visualization of the result
by \hyperlink{MI_Anzeige_EM_Relevanzen}{\emph{View - Single features - Show feature relevances (sorted table)}}).

\clearpage

\chapter{Menu items}\label{sec:menu}

\section{Menu items 'File'}
This point includes all file operations.
\subsection{Load project}
\hypertarget{MI_Laden}{}
A \textindex{SciXMiner project file} \emph{*.prjz} will be loaded and default values are added.
Necessary variables are ''d\_org'' or ''d\_orgs''. In addition, the variables ''code'',
''code\_alle'', ''dorgbez'', ''var\_bez'', ''zgf\_y\_bez'', ''bez\_code'', ''optionen'',
''projekt'', ''interpret\_merk'' are read.
\subsection{Save project}
\hypertarget{MI_Speichern}{}
saves the recent SciXMiner project \emph{*.prjz} using the same file name. This function saves the data set (but without
analysis results as designed classifiers) and optionally the current settings of the control elements (if chosen
in \hyperlink{CEO_Allg}{\emph{Control elements:  General options}}).
\subsection{Save project as...}
\hypertarget{MI_SpeichernUnter}{}
saves the SciXMiner project file \emph{*.prjz} with a new name (see \hyperlink{MI_Speichern}{\emph{File - Save project}}).
\subsection{Mean value project (based on the selected output variable)}
\hypertarget{MI_SaveMeanAggregation}{}
generates a new SciXMiner project with averaged values for all linguistic terms of the selected output variable. In the new project, only one data point per linguistic term exists.
\subsection{Export data}
\hypertarget{MI_Export_ASCII}{}
exports time series or single features in an ASCII file. The separators for columns and decimal numbers can be selected
in the opened configuration window. \index{export}
\begin{itemize}
\item {\bf Time series in multiple files...}: \newline  
The time series of the given project (variable ''d\_orgs'') will be exported into different files. We assume at least
one output variable. If more output variables exist, the first (n-1) output variables build a directory structure, the
information about the n-th output variable is encoded in the file name. Alternatively, all n output variables can be
encoded in the file name if the option ''Separator for output variables in file name'' is chosen. \index{export}
\hypertarget{MI_Export_Ordner}{}
\item {\bf Single features in a file...}: \newline  
The single feature of the given project (variable ''d\_org'') will be exported as file (all data
points in one file). \index{export}
\hypertarget{MI_Export_Datei}{}
\end{itemize}
\subsection{Import data}
\hypertarget{MI_Import_ASCII}{}
includes all functions for the \textindex{import} of data from files.
\begin{itemize}
\item {\bf From a directory...}: \newline  
imports data from a directory structure. Its definition is explained in
Section~\ref{sec:imundexport} and for the menu item \hyperlink{MI_Export_Datei}{\emph{File - Export data - Single features in a file...}}. The parallel import of
single features and time series is impossible. Output variables should be included as single
features in the file and can converted after import using \hyperlink{MI_Trans_Klasse2EM}{\emph{Edit - Convert - Selected output variable $\rightarrow$ Single feature}}.

Some import versions support more complex data formats including strings (see Section 4.2 ''Import and export of projects'' in the SciXMiner documentation). Such strings are converted in output variables.
\hypertarget{MI_Import_Ordner}{}
\item {\bf From a file...}: \newline  
imports data from a file. The filenames encode the output classes.  The parallel import of single
features and time series is impossible. Output variables should be included as single features in
the file and can converted after import using \hyperlink{MI_Trans_Klasse2EM}{\emph{Edit - Convert - Selected output variable $\rightarrow$ Single feature}}.

Some import versions support more complex data formats including strings (see Section 4.2 ''Import and export of projects'' in the SciXMiner documentation). Such strings are converted in output variables.
\hypertarget{MI_Import_Datei}{}
\item {\bf Interactive from a file}: \newline  
imports data with the MATLAB Import Wizard.
\hypertarget{MI_Import_DateiIntAktMATLAB}{}
\item {\bf Import of time series names from file}: \newline  
imports identifiers for time series from a file. It is expected that exactly one identifier exists
for every time series. The identifiers could either be separated by line breaks (i.e. one
identifier per row) or by a separating character (i.e. tabulator, comma, or semicolon).
\hypertarget{MI_Datei_ImportiereZRBezeichner}{}
\item {\bf Import of single feature names from file}: \newline  
see \hyperlink{MI_Datei_ImportiereZRBezeichner}{\emph{File - Import data - Import of time series names from file}}, but for single features instead of time series.
\hypertarget{MI_Datei_ImportiereEMBezeichner}{}
\end{itemize}
\subsection{Fusion of projects}
\hypertarget{MI_Project_Fusion}{}
contains functions for the fusion of existing SciXMiner projects to one joint project called
\emph{fusion.prjz}. It uses all projects from a directory selected by a configuration window.
Incompatible projects will be ignored. Four different types are available.
\begin{itemize}
\item {\bf Additional time series and single features}: \newline  
fuses data from all SciXMiner projects in a directory by adding the data as new time series and single
features. The number of data points, the length of time series, and the number and values of output variables
must be identical for all projects.

Example:

SciXMiner-Projekt a.prjz with 5 data points, 2 time series, 1 single feature, and 1 output variable

SciXMiner-Projekt b.prjz with 5 data points, 2 time series, 1 single feature, and 1 output variable

SciXMiner-Projekt fusion.prjz with 5 data points, 4 time series, 2 single feature, and 1 output variable
\hypertarget{MI_Project_Fusion_ZREM}{}
\item {\bf Additional data points}: \newline  
fuses data from all SciXMiner projects in a directory by adding the data as new data points. The
number and names of single features, time series and output variables and the length of time series
must be identical for all projects. Linguistic terms of the output variable are matched by their
names.

Example:

SciXMiner-Projekt a.prjz with 5 data points, 2 time series, 1 single feature, and 1 output variable

SciXMiner-Projekt b.prjz with 5 data points, 2 time series, 1 single feature, and 1 output variable

SciXMiner-Projekt fusion.prjz with 10 data points, 2 time series, 1 single feature, and 1 output variable
\hypertarget{MI_Project_Fusion_DT}{}
\item {\bf Additional time series and single features (tolerant for the selected output variable and data points, NaNs are deleted)}: \newline  
like \hyperlink{MI_Project_Fusion_ZREM}{\emph{File - Fusion of projects - Additional time series and single features}}, but more tolerant to differences between data points. The values of the selected output variable of the selected data points (e.g. an ID resp. identifier) in the open project are used as the base for the fusion. Each data point will be deleted if the corresponding data points in the other project for the fusion cannot be found. Non selected data points are not used for fusion.
\hypertarget{MI_Project_Fusion_ZREMTol}{}
\item {\bf Additional time series and single features (tolerant for the selected output variable and data points, NaNs are retained)}: \newline  
like \hyperlink{MI_Project_Fusion_ZREM}{\emph{File - Fusion of projects - Additional time series and single features}}, but more tolerant to differences between data points. The values of the selected output variable of the selected data points (e.g. an ID resp. identifier) in the open project are used as the base for the fusion. The fused time series resp. single features of each data point are set to NaN if the corresponding data points in the other project for the fusion cannot be found. Non selected data points are not used for fusion.
\hypertarget{MI_Project_Fusion_ZREMTolNaN}{}
\item {\bf Additional events, event types and data points}: \newline  

\hypertarget{MI_Project_Fusion_Events}{}
\end{itemize}
\subsection{Apply SciXMiner batch file}
\hypertarget{MI_Gaitbatch}{}
executes a \textindex{SciXMiner batch file} with the extension \emph{*.batch}. It can include one or more projects or even directories containing projects.
Optionally, a file with a list of control elements with the extension *.uihdg is loaded by \hyperlink{MI_Einstellungen_laden}{\emph{File - Options - Load options}}. For
each of these projects, a list of macros is executed. A recursive definition with other batch
projects is possible.
\subsection{Apply SciXMiner batch file (debug mode)}
\hypertarget{MI_GaitbatchDebug}{}
like \hyperlink{MI_Gaitbatch}{\emph{File - Apply SciXMiner batch file}}, but with a stop after errors and warnings. This option is useful for debugging SciXMiner batch files. After a stop, the processing can be continued by return.
\subsection{Apply SciXMiner batch file (step and debug mode)}
\hypertarget{MI_GaitbatchDebugStep}{}
like \hyperlink{MI_GaitbatchDebug}{\emph{File - Apply SciXMiner batch file (debug mode)}}, but with an additional step-wise execution of each macro in a SciXMiner batch file (continue with return).
\subsection{Normative data}
\hypertarget{MI_Datei_Norm}{}
contains all operation for the handling of normative data. Examples for \textindex{normative data}
are measurements of healthy people in medical data analysis or a desired process behavior for
technical diagnosis. Normative data are defined by mean value and standard deviations for all time
series and single features. They can be used for a comparison to a freely definable reference (e.g.
for a typical behavior of a plant).
\begin{itemize}
\item {\bf Load normative data}: \newline  
loads \textmainindex{normative data} (mean and standard deviation) for time series and single
features from a '*.norm' file.
\hypertarget{MI_ImportNorm}{}
\item {\bf Save normative data}: \newline  
saves \textindex{normative data} (mean values and standard deviations) for time series and single
features into a '*.norm' file. In detail, the variables ''dorgbez'', ''mstd'', ''mstd\_em'',
''my'', ''my\_em'', ''parameter'', ''titelzeile'', ''var\_bez'' will be saved. This function
requires an executing of \hyperlink{MI_Mittelwert_zu_Norm}{\emph{File - Normative data - Mean value to normative data}} for the generation of normative data.
\hypertarget{MI_ExportNorm}{}
\item {\bf Mean value to normative data}: \newline  
uses the recent data selection for the generation of \textindex{normative data}. Therefore, only one class
(=~linguistic value for the output variable) has to be selected (e.g. in medicine: healthy people, for technical
applications: error-free and normal behavior of a process or device).
\hypertarget{MI_Mittelwert_zu_Norm}{}
\end{itemize}
\subsection{Data mining}
\hypertarget{MI_Datei_DataMining}{}
This menu item contains functions for loading and saving of parameters from designed data mining
systems.
\begin{itemize}
\item {\bf Load fuzzy system}: \newline  
loads a designed \textindex{fuzzy system} from a file with the extension ''*.fuzzy''. The
assignment of input (single features) and output variables is based on the variable names. The file
must contain the variables ''gaitcad\_struct'', ''merkmale\_projekt'', and  ''mode''. The use of
files from other projects gives misleading results, especially in case of identical variable names.

The loaded fuzzy system is not yet directly usable as a fuzzy classifier. This application requires
an export by \hyperlink{MI_FUZZY2KLASS}{\emph{Data mining - Fuzzy systems - Export fuzzy system to classifier}}.
\hypertarget{MI_Fuzzy_RUBImport}{}
\item {\bf Load fuzzy system (only membership functions)}: \newline  
loads only the membership functions of a pre-designed fuzzy system in a file and sets \hyperlink{CE_Fuzzy_TypeZGF}{\emph{Control element: Data mining: Special methods - Type of membership function}} to ''Fix''.
\hypertarget{MI_Fuzzy_RUBImportMBF}{}
\item {\bf Save fuzzy system}: \newline  
exports a designed \textindex{fuzzy system} in a file. The results are saved in the variables
''gaitcad\_struct'', ''merkmale\_projekt'', and ''mode''.
\hypertarget{MI_Fuzzy_RUBExport}{}
\item {\bf Export ANSI-C-Code fuzzy rulebase}: \newline  
exports the recent fuzzy rulebase in a C file. This C file needs a pointer to a vector x containing the selected input variables (single features) and to a pointer of membership values of the output variable. The numbers of the input variables are explained in the function header. The output of the function is the linguistic term of the output variable. The feature numbers start with x[1] instead of x[0]!
\hypertarget{MI_Fuzzy2C}{}
\item {\bf Load classifier}: \newline  
loads a designed \textindex{classifier} from a file with the extension ''*.class''. It contains all settings specified
in \hyperlink{CEO_DM_Grund}{\emph{Control elements:  Data mining: Classification of single features}} and \hyperlink{CEO_DM_Klass}{\emph{Control elements:  Data mining: Special methods}} during the design. The assignment of input (single features) and output
variables is based on the variable names. The file must contain the variables ''gaitcad\_struct'',
''merkmale\_projekt'', and  ''mode''. The use of files from other projects might give misleading results, especially in
case of identical variable names.
\hypertarget{MI_Classifier_Import}{}
\item {\bf Save classifier}: \newline  
exports a designed \textindex{classifier} in a file. The results are saved in the variables ''gaitcad\_struct'',
''merkmale\_projekt'', and ''mode''. They contain all settings defined by the parameters in \hyperlink{CEO_DM_Grund}{\emph{Control elements:  Data mining: Classification of single features}} and
\hyperlink{CEO_DM_Klass}{\emph{Control elements:  Data mining: Special methods}}.
\hypertarget{MI_Classifier_Export}{}
\item {\bf Export ANSI C Code of the Bayes classifier}: \newline  
exports the recent Bayes classifier with feature selection and feature transformation into an \textindex{ANSI C file}.
\hypertarget{MI_Bayes2C}{}
\item {\bf Load regression model}: \newline  
loads a designed \textindex{regression model} from a file with the extension ''*.regression''. It contains
all steps specified by \hyperlink{CEO_DM_Regr}{\emph{Control elements:  Data mining: Regression}} and \hyperlink{CEO_DM_Klass}{\emph{Control elements:  Data mining: Special methods}} during the design. The assignment of input
(single features) and output variables is based on the variable names. The file must contain the variables
''gaitcad\_struct'', ''merkmale\_projekt'', and ''mode''. The use of files from other projects might give
misleading results, especially in case of identical variable names.
\hypertarget{MI_Regression_Import}{}
\item {\bf Save regression model}: \newline  
exports a designed \textindex{regression model} in a file. The results are saved in the variables
''gaitcad\_struct'', ''merkmale\_projekt'', and ''mode''. They contain all steps defined by the parameters in
\hyperlink{CEO_DM_Regr}{\emph{Control elements:  Data mining: Regression}} and \hyperlink{CEO_DM_Klass}{\emph{Control elements:  Data mining: Special methods}}.
\hypertarget{MI_Regression_Export}{}
\end{itemize}
\subsection{Options}
\hypertarget{MI_Datei_Einstellungen}{}
contains all functions for loading and saving SciXMiner options.
\begin{itemize}
\item {\bf Load options}: \newline  
restores all options for the control elements as defined in the loaded \textindex{option file} with
extension \emph{*.uihdg}.
\hypertarget{MI_Einstellungen_laden}{}
\item {\bf Save options}: \newline  
saves all options for the control elements and the command lines of the plugins into an option file with extension \emph{*.uihdg}.
\hypertarget{MI_Einstellungen_speichern}{}
\item {\bf Save standard options}: \newline  
saves all options for the control elements and the command lines of the plugins into an option file with extension \emph{standard\_options.uihdg}. This file will be loaded at each SciXMiner start.
\hypertarget{MI_StandardEinstellungen_speichern}{}
\item {\bf Load frequency list}: \newline  
loads a MATLAB file for the definition of frequency regions and overtones.
\hypertarget{MI_Frequenzliste_Laden}{}
\end{itemize}
\subsection{Exit}
\hypertarget{MI_Beenden}{}
exits the application.

\section{Menu items 'Edit'}
contains all functions to edit a data set. It includes the \textindex{selection of data points}, time series and single
features, the extraction of new single features from time series or other single features, several conversions,
functions for deletion and operations for a data preprocessing.
\subsection{Select}
\hypertarget{MI_Auswaehlen}{}
contains functions to select data points, single features and time series.
\begin{itemize}
\item {\bf All data points}: \newline  
selects all data points and removes any chosen \textindex{selection of data points}.
\hypertarget{MI_Datenauswahl_Alle}{}
\item {\bf Data points using classes ...}: \newline  
performs a \textindex{selection of data points} using a conjunction and disjunction of linguistic
terms of output variables. It includes

1. a disjunction of all selected terms of one output variable and

2. a conjunction of the so selected data points for all output variables.

A selection of 'ALL' means that no data point is excluded due to its value for this output
variable. The result should be evaluated using \hyperlink{MI_Anzeige_Datentupel}{\emph{View - Classes for selected data points}}.
\hypertarget{MI_Datenauswahl_Klassen}{}
\item {\bf Data points using numbers ...}: \newline  
enables a manual \textindex{selection of data points} using its numbers.
\hypertarget{MI_Datenauswahl_Nr}{}
\item {\bf Data point from GUI}: \newline  
selects the data points defined by \hyperlink{CE_Select_DataPoints}{\emph{Control element: General options - Manual selection of data points}}. This function is especially useful if the data point numbers for a desired selection are known.
\hypertarget{MI_Datenauswahl_GUI}{}
\item {\bf Random data points (defined percentage of selected data points)}: \newline  
selects a pre-defined percentage of the selected data points by chance. The percentage value is defined by \hyperlink{CE_Select_PercDataPoints}{\emph{Control element: General options - Percent for selection}}.
\hypertarget{MI_Datenauswahl_PercDataPoints}{}
\item {\bf Data points via most frequent terms}: \newline  
selects all data points belonging to frequent terms for the selected output variable. The minimal number of data points is defined by \hyperlink{CE_Select_MinDtClass}{\emph{Control element: Data preprocessing - Minimal number of data points in one class}}.

Example: 100 data points for o.k., 50 data points for Error Type A, 2 data points for Error Type B; Parameter 10; Selection of 150 data points for o.k. and Error Type A
\hypertarget{MI_Datenauswahl_FreqCode}{}
\item {\bf Data points via values of single features}: \newline  
selects all selected data points with logical true values in \hyperlink{CE_DatapointValueSelection}{\emph{Control element: Data preprocessing - Feature values for selection}} for the first selected single features. Multiple queries have to be executed step by step.

Example for the selection of all values between 5 and 10 for a single features: 1. Select single feature, 2. Write in \hyperlink{CE_DatapointValueSelection}{\emph{Control element: Data preprocessing - Feature values for selection}} ''>5'', 3. Execute \hyperlink{MI_Datenauswahl_ValueSelection}{\emph{Edit - Select - Data points via values of single features}}, 4. Write in \hyperlink{CE_DatapointValueSelection}{\emph{Control element: Data preprocessing - Feature values for selection}} ''<10'', 5. Execute \hyperlink{MI_Datenauswahl_ValueSelection}{\emph{Edit - Select - Data points via values of single features}}.
\hypertarget{MI_Datenauswahl_ValueSelection}{}
\item {\bf Deselect missing values for selected time series}: \newline  
deselects all data points from the selected data points if at least one selected time series contains missing values. The data points will no be deleted.
\hypertarget{MI_Datenauswahl_NonInfZR}{}
\item {\bf Deselect missing values for selected single features}: \newline  
deselects all data points from the selected data points if at least one selected single features contains missing values. The data points will no be deleted.
\hypertarget{MI_Datenauswahl_NonInfEM}{}
\item {\bf All time series}: \newline  
selects all time series of a project. Pushing the button ''ALL'' left from control element
\hyperlink{CE_Auswahl_ZR}{\emph{Control element: Time series: General options - Selection of time series (TS)}} gives the same result.

A selection of some time series can be done by \hyperlink{CE_Auswahl_ZR}{\emph{Control element: Time series: General options - Selection of time series (TS)}} or the related edit field.
\hypertarget{MI_Zeitreihen_Alle}{}
\item {\bf All single features}: \newline  
selects all single features of a project. Pushing the button ''ALL'' left from control element
\hyperlink{CE_Auswahl_EM}{\emph{Control element: Single features - Selection of single feature(s) (SF)}} gives the same result.

A selection of some single features can be done by \hyperlink{CE_Auswahl_EM}{\emph{Control element: Single features - Selection of single feature(s) (SF)}} or the related edit field.
\hypertarget{MI_Einzelmerkmal_Alle}{}
\end{itemize}
\subsection{Extract}
\hypertarget{MI_Extrahieren}{}
contains all functions for the extraction of time series from time series and single features from
time series or other single features.
\begin{itemize}
\item {\bf Time series $\rightarrow$ Time series, Time series $\rightarrow$ Single features...}: \newline  
opens a configuration window to extract new time series or single features from existing time
series. Here, the existing time series, the segment and the type of the extracted feature will be
selected. The extraction is done by \textindex{plugins} (see Section~\ref{sec:plugins}).
\index{Time series $\rightarrow$ Time series}\index{Time series $\rightarrow$ Single feature}
\hypertarget{MI_Extraktion_ZRZR}{}
\item {\bf Time series $\rightarrow$ Time series, Time series $\rightarrow$ Single features (via plugin sequence)...}: \newline  
similar to \hyperlink{MI_Extraktion_ZRZR}{\emph{Edit - Extract - Time series $\rightarrow$ Time series, Time series $\rightarrow$ Single features...}}, but uses a pre-defined sequence of plugins for extraction. In
the configuration windows, only the existing time series and the segments will be selected. The
feature extraction uses the sequence of plugins defined by \hyperlink{CE_Auswahl_Plugins}{\emph{Control element: Plugin sequence - Selection of plugins}}. The
scheduling of the plugins might be changed in the edit field from \hyperlink{CE_Auswahl_Plugins}{\emph{Control element: Plugin sequence - Selection of plugins}}. All
temporary results will be deleted. \index{Time series $\rightarrow$ Time series}\index{Time series $\rightarrow$ Single
feature}
\hypertarget{MI_Extraktion_ZRZR_Kombi}{}
\item {\bf Single features $\rightarrow$ Single features (with the selected feature aggregation from Options-Data Mining: Classification of single features)}: \newline  
extracts new single features from existing single features. It uses the parameters for feature
selection and aggregation from \hyperlink{MI_Anzeige_Klassifikation}{\emph{View - Classification}}.
 \index{Single feature $\rightarrow$ Single feature}\index{feature aggregation}
\hypertarget{MI_Extraktion_EMEMA}{}
\item {\bf Statistics for single features for linguistic terms of the selected output variable}: \newline  
extracts new single features for each selected single feature and each data point. These new features are minimum, maximum, median, mean value and number of all selected data points with the same linguistic term of the selected output variables.
\hypertarget{MI_ExtrStatSFTerm}{}
\item {\bf Term statistics for output variable...}: \newline  
extracts a new output variable for each output variable that is selected in the following configuration window. Each data point gets the most frequent term for all selected data points with the same linguistic term of \hyperlink{CE_Auswahl_Ausgangsgroesse}{\emph{Control element: Time series: General options - Selection of output variable}}. This function is especially useful to summarize data points that are replicas of the same measurement.
\hypertarget{MI_TermStat_Klasse}{}
\item {\bf Save time series segment for feature extraction}: \newline  
defines the selected segment of the time series as new segment for the feature extraction. The
selection is explained in \hyperlink{CE_Zeitreihen_Segment_Start}{\emph{Control element: Time series: General options - Time series segment from}} and \hyperlink{CE_Zeitreihen_Segment_Ende}{\emph{Control element: Time series: General options - to}}.
\hypertarget{MI_Extraktion_Segment2Einzug}{}
\end{itemize}
\subsection{Convert}
\hypertarget{MI_Transformieren}{}
contains all functions to convert time series, single features, classes, and data points to each
other.
\begin{itemize}
\item {\bf Selected single features $\rightarrow$ Output variables}: \newline  
converts the selected single features into output variables. The same values are used as linguistic
values if the features have discrete values with not more values than a defined maximal number
(defined by \hyperlink{CE_EM_Ausgangs}{\emph{Control element: Single features - Number of terms for Single feature$\rightarrow$Output variable}}) or if the option \hyperlink{CE_EM_Ausgangs_Alle}{\emph{Control element: Single features - All values}} was chosen.
Otherwise, a discretization is done by automatically designed fuzzy membership functions and a
subsequent generation of crisp membership functions. Its number of values is chosen by
\hyperlink{CE_EM_Ausgangs}{\emph{Control element: Single features - Number of terms for Single feature$\rightarrow$Output variable}}. The term name depends on the maximum of the fuzzy membership function (e.g.
''appr. 7.5''). \index{Single feature $\rightarrow$ Output variable}
\hypertarget{MI_EM_Klasse}{}
\item {\bf Selected single features $\rightarrow$ Time series}: \newline  
This function assumes a temporal order of all existing single features. It converts $s$ selected single features with
$N$ data points into $1$ time series with $K=s$ sample points. If the project yet contains time series, the number of
sample points must be equal to $s$. Alternatively, all existing time series might be deleted. \index{Single feature $\rightarrow$
Time series}
\hypertarget{MI_Trans_EM2ZR}{}
\item {\bf Selected single feature (Timestamp) $\rightarrow$ Single features (Year, month, day, hour, minute, second)}: \newline  
assumes that the selected single feature is a MATLAB timestamp. This function generates new single features with numbers for the year, month, day, hour, minute and second.
\hypertarget{MI_Trans_Timestamp2YearEtc}{}
\item {\bf Selected single feature (Timestamp) $\rightarrow$ Weekday}: \newline  
assumes that the selected single feature is a MATLAB time stamp. The result is converted in a new output variable containing the weekday.
\hypertarget{MI_Trans_Weekday}{}
\item {\bf All single features: Data point  $\rightarrow$ Time series}: \newline  
The function assumes a temporal order of all data points. The data points of all single features
are converted to sample points of time series. Existing time series will be deleted. These functions
make a visualization of data points as time series possible.

Example:

A project consists of 100 data points and 2 single features. The function generates a project with
2 time series and 100 sample points. \index{Data point $\rightarrow$ Time series}
\hypertarget{MI_Trans_DSEM2ZR}{}
\item {\bf All single features: Data point $\rightarrow$ Time series (class-wise)}: \newline  
The function assumes a temporal order of all data points with identical values for the selected
output variable. The data points of all single features are converted to sample points of time
series. Existing time series will be deleted. All output variables with identical values for all
new time series remain output variables. Output variables with different values are converted to
new time series. Incomplete sample points are set to zero. The number of valid sample points is
saved in  a new single feature. These functions make a visualization of data points as time series
possible.

Example:

A project consists of 102 data points and 2 single features. Blocks of 10 data points belong to a
sensor (sensor number in first output variable). The second output variable codes the number of a
measurement (1..10 for sensor 1..9, 1..12 for sensor 10).

The first output variable is selected. The function generates 10 new data points with 3 time series
(from both single features and the second output variable) and 12 sample points. The 11th and 12th
sample point is zero for the first ten data points. \index{Data point $\rightarrow$ Time series}
\hypertarget{MI_Trans_DSDTEM2ZR}{}
\item {\bf All single features: Data point $\rightarrow$ Time series (with sample point in output variable)}: \newline  
converts the data points into time series, each single feature will generate a separate time series. The selected output variable must contain the sample point number in its term numbers.
\hypertarget{MI_Trans_DSDTEM2ZRSAMPLE}{}
\item {\bf Data points $\rightarrow$ Single features and Single features $\rightarrow$ Data points}: \newline  
exchanges data points vs. single features in a SciXMiner project or vice versa (exchange $s$ vs. $N$, see Appendix Important Internal Data Structures). Existing time series will be deleted.
\hypertarget{MI_DT2Feat}{}
\item {\bf Selected time series: Sample points $\rightarrow$ Single features}: \newline  
generates a new single feature from each sample of a time series. The new features are named with
the time series name and  ''SP k'' for the $k$-th sample.

Example:

A project consists of one data point with two time series and 100 samples, the second and third
time series are selected. With \hyperlink{MI_Trans_ZR2EM}{\emph{Edit - Convert - Selected time series: Sample points $\rightarrow$ Single features}}, 200 new single features are generated by the
100 samples of the time series 2 and 3.

\index{Sample point $\rightarrow$ Single feature}
\hypertarget{MI_Trans_ZR2EM}{}
\item {\bf All time series: Sample points $\rightarrow$ Data points}: \newline  
converts the sample points of all time series into data points of single features. The number of
the sample point is saved as new feature. All single features of the old project will be deleted.
The results are saved as new project. This enables a \textindex{classification of time series} with
a separate class assignment of each sample point.

This function is an inverse function for \hyperlink{MI_Trans_DSEM2ZR}{\emph{Edit - Convert - All single features: Data point  $\rightarrow$ Time series}}.

Example:

A project consists of 1 data point with 2 time series and 100 sample points. The menu items
\hyperlink{MI_Trans_ZR2EM_SampleTupel}{\emph{Edit - Convert - All time series: Sample points $\rightarrow$ Data points}} generates a new project without time series, but with 3 new
single features (both time series and the number of the sample point) and 100 data points.
\index{Sample point $\rightarrow$ Data point}
\hypertarget{MI_Trans_ZR2EM_SampleTupel}{}
\item {\bf Time series  $\rightarrow$ Data points}: \newline  
converts time series into data points in a SciXMiner project (converts $s_z$ into $N$, see Appendix Important Internal Data Structures). This function is available if only one data point exist.
\hypertarget{MI_ZR2DT}{}
\item {\bf Separate time series with trigger events...}: \newline  
A time series will be segmented using a \textindex{trigger time series} and the resulting segments are written as new
time series into data points from the variable d\_orgs. In this way, events can be easily processed. A detail
description can be found in Section~\ref{sec:building}.

The trigger time series is expected to be the same for all data points.
\hypertarget{MI_Extraktion_Teile_ZR}{}
\item {\bf Shorten time series}: \newline  
contains all functions to \textindex{shorten time series}, e.g. for an easier handling of large
projects.
\hypertarget{MI_Kuerze_ZR}{}
\item {\bf Selected time series $\rightarrow$ Project-specific time scale}: \newline  
converts the selected time series into a \textindex{project-specific time scale}. Here, only one time series has to be selected. The conversion is based on the values for the first selected data point.
\hypertarget{MI_ZR2PTS}{}
\item {\bf Selected output variable $\rightarrow$ Single feature}: \newline  
copies the selected output variable into an additional single feature. \index{Output variable $\rightarrow$
Single feature}
\hypertarget{MI_Trans_Klasse2EM}{}
\item {\bf Selected output variable $\rightarrow$ Single features (requires digits in term names)}: \newline  
converts the selected output variable into a single feature if the names of the linguistic terms contain numbers.
\hypertarget{MI_Trans_Klasse2EMZ}{}
\item {\bf Selected output variable (with filenames) $\rightarrow$ Multiple output variables (path components, file)}: \newline  
assumes filenames containing the complete path as terms of the selected output variable. New output variables will be generated. Their terms result from splitting these filenames into separate subdirectories and the filenames. The file extension (e.g. .dat) is ignored.
\hypertarget{MI_Trans_Class2PathDirclasses}{}
\item {\bf Selected output variable (Date) $\rightarrow$ Single feature (Timestamp)}: \newline  
interprets the terms of the selected output variable as data, the date format is defined by \hyperlink{CE_DateFormat}{\emph{Control element: Data preprocessing - Date format}}.
A single feature is generated based of a conversion of date and time into a MATLAB \textindex{timestamp format}.
\hypertarget{MI_Trans_Date2Timestamp}{}
\item {\bf Selected output variable (with filenames) $\rightarrow$ Multiple output variables (path components, file, extension)}: \newline  
like \hyperlink{MI_Trans_Class2PathDirclasses}{\emph{Edit - Convert - Selected output variable (with filenames) $\rightarrow$ Multiple output variables (path components, file)}}, but with extension as separate output variable.
\hypertarget{MI_Trans_Class2PathDirExclasses}{}
\item {\bf Estimated output variable $\rightarrow$ Output variable}: \newline  
converts the estimated output variable generated by a classifier (\hyperlink{MI_EMKlassi_An}{\emph{Data mining - Classification - Apply}}) to a new output variable.
\hypertarget{MI_Trans_POS2Y}{}
\item {\bf Estimated output variable $\rightarrow$ Single feature}: \newline  
converts the result of the last classification into a new discrete-valued single feature. This
feature is called ''Estimation'' + the name of the selected output variable. This menu item can be
selected if a valid classification result exists resulting from an applied classifier.
\index{Output variable $\rightarrow$ Single feature}
\hypertarget{MI_Trans_POS2EM}{}
\item {\bf Estimated output variable (percental) $\rightarrow$ Single features}: \newline  
converts the estimated percental membership values (resulting from the last application of a classifier) to classes (linguistic terms) to single features.
\hypertarget{MI_Trans_PRZ2EM}{}
\item {\bf Errors of regression model $\rightarrow$ Single features}: \newline  
exports the error of the recent regression model into a single feature.
\hypertarget{MI_Trans_RegrErSF}{}
\item {\bf Add output with identical term for all data points}: \newline  
adds an additional output variable with a term value of 1 for all data points (useful for visualizations in a project with many output terms).
\hypertarget{MI_AddClassAll}{}
\item {\bf Class $\rightarrow$ Class: Combine terms (OR-operation with data point selection, name of the current output variable)}: \newline  
generates a new output variable with two linguistic terms. It results from the selected output
variables and the selected data points. The first linguistic term consists of all selected data
points, the second from all other data points. This is a comfortable way to combine multiple terms.
Later, they can be labeled by the renaming function.

Example:

 1. Select output variable ''Person'' with linguistic terms Anna, Maria, Thomas and Peter

 2. Select terms Anna and Maria in \hyperlink{MI_Datenauswahl_Klassen}{\emph{Edit - Select - Data points using classes ...}}

 3. Select menu item \hyperlink{MI_Kl_Kl_Oder}{\emph{Edit - Convert - Class $\rightarrow$ Class: Combine terms (OR-operation with data point selection, name of the current output variable)}}

 4. A new output variable ''Person (OR)'' with terms ''Anna, Maria'' and
 ''Rest: Thomas, Peter'' will be generated

 5. It can be renamed with \hyperlink{MI_Umbenennen}{\emph{Edit - Rename...}} in ''Sex'' with the new terms ''Female''
 and ''Male''.
 \index{Class $\rightarrow$ Class}
\hypertarget{MI_Kl_Kl_Oder}{}
\item {\bf Class $\rightarrow$ Class: conjunction (AND) of output variables ...}: \newline  
generates a new output variable. The linguistic terms results from a conjunction of a set of output
variables which are selected in the following configuration window.

Example:

Conjunction of ''Name'' (Terms: ''Anna'', ''Maria'', ''Thomas'', ''Peter'') and ''Examination''
(''Pre-therapeutic'', ''Post-therapeutic'') results in a new output variable ''AND: Name
Examination'' with 8 terms (''Anna Pre-therapeutic'', ..., ''Peter Post-therapeutic'').
\index{Class $\rightarrow$ Class}
\hypertarget{MI_Kl_Kl_Und}{}
\item {\bf Resort linguistic terms (order from GUI)}: \newline  
sorts the linguistic terms of the selected output variable using the new order from \hyperlink{CE_Anzeige_OrderofTerms}{\emph{Control element: General options - Order of linguistic terms}}.
\hypertarget{MI_Trans_Term_Order}{}
\item {\bf Clean up all variable names (start and end blanks)}: \newline  
deletes end blanks in the names of output variables, time series and single features.
\hypertarget{MI_Trans_CleanUpNames}{}
\end{itemize}
\subsection{Delete}
\hypertarget{MI_ImVid_Delete}{}
deletes data points, time series, single features, or output variables.
\begin{itemize}
\item {\bf Selected data points}: \newline  
deletes all selected data points. Finally, the data
points will be renumbered. \index{data point selection}
\hypertarget{MI_Loeschen_Datentupel}{}
\item {\bf Unselected data points}: \newline  
deletes all non-selected data points. Finally, the data
points will be renumbered. \index{data point selection}
\hypertarget{MI_Loeschen_Datentupel_NonSel}{}
\item {\bf Double single features, time series, and output variables}: \newline  
deletes doubles of single features, time series, and output variables with identical names. Such
doubles result normally from a multiple feature extraction.
\hypertarget{MI_Loeschen_doppelt}{}
\item {\bf Selected time series}: \newline  
deletes all selected time series.
\hypertarget{MI_Loesche_ZR}{}
\item {\bf Unselected time series}: \newline  
deletes all non-selected time series.
\hypertarget{MI_Loeschen_iZR}{}
\item {\bf Selected sample points}: \newline  
deletes all selected sample points out of all time series.
\hypertarget{MI_Loeschen_Kselect}{}
\item {\bf Unselected sample points}: \newline  
deletes all unselected sample points.
\hypertarget{MI_Loeschen_Knotselect}{}
\item {\bf All single features}: \newline  
deletes all single features.
\hypertarget{MI_Loeschen_alleEM}{}
\item {\bf Selected single features}: \newline  
deletes selected single features.
\hypertarget{MI_Loesche_EM}{}
\item {\bf Unselected single features}: \newline  
deletes unselected single features.
\hypertarget{MI_Loesche_iEM}{}
\item {\bf Output variable ...}: \newline  
deletes output variables. An additional configuration window is shown to select these output variables (multiple
selection by SHIFT + left mouse button or CTRL + left mouse button).
\hypertarget{MI_Loeschen_Klasse}{}
\item {\bf Non-existing terms of the output variable}: \newline  
deletes these terms of the output variables without any data point in the SciXMiner project. This makes some functions (e.g., the selection of data points based on terms) easier to use.
\hypertarget{MI_Loeschen_NonExTerms}{}
\item {\bf Missing data}: \newline  
removes time series or single features with NaN- or Inf values and time series containing only
zeros for all sample points of a data point. All cases are interpreted as missing values for these
data points and time series resp. single features.

Consequently, either the data point or the time series resp. the single feature will be deleted.
The preference is controlled by the threshold in \hyperlink{CE_Nullzr_Schwellwert}{\emph{Control element: Data preprocessing - Threshold for deleting [Perc. missing data]}}.

 Example: Threshold 20 \%

 1. All time series with \textindex{missing values} for more than 20 \% of the data points will be deleted.

 2. All single features with \textindex{missing values} for more than 20 \% of the data points will be deleted.

 3. All remaining data points with missing values will be deleted.

Tip: Misleading standard values for missing values of a single feature (e.g. -1) can be set to NaN
values using a macro to enable a unified handling.
\hypertarget{MI_Nullzr_loeschen}{}
\end{itemize}
\subsection{Sorting}
\hypertarget{MI_Sortieren}{}
alphabetic sorting of names in a project (i.e., very useful for fusing of projects with identical variable names in different order).
\begin{itemize}
\item {\bf Single features}: \newline  
sorts single features in a project alphabetically.
\hypertarget{MI_Sortieren_SingleFeatures}{}
\item {\bf Time series (TS)}: \newline  
sorts time series in a project alphabetically.
\hypertarget{MI_Sortieren_TimeSeries}{}
\item {\bf Output variables}: \newline  
sorts output variables in a project alphabetically.
\hypertarget{MI_Sortieren_OutputVariables}{}
\item {\bf Images}: \newline  
sorts images in a project alphabetically.
\hypertarget{MI_Sortieren_Images}{}
\end{itemize}
\subsection{Outlier detection}
\hypertarget{MI_Ausreisser}{}
generates a \textindex{classifier} to detect outliers for the selected single features and data
points. It can be applied to unknown data points. Some hints to the algorithms are given in
\cite{Burmeister06Dort}.
\begin{itemize}
\item {\bf Design data set (selected data points, options from classification)}: \newline  
The outlier detection bases on an one class classification problem. It decides if a data point belongs to this one
class or not.

However, the implemented methods for the feature selection and classification assume data points of at least two
classes. To solve this problem, an appropriate data set must be generated. A first step to prepare this data set is the
feature selection in \hyperlink{CE_Klassifikation_Merkmalsauswahl}{\emph{Control element: Data mining: Classification of single features - Selection of single features}}. Normally, the outlier detection uses the ''selected
features''.
\hypertarget{MI_Ausreisser_DatenEn}{}
\item {\bf Design (selected data points, designed data set)}: \newline  
designs a \textindex{classifier} for the outlier detection. This step uses the data set prepared by
\hyperlink{MI_Ausreisser_DatenEn}{\emph{Edit - Outlier detection - Design data set (selected data points, options from classification)}}. Three different methods might be used: an one-class method, a
distance-based method, and a density-based method. The method is chosen by \hyperlink{CEO_Datenvor}{\emph{Control elements:  Data preprocessing}}.

The one-class method uses a SVM optimization proposed by \cite{Campbell00}. The distance-based
method computes mean value and standard deviation for the selected data points and uses the
Mahalanobis distance to the mean value as criterion. The density-based method evaluates the number
of data points in a pre-defined neighborhood.
\hypertarget{MI_Ausreisser_En}{}
\item {\bf Apply (selected data points, designed data set)}: \newline  
applies the classifier design by \hyperlink{MI_Ausreisser_En}{\emph{Edit - Outlier detection - Design (selected data points, designed data set)}} for the \textindex{outlier detection}. The
parameter in \hyperlink{CE_Ausreisser_Schwelle}{\emph{Control element: Data preprocessing - Threshold for outliers}} tunes the threshold for this decision. The value of
\hyperlink{CE_Ausreisser_InKlasse}{\emph{Control element: Data preprocessing - Save result as output variable}} decides if the result is added as new output variable.

By the menu point \hyperlink{MI_Ausreisser_AlleKlassen}{\emph{Edit - Outlier detection - Compute class discriminant functions (selected data points, designed data set)}} the outlier detection is applied for all classes. It generates a
struct called ''ausreisser'' containing indices of outlier data points including the values for the decision (e.g.
distances). The indices address the data point numbers of all existing data points of the project.

The function does not select data points, e.g. to delete outliers. This functionality is done by
adding the outlier decision as new output variable (see  \hyperlink{CE_Ausreisser_InKlasse}{\emph{Control element: Data preprocessing - Save result as output variable}}) and the use
of data point selection in \hyperlink{MI_Datenauswahl_Klassen}{\emph{Edit - Select - Data points using classes ...}}.
\hypertarget{MI_Ausreisser_An}{}
\item {\bf Design and apply (selected data points, designed data set)}: \newline  
designs and applies the outlier detection.
\hypertarget{MI_Ausreisser_EnAn}{}
\item {\bf Compute class discriminant functions (selected data points, designed data set)}: \newline  
analyzes for all selected data points the linguistic terms of the output variable and computes for
each class limits for the \textindex{outlier detection}. The resulting parameters are saved in the
variable ''klassen\_grenzen'' for each class. The element ''lp\_detect'' describes the parameters
and the element ''ausreisser'' contains the application results to the selected data points.

If  ''New output variable'' or ''Replace identical output variable'' is chosen in
\hyperlink{CE_Ausreisser_InKlasse}{\emph{Control element: Data preprocessing - Save result as output variable}}, the result is saved as output variable. It can be used to select
outlier e.g. for a later visualization and deletion.
\hypertarget{MI_Ausreisser_AlleKlassen}{}
\end{itemize}
\subsection{Category}
\hypertarget{MI_Kategorie}{}
Categories summarize similar single features and time series for different analysis and
visualization functions (see Section~\ref{sec:ontologie}, \cite{Loose04}). They classify single
features and time series into different output classes of multiple output variables called
categories (e.g. according to the sensor type or to the kind of feature extraction). The matching
is done by category files with the extension *.categories.
\begin{itemize}
\item {\bf Select single features ...}: \newline  
enables a selection of single features by a conjunction or disjunction of multiple categories and
their linguistic terms. The function assumes computed categories using \hyperlink{MI_Kategorie_EMBer}{\emph{Edit - Category - Compute single features}}.
\hypertarget{MI_Kategorie_EMAusw}{}
\item {\bf Compute single features}: \newline  
assigns categories for all existing single features using their names. For safety reasons, the categories will be
deleted if the number of single features is changed (by adding or deleting single features).
\hypertarget{MI_Kategorie_EMBer}{}
\item {\bf Compute a priori relevances}: \newline  
computes \textindex{a priori relevances} of single features using the existing categories.
\hypertarget{MI_AprioriBer}{}
\item {\bf A priori relevances of selected features...}: \newline  
set the \textindex{a priori relevances} for all selected single features to a defined value.
\hypertarget{MI_AprioriEdit}{}
\item {\bf Selection of time series ...}: \newline  
enables a selection of time series by a conjunction or disjunction of multiple categories and their
linguistic terms. The function assumes computed categories using \hyperlink{MI_Kategorie_ZRBer}{\emph{Edit - Category - Compute time series}}.
\hypertarget{MI_Kategorie_ZRAusw}{}
\item {\bf Compute time series}: \newline  
assigns categories for all existing time series using their names. For safety reasons, the
categories will be deleted if the number of time series will be changed (by adding or deleting time
series).
\hypertarget{MI_Kategorie_ZRBer}{}
\item {\bf Search time series related to the selected single features}: \newline  
selects all time series belonging to the selected features. The criterion is an extraction of the
selected single features from the related time series.
\hypertarget{MI_Suche_ZR_zu_EM}{}
\end{itemize}
\subsection{Rename...}
\hypertarget{MI_Umbenennen}{}
This function opens a window to \textmainindex{rename} time series, single features, output
variables, and the corresponding linguistic terms (possible discrete values).

The upper element is used to select the type of variable to be renamed. In the element below, one
time series, single feature, or output variable will be selected represented by the (old) name. For
linguistic terms, an additional element is shown to the selected the term.

The new name must be written in the lowest element and saved by pressing ENTER.

\begin{figure}[h] 
 \includegraphics[width=0.9\columnwidth]{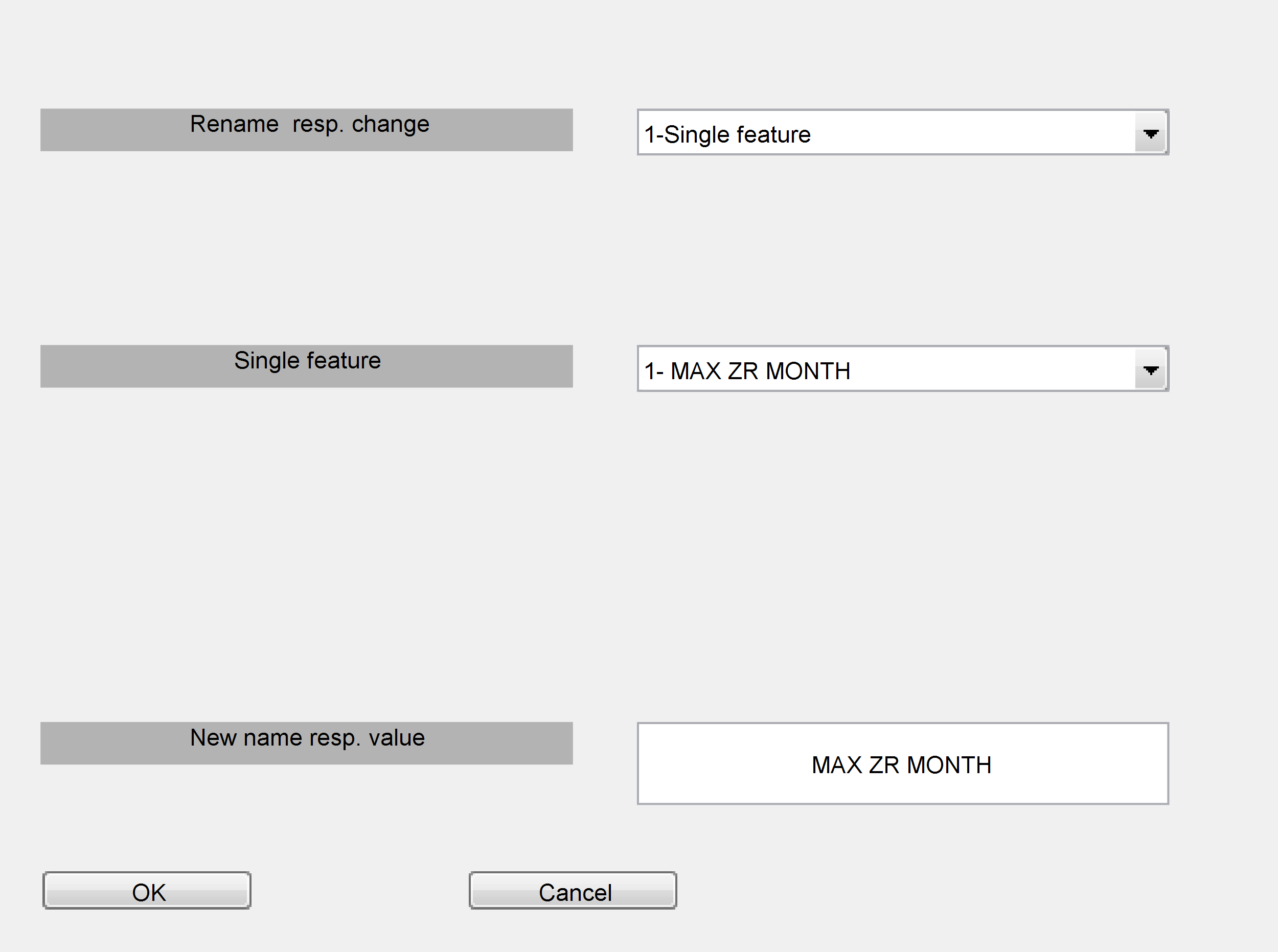}
 \caption{\label{fig:auswahlfensterMI_Umbenennen} Figure for \emph{''Edit - Rename...''}}
 \end{figure}

The corresponding figure shows Fig.~\ref{fig:auswahlfensterMI_Umbenennen}.
\subsection{Generating trigger time series}
\hypertarget{MI_Erstelle_Trigger}{}
shows a separate figure (\textindex{TS navigator}) to generate a \textmainindex{trigger time
series} (see Section~\ref{sec:edit_trigger} and Fig.~\ref{fig:trigger_grafik}).

The upper part of the figure shows time series and existing trigger events (red, dotted thin
lines). The selection of the time series is done in the element at the right bottom corner of the
figure. The class number for a trigger event is at the trigger line. The standard value is one. The
time series window can be zoomed by two control elements in the left bottom corner (left element:
length of time series segment, right: each n-th sample point will be shown). The grid is switched
on or off by the corresponding element. The slide bar below the time series enables a time shift.

A click to the time series part of the window selects a new sample point (black dashed line). A new
\textindex{trigger event} is added for this sample point by clicking the ''Add'' button. The list
of recent trigger events is shown with the syntax sample point (class), e.g. sample point k=23 and
class type 1: 23(1). The selected trigger event is marked. Its class label might be changed by
 ''+'' und ''-''. The use of ''$\rightarrow$'' resp. ''$\leftarrow$'' shifts the temporal position of the event.
It can be removed by the ''Remove'' button. ''Go to'' moves to the selected event.

''Import'' loads a variable with a pre-defined trigger time series from Matlab workspace. The
length must be equal to the length of time series of the project.

The button ''Save trigger time series'' saves the time series in the project and close the figure.
\subsection{Edit trigger time series}
\hypertarget{MI_Bearbeite_Trigger}{}
edits an existing trigger time series selected by \hyperlink{CE_Auswahl_ZR}{\emph{Control element: Time series: General options - Selection of time series (TS)}}. A wrong selection leads to
an error message or a time series with many trigger events. The handling is explained for
\hyperlink{MI_Erstelle_Trigger}{\emph{Edit - Generating trigger time series}}.
\subsection{Looking for missing data}
\hypertarget{MI_Anzeige_FehlDaten}{}
searches for missing values and shows the results for time series vs. data points and single
features vs. data points. See \hyperlink{MI_Nullzr_loeschen}{\emph{Edit - Delete - Missing data}} for the definition of missing values.

\section{Menu items 'View'}
contains all functions for visualization including many results of the applied algorithms.
\subsection{Classes for selected data points}
\hypertarget{MI_Anzeige_Datentupel}{}
shows the selected data points with their output variables as text file. \index{selection of data
points}
\subsection{Number of terms for selected data points}
\hypertarget{MI_Anzeige_Terms}{}
shows the number of terms for each output variable and all selected data points. Only frequent terms are listed in detail, terms with one or two data points are only summarized. This function is suited for large projects, because in contrast to \hyperlink{MI_Anzeige_Datentupel}{\emph{View - Classes for selected data points}} not all infrequent terms and data point descriptions are listed. The result is saved in a file called [project\_name]\_ind\_terms.txt.
\subsection{Time series (TS)}
\hypertarget{MI_Sortieren_TimeSeries}{}
contains all functions for the visualization of time series.

In addition, the figures can be used for the definition of segments from time series and of trigger
events. The definition of \textindex{segments} is done by zooming to the region of interest and
pushing the ''Select time series segment'' item in the menu bar of the figure. The segment is now
defined by the minimal and maximal value of the x-axis. A more detailed explanation is given in
\hyperlink{CE_Zeitreihen_Segment_Start}{\emph{Control element: Time series: General options - Time series segment from}} and \hyperlink{CE_Zeitreihen_Segment_Ende}{\emph{Control element: Time series: General options - to}}. \index{trigger event}
\begin{itemize}
\item {\bf Original time series}: \newline  
shows all selected data points of the selected time series (\hyperlink{CE_Auswahl_ZR}{\emph{Control element: Time series: General options - Selection of time series (TS)}}) vs. time.
\hypertarget{MI_Anzeige_ZR_Orig}{}
\item {\bf Mean time series}: \newline  
shows the mean values of all selected time series (\hyperlink{CE_Auswahl_ZR}{\emph{Control element: Time series: General options - Selection of time series (TS)}}) versus time for all
linguistic terms (classes) of the selected output variable.
\hypertarget{MI_Anzeige_ZR_MW}{}
\item {\bf Standard deviation time series}: \newline  
shows the standard deviations  of all selected time series (\hyperlink{CE_Auswahl_ZR}{\emph{Control element: Time series: General options - Selection of time series (TS)}}) versus time for
all linguistic terms (classes) of the selected output variable.
\hypertarget{MI_Anzeige_ZR_STD}{}
\item {\bf Mean and standard deviation time series}: \newline  
shows the mean values and standard deviations of all selected time series (\hyperlink{CE_Auswahl_ZR}{\emph{Control element: Time series: General options - Selection of time series (TS)}})
versus time for all linguistic terms (classes) of the selected output variable.
\hypertarget{MI_Anzeige_ZR_MWSTD}{}
\item {\bf Scatterplot time series}: \newline  
shows all sample points of the selected time series as \textindex{scatter plot}. The number of
selected time series must be even. The corresponding linguistic terms of the selected data points
are coded by color.

Example:

Four time series $x_1[k],...,  x_4[k]$ with $K=100$ sample points and $N=2$ data points are
selected. The results are two figures with $x_2 = f (x_1)$ resp. $x_4 = f (x_3)$ with 200 points
each.
\hypertarget{MI_Anzeige_ZRScatterplot}{}
\item {\bf Poincare plot time series (2D)}: \newline  
show the selected time series as 2D-Poincare plot (x-axis: x[k], y-axis x[k+1]). Subsequent time points can be plotted as connected or unconnected points, depending on the selection in \hyperlink{CE_Anzeige_PoincareLine}{\emph{Control element: View: Time series - Poincare plot: Connect points}}.
\hypertarget{MI_Anzeige_ZRPoincare2D}{}
\item {\bf Poincare plot time series (3D)}: \newline  
show the selected time series as 3D-Poincare plot (x-axis: x[k], y-axis x[k+1], z-axis x[k+2]). Subsequent time points can be plotted as connected or unconnected points, depending on the selection in \hyperlink{CE_Anzeige_PoincareLine}{\emph{Control element: View: Time series - Poincare plot: Connect points}}.
\hypertarget{MI_Anzeige_ZRPoincare3D}{}
\item {\bf Transformations vectors for feature extraction}: \newline  
defines if data-dependent transformations for the feature extractions are computed from the recent data, saved and/or loaded from file. The main difference is the handling of e.g. membership functions or a principal  component analysis. The saving in a file is necessary for a design with a training data set and a later usage for test data by applying the same parameters. The data is saved in a file called [project name].plugpar.
\hypertarget{MI_Anzeige_ExtractedFeaturesPlugpar}{}
\item {\bf Bode plot for filter in plugin}: \newline  
shows the Bode plot for the selected filter in the plugin called plugin\_filter.m. The related parameters are defined in \hyperlink{CE_Auswahl_PluginsCommandLine}{\emph{Control element: Plugin sequence - Plugin parameter}}. The plot is influenced by \hyperlink{CE_Zeitreihen_Abtastfreq}{\emph{Control element: Time series: General options - Sampling frequency of time series}} and \hyperlink{CE_Zeitreihen_Einheit_Abtastfreq}{\emph{Control element: Time series: General options - Unit}}.
\hypertarget{MI_Anzeige_BodeFilterPlugin}{}
\item {\bf Show all feature relevances (sorted table)}: \newline  
shows relevances of time series sorted by descending values.
\hypertarget{MI_Anzeige_ZRRelevanz}{}
\item {\bf Show all feature relevances (unsorted table)}: \newline  
shows relevances of time series sorted by time series numbers.
\hypertarget{MI_Anzeige_ZRRelevanz_un}{}
\end{itemize}
\subsection{Single features}
\hypertarget{MI_Sortieren_SingleFeatures}{}
contains all functions for the visualization of single features and their derived features.
\begin{itemize}
\item {\bf Single features vs. output classes}: \newline  
shows term-wise histograms for the first selected single feature on the x-axis versus the number of
the corresponding linguistic term of the selected output variable on y-axis.
\hypertarget{MI_Anzeige_EM_Klasse}{}
\item {\bf Single features vs. single features}: \newline  
shows a scatter plot of the selected single features. The functions shows pairwise scatter plots
($x_2=f(x_1)$, $x_4=f(x_3)$ etc.) if more than three features were selected.
\hypertarget{MI_Anzeige_EM}{}
\item {\bf Manual class assignment via single features}: \newline  
enables an \textindex{interactive assignment of linguistic terms of output variables} or the selection of data points with a figure containing two selected output variables. Here, a figure will be opened with different options in the menu bar.

''Mark'' contains all operations for the handling of a \textindex{list with interactively selected data points}. This list is initialized as empty list. Different menu items exist. ''Add'' adds all visible data points to the list depending on the recent axes scaling. With this option, data points can be selected by zooming and subsequent adding. ''Remove'' is used to remove all visible data points from the list. All marked data points are shown with a circle mark. ''Reset'' removes all data points and generates an empty list. Finally, the list can be transferred to an output variable (''Save as output variable'') or used to select data points for further analysis (''Select data points'').

Existing corresponding images for the visible data can be displayed using ''Show related images''.

Alternatively, ''Change: Term of the output variable ...'' can assign new linguistic terms to all visible data points for the selected output variable. Here, all existing terms or a new term (''New term'') are available.

The figure can be opened different times with different single features. ''Refresh'' updates all open figures with the new linguistic terms or the entries of the list.
\hypertarget{MI_Anzeige_SpecialSelection}{}
\item {\bf Discrete single features (2D histogram)}: \newline  
shows a two-dimensional histogram for the first two selected single features. Error messages are
generated if the single features have continuous values or if only one or more than two single
feature have been selected.
\hypertarget{MI_Anzeige_EM_Hist}{}
\item {\bf Discrete single features (2D histogram with output classes)}: \newline  
shows a two-dimensional histogram for two selected single features. In contrast to
\hyperlink{MI_Anzeige_EM_Hist}{\emph{View - Single features - Discrete single features (2D histogram)}}, separate bars are shown for different linguistic terms of the selected
output variable. Error messages are generated if the single features have continuous values or if
only one or more than two single feature have been selected.
\hypertarget{MI_Wertediskrete_Merkmale}{}
\item {\bf Mean and standard deviation}: \newline  
plots mean values and standard deviations for all selected single features in a file. The
computation is done for each class of the selected output variable separately.
\hypertarget{MI_Anzeige_MW_Streu}{}
\item {\bf Mean, standard deviation, minimum, maximum}: \newline  
like \hyperlink{MI_Anzeige_MW_Streu}{\emph{View - Single features - Mean and standard deviation}}, but with additional minimum and maximum values.
\hypertarget{MI_Anzeige_EM_Min_Max}{}
\item {\bf Median, Minimum, Maximum}: \newline  
like \hyperlink{MI_Anzeige_MW_Streu}{\emph{View - Single features - Mean and standard deviation}}, but with median, minimum and maximum values.
\hypertarget{MI_Anzeige_EM_Med_Min_Max}{}
\item {\bf Absolute values}: \newline  
writes the values of all selected single features and data points in a file \emph{tmp.txt} and
opens the file in an editor window. It contains also the corresponding value of the output
variable. The features and output variables are tabulator separated and can easily be exported to
other programs. \index{export}
\hypertarget{MI_Anzeige_EM_Abs}{}
\item {\bf Boxplot (with Statistics Toolbox!)}: \newline  
plots \textindex{boxplots} for all selected data points, single features and the selected output variable using the function boxplot.m of the Statistic Toolbox.
\hypertarget{MI_Anzeige_Gaitboxplot}{}
\item {\bf Membership function}: \newline  
plots a figure for each selected single feature containing the membership functions of the recent
\textindex{fuzzy system}. Such a fuzzy system is designed by \hyperlink{MI_Fuzzy_Einzelregel}{\emph{Data mining - Fuzzy systems - Design (single rules)}},
\hyperlink{MI_Fuzzy_Basis}{\emph{Data mining - Fuzzy systems - Design (rule base)}} or \hyperlink{MI_KLASS2FUZZY}{\emph{Data mining - Fuzzy systems - Import fuzzy system from classifier}}. New membership functions will be generated for this
figure if a fuzzy system does not exist or if the single feature is not a part of the fuzzy system.
\hypertarget{MI_Anzeige_ZGF}{}
\item {\bf Membership function and total histogram}: \newline  
like \hyperlink{MI_Anzeige_ZGF}{\emph{View - Single features - Membership function}}, but with an additional histogram describing the selected data
points of the single feature.
\hypertarget{MI_Anzeige_ZGF_Gesamthistogramm}{}
\item {\bf Membership function and class histogram}: \newline  
like \hyperlink{MI_Anzeige_ZGF}{\emph{View - Single features - Membership function}}, but with separate histograms for each class of the output variable
belonging to the selected data points of the single feature.
\hypertarget{MI_Anzeige_ZGF_Klassenhistogramm}{}
\item {\bf Entropy}: \newline  
shows the entropy balance  (\textindex{input entropy}, \textindex{output entropy}, and
\textindex{mutual information}) for each selected single feature vs. the selected output variable.
The single feature will be discretized using the parameters from \hyperlink{CE_Fuzzy_AnzLingTerme}{\emph{Control element: Data mining: Special methods - Number of linguistic terms}} and
\hyperlink{CE_Fuzzy_TypeZGF}{\emph{Control element: Data mining: Special methods - Type of membership function}}.
\hypertarget{MI_Anzeige_Entropie}{}
\item {\bf Correlation coefficients (Pearson)}: \newline  
writes relevant (linear) Pearson \textindex{correlation} coefficients between all selected output variables in a file. The
relevance is defined as a larger absolute value than the threshold in \hyperlink{CE_Krit_Koeff}{\emph{Control element: Data mining: Statistical options - Threshold for correlation coefficient}}. The function is only able
to show linear relations and assumes normal distributions of all features. Optionally, the computation can be restricted by \hyperlink{CE_Anzeige_CorrSelFeature}{\emph{Control element: Data mining: Statistical options - Show correlation only for one selected feature}}  to the selected single features defined by \hyperlink{CE_Regression_Output}{\emph{Control element: Data mining: Regression - Output variable of regression}}. The reporting of p-values for correlation coefficients unequal zero is switched on and off by \hyperlink{CE_Anzeige_CorrPValues}{\emph{Control element: Data mining: Statistical options - Show p values for correlation}}. A method for the correction of multiple tests can be defined by \hyperlink{CE_Corr_Bonferroni}{\emph{Control element: Data mining: Statistical options - Bonferroni correction}}.
\hypertarget{MI_Anzeige_EM_Korr}{}
\item {\bf Correlation coefficients (Spearman)}: \newline  
like \hyperlink{MI_Anzeige_EM_Korr}{\emph{View - Single features - Correlation coefficients (Pearson)}}, but with \textindex{Spearman correlation coefficients}. They
evaluate rank correlations of features. This function gives better results for nonlinear but
monotone relations between features.
\hypertarget{MI_Anzeige_EM_Spear}{}
\item {\bf Correlation coefficients (Kendall)}: \newline  
like \hyperlink{MI_Anzeige_EM_Korr}{\emph{View - Single features - Correlation coefficients (Pearson)}}, but with \textindex{Kendall''s tau coefficient}. They
evaluate rank correlations of features. This function gives better results for nonlinear but
monotone relations between features.
\hypertarget{MI_Anzeige_EM_Kendall}{}
\item {\bf Show correlation coefficients}: \newline  
shows linear \textindex{correlation} coefficients of selected single features as figure. Green
values visualize positive, red values negative correlations. The color intensity is a measure for
the absolute value of the correlation coefficient. Hereby, the correlation coefficient of a type according to \hyperlink{CE_Corr_Type}{\emph{Control element: Data mining: Statistical options - Type for correlation parameters}} is used.
\hypertarget{MI_Korrelationsvisualisierung}{}
\item {\bf Show feature relevances (graphic)}: \newline  
summarizes calculated feature relevances as figure(s). The main advantage is a first overview about
problems with a large set of single features.

For univariate relevances, the relevance is shown versus the feature number.

For multivariate relevances, two figures will be shown:

The first figure visualizes step-wise relevance improvements of each feature as bars versus the
feature number. The lowest bar shows the univariate relevance. The second bar contains improvements
for the multivariate relevance by the second feature after selecting the best univariate feature in
first step. All following bars show improvements of step-wise selections of the $s_m=3,4,...$
feature after preselecting $s_m-1$ feature in the steps before. Important features are
characterized by high bars.

The second figure contains the calculated multivariate relevances for the selection the 1-st,...,
$s_m$-th feature. The relevances assume also the step-wise preselection of $s_m-1$ best feature in
the steps before.
\hypertarget{MI_Anzeige_EM_Relevanzen_Grafik}{}
\item {\bf Show feature relevances (sorted table)}: \newline  
shows all computed feature relevances sorted by descending relevance values.
\hypertarget{MI_Anzeige_EM_Relevanzen}{}
\item {\bf Show feature relevances (unsorted table)}: \newline  
shows all computed feature relevances sorted by ascending feature numbers.
\hypertarget{MI_Anzeige_EM_Relevanzen_un}{}
\item {\bf Show a priori feature relevances (unsorted table)}: \newline  
shows all a priori relevances sorted by ascending feature numbers.
\hypertarget{MI_Anzeige_EM_APriori}{}
\item {\bf Add feature relevances to an archive}: \newline  
saves the recent feature relevances to a archive, e.g., to compare different methods for feature evaluation. This archive can be saved as separate SciXMiner project using \hyperlink{MI_Anzeige_SaveFeatArchive}{\emph{View - Single features - Save archive with feature relevances}}.
\hypertarget{MI_Anzeige_AddToFeatArchive}{}
\item {\bf Save archive with feature relevances}: \newline  
save the recent archive with feature relevances as separate SciXMiner project ([Name of the Project  '\_feature\_relevances.prjz']). In this SciXMiner project, the single features are represented as terms of an output variable and the values of the feature relevances are saved as single features.
\hypertarget{MI_Anzeige_SaveFeatArchive}{}
\end{itemize}
\subsection{Single features (multivariate)}
\hypertarget{MI_Einzelmerkmale_Multi}{}
groups multidimensional visualizations for single features.
\begin{itemize}
\item {\bf Parallel coordinates}: \newline  
plots the selected single features in parallel coordinates form.
\hypertarget{MI_Anzeige_EM_Parallel}{}
\item {\bf Andrews Plot}: \newline  
plots the selected single features as an "`Andrews Plot"'.
\hypertarget{MI_Anzeige_EM_Andrews}{}
\item {\bf Glyph Plot}: \newline  
plots a glyph plot of the selected features.
\hypertarget{MI_Anzeige_EM_Glyph}{}
\item {\bf Heatmap}: \newline  
plots a heatmap of the selected single features.
\hypertarget{MI_Anzeige_EM_Heatmap}{}
\end{itemize}
\subsection{Classification}
\hypertarget{MI_Anzeige_Klassifikation}{}
contains all functions for the visualization of classification results.
\begin{itemize}
\item {\bf Result}: \newline  
shows a scatterplot of the last classification result in the (optionally aggregated) feature space
of the classifier. The kind of chosen linguistic terms (e.g. true class assignment, result of the
classifier, a different output variable defined by \hyperlink{CE_Anzeige_DifferentClass}{\emph{Control element: View: Single features - Different output variable}} etc.) can be switched by the options in \hyperlink{CE_Anzeige_Klassenanzeige}{\emph{Control element: View: Classification and regression - Display classes for output variables}}.
\hypertarget{MI_Anzeige_Klassi_Erg}{}
\item {\bf 2D classification with covariance matrixes}: \newline  
like \hyperlink{MI_Anzeige_Klassi_Erg}{\emph{View - Classification - Result}}, but adds estimated covariance matrices for the linguistic terms
of the output variable. This menu item is only available if a \textindex{Bayes classifier} was
applied.
\hypertarget{MI_Anzeige_Klassi_Kova}{}
\item {\bf 2D plot classifier with support vectors}: \newline  
like \hyperlink{MI_Anzeige_Klassi_Erg}{\emph{View - Classification - Result}}, but marks \textindex{Support vectors} by triangles. This menu item is only available
if a Support Vector Machine was applied as classifier.\index{Support Vector Machines}
\hypertarget{MI_Anzeige_Klassi_SVM}{}
\item {\bf ROC curve...}: \newline  
computes a \textindex{Receiver-Operating-Characteristic} curve (\textindex{ROC curve}). It shows all
possible compromises between \textindex{sensitivity} and \textindex{specificity} with respect to the
classifier decision for one selected linguistic term of the output variable. This term is chosen by
a separate configuration window. The tuning parameter for the ROC curve is the gradual class
membership for a decision of the last applied classifier.
\hypertarget{MI_Anzeige_Klassi_ROC}{}
\item {\bf Time series classification error vs. time}: \newline  
shows the classification error of a time series classification vs. time.
\hypertarget{MI_Anzeige_ZRKlassi}{}
\item {\bf Used features of the time series classifier}: \newline  
shows a figure with the selected time series (marked by points) of a time series classifier vs.
time. The method of feature selection and the number of selected features are controlled by
\hyperlink{CE_Auswahl_EM}{\emph{Control element: Single features - Selection of single feature(s) (SF)}} resp. \hyperlink{CE_Anzahl_Merkmale}{\emph{Control element: Data mining: Classification of single features - Number of selected features}}.
\hypertarget{MI_Anzeige_ZRMerkAuswahl}{}
\end{itemize}
\subsection{Regression}
\hypertarget{MI_CV_RegrEM_Standard}{}
shows the results of the most recent applied \textindex{regression model}. The selection of data
points is fixed to the application of the regression model.
\begin{itemize}
\item {\bf Output variable and estimation}: \newline  
shows the output variable of the regression vs. the estimation by the \textindex{regression model}.
\hypertarget{MI_Anzeige_Regression_y_ydach}{}
\item {\bf Output variable and error}: \newline  
shows the output variable of the regression vs. the regression error.
\hypertarget{MI_Anzeige_Regression_y_Fehler}{}
\item {\bf Input variable(s), output variable, and regression function (2D resp. 3D)}: \newline  
shows the function of the estimated \textindex{regression model}. It is computed by a grid of
additional data points. The number of data points is defined by \hyperlink{CE_Anzeige_Gitterpunkte}{\emph{Control element: View: Classification and regression - Number of grid points}}. The
grid is tuned for the analysis of the interpolation and extrapolation behavior. In addition, the
values of the input variables vs. the output variables during the design are plotted as
scatterplot. If a normalization or aggregation of features was chosen, the visualization shows
these features instead of the original ones.

The function is only available for regression models with one or two inputs.
\hypertarget{MI_Anzeige_Regression_x_y_2D3D}{}
\item {\bf Input variable(s), output variable, and regression function (top view 3D)}: \newline  
like \hyperlink{MI_Anzeige_Regression_x_y_2D3D}{\emph{View - Regression - Input variable(s), output variable, and regression function (2D resp. 3D)}}, but with a two-dimensional projection (overhead) of the function.
The estimation of the output variable is color-coded.
\hypertarget{MI_Anzeige_Regression_x_y_Draufsicht}{}
\item {\bf Protocol regression model application into file}: \newline  
protocol the errors statistics of the regression model to a file.
\hypertarget{MI_Anzeige_RegrStatFile}{}
\item {\bf GUI for multidimensional visualization}: \newline  
opens a GUI to create a 3D extraction out of a \textindex{regression model}. Two input parameters can be set to variable, by typing ''x'' resp. ''y'' into the box. The input parameters left can be set to a constant value, by typing the value into the box. This Visualization works on the same principles as \hyperlink{MI_Anzeige_Regression_x_y_2D3D}{\emph{View - Regression - Input variable(s), output variable, and regression function (2D resp. 3D)}}. Moreover, the region for the scatterplot can be defined with the columns ''Min:'' and ''Max:''.

The function is only available for existing regression models with more than two inputs.
\hypertarget{MI_GUI_Regression_Multi_Dimensional}{}
\item {\bf Generate macro for multidimensional visualization}: \newline  
creates and opens a macro for 3D visualization of a \textindex{regression model}. Further information on handling can be found within the macro.
\hypertarget{MI_Erzeuge_Regression_Multi_Dimensional}{}
\item {\bf Apply macro for multidimensional visualization}: \newline  
executes the macro created by \hyperlink{MI_Erzeuge_Regression_Multi_Dimensional}{\emph{View - Regression - Generate macro for multidimensional visualization}}. Alternatively, the macro can be started the usual way via \hyperlink{MI_Makro_Ausfuehren}{\emph{Extras - Play macro...}}.
\hypertarget{MI_Anzeige_Regression_Multi_Dimensional}{}
\item {\bf Coefficients of the polynomial model}: \newline  
writes the coefficients of the recent polynomial model for the regression into a file called
\emph{*\_poly.tex} resp. \emph{*\_poly.txt}.
\hypertarget{MI_Anzeige_Koeff_Polynom}{}
\item {\bf Structure of the MLP net}: \newline  
shows the structure of a MLP net for a \textindex{regression model}. Excitatory links are green, inhibitory links red.
The color intensity is a measure for the strength of a link.
\hypertarget{MI_Anzeige_MLP}{}
\item {\bf Multidimensional response surface (RSM, with Statistics Toolbox)}: \newline  
opens the Matlab function rstools.m for the visualization of a \textindex{Multidimensional response surface (RSM)} with the selected single features and the selected output variable of the regression \hyperlink{CE_Regression_Output}{\emph{Control element: Data mining: Regression - Output variable of regression}}.
\hypertarget{MI_RSTools}{}
\end{itemize}
\subsection{Aggregated features}
\hypertarget{MI_Ansicht_Aggregation}{}
contain all functions for the visualization of aggregated single features. It includes the
transformation matrices for the dimension reduction and the related membership functions for the
aggregated single features. \index{feature aggregation}
\begin{itemize}
\item {\bf Eigenvectors and transformation vectors}: \newline  
shows the \textindex{factor loadings} resp. weights of the single features for the generation of
the aggregated features. For each aggregated features, a different color is chosen.
\hypertarget{MI_Ansicht_Aggregation_HD}{}
\item {\bf Factor loadings (2D eigenvectors or transformation vectors)}: \newline  
plots the elements of the first two transformation vectors for feature aggregation as scatterplot
in a two-dimensional space.
\hypertarget{MI_Ansicht_Aggregation_LD}{}
\item {\bf Membership function}: \newline  
plots for each aggregated feature a figure with the membership functions for all linguistic terms.
\hypertarget{MI_Anzeige_ZGFAggr}{}
\item {\bf Membership function and total histogram}: \newline  
like \hyperlink{MI_Anzeige_ZGFAggr}{\emph{View - Aggregated features - Membership function}}, but with an additional histogram for all selected data points.
\hypertarget{MI_Anzeige_ZGFAggr_Gesamthistogramm}{}
\item {\bf Membership function and class histogram}: \newline  
like \hyperlink{MI_Anzeige_ZGFAggr}{\emph{View - Aggregated features - Membership function}}, but with separate subfigures with histograms for each term of the
selected output variable.
\hypertarget{MI_Anzeige_ZGFAggr_Klassenhistogramm}{}
\end{itemize}
\subsection{Output variables}
\hypertarget{MI_Sortieren_OutputVariables}{}
contain all functions for the visualization of output variables.
\begin{itemize}
\item {\bf Qualitative output variables}: \newline  
plots the selected output variable vs. the number of the related data points.
\hypertarget{MI_Qual_Ausgangsgroessen}{}
\item {\bf 2D histogram...}: \newline  
shows the distribution of two output variables as histogram. Both output variables are selected by a
configuration window.
\hypertarget{MI_Ansicht_Ausgang2D}{}
\item {\bf Term number of output variable}: \newline  
writes the number of selected data points for all linguistic terms of the output variable in a protocol file.
\hypertarget{MI_Ansicht_OutputTerms}{}
\end{itemize}
\subsection{Spectrogram}
\hypertarget{MI_Spektrogramm}{}
contains all functions for the visualization of spectrograms.
\begin{itemize}
\item {\bf Compute and show}: \newline  
computes and shows a separate spectrogram for each selected data point and each selected time
series. Each spectrogram is normalized to values between zero and one (see
\hyperlink{MI_Spektrogramm_DSMitteln}{\emph{View - Spectrogram - Compute and show (mean for selected data points)}} resp. \hyperlink{MI_Spektrogramm_KlMitteln}{\emph{View - Spectrogram - Compute and show (mean for selected class)}} for mean values of
spectrograms). A warning is shown for very large numbers of spectrograms due to many selected data
points.

In many cases, the \textindex{interpretability} of spectrograms might be improved by adapting the
selection of \hyperlink{CE_Kennlinie}{\emph{Control element: View: Spectrogram, FFT, CCF - Function for color bar}} and the related parameters. This controls the color scaling
especially to a finer resolution for lower amplitudes. The compromise
between time and frequency resolution is tuned by \hyperlink{CE_Zeitreihen_Fenstergroesse}{\emph{Control element: View: Spectrogram, FFT, CCF - Window size [sample points]}}.
\hypertarget{MI_Spektrogramm_BerAnz}{}
\item {\bf Compute}: \newline  
similar to \hyperlink{MI_Spektrogramm_BerAnz}{\emph{View - Spectrogram - Compute and show}}, but only for computation of spectrograms. This saves time
especially for large project with a script-based generation.
\hypertarget{MI_Spektrogramm_Ber}{}
\item {\bf Show}: \newline  
similar to \hyperlink{MI_Spektrogramm_BerAnz}{\emph{View - Spectrogram - Compute and show}} but only for visualization. This saves time by avoiding a
new computation, e.g. during tuning of visualization parameters.
\hypertarget{MI_Spektrogramm_Anz}{}
\item {\bf Principal component analysis for spectrograms}: \newline  
makes a \textindex{Principal Component Analysis} on the recent computed \textindex{spectrogram}. The number of
principal components is defined by \hyperlink{CE_PCA_Spektrogramm}{\emph{Control element: View: Spectrogram, FFT, CCF - Number of principal components for spectrogram}}. The result are figures with eigenvectors vs. frequency,
percentages of eigenvalues, and aggregated features vs. time. The function is able to detect and visualize dominating
signals containing a mixture of different frequencies. The interpretation might be sophisticated due to window effects.
The window length is defined by (\hyperlink{CE_Zeitreihen_Fenstergroesse}{\emph{Control element: View: Spectrogram, FFT, CCF - Window size [sample points]}}).
\hypertarget{MI_Spektrogramm_PCA}{}
\item {\bf Compute and show (mean for selected data points)}: \newline  
computes firstly a separate spectrogram for each selected data point and each selected time series.
Each spectrogram is normalized to values between zero and one. In a next step, the mean value over
all data points of the computed spectrograms are computed and finally shown. The resulting ''mean
spectrogram'' gives only a qualitative impression due to the artifacts caused by separate
normalization.
\hypertarget{MI_Spektrogramm_DSMitteln}{}
\item {\bf Compute and show (mean for selected class)}: \newline  
similar to \hyperlink{MI_Spektrogramm_DSMitteln}{\emph{View - Spectrogram - Compute and show (mean for selected data points)}}, but with separate mean values of spectrograms for all existing
classes of the selected output variable. Due to the separate normalization, the result is rather
qualitatively.
\hypertarget{MI_Spektrogramm_KlMitteln}{}
\end{itemize}
\subsection{Morlet spectrogram}
\hypertarget{MI_MorletSpektro}{}
contains all functions for the visualization of Morlet spectrograms.
\begin{itemize}
\item {\bf Compute and show (selected data points and time series)}: \newline  
computes new time series for the selected time series by convolution with complex \textindex{Morlet
wavelets}. After the convolution, the new time series are filtered  by an \textindex{IIR-Filter}.
The IIR parameters are defined by \hyperlink{CE_Parameter_IIR}{\emph{Control element: Time series: Extraction - Parameters - Parameter IIR filter}}. The results are shown as
tree-dimensional plots (x axis: time, y axis: frequencies, color: coefficient of the
filtered signal).

The frequency range of the \textindex{Morlet wavelets} is defined by \hyperlink{CE_ZR_Filterfreq}{\emph{Control element: Time series: Extraction - Parameters - Frequencies (FIL, Morlet spectrogram)}} and
\hyperlink{CE_MorletWavelet_Eigenfreq}{\emph{Control element: Time series: Extraction - Parameters - Morlet wavelet: eigenfrequency}}. The parameters describe the lower and higher limit.
\hyperlink{CE_MorletSpekt_FreqSchritt}{\emph{Control element: View: Spectrogram, FFT, CCF - Morlet spectrogram: Frequency stepwidth}} could be used to adapt the step length (default: 1).

The spectrogram refers to absolute values or to relative values to a baseline. These parameters
could be modified in \hyperlink{CE_MorletSpekt_Relativ}{\emph{Control element: View: Spectrogram, FFT, CCF - Morlet spectrogram (relative to baseline)}} and \hyperlink{CE_MorletSpekt_Baseline}{\emph{Control element: View: Spectrogram, FFT, CCF - Sample points baseline}}.
\hypertarget{MI_MorletSpektro_BerAnz}{}
\item {\bf Compute and show (mean for selected class)}: \newline  
similar to \hyperlink{MI_MorletSpektro_BerAnz}{\emph{View - Morlet spectrogram - Compute and show (selected data points and time series)}}. Here, the mean values will be computed for the different
classes using the selected data points resulting in a mean \textindex{Morlet spectrogram} for each
linguistic term.
\hypertarget{MI_MorletSpektro_MWKlassen}{}
\item {\bf Plot only}: \newline  
shows again the computed \textindex{Morlet spectrogram}.
\hypertarget{MI_MorletSpektro_Anz}{}
\end{itemize}
\subsection{Cross and Auto Correlation Functions}
\hypertarget{MI_KKFAKF}{}
contains all functions to show Cross and Auto Correlation Functions.
\begin{itemize}
\item {\bf Separately for each data point}: \newline  
computes and shows auto resp. \textindex{Cross Correlation Functions} for each selected data point.
The selection of time series is done in a separate configuration window with two fields. The
selection of identical time series leads to auto correlation functions, different time series
produce Cross Correlation Functions. A multiple selection of time series is possible. See also
\hyperlink{MI_KKFAKF_ZR_MWDaten}{\emph{View - Cross and Auto Correlation Functions - Mean for selected data points}} and \hyperlink{MI_KKFAKF_ZR_MWKlassen}{\emph{View - Cross and Auto Correlation Functions - Mean for classes}} for a fast overview for many selected
data points.
\hypertarget{MI_AKFKKF_Anzeige}{}
\item {\bf Separately for selected data points (Short-time analysis)}: \newline  
similar to \hyperlink{MI_AKFKKF_Anzeige}{\emph{View - Cross and Auto Correlation Functions - Separately for each data point}}, but with a computation and visualization of a
\textindex{short-time correlation analysis}. The lengths of the time window is chosen in
\hyperlink{CE_Zeitreihen_Fenstergroesse}{\emph{Control element: View: Spectrogram, FFT, CCF - Window size [sample points]}}. This function deals with an overview about time-variant
changes of \textindex{correlation}.
\hypertarget{MI_AKFKKF_Kurzzeit}{}
\item {\bf Mean values of correlation coefficients (defined time shift)}: \newline  
similar to \hyperlink{MI_KKFAKF_ZR_MWDaten}{\emph{View - Cross and Auto Correlation Functions - Mean for selected data points}}, but extract only one sample of the computed Cross resp. Auto Correlation
Function. This sample is controlled by a time shift defined in \hyperlink{CE_ZR_Tau}{\emph{Control element: View: Spectrogram, FFT, CCF - Time shift Tau [SP]}}. Hereby, the correlation coefficient of a type according to \hyperlink{CE_Corr_Type}{\emph{Control element: Data mining: Statistical options - Type for correlation parameters}} is used. The results are displayed in a text
file for all relevant correlations with absolute values larger than \hyperlink{CE_Krit_Koeff}{\emph{Control element: Data mining: Statistical options - Threshold for correlation coefficient}}. Finally, all correlations are
shown in a color figure similar to \hyperlink{MI_Korrelationsvisualisierung}{\emph{View - Single features - Show correlation coefficients}}.

The main advantage of this function is a fast analysis of correlations between many time series.
\hypertarget{MI_KKFAKF_MWDaten}{}
\item {\bf Class mean values of correlation coefficients (with defined time shift)}: \newline  
similar to \hyperlink{MI_KKFAKF_MWDaten}{\emph{View - Cross and Auto Correlation Functions - Mean values of correlation coefficients (defined time shift)}}, but with separate computations for each class (see also
\hyperlink{MI_KKFAKF_ZR_MWKlassen}{\emph{View - Cross and Auto Correlation Functions - Mean for classes}}).
\hypertarget{MI_KKFAKF_MWKlassen}{}
\item {\bf Mean for selected data points}: \newline  
works as \hyperlink{MI_AKFKKF_Anzeige}{\emph{View - Cross and Auto Correlation Functions - Separately for each data point}}, but computes mean values for all selected data points. The function is only enabled
if at least two data points have been selected.
\hypertarget{MI_KKFAKF_ZR_MWDaten}{}
\item {\bf Mean for classes}: \newline  
works as \hyperlink{MI_AKFKKF_Anzeige}{\emph{View - Cross and Auto Correlation Functions - Separately for each data point}}, but computes mean values for all selected data points of each existing class using
the selected output variable. The function is only enabled if at least two data points have been selected.
\hypertarget{MI_KKFAKF_ZR_MWKlassen}{}
\item {\bf Correlation function for different data points and selected time series}: \newline  
computes and shows Cross Correlation Functions for all selected combinations of data points and time series. The
selection of data points is done by a separate configuration window. Only one data point should be selected in its
upper control element. In the lower control element, a multiple selection is possible  (CTRL + left mouse button). The
Cross Correlation Functions are generated for each possible combination of reference with comparison data points. The
selection of the time series is shown in the main window.

Example:

Reference data point: 1,

Comparison data points: 3, 4, 5,

Combinations: 1-3, 1-4, 1-5.
\hypertarget{MI_KKF_Datentupel}{}
\item {\bf Correlation function for different data points and selected time series (mean value)}: \newline  
like \hyperlink{MI_KKF_Datentupel}{\emph{View - Cross and Auto Correlation Functions - Correlation function for different data points and selected time series}}, but mean values for the correlation functions will be computed. As a
result, for each time series only one function is generated.
\hypertarget{MI_KKF_Datentupel_Mean}{}
\item {\bf Correlation coefficients selected data point and time series (pre-defined time-shift Tau)}: \newline  
like \hyperlink{MI_KKF_Datentupel}{\emph{View - Cross and Auto Correlation Functions - Correlation function for different data points and selected time series}}, but the data points are selected by the main window. A correlation
coefficient will be computed for each combination of data points and a predefined time shift
$\tau$. The result is shown as color matrix. Hereby, the correlation coefficient of a type according to \hyperlink{CE_Corr_Type}{\emph{Control element: Data mining: Statistical options - Type for correlation parameters}} is used.
\hypertarget{MI_KKoeff_Datentupel}{}
\item {\bf Mean correlation coefficients over time series for selected data points and time series (defined time shift Tau)}: \newline  
like \hyperlink{MI_KKoeff_Datentupel}{\emph{View - Cross and Auto Correlation Functions - Correlation coefficients selected data point and time series (pre-defined time-shift Tau)}}, but mean values for the different time series will be computed.
\hypertarget{MI_KKoeff_Datentupel_Mean}{}
\end{itemize}
\subsection{FFT}
\hypertarget{MI_FFTAll}{}
includes all functions for computation and analysis of the \textindex{Fast-Fourier-Transformation} (\textmainindex{FFT}).
\begin{itemize}
\item {\bf Compute and show FFT  (selected data points and time series)}: \newline  
computes a \textindex{Fast-Fourier-Transformation} (\textmainindex{FFT}) for all selected time
series and data points and plots the results. The control element \hyperlink{CE_Anzeige_FFTPeriode}{\emph{Control element: View: Spectrogram, FFT, CCF - Plot FFT vs. period length (instead of frequency)}}
switches between a plot vs. period lengths and frequency (default value).

The function reduces the number of sampling points to the next smaller power of two to compute a
FFT and to avoid window artifacts (e.g. 5000 sampling points $\rightarrow$ reduction to 4096).
\hypertarget{MI_FFT}{}
\item {\bf Show dominant frequencies (sorted by amplitude)}: \newline  
shows the most dominant frequencies of the FFT sorted by amplitudes. The number is defined under \hyperlink{CE_FFT_AnzFreq}{\emph{Control element: View: Spectrogram, FFT, CCF - Number of dominant frequencies for visualization}}.
\hypertarget{MI_FFT_HitlistSortAmp}{}
\item {\bf Show dominant frequencies (sorted by frequency)}: \newline  
shows the most dominant frequencies of the FFT sorted by frequencies. The number is defined under \hyperlink{CE_FFT_AnzFreq}{\emph{Control element: View: Spectrogram, FFT, CCF - Number of dominant frequencies for visualization}}.
\hypertarget{MI_FFT_HitlistSortFreq}{}
\end{itemize}
\subsection{Fuzzy systems}
\hypertarget{MI_Ansicht_Fuzzy}{}
contains all functions for the visualization of a \textindex{fuzzy system} designed by
\hyperlink{MI_Fuzzy_Einzelregel}{\emph{Data mining - Fuzzy systems - Design (single rules)}}, \hyperlink{MI_Fuzzy_Basis}{\emph{Data mining - Fuzzy systems - Design (rule base)}} or by a chosen fuzzy classifier.
\begin{itemize}
\item {\bf Show rules (with variable names)}: \newline  
shows a list of the recent fuzzy rules. It results from loading a rule base from file, from generating single rules or
rule bases. The input and output variables are written in form of the variable names.
\hypertarget{MI_Fuzzy_Anzeige_Bez}{}
\item {\bf Show rules (with variable numbers)}: \newline  
like \hyperlink{MI_Fuzzy_Anzeige_Bez}{\emph{View - Fuzzy systems - Show rules (with variable names)}}, but uses the numbers of input variables and a symbolic output
variable $y$.
\hypertarget{MI_Fuzzy_Anzeige_Nr}{}
\item {\bf Show rules (figure)}: \newline  
shows a fuzzy rules with a maximum number of three input variables in a figure.
\hypertarget{MI_Fuzzy_Grafik}{}

The corresponding figure shows Fig.~\ref{fig:auswahlfensterMI_Fuzzy_Grafik}.
\begin{figure}[h] 
 \includegraphics[width=0.9\columnwidth]{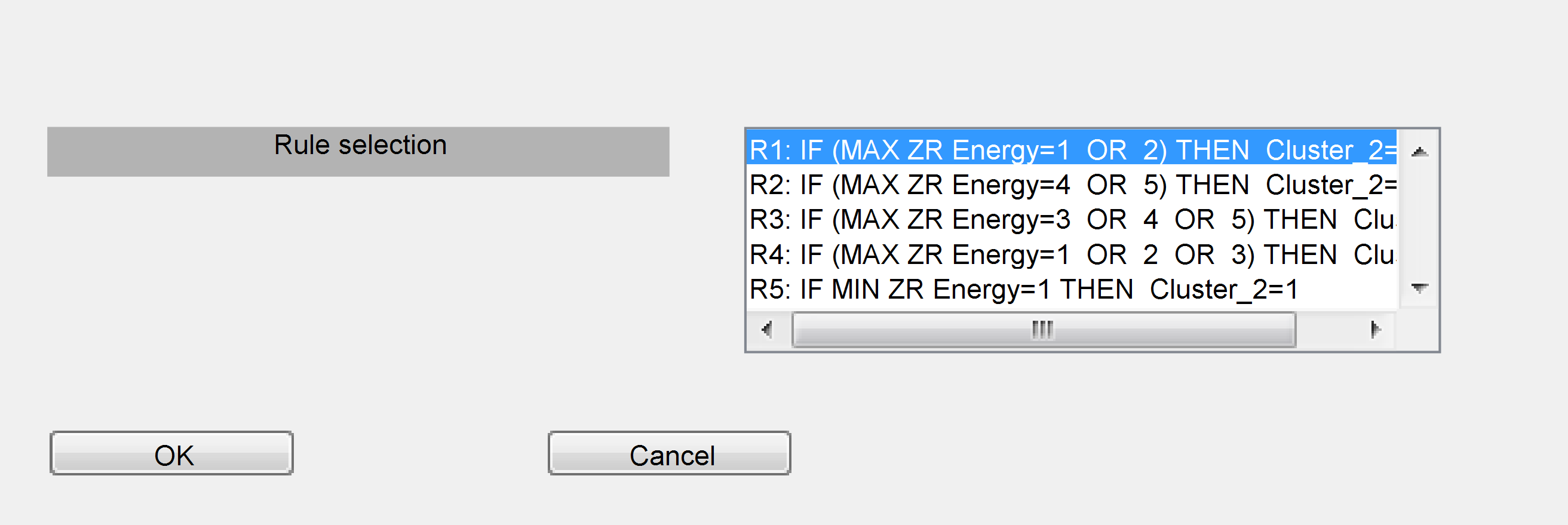}
 \caption{\label{fig:auswahlfensterMI_Fuzzy_Grafik} Figure for \emph{''View - Fuzzy systems - Show rules (figure)''}}
 \end{figure}
\end{itemize}
\subsection{Decision tree (LaTeX)}
\hypertarget{MI_Ansicht_Ebaum}{}
describes a \textindex{decision tree} of the last designed classifier in a LaTeX table and writes
the result in a file.
\subsection{Clustering}
\hypertarget{MI_Ansicht_Clustern}{}
contains all functions for clustering.
\begin{itemize}
\item {\bf Cluster memberships (sorted by clusters)}: \newline  
shows the recent \textindex{cluster membership values} in a figure and sorts the results by the
number of the cluster. This figure enables an evaluation of the quality of the found clusters. A
good quality is characterized by cluster membership values near zero and near one for most data
points. A possible reason for an unclear membership to different clusters is an inappropriate
number of clusters.
\hypertarget{MI_Ansicht_ClusterZGH_sortiert}{}
\item {\bf Cluster memberships (sorted by data points)}: \newline  
shows the recent \textindex{cluster membership values} in a figure and sorts the results by the
number of the data point. This figure enables an evaluation of possible (e.g. temporal) structures
depending on the number of data points.
\hypertarget{MI_Ansicht_ClusterZGH_unsortiert}{}
\item {\bf Dendrogram (Statistic Toolbox)}: \newline  
shows the similarities between data points as a tree (\textindex{dendrogram}). This visualization is only
available after performing \textindex{hierarchical cluster algorithms} using the \textindex{Statistic
Toolbox} of Matlab by \hyperlink{MI_Cluster_matlab}{\emph{Data mining - Clustering - Design and apply (Statistic Toolbox)}}.
\hypertarget{MI_Ansicht_ClusterZGH_Dendro}{}
\item {\bf Pair-wise distances for dendrogram (STAT toolbox)}: \newline  
shows the pairwise distances between all selected data points. Such distances are used for the computation of
dendrograms.
\hypertarget{MI_Ansicht_ClusterZGH_DendroDistPair}{}
\item {\bf Clustergram}: \newline  
computes a \textindex{two-dimensional clustering} for the selected single features and data points. This command executes the function ''clustergram'' of the MATLAB Bioinformatics Toolbox.
\hypertarget{MI_Ansicht_Clustergramm}{}
\end{itemize}
\subsection{Self Organizing Maps...}
\hypertarget{MI_Ansicht_SOM}{}
contains all functions for the visualization of \textindex{Self Organizing Maps}.
\begin{itemize}
\item {\bf Positions}: \newline  
shows the positions of the weight vectors of Self Organizing Maps in the (single) feature space (for 2D using the MATLAB function plotsompos.m).
\hypertarget{MI_Ansicht_SOMPos}{}
\item {\bf Hits}: \newline  
show the number of hits per neuron for Self-Organizing Maps (based on the design).
\hypertarget{MI_Ansicht_SOMHits}{}
\end{itemize}
\subsection{Data point distances}
\hypertarget{MI_Ansicht_PairDistDat}{}
contains all functions to compute \textindex{data point distances}.
\begin{itemize}
\item {\bf Compute (selected data points)}: \newline  
computes pairwise distances between all selected data points. Here, the Frobenius norm for all selected single features will be applied. In addition, an optional normalization defined by \hyperlink{CE_Normierung_Merkmale}{\emph{Control element: Data mining: Classification of single features - Normalization of single features}} is used.
\hypertarget{MI_Ansicht_PairDistDatCompInd}{}
\item {\bf Compute (selected data points vs. manual selection)}: \newline  
computes pairwise distances between all selected data points vs. all data points in \hyperlink{CE_Select_DataPoints}{\emph{Control element: General options - Manual selection of data points}}.
\hypertarget{MI_Ansicht_PairDistDatCompManu}{}
\item {\bf Compute (searching for neighbors of the first element of the manual selection)}: \newline  
computes the nearest neighbors for the first data point in \hyperlink{CE_Select_DataPoints}{\emph{Control element: General options - Manual selection of data points}} using the data point distances to the selected data points.
\hypertarget{MI_Ansicht_PairDistDatCompNeighb}{}
\item {\bf Sort (VAT algorithm)}: \newline  
resorts data point in the plot based on VAT (visual assessment of tendency) algorithm to improve the interpretation of data point distance matrices (similar data points are grouped near together).
\hypertarget{MI_Ansicht_PairDistSortVAT}{}
\item {\bf Show vector}: \newline  
shows the computed data point distances in vector form (x: data point number, y: distance).
\hypertarget{MI_Ansicht_PairDistDatVect}{}
\item {\bf Show matrix}: \newline  
shows the computed data point distances in matrix form (x,y: data point numbers, color: distance).
\hypertarget{MI_Ansicht_PairDistDatMat}{}
\item {\bf Show neighbors}: \newline  
plots the k nearest neighbors computed by \hyperlink{MI_Ansicht_PairDistDatCompNeighb}{\emph{View - Data point distances - Compute (searching for neighbors of the first element of the manual selection)}} in the MATLAB workspace. The value of k is chosen by \hyperlink{CE_kNN_k}{\emph{Control element: Data mining: Special methods - k}}.
\hypertarget{MI_Ansicht_PairDistDatNeighb}{}
\end{itemize}
\subsection{Project report}
\hypertarget{MI_Projektreport}{}
shows the information about the recent project (name, number of data points etc.) and all selected
parameters in the control elements.
\subsection{Recoding table}
\hypertarget{MI_Umkodierung}{}
shows modified class numbers if an output variable was not original numbered with 1 ... n in the
code variable.

\section{Menu items 'Data mining'}
contains all functions for the supervised and unsupervised \textindex{classification} and for the
\textindex{regression} of single features and time series.
\subsection{Selection and evaluation of single features}
\hypertarget{MI_Bewertung_EM}{}
contains all functions for the evaluation and selection of single features using data mining
methods.
\begin{itemize}
\item {\bf ANOVA, univariate}: \newline  
computes feature relevances with the \textindex{ANOVA} method. In contrast to the \textindex{MANOVA} method
redundancies between features are not investigated. The method assumes a normal distribution for each output class
(linguistic term) in the input space.
\hypertarget{MI_EMAusw_ANOVA}{}
\item {\bf MANOVA, multivariate}: \newline  
computes feature relevances with the \textindex{MANOVA} method. It performs a step-wise selection by adding one best
single feature to an existing set of selected features. The selection terminates if the desired number of features in
\hyperlink{CE_Anzahl_Merkmale}{\emph{Control element: Data mining: Classification of single features - Number of selected features}} is reached or if the next feature does not increase the feature relevance of the set of
features. The method takes redundancies between features into account.
\hypertarget{MI_EMAusw_MANOVA}{}
\item {\bf Information theoretic measures}: \newline  
similar to \hyperlink{MI_EMAusw_ANOVA}{\emph{Data mining - Selection and evaluation of single features - ANOVA, univariate}}, but the method does not assume a normal distribution.

All features are fuzzified and the \textindex{mutual information} per \textindex{output entropy}
for the selected output variable is computed (evaluation measure of the \textindex{ID3 method}).
However, a split by existing membership functions is used instead of a new computed binary split in
each node \cite{Mikut05}. \index{information measures}
\hypertarget{MI_EMAusw_Inform}{}
\item {\bf Classification accuracy}: \newline  
similar to \hyperlink{MI_EMAusw_MANOVA}{\emph{Data mining - Selection and evaluation of single features - MANOVA, multivariate}}, but computes a relative classification error as performance
measure (see e.g. Eq. (3.2) in \cite{Reischl06}). In each selection step, the design Bayes classifier bases on
the pre-selected features for former steps and a new candidate feature.
\hypertarget{MI_EMAusw_Klassifikation}{}
\item {\bf Classification accuracy (univariate)}: \newline  
like \hyperlink{MI_EMAusw_Klassifikation}{\emph{Data mining - Selection and evaluation of single features - Classification accuracy}}, but only for the selection of one single feature.
\hypertarget{MI_Merkausklassuni}{}
\item {\bf Fuzzy classification accuracy}: \newline  
like \hyperlink{MI_EMAusw_Klassifikation}{\emph{Data mining - Selection and evaluation of single features - Classification accuracy}}, but with the mean membership value resp. probability for the
true classification decision as evaluation measure (see e.g. Eq. (3.4) in \cite{Reischl06}).

Example:

Data point 1: true class 1, gradual decision with membership degree 0.9 for class 1

Data point 2: true class 2, gradual decision with membership degree 0.6 for class 2

Evaluation  = (0.9+0.6)/2 = 0.75 (instead of 1.0 using \hyperlink{MI_EMAusw_Klassifikation}{\emph{Data mining - Selection and evaluation of single features - Classification accuracy}})
\hypertarget{MI_EMAusw_Fuzzy}{}
\item {\bf Fuzzy classification accuracy (univariate)}: \newline  
like \hyperlink{MI_EMAusw_Fuzzy}{\emph{Data mining - Selection and evaluation of single features - Fuzzy classification accuracy}}, but only for the selection of one feature.
\hypertarget{MI_Merkausklassuniunsch}{}
\item {\bf Weighted fuzzy classification accuracy}: \newline  
like \hyperlink{MI_EMAusw_Klassifikation}{\emph{Data mining - Selection and evaluation of single features - Classification accuracy}}, but with the weighted mean membership value resp. probability
for the true classification decision as evaluation measure (see e.g. Eq. (3.6) and (3.7) in
\cite{Reischl06}). The weighting leads to a higher preference of true classification results in
comparison to \hyperlink{MI_EMAusw_Fuzzy}{\emph{Data mining - Selection and evaluation of single features - Fuzzy classification accuracy}}.
\hypertarget{MI_EMAusw_ModFuzzy}{}
\item {\bf Weighted fuzzy classification accuracy (univariate)}: \newline  
like \hyperlink{MI_EMAusw_ModFuzzy}{\emph{Data mining - Selection and evaluation of single features - Weighted fuzzy classification accuracy}}, but only for the selection of one feature.
\hypertarget{MI_Merkausklassunigewunsch}{}
\item {\bf Compute T-test (with Statistics Toolbox!)}: \newline  
tests the selected single features for statistical relevant difference between all pairs of
linguistic terms of the selected output variable and shows the results. Only selected data points
are used. A method for the correction of multiple tests can be defined by \hyperlink{CE_Corr_Bonferroni}{\emph{Control element: Data mining: Statistical options - Bonferroni correction}}.
The Statistical Toolbox for Matlab is required for this function.
\hypertarget{MI_TTest}{}
\item {\bf Paired T-Test (with Statistics Toolbox!)}: \newline  
similar to \hyperlink{MI_TTest}{\emph{Data mining - Selection and evaluation of single features - Compute T-test (with Statistics Toolbox!)}}, but performs a paired test between two classes expressed by linguistic
terms of an output variable. Therefor, for each data point belonging to a linguistic term a unique
relationship to a particular data point of the other linguistic term is assumed. This relation is
defined by identical values for all other output variables (e.g. patient ID,...). A method for the correction of multiple tests can be defined by \hyperlink{CE_Corr_Bonferroni}{\emph{Control element: Data mining: Statistical options - Bonferroni correction}}.

Typical examples are pre-therapeutic and post-therapeutic data of a collective of patients. For the
test, exactly one pre-therapeutic and one post-therapeutic data point must exist for each included
patient. In general, this pairwise relation is not given for all data points of a project (e.g.
containing healthy subjects without post-therapeutic data points). For a valid test, it can be
generated by an appropriate selection of all patient data points which have to be included.
\hypertarget{MI_Paired_TTest}{}
\item {\bf Compute Wilcoxon ranksum test (with Statistics Toolbox!)}: \newline  
similar to \hyperlink{MI_TTest}{\emph{Data mining - Selection and evaluation of single features - Compute T-test (with Statistics Toolbox!)}}, but performs a non-parametric \textindex{Wilcoxon-Ranksum-Test}. The test results are also valid for data with other distributions beyond normal distributions. A method for the correction of multiple tests can be defined by \hyperlink{CE_Corr_Bonferroni}{\emph{Control element: Data mining: Statistical options - Bonferroni correction}}.
\hypertarget{MI_Wilcoxon}{}
\item {\bf Test of normal distribution}: \newline  
tests the selected single features to normal distribution and protocols the result, for all selected data points and for each single class of the selected output variable. Here, the test defined by \hyperlink{CE_Normtest}{\emph{Control element: Data mining: Statistical options - Test of normal distribution}}
is applied. The Statistical Toolbox for Matlab is required for this function.
\hypertarget{MI_NormTest}{}
\item {\bf Chi-Square test output variable vs. discrete-valued single feature}: \newline  
like \hyperlink{MI_ChiSquareCrosstabAllOut}{\emph{Data mining - Evaluation of output variables - Chi-Square test for contingency tables of output variables}}, but for the selected output variable and the selected single features. The latter one are assumed to have discrete values.
\hypertarget{MI_ChiSquareCrosstabOutSingle}{}
\item {\bf Linear regression coefficients (univariate)}: \newline  
absolute values for univariate linear regression coefficients (square root of the coefficient of determination $R^2$). They deal with the prognosis of the single feature defined by
\hyperlink{CE_Regression_Output}{\emph{Control element: Data mining: Regression - Output variable of regression}}. The input variables are all single features except the output variable and all previous estimations.

Univariate relevances ignore correlations between different input variables.

A better combination might be found by \hyperlink{MI_EMAusw_RegrMulti}{\emph{Data mining - Selection and evaluation of single features - Linear regression coefficients (multivariate)}}. If \hyperlink{CE_Regression_Preselection}{\emph{Control element: Data mining: Regression - Preselection of features}} is marked, all selected input variables are preselected as first features.

The computed relevances can be shown e.g. by
\hyperlink{MI_Anzeige_EM_Relevanzen}{\emph{View - Single features - Show feature relevances (sorted table)}}.
\hypertarget{MI_EMAusw_RegrUni}{}
\item {\bf Linear regression coefficients (multivariate)}: \newline  
like \hyperlink{MI_EMAusw_RegrUni}{\emph{Data mining - Selection and evaluation of single features - Linear regression coefficients (univariate)}}, but with multivariate instead of univariate regression coefficients.
\hypertarget{MI_EMAusw_RegrMulti}{}
\item {\bf Regression accuracy (univariate)}: \newline  
like \hyperlink{MI_EMAusw_RegrUni}{\emph{Data mining - Selection and evaluation of single features - Linear regression coefficients (univariate)}}, but with complete regression models defined by \hyperlink{CEO_DM_Regr}{\emph{Control elements:  Data mining: Regression}} instead of univariate regression coefficients. The evaluation measure is the regression coefficient between the true value and the estimation generated by the regression model.
\hypertarget{MI_EMAusw_RegKomplUni}{}
\item {\bf Regression accuracy (multivariate)}: \newline  
like \hyperlink{MI_EMAusw_RegrUni}{\emph{Data mining - Selection and evaluation of single features - Linear regression coefficients (univariate)}}, but with regression models with multiple inputs.
\hypertarget{MI_EMAusw_RegKomplMulti}{}
\item {\bf Regression accuracy via improvement of the mean error (univariate)}: \newline  
evaluates the fitness of a regression model via the relative improvement of the mean absolute error in contrast to the trivial model resulting from the mean value of the output variable.
\hypertarget{MI_EMAusw_RegKomplUniMeanErr}{}
\item {\bf Regression accuracy via improvement of the mean error (multivariate)}: \newline  
like \hyperlink{MI_EMAusw_RegKomplUniMeanErr}{\emph{Data mining - Selection and evaluation of single features - Regression accuracy via improvement of the mean error (univariate)}}, but with regression models with multiple inputs.
\hypertarget{MI_EMAusw_RegKomplMultiMeanErr}{}
\end{itemize}
\subsection{Evaluation of time series}
\hypertarget{MI_Bewertung_ZR}{}
contains all functions for an automatic evaluation of time series.
\begin{itemize}
\item {\bf ANOVA (with sample point for Time series $\rightarrow$ Single feature)}: \newline  
computes for all time series a relevance using a univariate variance analysis \textindex{ANOVA} and
selects the best time series. The analysis is only performed for the sampling point chosen by
\hyperlink{CE_ZR_EM_Abtastpunkt}{\emph{Control element: Time series: Extraction - Parameters - Sample point for Time series$\rightarrow$Single feature}}. The number of selected time series is defined by
\hyperlink{CE_Anzahl_Merkmale}{\emph{Control element: Data mining: Classification of single features - Number of selected features}}.
\hypertarget{MI_ZRAusw_ANOVA}{}
\item {\bf MANOVA (with sample point for Time series $\rightarrow$ Single feature)}: \newline  
see \hyperlink{MI_ZRAusw_ANOVA}{\emph{Data mining - Evaluation of time series - ANOVA (with sample point for Time series $\rightarrow$ Single feature)}} but using a multivariate analysis of variances (\textindex{MANOVA}).
\hypertarget{MI_ZRAusw_MANOVA}{}
\item {\bf Compute best sample points}: \newline  
contains all functions to search for the best sample points of a time series for a subsequent
\textindex{feature selection}.
\hypertarget{MI_BestAbtast}{}
\item {\bf Feature maps (all time series, selected data points)}: \newline  
computes feature relevances of all time series and shows the results as scatter plot (x axis: Time,
y axis: number of time series, color: feature relevance) \cite{Burmeister06Dort}. The resulting
so-called \textindex{feature map} visualizes the contained information to predict the selected
output variable from the selected data points using different methods.
\hypertarget{MI_MKarten}{}
\item {\bf Linear regression coefficients (univariate)}: \newline  
computes absolute values for univariate linear regression coefficients (square root of the coefficient of determination $R^2)$. They deal with the prognosis of the time series defined
by \hyperlink{CE_Regression_Output}{\emph{Control element: Data mining: Regression - Output variable of regression}}. The input variables are past ($[k-i]$) and/or present $[k]$ sample points of
all or selected time series defined by \hyperlink{CE_Regression_Merkmalsauswahl}{\emph{Control element: Data mining: Regression - Feature selection}} resp.
\hyperlink{CE_Regression_Abtastpunkte}{\emph{Control element: Data mining: Regression - Sample points}}. Univariate relevances ignore correlations between different input
variables.

A better combination might be found by \hyperlink{MI_ZRAusw_RegrMulti}{\emph{Data mining - Evaluation of time series - Linear regression coefficients (multivariate)}}. The value of \hyperlink{CE_Regression_Preselection}{\emph{Control element: Data mining: Regression - Preselection of features}} is ignored, no preselection is done.

The computed relevances can be
shown e.g. by \hyperlink{MI_Anzeige_ZRRelevanz}{\emph{View - Time series (TS) - Show all feature relevances (sorted table)}}.

Only one data point can be selected.
\hypertarget{MI_ZRAusw_RegrUni}{}
\item {\bf Linear regression coefficients (multivariate)}: \newline  
like \hyperlink{MI_ZRAusw_RegrUni}{\emph{Data mining - Evaluation of time series - Linear regression coefficients (univariate)}}, but with multivariate instead of univariate regression coefficients.
\hypertarget{MI_ZRAusw_RegrMulti}{}
\end{itemize}
\subsection{Evaluation of output variables}
\hypertarget{MI_Bewertung_Output}{}
contains functions for the evaluation of output variables.
\begin{itemize}
\item {\bf Chi-Square test for contingency tables of output variables}: \newline  
performs a Pearson Chi-Square-Test (using the function ''crosstab'' of the MATLAB Statistics Toolbox) that the output variables are pairwise independent (null hypothesis). The p-value indicates that this null hypothesis is rejected by chance. The results are valid for sample size >5 data points in each cell of the cross tabulation. Otherwise for cross tabulations with only two values 0 and 1, the (more conservative) Two-tail Exact Fisher Test is used in the implementation of \cite{Trujillo-Ortiz04} (function ''fisherextest''). A method for the correction of multiple tests can be defined by \hyperlink{CE_Corr_Bonferroni}{\emph{Control element: Data mining: Statistical options - Bonferroni correction}}.
\hypertarget{MI_ChiSquareCrosstabAllOut}{}
\end{itemize}
\subsection{Classification}
\hypertarget{MI_Anzeige_Klassifikation}{}
contains all functions for a complete classification run based on single features (incl. feature
selection and aggregation).
\begin{itemize}
\item {\bf Design}: \newline  
trains a classifier based on the selected data points. The output variable for the supervised
classification is defined by \hyperlink{CE_Auswahl_Ausgangsgroesse}{\emph{Control element: Time series: General options - Selection of output variable}} (also found in other windows, see
e.g. \hyperlink{CEO_DM_Grund}{\emph{Control elements:  Data mining: Classification of single features}}). This run includes all steps defined by \hyperlink{CEO_DM_Grund}{\emph{Control elements:  Data mining: Classification of single features}} (optional
\textindex{feature selection}, normalization, \textindex{feature aggregation}, normalization of
aggregated features, and classification itself). The run is defined by all parameters from
\hyperlink{CEO_DM_Grund}{\emph{Control elements:  Data mining: Classification of single features}} and \hyperlink{CEO_DM_Klass}{\emph{Control elements:  Data mining: Special methods}}.
\hypertarget{MI_EMKlassi_En}{}
\item {\bf Apply}: \newline  
applies a classifier design by \hyperlink{MI_EMKlassi_En}{\emph{Data mining - Classification - Design}} to all selected data points. This application
includes all designed steps for \textindex{feature selection}, normalization, \textindex{feature
aggregation}, normalization of aggregated features, and the classification itself.
\hypertarget{MI_EMKlassi_An}{}
\item {\bf Design and apply}: \newline  
performs a subsequent processing of \hyperlink{MI_EMKlassi_En}{\emph{Data mining - Classification - Design}} and \hyperlink{MI_EMKlassi_An}{\emph{Data mining - Classification - Apply}}.
\hypertarget{MI_EMKlassi_EnAn}{}
\end{itemize}
\subsection{Time series classification}
\hypertarget{MI_CV_ZR_Standard}{}
contains all functions to classify time series.
\begin{itemize}
\item {\bf Design}: \newline  
designs a \textindex{time series classifier} using the parameters in \hyperlink{CEO_DM_ZRKlass}{\emph{Control elements:  Data mining: Classification of time series}}. The
different classifier types are described in \cite{Burmeister06}. The detailed parameters for the
classifiers are defined by \hyperlink{CEO_DM_Klass}{\emph{Control elements:  Data mining: Special methods}}.
\hypertarget{MI_ZRKlassi_En}{}
\item {\bf Apply}: \newline  
applies a classifier designed by \hyperlink{MI_ZRKlassi_En}{\emph{Data mining - Time series classification - Design}} to the selected data points.
\hypertarget{MI_ZRKlassi_An}{}
\item {\bf Design and apply}: \newline  
performs a subsequent processing of \hyperlink{MI_ZRKlassi_En}{\emph{Data mining - Time series classification - Design}} and \hyperlink{MI_ZRKlassi_An}{\emph{Data mining - Time series classification - Apply}}.
\hypertarget{MI_ZRKlassi_EnAn}{}
\item {\bf Time aggregation}: \newline  
aggregates classifier decisions over the time using the filter parameters in
(\hyperlink{CE_ZRKlassi_Filter}{\emph{Control element: Data mining: Classification of time series - Filtering of results}}) resulting in modified classifier decisions.
\hypertarget{MI_ZRKlassi_ZeitlAggr}{}
\item {\bf Time aggregation with result plot}: \newline  
like \hyperlink{MI_ZRKlassi_ZeitlAggr}{\emph{Data mining - Time series classification - Time aggregation}}, but with an additional plot of the results.
\hypertarget{MI_ZRKlassi_ZeitlAggrPlot}{}
\end{itemize}
\subsection{Hierarchical Bayes classifier}
\hypertarget{MI_Hierar_Klassi}{}
contains functions for the design and application of hierarchical Bayes classifiers.
\begin{itemize}
\item {\bf Design}: \newline  
designs a \textindex{Hierarchical Bayes classifier} by a step-wise separation of single classes
from all other classes \cite{Reischl06}. The maximum number of selected and aggregated features are
defined by (\hyperlink{CE_Anzahl_Merkmale}{\emph{Control element: Data mining: Classification of single features - Number of selected features}}) resp. \hyperlink{CE_Anzahl_Aggregiert}{\emph{Control element: Data mining: Classification of single features - Number of aggregated features}}. The feature aggregation
is done by a \textindex{Discriminant Analysis with optimization}. This function always uses a
\textindex{Bayes classifier}.
\hypertarget{MI_EMHierchKlassi_En}{}
\item {\bf Apply}: \newline  
applies a classifier designed  by \hyperlink{MI_EMHierchKlassi_En}{\emph{Data mining - Hierarchical Bayes classifier - Design}} to all selected data points.
\hypertarget{MI_EMHierchKlassi_An}{}
\item {\bf Design and apply}: \newline  
performs a subsequent processing of \hyperlink{MI_EMHierchKlassi_En}{\emph{Data mining - Hierarchical Bayes classifier - Design}} and \hyperlink{MI_EMHierchKlassi_An}{\emph{Data mining - Hierarchical Bayes classifier - Apply}}.
\hypertarget{MI_EMHierchKlassi_EnAn}{}
\end{itemize}
\subsection{Fuzzy systems}
\hypertarget{MI_Ansicht_Fuzzy}{}
contain all functions for the design and application of fuzzy systems. The design algorithms are
described in \cite{Mikut05}.
\begin{itemize}
\item {\bf Design (single rules)}: \newline  
searches for relevant fuzzy rules using only the selected single features and the selected output
variable. This design does not consider the cooperation of the designed rules in a rule base (e.g.
to avoid redundancies). Preexisting rules will be deleted.
\hypertarget{MI_Fuzzy_Einzelregel}{}
\item {\bf Design (rule base)}: \newline  
searches for a relevant fuzzy rule base using only the selected single features and the selected
output variable. This design searches in first step for single rules and collects a small set of
cooperating rules by an optimization. The evaluation measure prefers rules without redundancies and
a complete coverage of the input space. Preexisting rules will be deleted.
\hypertarget{MI_Fuzzy_Basis}{}
\item {\bf Delete rules}: \newline  
open a window to delete fuzzy rules manually.
\hypertarget{MI_Fuzzy_Loeschen}{}
\item {\bf Design membership functions for single features}: \newline  
designs membership functions for all single features using the method specified by \hyperlink{CE_Fuzzy_TypeZGF}{\emph{Control element: Data mining: Special methods - Type of membership function}}.
\hypertarget{MI_Fuzzy_DesignMBF}{}
\item {\bf Export fuzzy system to classifier}: \newline  
exports a designed fuzzy system as \textindex{classifier}. A design by \hyperlink{MI_Fuzzy_Basis}{\emph{Data mining - Fuzzy systems - Design (rule base)}}
corresponds to a classifier design using the selected features without an aggregation. A design by
\hyperlink{MI_Fuzzy_Einzelregel}{\emph{Data mining - Fuzzy systems - Design (single rules)}} is also possible, but it leads in many cases to bad results due to
redundant rules and missing priorities of good single rules.
\hypertarget{MI_FUZZY2KLASS}{}
\item {\bf Import fuzzy system from classifier}: \newline  
imports a fuzzy system from a designed fuzzy classifier. This function is only available if
''\textindex{fuzzy classifier}'' was chosen in \hyperlink{CE_ZRKlassi_Typ}{\emph{Control element: Data mining: Classification of time series - Classifier type}} during last design. It is
useful to list and to visualize fuzzy rules from classifiers (see also \hyperlink{MI_Ansicht_Fuzzy}{\emph{View - Fuzzy systems}}).
\hypertarget{MI_KLASS2FUZZY}{}
\item {\bf Import fuzzy system from regression model}: \newline  
imports a fuzzy system from a regression model designed by \hyperlink{MI_Regression_Entwurf}{\emph{Data mining - Regression - Design}} resp. \hyperlink{MI_Regression_EnAn}{\emph{Data mining - Regression - Design and apply}}. Before designing, ''Fuzzy system'' has to be selected in \hyperlink{CE_Regression_Typ}{\emph{Control element: Data mining: Regression - Type}}.
\hypertarget{MI_REGR2FUZZY}{}
\end{itemize}
\subsection{Clustering}
\hypertarget{MI_Ansicht_Clustern}{}
contains all functions for the design and application of clusters.
\begin{itemize}
\item {\bf Design and apply}: \newline  
contains different methods to search for subgroups with \textindex{fuzzy cluster methods} (FCM) for
the selected time series and resp. or single features (see e.g. \cite{Loose04}). The design methods
use only the selected data points. The application for the cluster assignment is done for all data
points. The necessary parameters and methods as well as the options for the conversion of clusters
into new output variables are defined in \hyperlink{CEO_DM_Clustern}{\emph{Control elements:  Data mining: Clustering}}. Here, own clustering functions are
applied.
\hypertarget{MI_Cluster_Ber}{}
\item {\bf Design and apply (Statistic Toolbox)}: \newline  
computes clusters for the selected single features and resp. or time series. The necessary
parameters and methods as well as the options for the conversion of clusters into new output
variables are defined in \hyperlink{CEO_DM_Clustern}{\emph{Control elements:  Data mining: Clustering}}. In contrast to \hyperlink{MI_Cluster_Ber}{\emph{Data mining - Clustering - Design and apply}}, the crisp
and \textindex{hierarchical cluster algorithms} of the \textindex{Statistic Toolbox} of Matlab
(e.g. ''pdist'', ''linkage'' and ''cluster'')  are used. A detailed description can be found in the
handbook of this toolbox.
\hypertarget{MI_Cluster_matlab}{}
\end{itemize}
\subsection{Association analysis}
\hypertarget{MI_Association_Rules}{}
performs an \textindex{Association analysis} for all output variables. Here, an implementation of Narine Manukyan was integrated. The results are saved as file. If this analysis should include single features, these single features have to be converted to output variables by \hyperlink{MI_EM_Klasse}{\emph{Edit - Convert - Selected single features $\rightarrow$ Output variables}}.

Output variables with too many linguistic terms will be ignored, the maximum of term numbers is defined by \hyperlink{CE_AssociationIgnoreFrequent}{\emph{Control element: Data mining: Special methods - Ignore output variables with many terms}}.
Only rules with a minimum confidence and support are shown (Definition in \hyperlink{CE_AssociationMinKonf}{\emph{Control element: Data mining: Special methods - Minimum confidence}} resp. \hyperlink{CE_AssociationMinSupp}{\emph{Control element: Data mining: Special methods - Minimum support}}).
\subsection{Self Organizing Maps}
\hypertarget{MI_Som}{}
contains the functions for the design and application of \textindex{Self Organizing Maps}. Here, the function selforgmap.m from MATLAB \textindex{Neural Network Toolbox} is used.
\begin{itemize}
\item {\bf Design}: \newline  
designs \textindex{Self Organizing Maps} for all selected single features with the function selforgmap.m from MATLAB \textindex{Neural Network Toolbox}. The dimensionality is defined by \hyperlink{CE_SomDimension}{\emph{Control element: Data mining: Special methods - Dimension}}, the number of neurons per dimension by \hyperlink{CE_SomNeurons}{\emph{Control element: Data mining: Special methods - Number of neurons}} and the number of training epochs by \hyperlink{CE_MLP_Lernepochen}{\emph{Control element: Data mining: Special methods - Number of learning epochs}}.
\hypertarget{MI_SomEn}{}
\item {\bf Apply}: \newline  
applies a Self Organizing Map designed by \hyperlink{MI_SomEn}{\emph{Data mining - Self Organizing Maps - Design}} to the selected data points. A new output variable with the number of winner neurons in the terms is generated.
\hypertarget{MI_SomAn}{}
\end{itemize}
\subsection{Regression}
\hypertarget{MI_CV_RegrEM_Standard}{}
contains all functions for the design and application of regression models.
\begin{itemize}
\item {\bf Design}: \newline  
designs a \textindex{regression model} using the selected data points and the parameters chosen in
\hyperlink{CEO_DM_Regr}{\emph{Control elements:  Data mining: Regression}}.

The related options (regression of time series or single features, feature selection and aggregation,
normalization, regression method) are chosen in \hyperlink{CEO_DM_Regr}{\emph{Control elements:  Data mining: Regression}}. The methods can be selected and
parameterized using \hyperlink{CEO_DM_Klass}{\emph{Control elements:  Data mining: Special methods}}.
\hypertarget{MI_Regression_Entwurf}{}
\item {\bf Apply}: \newline  
applies the recent \textindex{regression model} to the selected data points. The recent model
results from the last design step or a loaded regression model.

Hereby, all related operations (regression of time series or single features, feature selection, and
aggregation, normalization, regression method) are executed.
\hypertarget{MI_Regression_Anwendung}{}
\item {\bf Design and apply}: \newline  
performs a subsequent processing of \hyperlink{MI_Regression_Entwurf}{\emph{Data mining - Regression - Design}} and \hyperlink{MI_Regression_Anwendung}{\emph{Data mining - Regression - Apply}}.
\hypertarget{MI_Regression_EnAn}{}
\end{itemize}
\subsection{Validation}
\hypertarget{MI_Validierung}{}
validates the \textindex{classification} of single features and time series.
\begin{itemize}
\item {\bf Classification of single features}: \newline  
validates classifiers. Possible \textindex{validation strategies} are a
\textindex{cross-validation} (with diverse subtypes) and a \textindex{bootstrap} method (selection
by \hyperlink{CE_CV_Typ}{\emph{Control element: Data mining: Validation - Validation strategy}}). The menu item \hyperlink{MI_EMKlassi_En}{\emph{Data mining - Classification - Design}} is performed for each produced training
data set using the defined options. The resulting classifier is applied for each validation data
set by means of \hyperlink{MI_EMKlassi_An}{\emph{Data mining - Classification - Apply}}. The results are aggregated and written into protocol files.
\hypertarget{MI_CV_EM_Standard}{}
\item {\bf Classification of single features (selected macros)}: \newline  
similar to \hyperlink{MI_CV_EM_Standard}{\emph{Data mining - Validation - Classification of single features}} but with macros for classifier design and application instead
of the corresponding menu items. The macros are defined by a configuration window or by the values
of the variables ''makro\_lern'' (design) resp. ''makro\_test'' (application) for a script-based
technique.
\hypertarget{MI_CV_EM}{}
\item {\bf Hierarchical Bayes Classifier for single features}: \newline  
similar to \hyperlink{MI_CV_EM_Standard}{\emph{Data mining - Validation - Classification of single features}} but for Hierarchical Bayes classifiers designed by \hyperlink{MI_EMHierchKlassi_En}{\emph{Data mining - Hierarchical Bayes classifier - Design}}.
\hypertarget{MI_CV_EMHier_Standard}{}
\item {\bf Time series classification}: \newline  
see \hyperlink{MI_CV_ZR}{\emph{Data mining - Validation - Time series classification (selected macros)}}. Here, the design and application of the time series classifiers are executed.
A selection of user-defined macros is not possible.
\hypertarget{MI_CV_ZR_Standard}{}
\item {\bf Time series classification (selected macros)}: \newline  
validates time series classifiers. Two macros are executed for the classifier design and its
application. Both macros are selected by a configuration window. Different types of
\textindex{cross-validation} and \textindex{bootstrap} are available as \textindex{validation
strategies}. The parameters are tuned by \hyperlink{CEO_DM_Vali}{\emph{Control elements:  Data mining: Validation}}. The results are separately computed
for each trial. In addition, the mean value and the standard deviation for all trials are shown.

For the cross-validation of time series, a variable \verb=zr_fehl_proz= is introduced in the struct
\verb=relevanz_cv_alle=. It contains the classification error vs. time. The mean value and the
standard deviation is defined by the minimum error of each trial.
\hypertarget{MI_CV_ZR}{}
\item {\bf Regression}: \newline  
see \hyperlink{MI_CV_RegrEM}{\emph{Data mining - Validation - Regression (own macros)}}. Here, the design and application of the regression models are executed using the standard macros ''regr\_em\_en.makrog'' (design) and ''regr\_em\_an.makrog'' (application).
A selection of user-defined macros is not possible.
\hypertarget{MI_CV_RegrEM_Standard}{}
\item {\bf Regression (own macros)}: \newline  
validates regression models. Two macros are executed for the regression model design and its
application. Each \textindex{macro} is selected by a configuration window. Different types of
\textindex{cross-validation} and \textindex{bootstrap} are available as \textindex{validation
strategies}. The parameters are tuned by \hyperlink{CEO_DM_Vali}{\emph{Control elements:  Data mining: Validation}}. The results (\textindex{regression error} and (Pearson) \textindex{correlation coefficient}) are separately computed
for each trial. In addition, the mean value and the standard deviation for all trials are shown.
\hypertarget{MI_CV_RegrEM}{}
\end{itemize}

\section{Menu items 'Extras'}
contains all functions to work with macros and for the administration of application-specific
extension packages.
\subsection{Play macro...}
\hypertarget{MI_Makro_Ausfuehren}{}
loads and runs a recorded \textmainindex{macro}. If an error message about a missing macro in the search path
occurs, possible reasons are deleted macros or errors during the last macro run. In latter case,
the use of \hyperlink{MI_Makro_Loeschen}{\emph{Extras - Reset macro names}} might help.
\subsection{Play macro (debug mode)...}
\hypertarget{MI_Makro_ExecuteMakroMFile}{}
executes a macro in debug mode. The selected macro is copied into a m-file called  ''makro\_m\_file.m'' and it is started. With this option, the complete debugging functionality MATLAB is available. After a new start, SciXMiner checks if ''makro\_m\_file.m'' was changed manually. If yes, it asks if these changes should be transferred also to the macro.
\subsection{Record macro...}
\hypertarget{MI_Makro_Simultan}{}
records a sequence of clicked menu items and control elements as \textindex{macro file}
\emph{*.makrog}. A manual modification of this file is possible due to its textual Matlab syntax.
The macro name is defined by an additional window. The background of the SciXMiner window is yellow
during recording. All menu items are executed in parallel. The recording is stopped by
\hyperlink{MI_Makro_Beenden}{\emph{Extras - Stop macro record}}. Some functions (especially file operations) are not available during macro
recording. They can be added manually by using the related callbacks from the file menu\_elements.m. A critical evaluation of the recorded macros is strongly recommended especially for
complex tasks.

A modified assignment of plotted figures (e.g. to plot multiple figures into subplots instead of
separate windows) is possible by \hyperlink{CE_Aktuelle_Figure}{\emph{Control element: General options - For macros: plot always in current figure}}. Such commands (e.g. subplot) should be
placed directly before the menu item in the macro to avoid a plot into wrong figures (especially in
the SciXMiner main window). \index{menu items}\index{control elements}
\subsection{Stop macro record}
\hypertarget{MI_Makro_Beenden}{}
terminates the record of a macro.
\subsection{Edit macro...}
\hypertarget{MI_Makro_Bearb}{}
opens an existing macro in an editor window for a manual modification.
\subsection{Reset macro names}
\hypertarget{MI_Makro_Loeschen}{}
resets the names of the recent macro in the Matlab workspace. This function is important after an
occurring error during the macro execution. Otherwise, problems during the execution of different
macros are possible.
\subsection{Execute M-file...}
\hypertarget{MI_MFile_Ausfuehren}{}
executes a M file.
\subsection{Edit M-file...}
\hypertarget{MI_MFile_Bearb}{}
opens an existing M-file in an editor window for a manual modification.
\subsection{Edit SciXMiner batch file ...}
\hypertarget{MI_GaitbatchBearb}{}
opens an existing SciXMiner batch file in an editor window for a manual modification.
\subsection{Generate PDF project report (needs  Latex)}
\hypertarget{MI_Extras_Autoreport}{}
generates PDF reports using Latex for the given project or all projects within a directory. Latex must be installed and the functions texify and dvipdfm must be found in the Windows search path.

The function generates a subdirectory ''report'' in the recent working directory. Here, all new or modified files for the \textindex{project report} are saved. The title page is generated from the Latex file ''reporttemplate.tex'', located in the directory ''standardmakros'' of SciXMiner. It can be modified for individual needs. Only for the first report, the template content is copied to a new file ''project\_results\_*.*'' in the ''report'' directory.  After this, it can be also modified to individual needs only for this project or directory.

The following contents are added if they exist:

Once per PDF:

1. File ''general\_comments.tex''. Here, all project related metadata can be added (e.g. experiment planning).

2. Error protocol ''error.log''.

3. All jpeg images located in a subdirectory ''images''.

Once per project:

1. All jpeg image generated by SciXMiner for a project.

2. All Latex files generated by SciXMiner for a project. Here, only the parts between the begin and the end of the document are used. Optional sections SciXMiner parameters are removed starting from the first   subsection\{Parameter\}.

All modified files  (and only these, not the images and unmodified Latex files) are written in the subdirectory ''report''. They should not be modified manually to an overwriting by the next generation of a project report.

\begin{itemize}
\item {\bf for the current project}: \newline  
generated the PDF project report only for the recent project.
\hypertarget{MI_Extras_AutoreportProjekt}{}
\item {\bf for all projects in the directory}: \newline  
generated the PDF project report for all projects in the recent directory. The different projects are included as sections. The name of the report is set to the name of the directory.
\hypertarget{MI_Extras_AutoreportPath}{}
\end{itemize}
\subsection{Translate German Gait-CAD m-files and macros into English}
\hypertarget{MI_Extras_TranslateGermanGaitCAD}{}
this menu item supports the translation of German Gait-CAD m-files and macros into English. It converts strings in all macros and m-files of a chosen directory using a dictionary. The related directory and a further directory for a safety copy are selected by a separate window.
\subsection{Choose application-specific extension packages...}
\hypertarget{MI_Extras_Erweiterungen}{}
switch packages with application-specific extensions on or off. This selection will be done after
the next SciXMiner start. At the moment, the extension packages ''Peptides'', ''\textindex{Images and Videos}'' and ''\textindex{Tracking}'' are available. The beta versions of the extension packages ''\textindex{Gait analysis}'' and ''\textindex{Text mining}'' are available on request.\index{gait analysis}\index{Application-specific extension packages}
\subsection{Search path for m-files and plugins}
\hypertarget{MI_Extra_AddSearchPath}{}
adds a directory to the MATLAB search path to use m-files and plugins in this directory in SciXMiner.
\begin{itemize}
\item {\bf Permanent}: \newline  
adds a directory permanently to the MATLAB search path to use m-files and plugins in this directory in SciXMiner. This information is saved in read\_gaitcad\_searchpath.m.
\hypertarget{MI_Extra_AddSearchPathPerm}{}
\item {\bf Temporary for the session}: \newline  
adds a directory for the recent SciXMiner session to the MATLAB search path to use m-files and plugins in this directory in SciXMiner.
\hypertarget{MI_Extra_AddSearchPathTemp}{}
\item {\bf Reset (permanent search path)}: \newline  
resets the permanent search path, that was added by \hyperlink{MI_Extra_AddSearchPathPerm}{\emph{Extras - Search path for m-files and plugins - Permanent}} and saved in read\_gaitcad\_searchpath.m.
\hypertarget{MI_Extra_AddSearchPathResetPerm}{}
\item {\bf Reset (only temporarily for the session)}: \newline  
resets the search path for the SciXMiner session, that was defined by \hyperlink{MI_Extra_AddSearchPathTemp}{\emph{Extras - Search path for m-files and plugins - Temporary for the session}}.
\hypertarget{MI_Extra_AddSearchPathResetTemp}{}
\end{itemize}
\subsection{Matlab Parallel}
\hypertarget{MI_Extra_Parallel}{}
controls the use of configurations of the \textindex{MATLAB Parallel Computing Toolbox}.
\begin{itemize}
\item {\bf Start}: \newline  
starts the use of the selected configuration of the \textindex{MATLAB Parallel Computing Toolbox}. Here, the configuration with the name defined by \hyperlink{CE_AllgParConfig}{\emph{Control element: General options - Configuration name for MATLAB parallel}} is used (see also Parallel - Manage Configurations in the menu of the MATLAB command window).
\hypertarget{MI_Extra_StartParallel}{}
\item {\bf Stop}: \newline  
stops the use of the recent configuration of the \textindex{MATLAB Parallel Computing Toolbox}.
\hypertarget{MI_Extra_StopParallel}{}
\end{itemize}

\section{Menu items 'Favorites'}
contains all functions for a fast access to frequently used or user-defined menu points and macros.
All deactivated entries (e.g. due to missing single features) are shown with a light gray color.
\subsection{Edit user-defined favorites}
\hypertarget{MI_Manuelle_Favoriten}{}
opens a configuration window to add and delete menu points from the favorite list.
\subsection{Delete all favorites}
\hypertarget{MI_Loesche_Favoriten}{}
deletes all favorites.

\section{Menu items 'Window'}
activates or closes all open Matlab figures except the SciXMiner main window.
\subsection{Close figures}
\hypertarget{MI_Schliessen}{}
closes all figure except the SciXMiner main window.
\subsection{Arrange figures}
\hypertarget{MI_Fenster_Anordnen}{}
contains different options for the placement of figures.
\begin{itemize}
\item {\bf Horizontal}: \newline  
places all figures horizontally.
\hypertarget{MI_Fenster_Anordnen_Horizontal}{}
\item {\bf Vertical}: \newline  
places all figures vertically.
\hypertarget{MI_Fenster_Anordnen_Vertikal}{}
\item {\bf Cascade}: \newline  
places all figures as a cascade.
\hypertarget{MI_Fenster_Anordnen_Kaskade}{}
\item {\bf Position of the current figure}: \newline  
places all figures to the position of the last opened figure. This command is especially useful for
the comparison of figures with small differences by switching between figures with the same size
and position.
\hypertarget{MI_Fenster_Anordnen_LetztesBild}{}
\end{itemize}
\subsection{Logarithmic scaling of current figures}
\hypertarget{MI_Fenster_Log}{}
applies a logarithmic scale to all open MATLAB figures.
\begin{itemize}
\item {\bf only x axis}: \newline  
applies a logarithmic scale to the x-axis of all open MATLAB figures.
\hypertarget{MI_Fenster_Logx}{}
\item {\bf only y-axis}: \newline  
applies a logarithmic scale to the y-axis of all open MATLAB figures.
\hypertarget{MI_Fenster_Logy}{}
\item {\bf x- and y axis}: \newline  
applies a logarithmic scale to the x- and the y-axis of all open MATLAB figures.
\hypertarget{MI_Fenster_Logxy}{}
\end{itemize}
\subsection{Remove Latex codes in MATLAB figures}
\hypertarget{MI_Fenster_TexReset}{}
removes Latex-style variables interpreted as equation (e.g. $x_1$ instead of x\_1) from all open MATLAB figures.
\subsection{Update font and font size in figures}
\hypertarget{MI_Fenster_FontNameSize}{}
sets font type and size in all open MATLAB figures to the values defined by \hyperlink{CE_AllgFontName}{\emph{Control element: View: Single features - Font}} and \hyperlink{CE_AllgFontSize}{\emph{Control element: View: Single features - Font size}}.
\subsection{Plot all figures as images in files}
\hypertarget{MI_Fenster_Datei}{}
plots the content of all figures in files. The filename is generated from the project and the figure name.
The file type is defined by \hyperlink{CE_Allgemein_FileTypeImage}{\emph{Control element: General options - File type for images}}. The image in the file depends on the size on
the monitor.

\section{Menu items 'Help'}
contains SciXMiner version and license information as well as SciXMiner handbooks.
\subsection{Show SciXMiner documentation (PDF)}
\hypertarget{MI_Hilfe_PDF}{}
opens the SciXMiner handbook and help file as PDF.
\subsection{About SciXMiner}
\hypertarget{MI_Hilfe_About}{}
contains the version information and the contact address to the developer team.
\subsection{License information}
\hypertarget{MI_Hilfe_Lizenz}{}
shows the license file for the GNU licence.
\chapter{Control elements}\label{sec:optionen}

\section{Control elements for 'Project overview'}

 \begin{figure}[h] 
 \includegraphics[width=0.9\columnwidth]{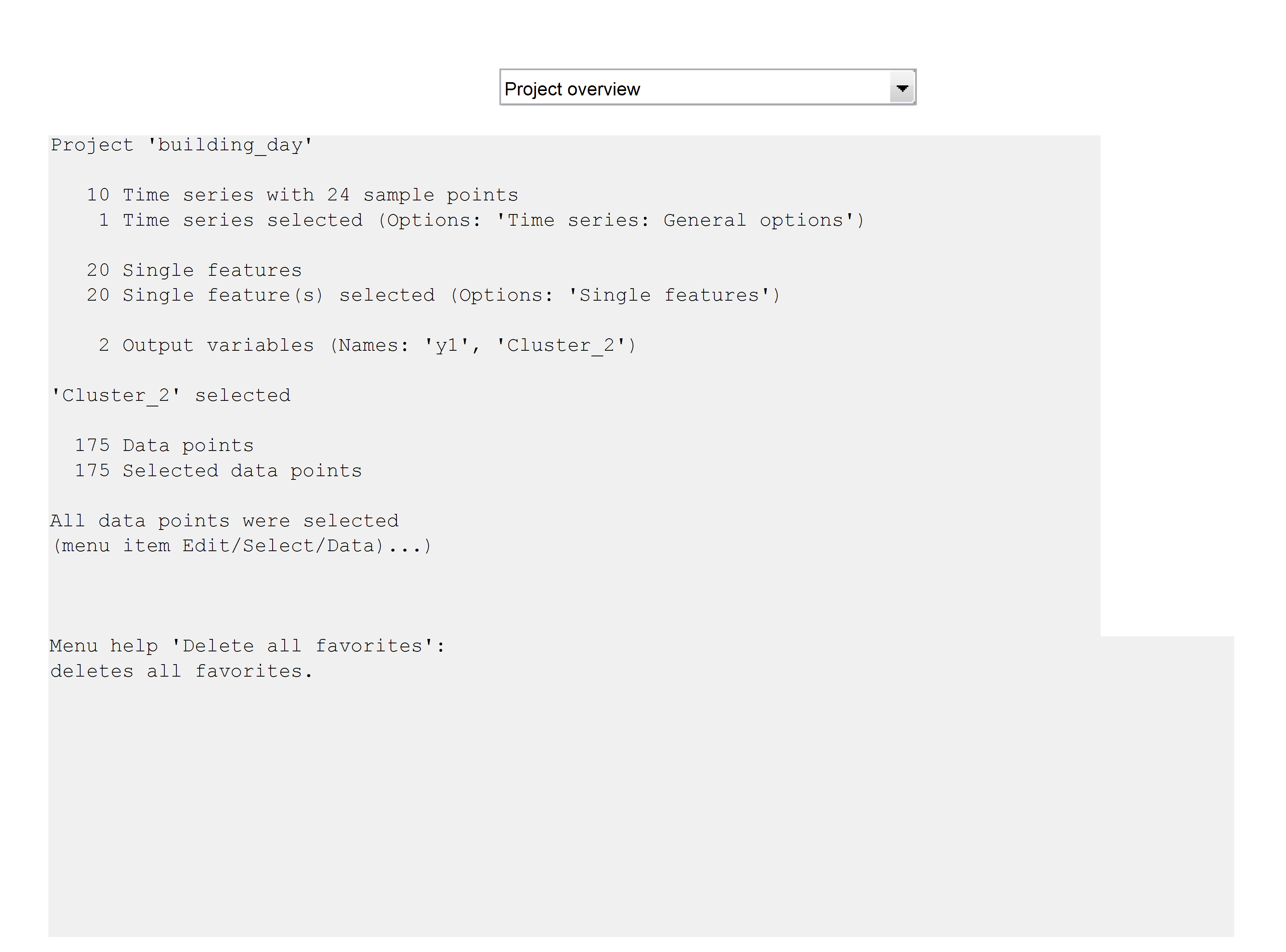}
 \caption{\label{fig:optionen1-2} Control elements for Project overview - Show menu tips}
 \end{figure}

\clearpage\clearpage

\section{Control elements for 'Time series: General options'}

 \begin{figure}[h] 
 \includegraphics[width=0.9\columnwidth]{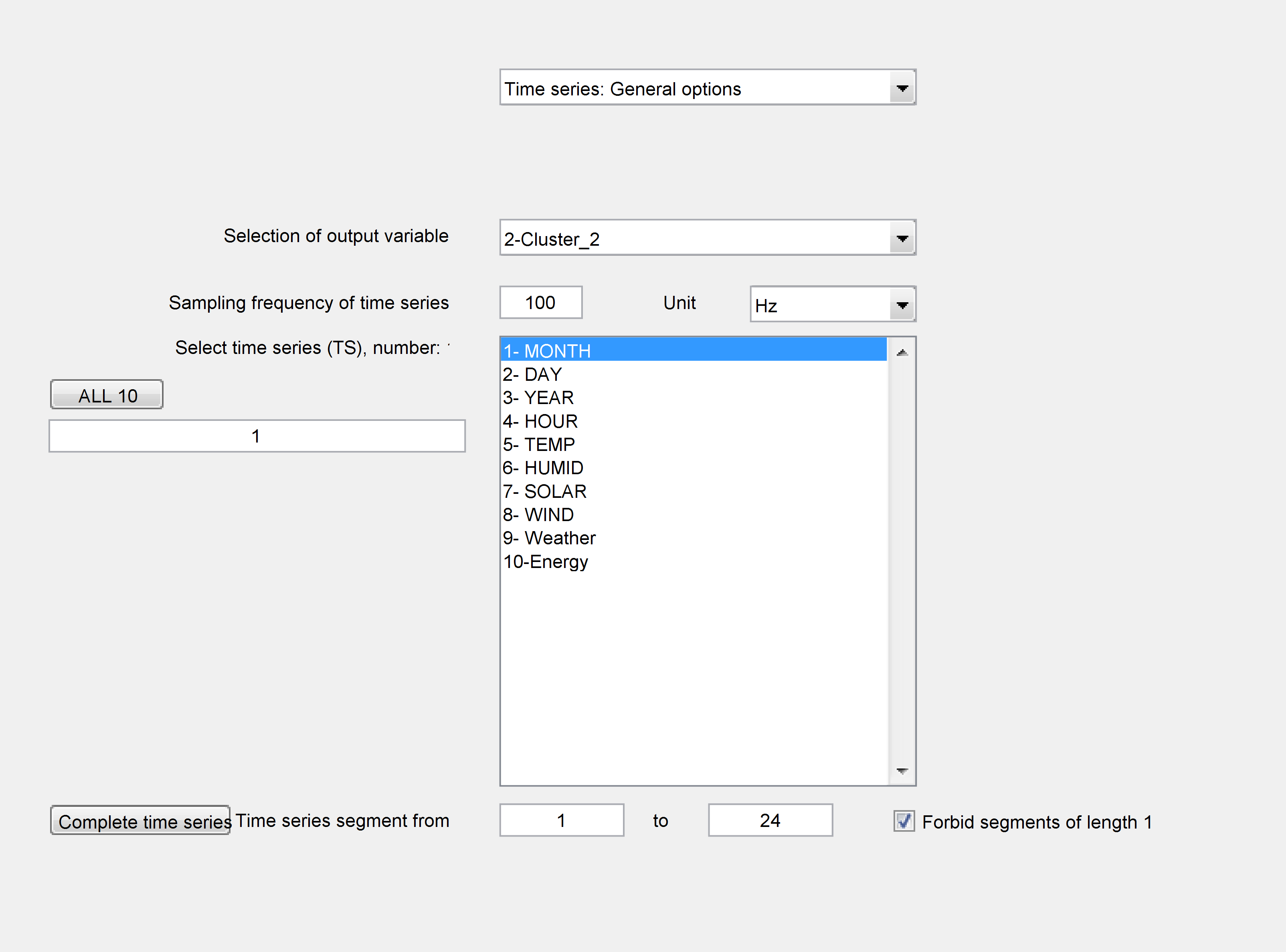}
 \caption{\label{fig:optionen2-3} Control elements for Time series: General options }
 \end{figure}
\begin{itemize}
\item {\bf Selection of output variable}: \hypertarget{CE_Auswahl_Ausgangsgroesse}{} \newline
selects an output variable. This selection influences many functions as the evaluation of single
features and time series, the design of classifiers, almost all visualization functions etc.
\item {\bf Sampling frequency of time series}: \hypertarget{CE_Zeitreihen_Abtastfreq}{} \newline
defines the sampling frequency for the time series in the data set. It is assumed that this value
is equal for all time series. The parameter mainly influences the visualization of time series using
time or frequency.
\item {\bf Unit}: \hypertarget{CE_Zeitreihen_Einheit_Abtastfreq}{} \newline
defines the physical quantity for time series (e.g. sampling frequency in kHz, per day etc.). The
parameter mainly influences the visualization of time series or spectrograms using time or
frequency.
\item {\bf Selection of time series (TS)}: \hypertarget{CE_Auswahl_ZR}{} \newline
selects time series for visualization and all following processing steps. The selection can be done
by mouse clicks in the listbox, by writing the numbers in the edit field on the left hand side or
by the \emph{'ALL'} button to select all time series. The different fields will be synchronized
after the input.

TIP: An input in the edit field enables a selection with a different order (e.g. 2-4-1 for a
visualization) in contrast to the listbox.
\item {\bf Complete time series}: \hypertarget{CE_Zeitreihen_Segment_Komplett}{} \newline
selects all sample points of the time series.
\item {\bf Time series segment from}: \hypertarget{CE_Zeitreihen_Segment_Start}{} \newline
defines the begin of a time series segment for the visualization of sample points. This parameter
might also defined by a zooming in a time series visualization (\hyperlink{MI_Zeitreihen}{\emph{View - Time series (TS)}}) and pushing
the button ''\textindex{Select time series segment}'' in the menu bar. The begin is the left limit
of the current x-axis.

The button ''Complete time series'' set the \textindex{time series segment} to all sample points.
\item {\bf to}: \hypertarget{CE_Zeitreihen_Segment_Ende}{} \newline
defines the end of a time series segment for the visualization of sample points. This parameter
might also defined by a zooming in a time series visualization (\hyperlink{MI_Zeitreihen}{\emph{View - Time series (TS)}}) and pushing
the button ''\textindex{Select time series segment}'' in the menu bar. The end is the right limit
of the current x-axis.

The button ''Complete time series'' set the \textindex{time series segment} to all sample points.
\item {\bf Forbid segments of length 1}: \hypertarget{CE_TimeSeriesForbidLength1}{} \newline
forbid time series segments of length 1. This option is almost always useful except some special cases for feature extraction.
\end{itemize}
\clearpage

\section{Control elements for 'Time series: Extraction - Parameters'}

 \begin{figure}[h] 
 \includegraphics[width=0.9\columnwidth]{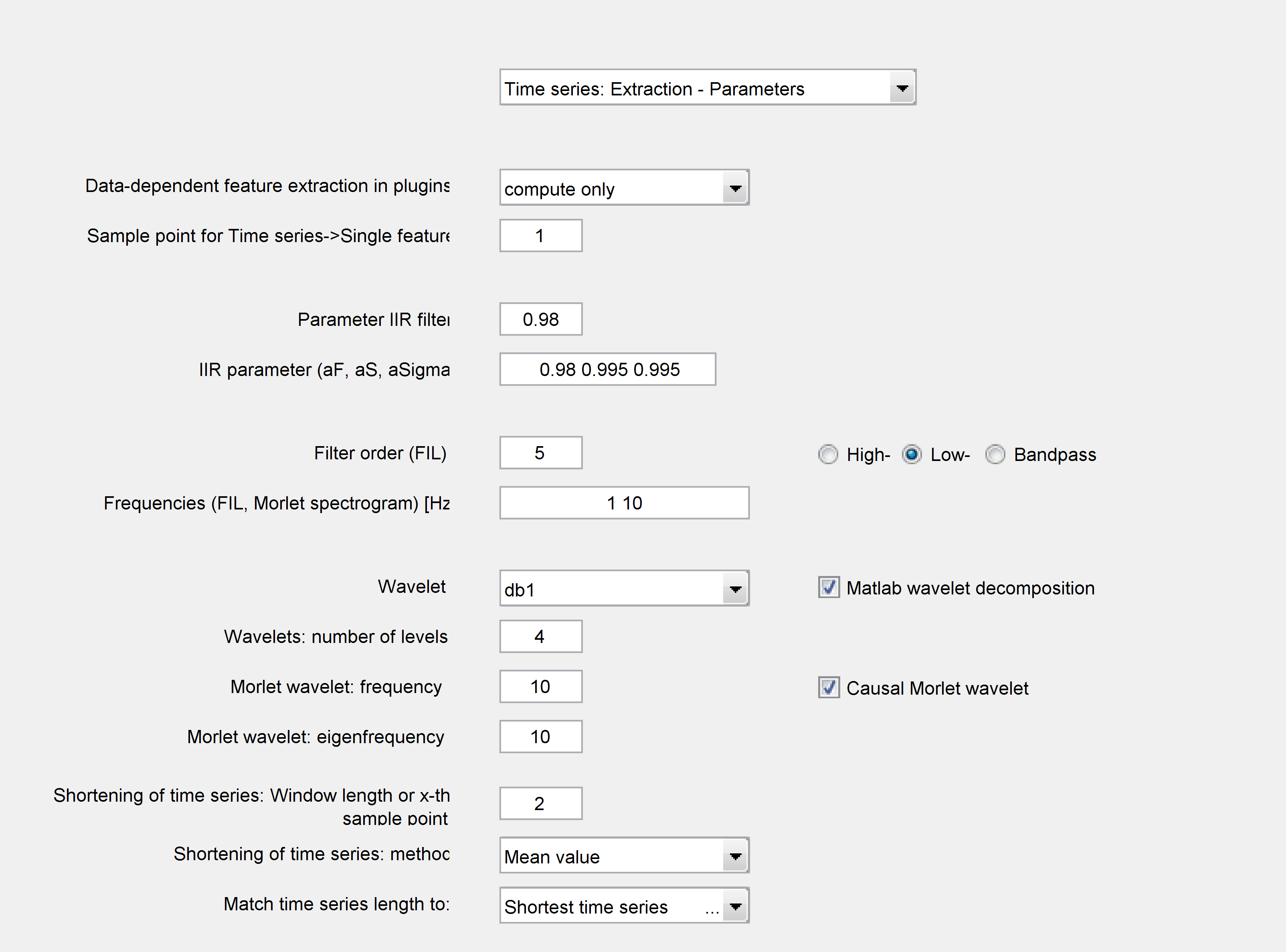}
 \caption{\label{fig:optionen3-4} Control elements for Time series: Extraction - Parameters }
 \end{figure}
\begin{itemize}
\item {\bf Data-dependent feature extraction in plugins}: \hypertarget{CE_FilePlugins}{} \newline
defines the handling of plugins with data-dependent transformation functions (and not only data-dependent
output variables for time series and single features!). Typical examples are a transformation matrix for a
\textindex{Principal Component Analysis} or fuzzy membership functions. Possible options are a computation
with and without saving the results into a *.plugpar file with the same project name or a loading from these
file. By loading, the name of the time series, the plugin description, and the segments must be the same. The
handling has to be separately implemented for each plugin!
\item {\bf Sample point for Time series$\rightarrow$Single feature}: \hypertarget{CE_ZR_EM_Abtastpunkt}{} \newline
select the sample point for all functions using only one sample point for a conversion from time
series to single features. \index{Time series $\rightarrow$ Single feature}
\item {\bf Parameter IIR filter}: \hypertarget{CE_Parameter_IIR}{} \newline
defines the filter parameter $a$ for an Infinite-Impulse-Response (IIR) filter. The filter is a
first-order lowpass with $x_f [k+1] = a * x_f [k]  + (1-a) * x [k]$. The parameter defines the
influence of old values. A higher value leads to a slower behavior due to a higher weighting of
past values. Some functions and \textindex{plugins} for the feature extraction use this parameter.
For stability reasons, the value is limited to $[0, 1]$. \index{IIR filter}
\item {\bf IIR parameter (aF, aS, aSigma)}: \hypertarget{CE_Parameter_aSaLaSigma}{} \newline
defines the parameters for three different \textindex{IIR filters} for trend and standard deviation
computation (see \cite{Reischl02}).
\item {\bf Filter order (FIL)}: \hypertarget{CE_ZR_Filterordnung}{} \newline
defines the order of the filter for feature extraction.
\item {\bf High-}: \hypertarget{CE_ZR_Filtertyp_Hoch}{} \newline
switches to a \textindex{high-pass filter} for feature extraction.
\item {\bf Low-}: \hypertarget{CE_ZR_Filtertyp_Tief}{} \newline
switches to a \textindex{low-pass filter} for feature extraction.
\item {\bf Bandpass}: \hypertarget{CE_ZR_Filtertyp_Band}{} \newline
switches to a \textindex{band-pass filter} for feature extraction.
\item {\bf Frequencies (FIL, Morlet spectrogram)}: \hypertarget{CE_ZR_Filterfreq}{} \newline
defines the \textindex{cutoff frequencies} for different filters. The parameter is used for a
\textindex{Butterworth filter} in feature extraction and for Morlet spectrograms. A
\textindex{high-pass filter} and a \textindex{low-pass filter} use only the first value, whereas the
\textindex{band-pass filter} uses both values. \index{Morlet spectrogram}
\item {\bf Wavelet}: \hypertarget{CE_Wavelets_Wavelet}{} \newline
selects the wavelet type for a \textindex{wavelet decomposition} (see e.g. \cite{Lipovei04} or the
documentation of the Matlab Wavelet Toolbox).
\item {\bf Matlab wavelet decomposition}: \hypertarget{CE_Wavelets_MatlabWavelet}{} \newline
uses the Matlab implementation instead of the implementation of \cite{Lipovei04} for the
\textindex{wavelet decomposition}. It requires the Matlab Wavelet Toolbox, but it is faster
especially in case of many data points and long time series.
\item {\bf Wavelets: number of levels}: \hypertarget{CE_Wavelets_Level}{} \newline
defines the number of levels for a \textindex{wavelet decomposition}. It computes for the half of
the sampling frequency $f_A$ an \textindex{high-pass filter} and a \textindex{low-pass filter}. The
high-pass in level $i$ has a cutoff frequency of ($i=1, \hdots$) with $$\frac{f_A}{2^{i+1}}.$$ The
cut-off frequency for the low-pass filter is computed accordingly.
\item {\bf Morlet wavelet: frequency}: \hypertarget{CE_MorletWavelet_Freq}{} \newline
defines the frequency for the generation of new time series by means of complex \textindex{Morlet
wavelets}. This frequency describes the center of the frequency range which will not be damped by
the wavelet. The width of this range is defined by the eigenfrequency of the Morlet wavelet
(\hyperlink{CE_MorletWavelet_Eigenfreq}{\emph{Control element: Time series: Extraction - Parameters - Morlet wavelet: eigenfrequency}}). For the computation of Morlet spectrograms, the parameter from
\hyperlink{CE_ZR_Filterfreq}{\emph{Control element: Time series: Extraction - Parameters - Frequencies (FIL, Morlet spectrogram)}} is used instead of this parameter.
\item {\bf Causal Morlet wavelet}: \hypertarget{CE_MorletWavelet_Kausal}{} \newline
temporal shift the complex Morlet wavelet for a causal filtering. In addition, the wavelet will be
limited in time.
\item {\bf Morlet wavelet: eigenfrequency}: \hypertarget{CE_MorletWavelet_Eigenfreq}{} \newline
see \hyperlink{CE_MorletWavelet_Freq}{\emph{Control element: Time series: Extraction - Parameters - Morlet wavelet: frequency}}.
\item {\bf Shortening of time series: Window length or x-th sample point}: \hypertarget{CE_Umwandeln_Fensterbreite}{} \newline
defines the window lengths (\hyperlink{MI_Kuerze_ZR_Fenster}{\emph{Edit - Convert - Shorten time series - Shortening by window methods}}) resp. the step width
(\hyperlink{MI_Kuerze_ZR_JederXte}{\emph{Edit - Convert - Shorten time series - Use only each x-th sample point}}) to \textindex{short time series}.
\item {\bf Shortening of time series: method}: \hypertarget{CE_Umwandeln_Verfahren}{} \newline
selects the method for shortening all time series in a project using
\hyperlink{CE_Umwandeln_Fensterbreite}{\emph{Control element: Time series: Extraction - Parameters - Shortening of time series: Window length or x-th sample point}}. First, a window of with a length defined by
\hyperlink{MI_Kuerze_ZR_Fenster}{\emph{Edit - Convert - Shorten time series - Shortening by window methods}} is chosen. Second, the new value for the short time series is computed
by the mean, median, minimum, or maximum value of these windows. The windows do not overlap.
\item {\bf Match time series length to}: \hypertarget{CE_Umwandeln_NormierteLaenge}{} \newline
defines the length for the shortening of the time series via \hyperlink{MI_Kuerze_ZR_AufKuerzeste}{\emph{Edit - Convert - Shorten time series - Match time series lengths (use 0 and NaN as undefined values)}}. The new length of the
time series is fixed (percent: 100, per thousand 1000) or the shortest time series in the data set is used. The length of the
shortest time series is defined by the sample after which the time series is only zero, Inf or NaN.
\end{itemize}
\clearpage

\section{Control elements for 'Plugin sequence'}

 \begin{figure}[h] 
 \includegraphics[width=0.9\columnwidth]{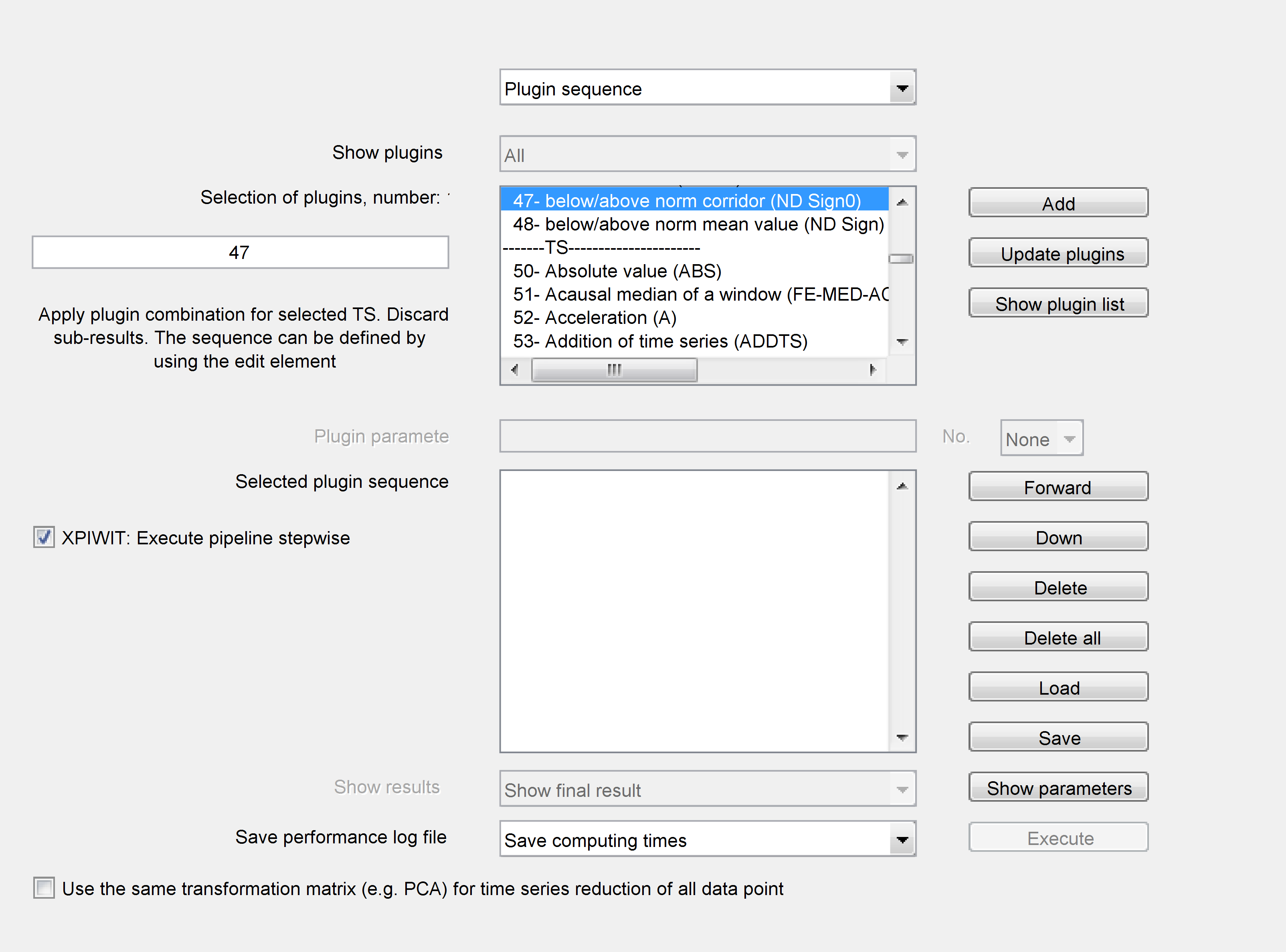}
 \caption{\label{fig:optionen4-5} Control elements for Plugin sequence }
 \end{figure}
\begin{itemize}
\item {\bf Show plugins}: \hypertarget{CE_Plugins_Anzeige}{} \newline
select the shown plugins depending on type (e.g. time series $\rightarrow$ time series).
\item {\bf Add}: \hypertarget{CE_PlugListAdd}{} \newline
adds the selected plugins from \hyperlink{CE_Auswahl_Plugins}{\emph{Control element: Plugin sequence - Selection of plugins}} to a plugin sequence.
\item {\bf Update plugins}: \hypertarget{CE_PlugListUpdate}{} \newline
reads the available plugins from all plugin files. It is useful for the development of plugins to avoid SciXMiner restarts.
\item {\bf Show plugin list}: \hypertarget{CE_PlugShowTxt}{} \newline
writes the characteristics and descriptions of all available plugins in a file and displays this file.
\item {\bf Selection of plugins}: \hypertarget{CE_Auswahl_Plugins}{} \newline
selects the plugins used by \hyperlink{MI_Extraktion_ZRZR_Kombi}{\emph{Edit - Extract - Time series $\rightarrow$ Time series, Time series $\rightarrow$ Single features (via plugin sequence)...}}. Only a definition in the edit window
guarantees the selected order of the plugins. A definition in the list box ignores this order and
sorts the plugins by their numbers.
\item {\bf Plugin parameter}: \hypertarget{CE_Auswahl_PluginsCommandLine}{} \newline
set the parameters of the selected plugin from \hyperlink{CE_Auswahl_Plugins}{\emph{Control element: Plugin sequence - Selection of plugins}} resp. \hyperlink{CE_Auswahl_PluginsList}{\emph{Control element: Plugin sequence - Selected plugin sequence}}. The switch between multiple parameters is done by \hyperlink{CE_Plugins_ParameterNumber}{\emph{Control element: Plugin sequence - No.}}.
\item {\bf No.}: \hypertarget{CE_Plugins_ParameterNumber}{} \newline
selects a specific parameter for plugins with multiple parameters.
\item {\bf Forward}: \hypertarget{CE_PlugListUp}{} \newline
moves the selected plugins in \hyperlink{CE_Auswahl_PluginsList}{\emph{Control element: Plugin sequence - Selected plugin sequence}} to earlier positions in the sequence.
\item {\bf XPIWIT: Execute pipeline stepwise}: \hypertarget{CE_Anzeige_XPIWITPipelineCall}{} \newline
execute plugins in XPIWIT separately. In this option, for each plugin a separate call of XPIWIT.exe is executed. This option is only used by the extension package ''Images and Videos''.
\item {\bf Down}: \hypertarget{CE_PlugListDown}{} \newline
moves the selected plugins in \hyperlink{CE_Auswahl_PluginsList}{\emph{Control element: Plugin sequence - Selected plugin sequence}} to later positions in the sequence.
\item {\bf Delete}: \hypertarget{CE_PlugListDel}{} \newline
deletes the recent plugin from the plugin sequence.
\item {\bf Delete all}: \hypertarget{CE_PlugListDelAll}{} \newline
deletes all plugins from a plugin sequence.
\item {\bf Load}: \hypertarget{CE_PlugListLoad}{} \newline
loads a plugin sequence from a \textindex{plugin sequence file} (\emph{*.plugseq}). The matching of the plugins is done by the function names.
\item {\bf Selected plugin sequence}: \hypertarget{CE_Auswahl_PluginsList}{} \newline
shows the recent \textindex{plugin sequence}. The plugin sequence is performed step by step using \hyperlink{CE_PlugListExec}{\emph{Control element: Plugin sequence - Execute}} resp. \hyperlink{MI_Extraktion_ZRZR_Kombi}{\emph{Edit - Extract - Time series $\rightarrow$ Time series, Time series $\rightarrow$ Single features (via plugin sequence)...}} (plugins for time series) or
\hyperlink{MI_ImVid_ApplyPlugImgs}{\emph{Images and Videos - Apply plugin sequence to the selected images}} (plugins for images). Existing parameters of the plugins are set by \hyperlink{CE_Auswahl_PluginsCommandLine}{\emph{Control element: Plugin sequence - Plugin parameter}}. The order of plugins can be changed by \hyperlink{CE_PlugListDown}{\emph{Control element: Plugin sequence - Down}} etc.

WARNING! If the plugin sequence is recorded by a macro, each plugin must be added separately and its parameters have to be chosen before the next plugin can be added. Otherwise, the macros recording might cause problems if the plugin sequence contains more than one plugin with the same name. In this case, the parameter of the last plugin with the same name are adapted.
\item {\bf Save}: \hypertarget{CE_PlugListSave}{} \newline
saves a plugin sequence with the defined plugin parameters in a \textindex{plugin sequence file} (\emph{*.plugseq}).
\item {\bf Show results}: \hypertarget{CE_Plugins_IgnoreIntermediates}{} \newline
defines the visualization and save options for plugin sequences of images. Intermediate results are the result of all plugins except the last one. ''Show'' means a visualization as image in a separate Matlab figure, ''Save'' the generation of a new image file and the generation of a new image type.

Plugins in projects without images ignores this option and set it to ''Save final result''.
\item {\bf Show parameters}: \hypertarget{CE_PlugListShowPar}{} \newline
shows all used parameters in a plugin sequence.
\item {\bf Save performance log file}: \hypertarget{CE_Plugins_PerformanceLogSettings}{} \newline
defines if computing times for single plugins or complete plugin sequences should be written in a log file ''*\_PerformanceLog.csv'' in the recent project directory.
\item {\bf Execute}: \hypertarget{CE_PlugListExec}{} \newline
executes the recent plugin sequence in \hyperlink{CE_Auswahl_PluginsList}{\emph{Control element: Plugin sequence - Selected plugin sequence}}.
\item {\bf Use the same transformation matrix (e.g. PCA) for time series reduction of all data points}: \hypertarget{CE_ZR_GlTrans}{} \newline
influences the reduction of time series by the Principal Component Analysis. If the option is
marked, an identical transformation matrix is used for all data points.  Otherwise, a
separate transformation matrix is computed for each data point.
\end{itemize}
\clearpage

\section{Control elements for 'Single features'}

 \begin{figure}[h] 
 \includegraphics[width=0.9\columnwidth]{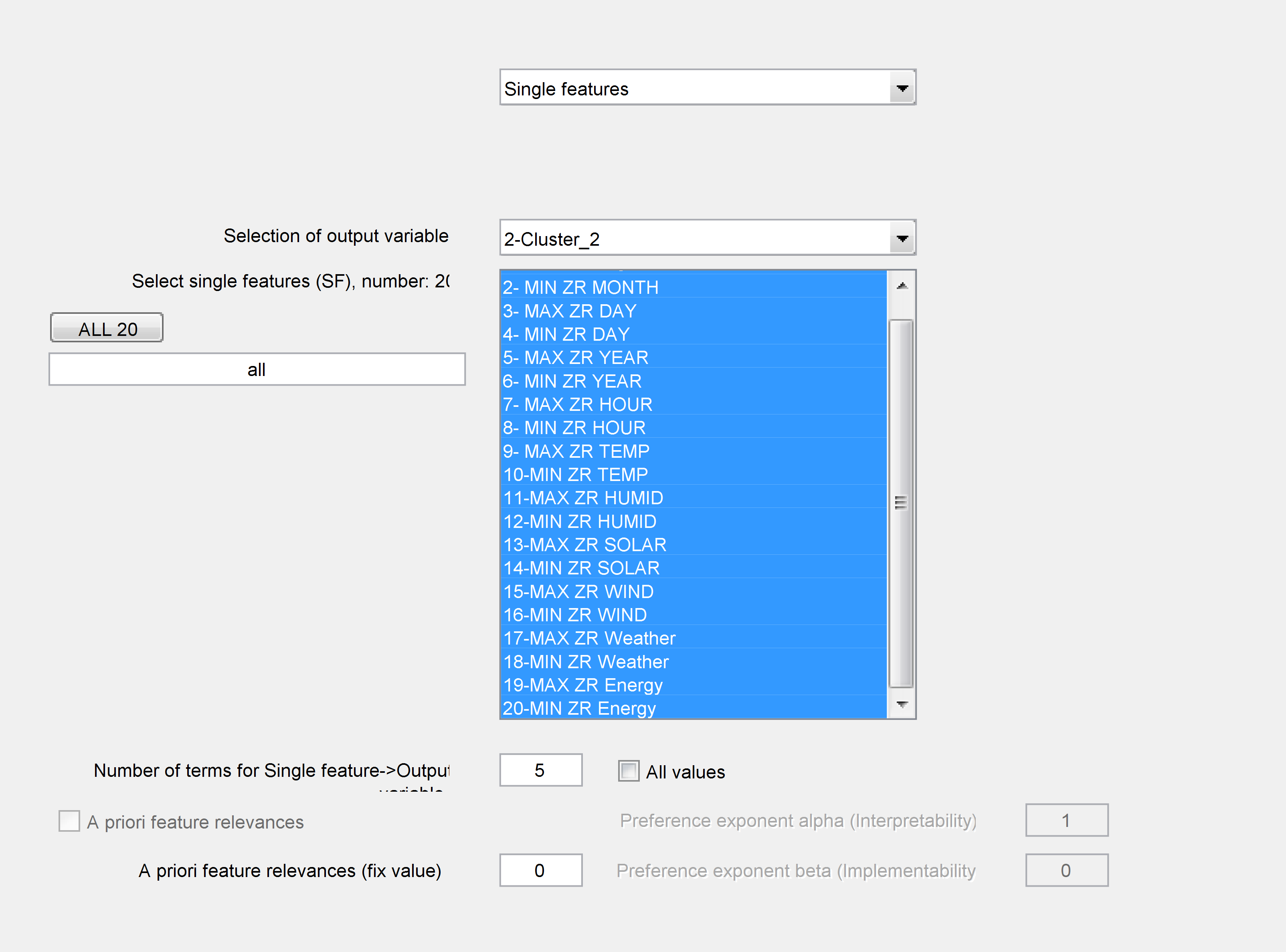}
 \caption{\label{fig:optionen5-6} Control elements for Single features }
 \end{figure}
\begin{itemize}
\item {\bf Selection of output variable}: \hypertarget{CE_Auswahl_Ausgangsgroesse}{} \newline
selects an output variable. This selection influences many functions as the evaluation of single
features and time series, the design of classifiers, almost all visualization functions etc.
\item {\bf Selection of single feature(s) (SF)}: \hypertarget{CE_Auswahl_EM}{} \newline
selects single features for visualization and all following processing steps. The selection can be
done by mouse clicks in the listbox, by writing the numbers in the edit field on the left hand side
or by the \emph{'ALL'} button to select all single features. The different fields will be
synchronized after the input.

TIP: An input in the edit field enables a selection with a different order (e.g. 2-4-1 for a
visualization) in contrast to the listbox.
\item {\bf Number of terms for Single feature$\rightarrow$Output variable}: \hypertarget{CE_EM_Ausgangs}{} \newline
defines the number of terms for a new output variable generated by \hyperlink{MI_EM_Klasse}{\emph{Edit - Convert - Selected single features $\rightarrow$ Output variables}} (see also
\hyperlink{CE_EM_Ausgangs_Alle}{\emph{Control element: Single features - All values}}). The transformation is done by a maximum defuzzification. The
membership functions are computed by a cluster method.  \index{Single feature $\rightarrow$ Output variable}
\item {\bf All values}: \hypertarget{CE_EM_Ausgangs_Alle}{} \newline
uses all different values of a feature as separate linguistic terms (see \hyperlink{MI_EM_Klasse}{\emph{Edit - Convert - Selected single features $\rightarrow$ Output variables}}). The
value of \hyperlink{CE_EM_Ausgangs}{\emph{Control element: Single features - Number of terms for Single feature$\rightarrow$Output variable}} will be ignored.

Example:

A single feature has the values 1, 2.5, 3.5 and 7. The result is an output variable with four terms
called  ''ca. 1'', ''ca. 2.5'', ''ca. 3.5'' and ''ca. 7''. In contrast, new terms with the number
defined in \hyperlink{CE_EM_Ausgangs}{\emph{Control element: Single features - Number of terms for Single feature$\rightarrow$Output variable}} will be computed if the checkbox is deactivated.
\item {\bf A priori feature relevances}: \hypertarget{CE_A_Priori}{} \newline
switches on the use of \textmainindex{a priori relevances} of single features. For feature
selection, features with higher a priori relevances are preferred. The values are between zero (bad
feature) and one (preferable feature).

They can be

 - defined in the project file (variable \texttt{interpret\_merk})

 - generated by means of categories, or

 - manually modified  (e.g. by \hyperlink{CE_A_Priori_BestimmterWert}{\emph{Control element: Single features - A priori feature relevances (fix value)}}).
\item {\bf Preference exponent alpha (Interpretability)}: \hypertarget{CE_A_Priori_Alpha}{} \newline
defines an exponent for feature relevances using \textindex{a priori relevances}. The exponent
weights the first value of the a priori relevances mostly associated with an interpretability
value. A value of zero ignores the interpretability, large values prefer features with higher
interpretability.
\item {\bf A priori feature relevances (fix value)}: \hypertarget{CE_A_Priori_BestimmterWert}{} \newline
sets a defined value for a priori relevances for a manual setting using
\hyperlink{MI_Apriori_AufBestimmtenWert}{\emph{Edit - Category - A priori relevances of selected features... - set to a fix value (from GUI)}} for all selected single features.
\item {\bf Preference exponent beta (Implementability)}: \hypertarget{CE_A_Priori_Beta}{} \newline
defines an exponent for feature relevances using \textindex{a priori relevances}. The exponent
weights the second value of the a priori relevances mostly associated with the
\textindex{implementability}. A value of zero ignores the implementability, large values prefer
features with higher implementability.
\end{itemize}
\clearpage

\section{Control elements for 'Data preprocessing'}

 \begin{figure}[h] 
 \includegraphics[width=0.9\columnwidth]{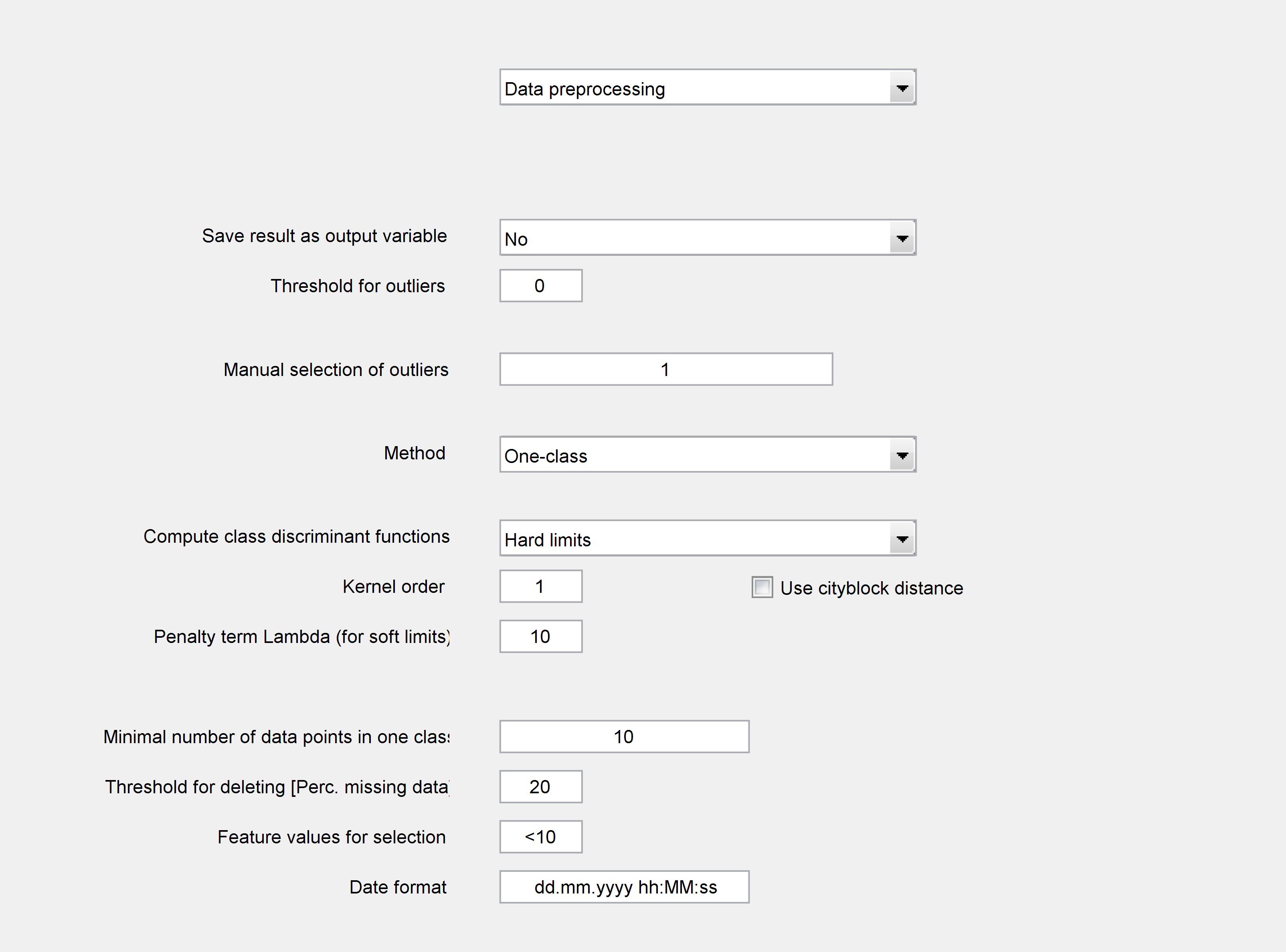}
 \caption{\label{fig:optionen6-7} Control elements for Data preprocessing - One-class}
 \end{figure}
\begin{itemize}
\item {\bf Save result as output variable}: \hypertarget{CE_Ausreisser_InKlasse}{} \newline
save the result of the \textindex{outlier detection} as a new output variable.

The option ''new output variable'' adds always a new output variable with the result of the outlier
detection. The variable name is ''Outlier'' with the name of the used method.

The option ''Replace identical output variable'' replace the results in an existing output variable
computed by the same method. Otherwise, a new output variable is added.
\item {\bf Threshold for outliers}: \hypertarget{CE_Ausreisser_Schwelle}{} \newline
defines a threshold parameter if a data point is classified as outlier or not (see
\hyperlink{MI_Ausreisser_An}{\emph{Edit - Outlier detection - Apply (selected data points, designed data set)}}).

The meaning of the parameter depends on the method:

SVM - a value smaller than zero (tolerated distance to the class border)

distance-based: distance value

density-based: number of neighbors

The level curves in the visualization are helpful for the parameter definition (see
\hyperlink{CE_Anzeige_Ausreisser}{\emph{Control element: View: Classification and regression - Show outlier detection}}).

\item {\bf Manual selection of outliers}: \hypertarget{CE_Ausreisser_Manuell}{} \newline
removes the data points with the indices in the edit field from the training data set. This
function supports the outlier detection in complicated cases because these data points are not
evaluated as good measurements. Depending on the related positions in feature space, it normally
increases the probability of an automatic outlier detection of these data points by applying the
outlier detection. All selected data points are used if the field is empty or ''[]'' is written in
the field.

During the application, all data points will be evaluated.

The number of data points is shown if \hyperlink{CE_Anzeige_NrDatentupel}{\emph{Control element: View: Single features - Show data point number}} is marked.
\item {\bf Method}: \hypertarget{CE_Ausreisser_Verfahren}{} \newline
selects the classifier type for the outlier detection.

The one-class method optimizes the coefficients $a_i$ and $b$ of the function: $f(\mathbf{z}) =
\sum_i a_i K(\mathbf{z}, \mathbf{x}_i) + b,$ with an unknown data point $\mathbf{z}$ and the data
points $\mathbf{x}_i$ of the training data set. $K$ is a kernel function known from
\textindex{Support Vector Machines}. A data point $\mathbf{z}$ is handled as outlier if $f(z) <$
Threshold (see \cite{Campbell00} for algorithm details). The threshold is defined by
\hyperlink{CE_Ausreisser_Schwelle}{\emph{Control element: Data preprocessing - Threshold for outliers}}.

The distance-based method uses the Mahalanobis distance to the mean value of the training data set.
All data points with a distance larger than \hyperlink{CE_Ausreisser_Schwelle}{\emph{Control element: Data preprocessing - Threshold for outliers}} (positive value) are
handled as outliers.

The density-based method classifies all data points without a minimal number of neighbors (defined
by \hyperlink{CE_Ausreisser_Schwelle}{\emph{Control element: Data preprocessing - Threshold for outliers}}) as outliers. The neighborhood is defined by a maximal
([0,1]-normalized Euclidean) distance. The distance is tuned by \hyperlink{CE_kNN_MaxAbstand}{\emph{Control element: Data preprocessing - Max. distance}}.
\item {\bf Compute class discriminant functions}: \hypertarget{CE_OneClass_HartWeich}{} \newline
switches between hard and soft limits for the one-class method. For hard limits, all training data
are handled as class members. For soft limits, some training data could be automatically classified
as outliers. It makes especially sense if some data points in the training data set look
suspicious.
\item {\bf Kernel order}: \hypertarget{CE_SVM_Ordnung}{} \newline
set the order of the kernel of the SVM. It is used for the classification with the SVM and by
the SVM-based outlier detection.
\item {\bf Use cityblock distance}: \hypertarget{CE_OneClass_Cityblock}{} \newline
switches on the 1-norm (city block distance) instead of the Euclidean distance for the Gaussian RBF
kernel in the one-class method of the outlier detection (see \hyperlink{CE_SVM_Kernel}{\emph{Control element: Data mining: Special methods - Kernel}}). This results
normally in rougher class borders.
\item {\bf Penalty term Lambda (for soft limits)}: \hypertarget{CE_OneClass_Strafterm}{} \newline
tunes the number of training data accepted as outliers. A larger value reduces the number of
outliers.

This parameter is only used if ''soft limits'' are selected in \hyperlink{CE_OneClass_HartWeich}{\emph{Control element: Data preprocessing - Compute class discriminant functions}}.
\item {\bf Minimal number of data points in one class}: \hypertarget{CE_Select_MinDtClass}{} \newline
minimal number of data points for the selection with  \hyperlink{MI_Datenauswahl_FreqCode}{\emph{Edit - Select - Data points via most frequent terms}}.
\item {\bf Threshold for deleting [Perc. missing data]}: \hypertarget{CE_Nullzr_Schwellwert}{} \newline
defines the percental threshold in \textindex{data preprocessing} for the deletion of time series
or single features in case of missing values. Otherwise, each data point with at least one missing
value in any single feature or data point will be deleted. A time series contains missing values if
at least one sample point is missing. Missing values must be coded with \textindex{NaN values} or
\textindex{Inf values}. Time series with zero values for all sample points are also considered as
\textindex{missing values}.

Example:

Project with 100 data points and 15 \% missing values (data point 1-15) for time series  x1 and 5
\% missing values for time series x2 (data point 96-100) and threshold of 10 \%: Time series
 x1 will be completely deleted, data points 96-100 will be deleted.
\item {\bf Feature values for selection}: \hypertarget{CE_DatapointValueSelection}{} \newline
value range for the selection of data points based on single feature values in \hyperlink{CE_DatapointValueSelection}{\emph{Control element: Data preprocessing - Feature values for selection}}.
\item {\bf Date format}: \hypertarget{CE_DateFormat}{} \newline
defines the date format for the conversion of dates and hours in an output variable into a MATLAB \textindex{timestamp format} using \hyperlink{MI_Trans_Date2Timestamp}{\emph{Edit - Convert - Selected output variable (Date) $\rightarrow$ Single feature (Timestamp)}}.
\end{itemize}

\clearpage
 \begin{figure}[h] 
 \includegraphics[width=0.9\columnwidth]{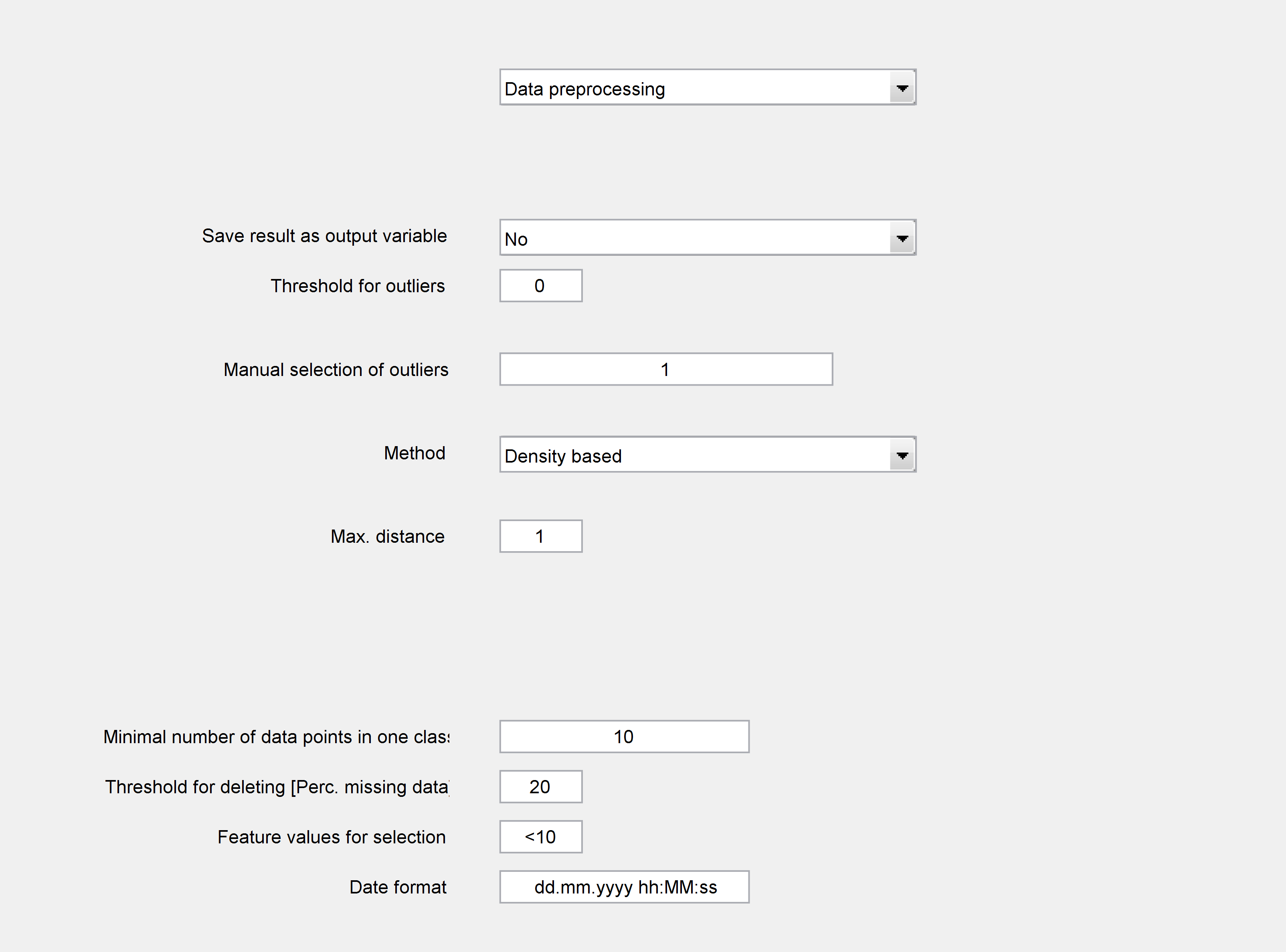}
 \caption{\label{fig:optionen6-8} Control elements for Data preprocessing - Density based}
 \end{figure}
\begin{itemize}
\item {\bf Save result as output variable}: \hypertarget{CE_Ausreisser_InKlasse}{} \newline
save the result of the \textindex{outlier detection} as a new output variable.

The option ''new output variable'' adds always a new output variable with the result of the outlier
detection. The variable name is ''Outlier'' with the name of the used method.

The option ''Replace identical output variable'' replace the results in an existing output variable
computed by the same method. Otherwise, a new output variable is added.
\item {\bf Threshold for outliers}: \hypertarget{CE_Ausreisser_Schwelle}{} \newline
defines a threshold parameter if a data point is classified as outlier or not (see
\hyperlink{MI_Ausreisser_An}{\emph{Edit - Outlier detection - Apply (selected data points, designed data set)}}).

The meaning of the parameter depends on the method:

SVM - a value smaller than zero (tolerated distance to the class border)

distance-based: distance value

density-based: number of neighbors

The level curves in the visualization are helpful for the parameter definition (see
\hyperlink{CE_Anzeige_Ausreisser}{\emph{Control element: View: Classification and regression - Show outlier detection}}).

\item {\bf Manual selection of outliers}: \hypertarget{CE_Ausreisser_Manuell}{} \newline
removes the data points with the indices in the edit field from the training data set. This
function supports the outlier detection in complicated cases because these data points are not
evaluated as good measurements. Depending on the related positions in feature space, it normally
increases the probability of an automatic outlier detection of these data points by applying the
outlier detection. All selected data points are used if the field is empty or ''[]'' is written in
the field.

During the application, all data points will be evaluated.

The number of data points is shown if \hyperlink{CE_Anzeige_NrDatentupel}{\emph{Control element: View: Single features - Show data point number}} is marked.
\item {\bf Method}: \hypertarget{CE_Ausreisser_Verfahren}{} \newline
selects the classifier type for the outlier detection.

The one-class method optimizes the coefficients $a_i$ and $b$ of the function: $f(\mathbf{z}) =
\sum_i a_i K(\mathbf{z}, \mathbf{x}_i) + b,$ with an unknown data point $\mathbf{z}$ and the data
points $\mathbf{x}_i$ of the training data set. $K$ is a kernel function known from
\textindex{Support Vector Machines}. A data point $\mathbf{z}$ is handled as outlier if $f(z) <$
Threshold (see \cite{Campbell00} for algorithm details). The threshold is defined by
\hyperlink{CE_Ausreisser_Schwelle}{\emph{Control element: Data preprocessing - Threshold for outliers}}.

The distance-based method uses the Mahalanobis distance to the mean value of the training data set.
All data points with a distance larger than \hyperlink{CE_Ausreisser_Schwelle}{\emph{Control element: Data preprocessing - Threshold for outliers}} (positive value) are
handled as outliers.

The density-based method classifies all data points without a minimal number of neighbors (defined
by \hyperlink{CE_Ausreisser_Schwelle}{\emph{Control element: Data preprocessing - Threshold for outliers}}) as outliers. The neighborhood is defined by a maximal
([0,1]-normalized Euclidean) distance. The distance is tuned by \hyperlink{CE_kNN_MaxAbstand}{\emph{Control element: Data preprocessing - Max. distance}}.
\item {\bf Max. distance}: \hypertarget{CE_kNN_MaxAbstand}{} \newline
defines the accepted maximum distance for a neighborhood (see \hyperlink{CE_kNN_NachbarnPruefen}{\emph{Control element: Data mining: Special methods - Evaluate minimum number of neighbors}}).
\item {\bf Minimal number of data points in one class}: \hypertarget{CE_Select_MinDtClass}{} \newline
minimal number of data points for the selection with  \hyperlink{MI_Datenauswahl_FreqCode}{\emph{Edit - Select - Data points via most frequent terms}}.
\item {\bf Threshold for deleting [Perc. missing data]}: \hypertarget{CE_Nullzr_Schwellwert}{} \newline
defines the percental threshold in \textindex{data preprocessing} for the deletion of time series
or single features in case of missing values. Otherwise, each data point with at least one missing
value in any single feature or data point will be deleted. A time series contains missing values if
at least one sample point is missing. Missing values must be coded with \textindex{NaN values} or
\textindex{Inf values}. Time series with zero values for all sample points are also considered as
\textindex{missing values}.

Example:

Project with 100 data points and 15 \% missing values (data point 1-15) for time series  x1 and 5
\% missing values for time series x2 (data point 96-100) and threshold of 10 \%: Time series
 x1 will be completely deleted, data points 96-100 will be deleted.
\item {\bf Feature values for selection}: \hypertarget{CE_DatapointValueSelection}{} \newline
value range for the selection of data points based on single feature values in \hyperlink{CE_DatapointValueSelection}{\emph{Control element: Data preprocessing - Feature values for selection}}.
\item {\bf Date format}: \hypertarget{CE_DateFormat}{} \newline
defines the date format for the conversion of dates and hours in an output variable into a MATLAB \textindex{timestamp format} using \hyperlink{MI_Trans_Date2Timestamp}{\emph{Edit - Convert - Selected output variable (Date) $\rightarrow$ Single feature (Timestamp)}}.
\end{itemize}

\clearpage\clearpage

\section{Control elements for 'View: Single features'}

 \begin{figure}[h] 
 \includegraphics[width=0.9\columnwidth]{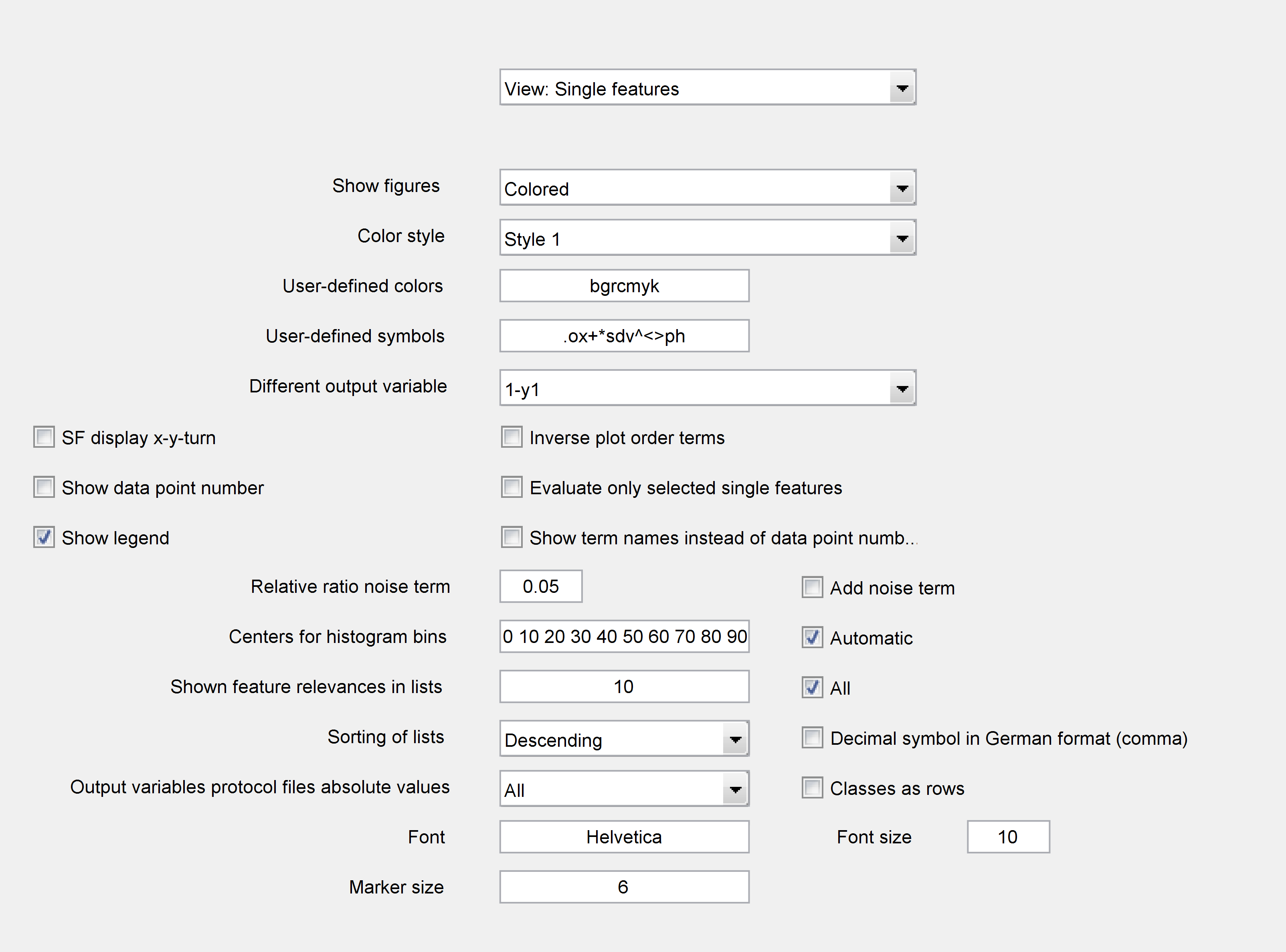}
 \caption{\label{fig:optionen7-9} Control elements for View: Single features }
 \end{figure}
\begin{itemize}
\item {\bf Show figures}: \hypertarget{CE_Anzeige_Grafiken}{} \newline
switches between color-coded, number-coded (for the number of the linguistic term of the output
variable), and (black and white) symbol-coded features and time series.
\item {\bf Color style}: \hypertarget{CE_Anzeige_FarbStyle}{} \newline
modifies the visualization of output variables by means of different color and style combinations
for time series and single features. It defines the mapping of linguistic terms to colors and
styles.
\item {\bf User-defined colors}: \hypertarget{CE_Anzeige_UsrDefCol}{} \newline
defines the order of user-defined colors for the visualization of linguistic terms of the selected output variable in time series and single feature plots. Here, the color abbreviations of Matlab are used (see help text for the function plot). The option is only active if \emph{User-defined (color or symbol)} resp. \emph{User-defined (color and symbol))} is selected in  \hyperlink{CE_Anzeige_FarbStyle}{\emph{Control element: View: Single features - Color style}}
and \emph{Colored} is selected in \hyperlink{CE_Anzeige_Grafiken}{\emph{Control element: View: Single features - Show figures}}.
\item {\bf User-defined symbols}: \hypertarget{CE_Anzeige_UsrDefSym}{} \newline
defines the order of user-defined symbols for the visualization of linguistic terms of the selected output variable in time series and single feature plots. Here, the symbol abbreviations of Matlab are used (see help text for the function plot). The option is only active if \emph{User-defined (color or symbol)} resp. \emph{User-defined (color and symbol))} is selected in  \hyperlink{CE_Anzeige_FarbStyle}{\emph{Control element: View: Single features - Color style}}
and \emph{Black-and-white symbol} is selected in \hyperlink{CE_Anzeige_Grafiken}{\emph{Control element: View: Single features - Show figures}}.
\item {\bf Different output variable}: \hypertarget{CE_Anzeige_DifferentClass}{} \newline
shows the results of classification and regression with a free selectable output variable if ''Different output variable'' was chosen in \hyperlink{CE_Anzeige_Klassenanzeige}{\emph{Control element: View: Classification and regression - Display classes for output variables}}.
\item {\bf SF display x-y-turn}: \hypertarget{CE_Anzeige_XYTausch}{} \newline
rotates a figure by switching the x- and y-axes for a two-dimensional plot.
\item {\bf Inverse plot order terms}: \hypertarget{CE_ZR_ShowInverseClassOrder}{} \newline
plots all data points in reverse order of the terms of the output variables.
\item {\bf Show data point number}: \hypertarget{CE_Anzeige_NrDatentupel}{} \newline
switches the display of data point numbers on or off.
\item {\bf Evaluate only selected single features}: \hypertarget{CE_Anzeige_RelSelFeat}{} \newline
evaluates only the selected single features.
\item {\bf Show legend}: \hypertarget{CE_Anzeige_Legende}{} \newline
plot the legend with the linguistic terms of the output variable.
\item {\bf Show term names instead of data point numbers}: \hypertarget{CE_Anzeige_Termnames}{} \newline
shows the term names instead of the data point number in figures. The option is only active if \hyperlink{CE_Anzeige_NrDatentupel}{\emph{Control element: View: Single features - Show data point number}} is activated.
\item {\bf Relative ratio noise term}: \hypertarget{CE_Anzeige_VisuNoiseValue}{} \newline
relative ratio (relative to the axis scale of each single feature in the original plot) for an additional noise (see \hyperlink{CE_Anzeige_VisuNoise}{\emph{Control element: View: Single features - Add noise term}}).
\item {\bf Add noise term}: \hypertarget{CE_Anzeige_VisuNoise}{} \newline
adds a uniformly distributed noise with the factor from \hyperlink{CE_Anzeige_VisuNoiseValue}{\emph{Control element: View: Single features - Relative ratio noise term}} to the shown single features. This make sense for the easier visualization of discrete-valued single features.
\item {\bf Centers for histogram bins}: \hypertarget{CE_Anzeige_Hist}{} \newline
defines bin centers for histograms manually (e.g. 0:10:100). This bin centers are used if \hyperlink{CE_Anzeige_HistAuto}{\emph{Control element: View: Single features - Automatic}} is deactivated.
\item {\bf Automatic}: \hypertarget{CE_Anzeige_HistAuto}{} \newline
defines the bin centers for histograms automatically.
\item {\bf Shown feature relevances in lists}: \hypertarget{CE_Anzeige_FeatureNumber}{} \newline
defines the number of shown features and their relevance values in relevance lists. The value is only used if  \hyperlink{CE_Anzeige_FeatureNumberCheck}{\emph{Control element: View: Single features - All}} is switched off. The reduction of the numbers allows more compact lists for documentation purposes.
\item {\bf All}: \hypertarget{CE_Anzeige_FeatureNumberCheck}{} \newline
shows all features and their relevance values in relevance lists.
\item {\bf Sorting of lists}: \hypertarget{CE_Anzeige_ListSorting}{} \newline
defines the order of single features in lists and text files: ascending or descending sorted resp. unsorted.
\item {\bf Decimal symbol in German format (comma)}: \hypertarget{CE_Anzeige_GermanNumbers}{} \newline
defines a German style number format with commas as decimal point.
\item {\bf Output variables protocol files absolute values}: \hypertarget{CE_Anzeige_Output4EM}{} \newline
defines the protocol option of output variables in files generated by \hyperlink{MI_Anzeige_EM_Abs}{\emph{View - Single features - Absolute values}}. Possible values are all, the selected or none output variables.
\item {\bf Classes as rows}: \hypertarget{CE_Anzeige_ClassesAsLines}{} \newline
switches the style of protocol files for mean values etc. as generated by \hyperlink{MI_Anzeige_EM_Min_Max}{\emph{View - Single features - Mean, standard deviation, minimum, maximum}}. If the option is activated, each class is written in a row and each single feature in a column. Otherwise, each single feature is written in a row and each class in a column.
\item {\bf Font}: \hypertarget{CE_AllgFontName}{} \newline
defines the font type, that is set by \hyperlink{MI_Fenster_FontNameSize}{\emph{Window - Update font and font size in figures}} for all open MATLAB figures.
\item {\bf Font size}: \hypertarget{CE_AllgFontSize}{} \newline
defines the font size, that is set by \hyperlink{MI_Fenster_FontNameSize}{\emph{Window - Update font and font size in figures}} for all open MATLAB figures.
\item {\bf Marker size}: \hypertarget{CE_MarkerSize}{} \newline
defines the marker size in plots of single features.
\end{itemize}
\clearpage

\section{Control elements for 'View: Time series'}

 \begin{figure}[h] 
 \includegraphics[width=0.9\columnwidth]{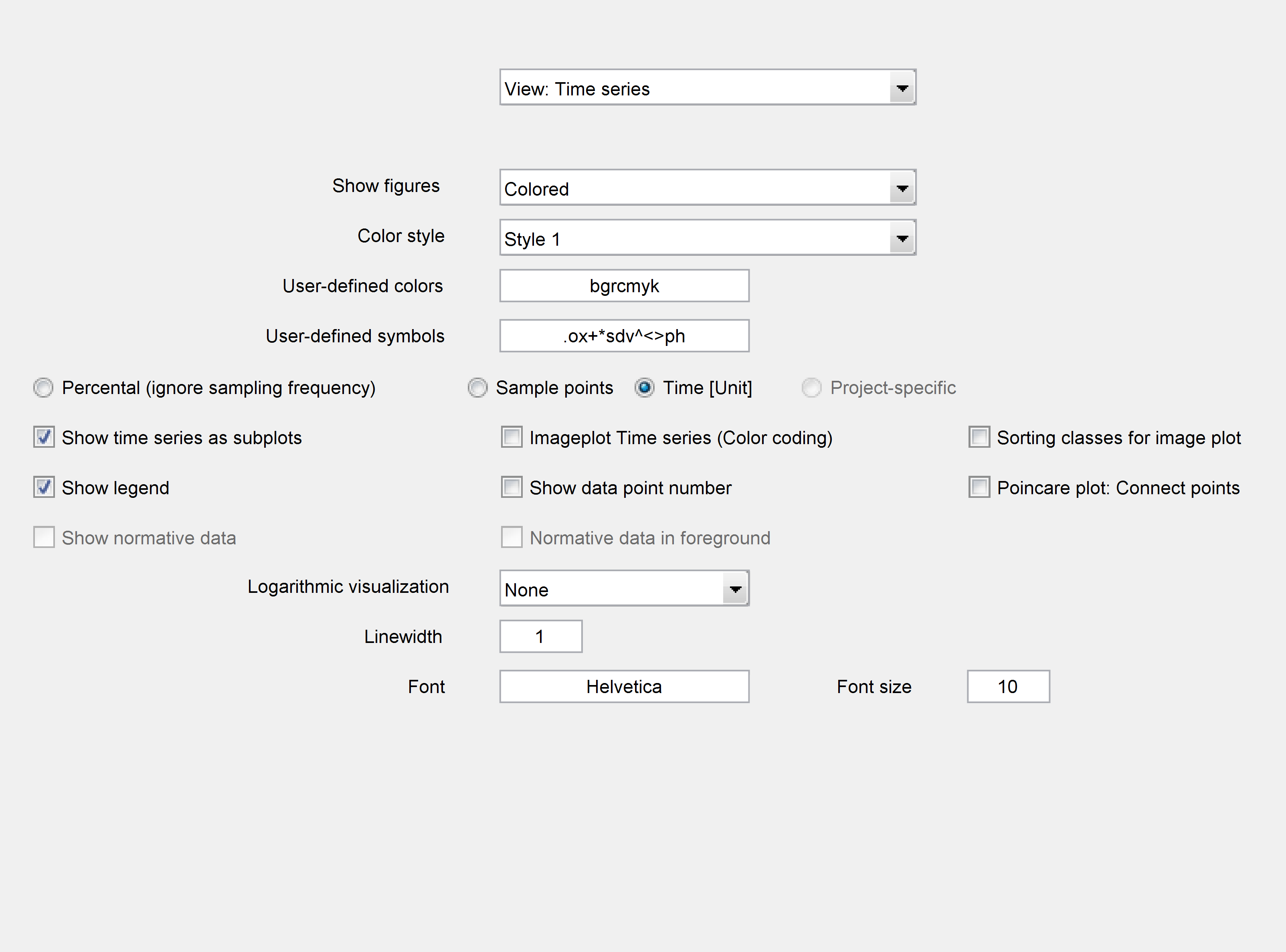}
 \caption{\label{fig:optionen8-10} Control elements for View: Time series }
 \end{figure}
\begin{itemize}
\item {\bf Show figures}: \hypertarget{CE_Anzeige_Grafiken}{} \newline
switches between color-coded, number-coded (for the number of the linguistic term of the output
variable), and (black and white) symbol-coded features and time series.
\item {\bf Color style}: \hypertarget{CE_Anzeige_FarbStyle}{} \newline
modifies the visualization of output variables by means of different color and style combinations
for time series and single features. It defines the mapping of linguistic terms to colors and
styles.
\item {\bf User-defined colors}: \hypertarget{CE_Anzeige_UsrDefCol}{} \newline
defines the order of user-defined colors for the visualization of linguistic terms of the selected output variable in time series and single feature plots. Here, the color abbreviations of Matlab are used (see help text for the function plot). The option is only active if \emph{User-defined (color or symbol)} resp. \emph{User-defined (color and symbol))} is selected in  \hyperlink{CE_Anzeige_FarbStyle}{\emph{Control element: View: Single features - Color style}}
and \emph{Colored} is selected in \hyperlink{CE_Anzeige_Grafiken}{\emph{Control element: View: Single features - Show figures}}.
\item {\bf User-defined symbols}: \hypertarget{CE_Anzeige_UsrDefSym}{} \newline
defines the order of user-defined symbols for the visualization of linguistic terms of the selected output variable in time series and single feature plots. Here, the symbol abbreviations of Matlab are used (see help text for the function plot). The option is only active if \emph{User-defined (color or symbol)} resp. \emph{User-defined (color and symbol))} is selected in  \hyperlink{CE_Anzeige_FarbStyle}{\emph{Control element: View: Single features - Color style}}
and \emph{Black-and-white symbol} is selected in \hyperlink{CE_Anzeige_Grafiken}{\emph{Control element: View: Single features - Show figures}}.
\item {\bf Percental (ignore sampling frequency)}: \hypertarget{CE_ZRAnzeige_Proz}{} \newline
switches to a percental x-axis for the visualization of time series. The time series length is set
to 100\%. The chosen sampling frequency will be ignored.
\item {\bf Sample points}: \hypertarget{CE_ZRAnzeige_Abtastpkt}{} \newline
switches to the number of sample points on the x-axis for the visualization of time series. The
chosen sampling frequency will be ignored.
\item {\bf Time [Unit]}: \hypertarget{CE_ZRAnzeige_Zeit}{} \newline
switches to the time on the x-axis for the visualization of time series. The chosen sampling
frequency in \hyperlink{CE_Zeitreihen_Abtastfreq}{\emph{Control element: Time series: General options - Sampling frequency of time series}} will be used to transform sample points to the time.
The time starts always with zero.
\item {\bf Project-specific}: \hypertarget{CE_ZRAnzeige_Projekt}{} \newline
uses a \textindex{project-specific time scale} defined by  (parameter.)projekt.timescale for the visualization of time series. This struct contains the elements .name for the name of the time scale and .time for the time values of each sample point. This option is useful for non-equidistant time scales or other one-dimensional scales (e.g. mass for mass spectroscopy). An existing time series can be converted to a project-specific time scale by \hyperlink{MI_ZR2PTS}{\emph{Edit - Convert - Selected time series $\rightarrow$ Project-specific time scale}}.
\item {\bf Show time series as subplots}: \hypertarget{CE_Zeitreihen_Subplots}{} \newline
plots selected time series to individual axes, if the option is activated.

Otherwise, all time series are plotted to the same axis. Different time series are displayed with
different colors. The color mapping to linguistic terms of output values gets lost.
\item {\bf Imageplot Time series (Color coding)}: \hypertarget{CE_Anzeige_ZRImagePlot}{} \newline
shows time series in a color code. The scale is tuned by \hyperlink{CE_Anzeige_Farbachse}{\emph{Control element: View: Spectrogram, FFT, CCF - Limits for color axis}} and \hyperlink{CE_Kennlinie}{\emph{Control element: View: Spectrogram, FFT, CCF - Function for color bar}}, the colormap by \hyperlink{CE_Anzeige_Colormap}{\emph{Control element: View: Spectrogram, FFT, CCF - Colormap}}. The selected data points are sorted by number or by the linguistic terms of the selected output variable depending on the value of  \hyperlink{CE_Anzeige_ZRImagePlotClassSort}{\emph{Control element: View: Time series - Sorting classes for image plot}}.

If only one time series is visualized, the option is switched off temporarily.
\item {\bf Sorting classes for image plot}: \hypertarget{CE_Anzeige_ZRImagePlotClassSort}{} \newline
switches between sorted by number or class-wise by the linguistic terms of the selected output variable. The selection is only relevant if \hyperlink{CE_Anzeige_ZRImagePlot}{\emph{Control element: View: Time series - Imageplot Time series (Color coding)}} is activated.
\item {\bf Show legend}: \hypertarget{CE_Anzeige_Legende}{} \newline
plot the legend with the linguistic terms of the output variable.
\item {\bf Show data point number}: \hypertarget{CE_Anzeige_NrDatentupel}{} \newline
switches the display of data point numbers on or off.
\item {\bf Poincare plot: Connect points}: \hypertarget{CE_Anzeige_PoincareLine}{} \newline
defines if the sample points in a Poincare plot will be connected by lines or not.
\item {\bf Show normative data}: \hypertarget{CE_Anzeige_Normdaten}{} \newline
shows \textindex{normative data} as light-gray region in the visualization of time series.
Normative data has to be produced by \hyperlink{MI_Mittelwert_zu_Norm}{\emph{File - Normative data - Mean value to normative data}} (internally from the project) or
by \hyperlink{MI_ImportNorm}{\emph{File - Normative data - Load normative data}} from a file.
\item {\bf Normative data in foreground}: \hypertarget{CE_Normdaten_Vordergrund}{} \newline
the visualization of many original time series may overlay the \textindex{normative data}. Activate this option to plot the normative data in the foreground.
\item {\bf Logarithmic visualization}: \hypertarget{CE_ZRAnzeige_Log}{} \newline
select an optional logarithmic scaling of times and resp. or amplitudes of time series for the visualization.
\item {\bf Linewidth}: \hypertarget{CE_Anzeige_Linienstaerke}{} \newline
modifies the line width for the visualization of time series. It makes sense e.g. for the export of
figures in \LaTeX figures with values of e.g. $1.5$ or $2$.
\item {\bf Font}: \hypertarget{CE_AllgFontName}{} \newline
defines the font type, that is set by \hyperlink{MI_Fenster_FontNameSize}{\emph{Window - Update font and font size in figures}} for all open MATLAB figures.
\item {\bf Font size}: \hypertarget{CE_AllgFontSize}{} \newline
defines the font size, that is set by \hyperlink{MI_Fenster_FontNameSize}{\emph{Window - Update font and font size in figures}} for all open MATLAB figures.
\end{itemize}
\clearpage

\section{Control elements for 'View: Spectrogram, FFT, CCF'}

 \begin{figure}[h] 
 \includegraphics[width=0.9\columnwidth]{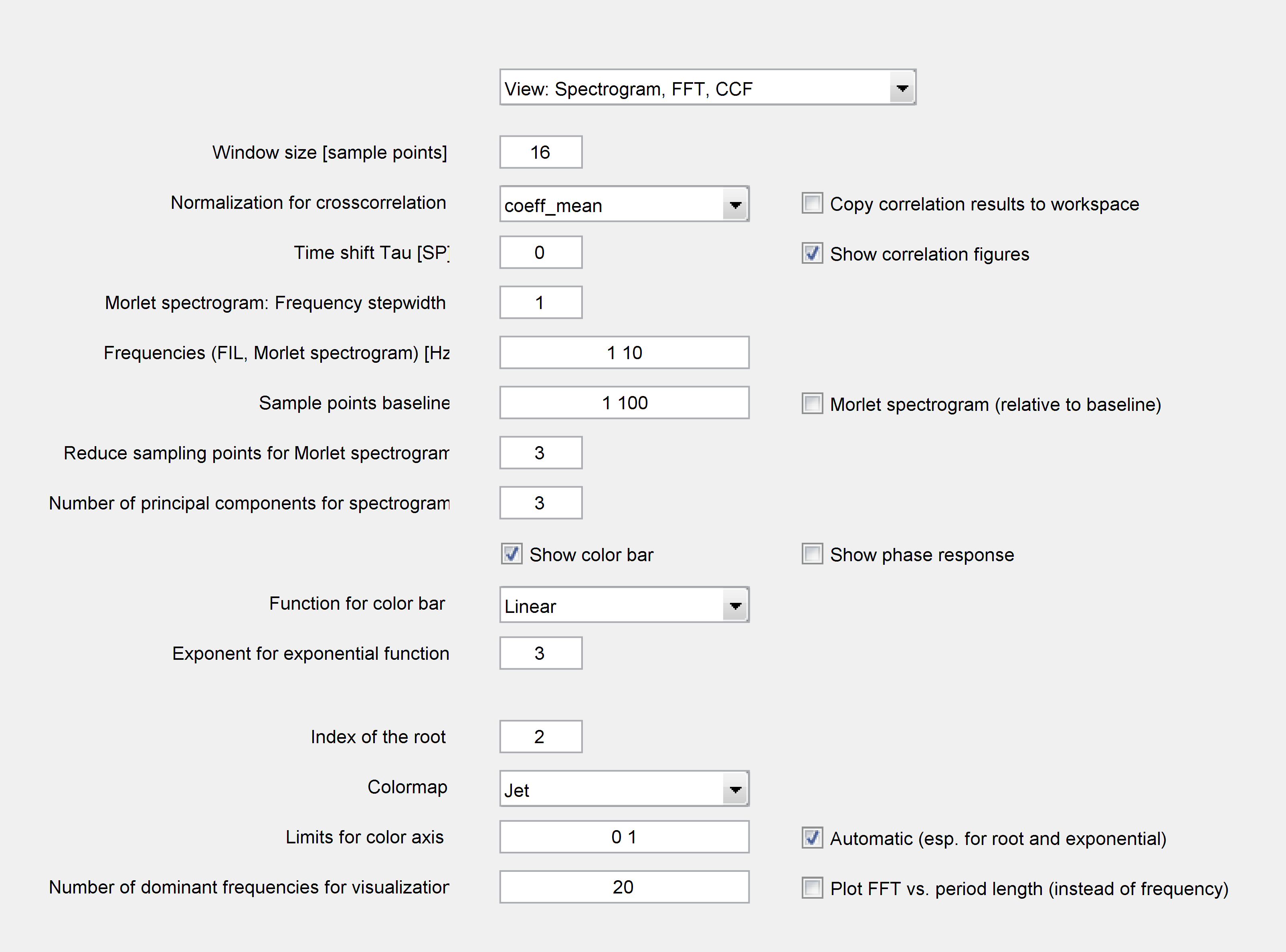}
 \caption{\label{fig:optionen9-11} Control elements for View: Spectrogram, FFT, CCF }
 \end{figure}
\begin{itemize}
\item {\bf Window size [sample points]}: \hypertarget{CE_Zeitreihen_Fenstergroesse}{} \newline
defines the window size for the computation of spectrograms. It controls the compromise between a
temporal and frequency resolution. A small value prefers a better temporal resolution. The value is
automatically reduced to the next power of two to preserve the efficiency of the FFT algorithm
(e.g. $2^7=128$, $2^{10}=$1024 etc.). The value is limited by the number of sample points.
\item {\bf Normalization for crosscorrelation}: \hypertarget{CE_ScaleOptCrossCorr}{} \newline
defines the mode of normalization for Auto and Cross Correlation Functions:

- ''biased'', scales to the length of the time series, see MATLAB function xcorr

- ''unbiased'', scales to (length of time series - lag), see MATLAB function xcorr (less robust results for large lags)

- ''coeff'', normalizes the result to autocorrelation 1 with lag 0, see MATLAB function xcorr

- ''coeff\_local'' - as ''biased'', but with an additional mean correction and variance normalization to one (identical to ''coeff\_mean'', hold for compatibility reasons)

- ''none'' - without normalization, see MATLAB function xcorr

In addition, options with mean value correction (+''\_mean'') and linear trend compensation (+''\_detrend'') of the time series exist.
\item {\bf Copy correlation results to workspace}: \hypertarget{CE_ZR_ParKorrWorkspace}{} \newline
defines if the auto or cross-correlation functions compute by \hyperlink{MI_AKFKKF_Anzeige}{\emph{View - Cross and Auto Correlation Functions - Separately for each data point}} are saved into a workspace variable kkfs.
\item {\bf Time shift Tau [SP]}: \hypertarget{CE_ZR_Tau}{} \newline
sets the time shift for the visualization of correlation coefficients in \hyperlink{MI_KKFAKF_MWDaten}{\emph{View - Cross and Auto Correlation Functions - Mean values of correlation coefficients (defined time shift)}}
and \hyperlink{MI_KKFAKF_MWKlassen}{\emph{View - Cross and Auto Correlation Functions - Class mean values of correlation coefficients (with defined time shift)}}. A value of zero uses the same sample point for both time series.
\item {\bf Show correlation figures}: \hypertarget{CE_ZR_ShowFigKorr}{} \newline
defines if the figures of correlation functions are shown or not. The disabling is useful if the results are saved into a variable (see \hyperlink{CE_ZR_ParKorrWorkspace}{\emph{Control element: View: Spectrogram, FFT, CCF - Copy correlation results to workspace}}).
\item {\bf Morlet spectrogram: Frequency stepwidth}: \hypertarget{CE_MorletSpekt_FreqSchritt}{} \newline
defines distances for frequencies, e.g. 1:100:10000.
\item {\bf Frequencies (FIL, Morlet spectrogram)}: \hypertarget{CE_ZR_Filterfreq}{} \newline
defines the \textindex{cutoff frequencies} for different filters. The parameter is used for a
\textindex{Butterworth filter} in feature extraction and for Morlet spectrograms. A
\textindex{high-pass filter} and a \textindex{low-pass filter} use only the first value, whereas the
\textindex{band-pass filter} uses both values. \index{Morlet spectrogram}
\item {\bf Sample points baseline}: \hypertarget{CE_MorletSpekt_Baseline}{} \newline
see \hyperlink{CE_MorletSpekt_Relativ}{\emph{Control element: View: Spectrogram, FFT, CCF - Morlet spectrogram (relative to baseline)}}.
\item {\bf Morlet spectrogram (relative to baseline)}: \hypertarget{CE_MorletSpekt_Relativ}{} \newline
computes the Morlet spectrogram relatively to a baseline if the option is activated (see
\hyperlink{MI_MorletSpektro_BerAnz}{\emph{View - Morlet spectrogram - Compute and show (selected data points and time series)}}). Instead of absolute value, the temporal changes will be shown. The
limits of the baseline are defined by \hyperlink{CE_MorletSpekt_Baseline}{\emph{Control element: View: Spectrogram, FFT, CCF - Sample points baseline}}. \index{Morlet spectrogram}
\item {\bf Reduce sampling points for Morlet spectrogram}: \hypertarget{CE_Zeitreihen_RedAbtast}{} \newline
reduces the number of shown sample points, because exported figures with Morlet spectrograms need a
lot of memory. If a value $k$ is greater than one,  only each $k$-th sample point will be shown.
The loss of information is usually acceptable for small values.
\item {\bf Number of principal components for spectrogram}: \hypertarget{CE_PCA_Spektrogramm}{} \newline
defines the number of principal components for \hyperlink{MI_Spektrogramm_PCA}{\emph{View - Spectrogram - Principal component analysis for spectrograms}}.
\item {\bf Show color bar}: \hypertarget{CE_Anzeige_Colorbar}{} \newline
shows the mapping between numbers and the color code for some visualizations as e.g. spectrograms.
\item {\bf Show phase response}: \hypertarget{CE_Anzeige_Phasengang}{} \newline
shows the \textindex{phase response} in addition to the amplitudes of a spectrogram.
\item {\bf Function for color bar}: \hypertarget{CE_Kennlinie}{} \newline
changes the color code for the mapping of numbers (e.g. for spectrograms). It is useful especially
for a better resolution for small values of amplitudes. Here, the roots square or the inverse
exponential function increase the resolution for small amplitude values. The parameters
\hyperlink{CE_Kennlinie_Exp}{\emph{Control element: View: Spectrogram, FFT, CCF - Exponent for exponential function}} resp. \hyperlink{CE_Kennline_Sqrt}{\emph{Control element: View: Spectrogram, FFT, CCF - Index of the root}} control the quantitative characteristic.
\item {\bf Exponent for exponential function}: \hypertarget{CE_Kennlinie_Exp}{} \newline
see \hyperlink{CE_Kennlinie}{\emph{Control element: View: Spectrogram, FFT, CCF - Function for color bar}} (exponent for exponential characteristic). A larger value increases the resolution for smaller amplitudes.
\item {\bf Index of the root}: \hypertarget{CE_Kennline_Sqrt}{} \newline
see \hyperlink{CE_Kennlinie}{\emph{Control element: View: Spectrogram, FFT, CCF - Function for color bar}} (n-fold root). A larger value increases the resolution for smaller amplitudes.
\item {\bf Colormap}: \hypertarget{CE_Anzeige_Colormap}{} \newline
changes the color style for some functions (e.g. for spectrograms). The colormaps are explained in
the Matlab documentation (e.g. for Jet: \verb=help jet=). The colormap \verb=1 - Gray values=
inverts the gray map: 0 is now white and 1 black.
\item {\bf Limits for color axis}: \hypertarget{CE_Anzeige_Farbachse}{} \newline
defines user-defined limits for some visualizations (e.g. for spectrograms). Larger and lower
values are limited to the minimal resp. maximal value. All values between are scaled to identical
values. This option is useful for an identical color code in different figures with different
minimal or maximal values. The option is switched off by \hyperlink{CE_ZRAnzeige_CAxis}{\emph{Control element: View: Spectrogram, FFT, CCF - Automatic (esp. for root and exponential)}}.
\item {\bf Automatic (esp. for root and exponential)}: \hypertarget{CE_ZRAnzeige_CAxis}{} \newline
uses an automatic scaling (minimum to maximum value) to compute the colormap. Otherwise, the
values from \hyperlink{CE_Anzeige_Farbachse}{\emph{Control element: View: Spectrogram, FFT, CCF - Limits for color axis}} are used as limits.
\item {\bf Number of dominant frequencies for visualization}: \hypertarget{CE_FFT_AnzFreq}{} \newline
defines the number of shown frequencies in a \textindex{frequency list file} (\emph{*.freq}).
\item {\bf Plot FFT vs. period length (instead of frequency)}: \hypertarget{CE_Anzeige_FFTPeriode}{} \newline
plot the \textindex{FFT} results vs. the period length. If the option is deactivated, the results
are shown vs. frequency.
\end{itemize}
\clearpage

\section{Control elements for 'View: Classification and regression'}

 \begin{figure}[h] 
 \includegraphics[width=0.9\columnwidth]{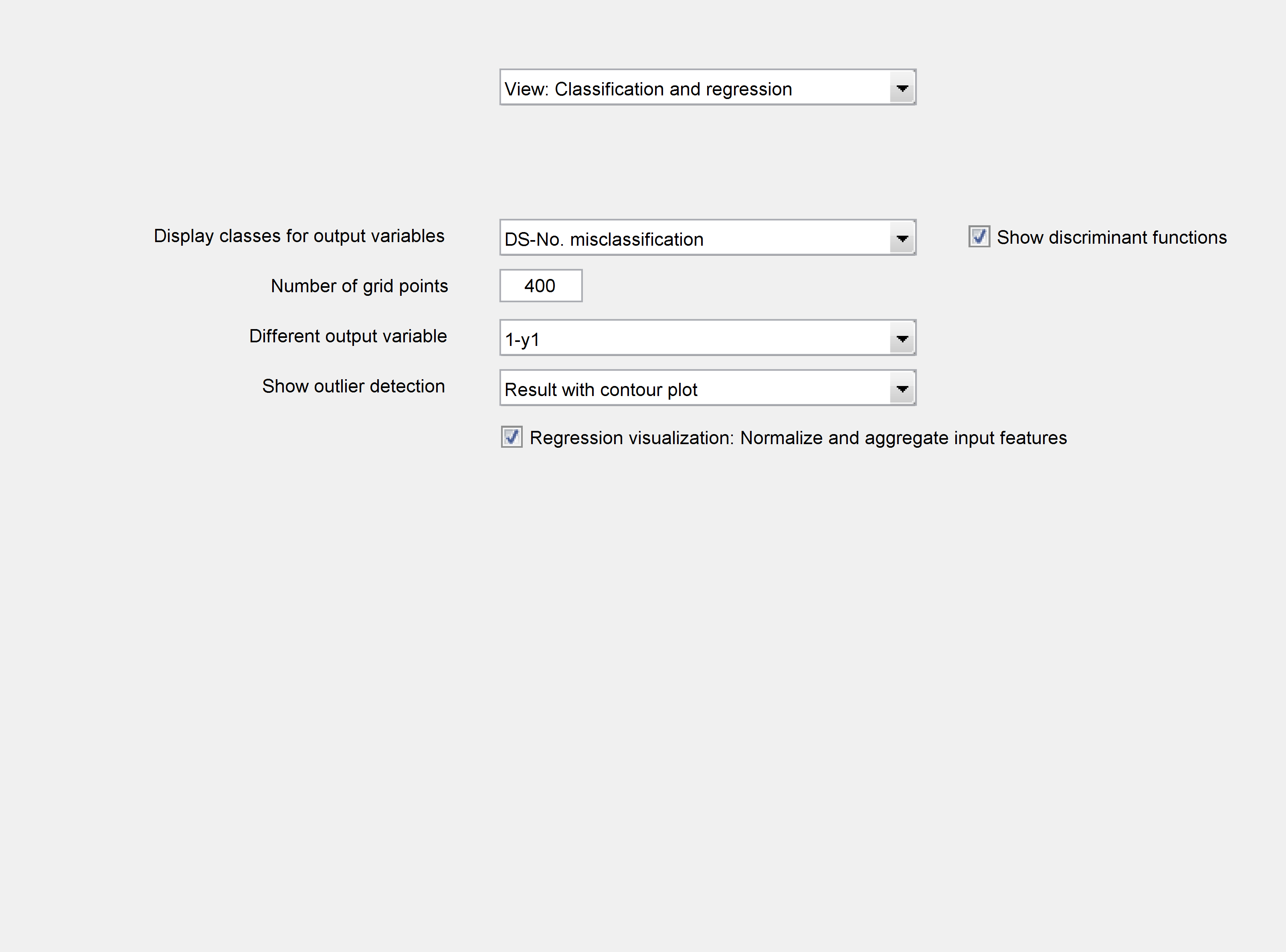}
 \caption{\label{fig:optionen10-12} Control elements for View: Classification and regression }
 \end{figure}
\begin{itemize}
\item {\bf Display classes for output variables}: \hypertarget{CE_Anzeige_Klassenanzeige}{} \newline
switches the visualization of classification results with \hyperlink{MI_Anzeige_Klassi_Erg}{\emph{View - Classification - Result}} and similar
functions:

''only learning data:'' shows the true class assignments for the learning data set.

''only classification:'' shows the classifier decisions.

''DS-No misclassification:'' shows the true class assignments for the learning data set with the
data point number for misclassified data points.

''Class-No misclassification:'' shows the true class assignments for the learning data set with the
number of the class decision for misclassified data points.
\item {\bf Show discriminant functions}: \hypertarget{CE_Anzeige_Trennflaechen}{} \newline
plots the discrimination functions for the class borders using the recent classifier in the feature
space.
\item {\bf Number of grid points}: \hypertarget{CE_Anzeige_Gitterpunkte}{} \newline
defines the number of grid points (in x- and y-direction) for the computation of regression
functions or discriminant functions for class borders. The function will be applied to each grid
point, resulting in (number of grid points)*(number of grid points) computations for
two-dimensional problems.
\item {\bf Different output variable}: \hypertarget{CE_Anzeige_DifferentClass}{} \newline
shows the results of classification and regression with a free selectable output variable if ''Different output variable'' was chosen in \hyperlink{CE_Anzeige_Klassenanzeige}{\emph{Control element: View: Classification and regression - Display classes for output variables}}.
\item {\bf Show outlier detection}: \hypertarget{CE_Anzeige_Ausreisser}{} \newline
contains the options for the \textindex{outlier detection}.

''None:'' computes the result, but does not show any visualization.

''only result:'' shows the classification of the selected data points as outliers or normal data
points.

''Result with contour plot:'' shows the classification of the selected data points as outliers or
normal data points. In addition, lines with the values of the decision functions are plotted. The
values of the feature space are expanded. The corresponding parameter can be changed in
\hyperlink{CE_Anzeige_Gitterpunkte}{\emph{Control element: View: Classification and regression - Number of grid points}}.

''only discriminant function:'' shows only the class borders in form of discriminant functions.

\item {\bf Regression visualization: Normalize and aggregate input features}: \hypertarget{CE_Anzeige_RegrInputNormAggr}{} \newline
show the results of the estimated output variable of the regression model in \hyperlink{MI_Anzeige_Regression_x_y_2D3D}{\emph{View - Regression - Input variable(s), output variable, and regression function (2D resp. 3D)}} as a function of normalized and aggregated single features if such features were chosen in the design of the regression, see parameters in \hyperlink{CEO_DM_Regr}{\emph{Control elements:  Data mining: Regression}}. If not, the option does not influence the figure. If the option is switched off, the estimated output variable is shown as function of the original single features without normalization and aggregation.  For more than two single features, \hyperlink{MI_GUI_Regression_Multi_Dimensional}{\emph{View - Regression - GUI for multidimensional visualization}} has to be used.
\end{itemize}
\clearpage

\section{Control elements for 'Data mining: Classification of single features'}
\hypertarget{CEO_DM_Grund}{}
contains the most important elements for the parameterization of data mining methods. The
parameterization of the chosen classifier is done using \hyperlink{CEO_DM_Klass}{\emph{Control elements:  Data mining: Special methods}}.

 \begin{figure}[h] 
 \includegraphics[width=0.9\columnwidth]{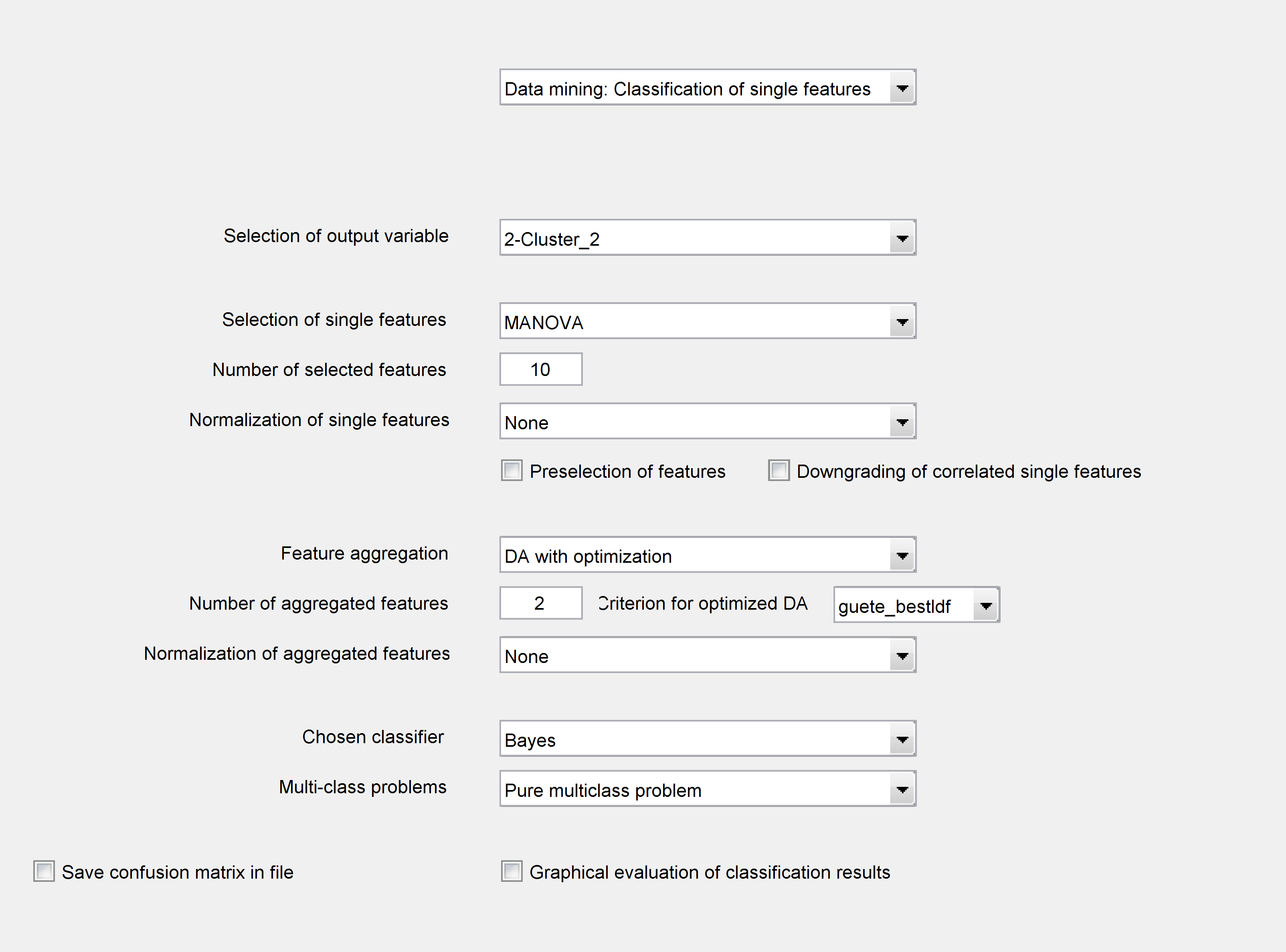}
 \caption{\label{fig:optionen11-13} Control elements for Data mining: Classification of single features - DA with optimization}
 \end{figure}
\begin{itemize}
\item {\bf Selection of output variable}: \hypertarget{CE_Auswahl_Ausgangsgroesse}{} \newline
selects an output variable. This selection influences many functions as the evaluation of single
features and time series, the design of classifiers, almost all visualization functions etc.
\item {\bf Selection of single features}: \hypertarget{CE_Klassifikation_Merkmalsauswahl}{} \newline
defines the method for the \textindex{feature selection} (e.g. use the recent selected features,
automatic selection by ANOVA), see \hyperlink{MI_Bewertung_EM}{\emph{Data mining - Selection and evaluation of single features}}.
\item {\bf Number of selected features}: \hypertarget{CE_Anzahl_Merkmale}{} \newline
defines the maximum number of single features for an automatic feature selection.
\item {\bf Normalization of single features}: \hypertarget{CE_Normierung_Merkmale}{} \newline
performs a normalization of features before a classification. It improves the results for all
methods which are sensitive against different scaled features. Two approaches are implemented; to
the interval $[0,1]$ or to a mean value of zero and a standard deviation of one under the
assumption of a normal distribution.
\item {\bf Preselection of features}: \hypertarget{CE_Klassifikation_Preselection}{} \newline
preselects all manually selected features (by \hyperlink{CE_Auswahl_EM}{\emph{Control element: Single features - Selection of single feature(s) (SF)}}) for an automatic
\textindex{feature selection}.

Example:

Features x5 and x7 are manually selected, selection of four features by ANOVA, best features are
x8, x9, x10, x13 leads to the selection of x5, x7, x8, x9.
\item {\bf Downgrading of correlated single features}: \hypertarget{CE_Klassifikation_Rueckstufung}{} \newline
downgrades a feature to a relevance of zero if it correlates significantly to at least one better
feature. This significance is measured by a positive (linear) Pearson correlation coefficient with a value
larger than a threshold define by the \hyperlink{CE_Krit_Koeff}{\emph{Control element: Data mining: Statistical options - Threshold for correlation coefficient}}). The aim of this function is the
generation of a short list with different features. Otherwise, some very similar features might
dominate at the first places of the list. \index{threshold for the correlation coefficient}
\item {\bf Feature aggregation}: \hypertarget{CE_Klassifikation_Merkmalsaggregation}{} \newline
defines the method for the linear \textindex{feature aggregation} by means of a multiplication with a
weighting matrix. Possible options are ''No aggregation'',  a (linear) \textindex{Discriminant Analysis}, a
Discriminant Analysis followed by a numerical optimization (''DA with optimization'', see
\hyperlink{CE_Klassifikation_OptDA}{\emph{Control element: Data mining: Classification of single features - Criterion for optimized DA}} for the optimization criterion),  a ''\textindex{Principal Component Analysis}
(PCA)'', an ''Independent Component Analysis (ICA)'' or the sum resp. mean value.
\item {\bf Number of aggregated features}: \hypertarget{CE_Anzahl_Aggregiert}{} \newline
defines the number of aggregated features ($s_d$) in a classification (see e.g. \cite{Reischl06}). The value
is ignored if ''No aggregation'' is chosen in \hyperlink{CE_Klassifikation_Merkmalsaggregation}{\emph{Control element: Data mining: Classification of single features - Feature aggregation}}. A value of
$s_d=1$ is always used for ''Sum'' and ''Mean value''.
\item {\bf Criterion for optimized DA}: \hypertarget{CE_Klassifikation_OptDA}{} \newline
specifies the criterion for a Discriminant Analysis with (numerical) optimization (see also
\hyperlink{CE_Klassifikation_Merkmalsaggregation}{\emph{Control element: Data mining: Classification of single features - Feature aggregation}}). Available options are an a maximization of the
classification accuracy  (''best\_class'') or a maximization of the minimal distance between two
classes using class-specific covariance matrixes as metric (''best\_ldf'') \cite{Reischl06}.
\item {\bf Normalization of aggregated features}: \hypertarget{CE_Normierung_Aggregiert}{} \newline
performs a normalization of the aggregated features before a classification. It improves the
results for all methods which are sensitive against different scaled features. Two approaches are
implemented; to the interval $[0,1]$ or to a mean value of zero and a standard deviation of one
under the assumption of a normal distribution.
\item {\bf Chosen classifier}: \hypertarget{CE_Klassifikation_Klassifikator}{} \newline
defines the type of the \textindex{classifier}. Possible options are a ''\textindex{Bayes
classifier}'', an ''\textindex{Artificial Neural Network}'', a ''Support vector machine'', a
''\textindex{$k$-nearest neighbor}'' classifier in two different implementation options, a ''Fuzzy
classifier'', a ''Decision tree'' and \textindex{Ensemble learning (fitensemble)}. Due to incompatibilities, the parameter of
\hyperlink{CE_Klassifikation_Mehrklassen}{\emph{Control element: Data mining: Classification of single features - Multi-class problems}} might be modified as well. Here, an additional warning is
shown in the Matlab command window. \index{decision tree}

The selected classifier type is parameterized in more detail by \hyperlink{CEO_DM_Klass}{\emph{Control elements:  Data mining: Special methods}}.
\item {\bf Multi-class problems}: \hypertarget{CE_Klassifikation_Mehrklassen}{} \newline
selects the approach for the optional decomposition of problems with more than two classes of the
output variable:

1. Pure \textindex{multi-class problems} solve the problem without a decomposition.

2. The \textindex{one-against-all} method decomposes a problem with $C$ classes in $C$ two-class
problems with class $c$ vs. the union of all remaining classes.

3. The \textindex{one-against-one} method decomposes a problem with $C$ classes in all pairwise
problems with class $c$ vs. class $j$ ($j \not= c$).

For the cases 2 and 3, a fusion of the single results is automatically done.
\item {\bf Save confusion matrix in file}: \hypertarget{CE_Konfusion_Datei}{} \newline
saves the confusion matrix for the training data set of the recent classifier in a file, if the
option is activated.
\item {\bf Graphical evaluation of classification results}: \hypertarget{CE_Anzeige_KlassiErg}{} \newline
opens a figure with classification results in the feature space of the classifier (if the option was activated).
\end{itemize}

\clearpage\clearpage

\section{Control elements for 'Data mining: Classification of time series'}

 \begin{figure}[h] 
 \includegraphics[width=0.9\columnwidth]{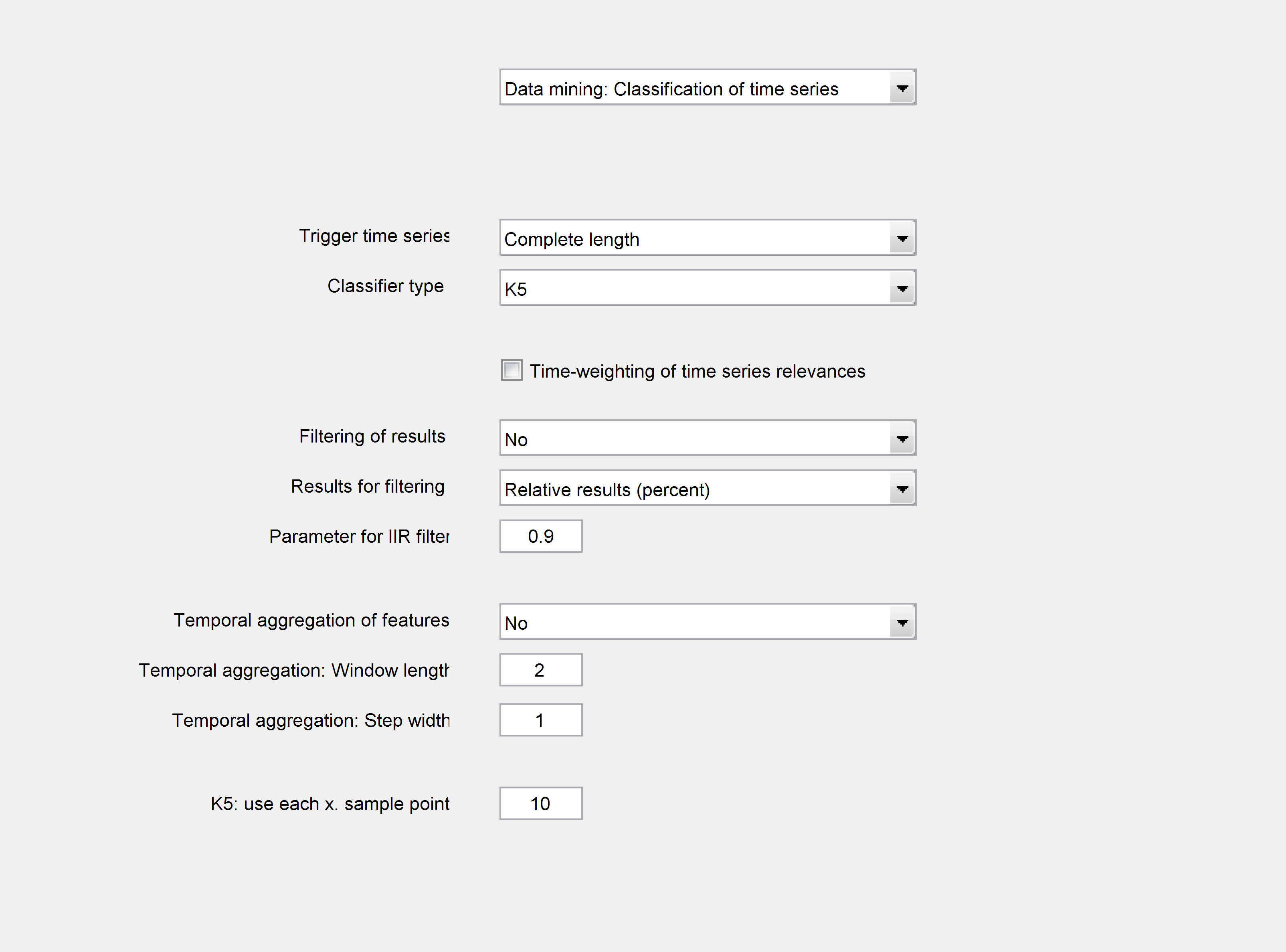}
 \caption{\label{fig:optionen12-14} Control elements for Data mining: Classification of time series - K5}
 \end{figure}
\begin{itemize}
\item {\bf Trigger time series}: \hypertarget{CE_ZRKlassi_Trigger}{} \newline
selects the \textindex{trigger time series} out of the existing time series. It is zero until the
\textindex{trigger} event, contains increasing values to measure the number of samples since the
trigger and zeros after the end of the recent event. If no trigger time series was selected, the
sample numbers of the complete time series are used: It starts with a value of one for the first
sample point and ends up with the length of the time series.
\item {\bf Classifier type}: \hypertarget{CE_ZRKlassi_Typ}{} \newline
selects a \textindex{classifier} type for the time series:

K1 contains only one classifier and ignores the trigger events. K5 is similar, but it uses the time
since the trigger event as additional input variable. K2 is a special kind of single feature
classifier for all sample points and aggregates results via a discriminant analysis. K3 and K4 have
separate sub-classifiers for each sample point since the trigger event. K3 has a fix selection of
time series, K4 a variable one for each sample point. TSK-Fuzzy is similar to K4, but it aggregates
similar sample points by means of fuzzy sets.

\cite{Burmeister06,Mikut08IEEE} give a detailed description of the classifiers.

The feature selection and the underlying classifiers (e.g. Bayes classifiers) are parameterized
using \hyperlink{CEO_DM_Grund}{\emph{Control elements:  Data mining: Classification of single features}} and \hyperlink{CE_Spezielle_Verfahren}{\emph{Control element: Data mining: Special methods - Method}}.
\item {\bf Time-weighting of time series relevances}: \hypertarget{CE_ZRKlassi_Rel_gewichten}{} \newline
gives the opportunity to prefer an earlier classification, e.g. for the feature selection of the
\textindex{K1} and \textindex{K3} classifier. The method modifies the feature relevances by
 $$R_k=\frac{R}{k-k_{trig}}, \text{ for } k > (k_{trig} + 0.15 \cdot k_{trig})$$
$k_{trig}$ is the sample point of the last trigger event. It increases feature relevances near the
\textindex{trigger event} in contrast to later ones.
\item {\bf Filtering of results}: \hypertarget{CE_ZRKlassi_Filter}{} \newline
select the method for filtering. The following options are available:

''No'' filter is used..

''\textindex{IIR-Filter}'' uses the filter parameters in \hyperlink{CE_ZRKlassi_IIR}{\emph{Control element: Data mining: Classification of time series - Parameter for IIR filter}}.

''Classification error for training data'' prefers samples with a better classification accuracy in
the training data set in contrast to samples with a lower accuracy. This results in time-variant
filter constants.

The ''Most frequent decision in a window'' selects the estimated class with the highest frequency
in a sliding window. The window length is defined in \hyperlink{CE_ZRKlassi_Aggregation}{\emph{Control element: Data mining: Classification of time series - Temporal aggregation of features}}. The first
sample points with a sample number up to the window length use all sample points up to this time
for decision.
\item {\bf Results for filtering}: \hypertarget{CE_ZRKlassi_Filtererg}{} \newline
switches between absolute and relative values as input values for the fusion by filtering. As an
example, absolute values of a \textindex{Bayes classifier} are the estimated  a-posteriori
probabilities, relative values the percentage of the estimated a-posteriori probabilities for the
given class relative to all classes.
\item {\bf Parameter for IIR filter}: \hypertarget{CE_ZRKlassi_IIR}{} \newline
set the parameter for the IIR filtering for the temporal fusion of classification results in time
series classification (see \hyperlink{CE_ZRKlassi_Filter}{\emph{Control element: Data mining: Classification of time series - Filtering of results}}).
\item {\bf Temporal aggregation of features}: \hypertarget{CE_ZRKlassi_Aggregation}{} \newline
tunes the temporal aggregation of sample points before the classification. The number and the names
of time series remain unchanged.

A selection of ''No'' does not use a temporal aggregation.

The ''Mean value'' is computed in a sliding window defined by
\hyperlink{CE_ZRKlassi_Aggregation_Groesse}{\emph{Control element: Data mining: Classification of time series - Temporal aggregation: Window length}} and \hyperlink{CE_ZRKlassi_Aggregation_Schritt}{\emph{Control element: Data mining: Classification of time series - Temporal aggregation: Step width}}.

''Minimum'', ''Maximum'' and ''ROM'' (Range of Motion) work with the same window, but with
different operators.

The aggregation is always causal, i.e. only past sample points are aggregated. The first
window does not start before the trigger event.
\item {\bf Temporal aggregation: Window length}: \hypertarget{CE_ZRKlassi_Aggregation_Groesse}{} \newline
defines the length of the sliding window for the temporal aggregation of features (see also
\hyperlink{CE_ZRKlassi_Aggregation}{\emph{Control element: Data mining: Classification of time series - Temporal aggregation of features}}). The aggregation is always causal, i.e. only past sample
points are aggregated. The first window does not start before the trigger event. Consequently, the
first window terminates at ''Trigger event + Window length - 1''.
\item {\bf Temporal aggregation: Step width}: \hypertarget{CE_ZRKlassi_Aggregation_Schritt}{} \newline
step width for the distance between two sliding windows. If all sample points should be used, a
value of~''1'' must be chosen.
\item {\bf K5: use each x. sample point}: \hypertarget{CE_ZRKlassi_K5JederXTe}{} \newline
The classifier \textindex{K5} handles a training data set of a classification problem for time
series as a sample-wise classification problem with the time since the \textindex{trigger event} as
an additional feature. As a consequence, it has a high memory consumption because all sample points
are handled as separate data points. It might cause problems up to Matlab crashes especially for
\textindex{Support Vector Machines} (SVM). The option reduces the number of sample points for the
training of a K5 classifier.
\end{itemize}

\clearpage
 \begin{figure}[h] 
 \includegraphics[width=0.9\columnwidth]{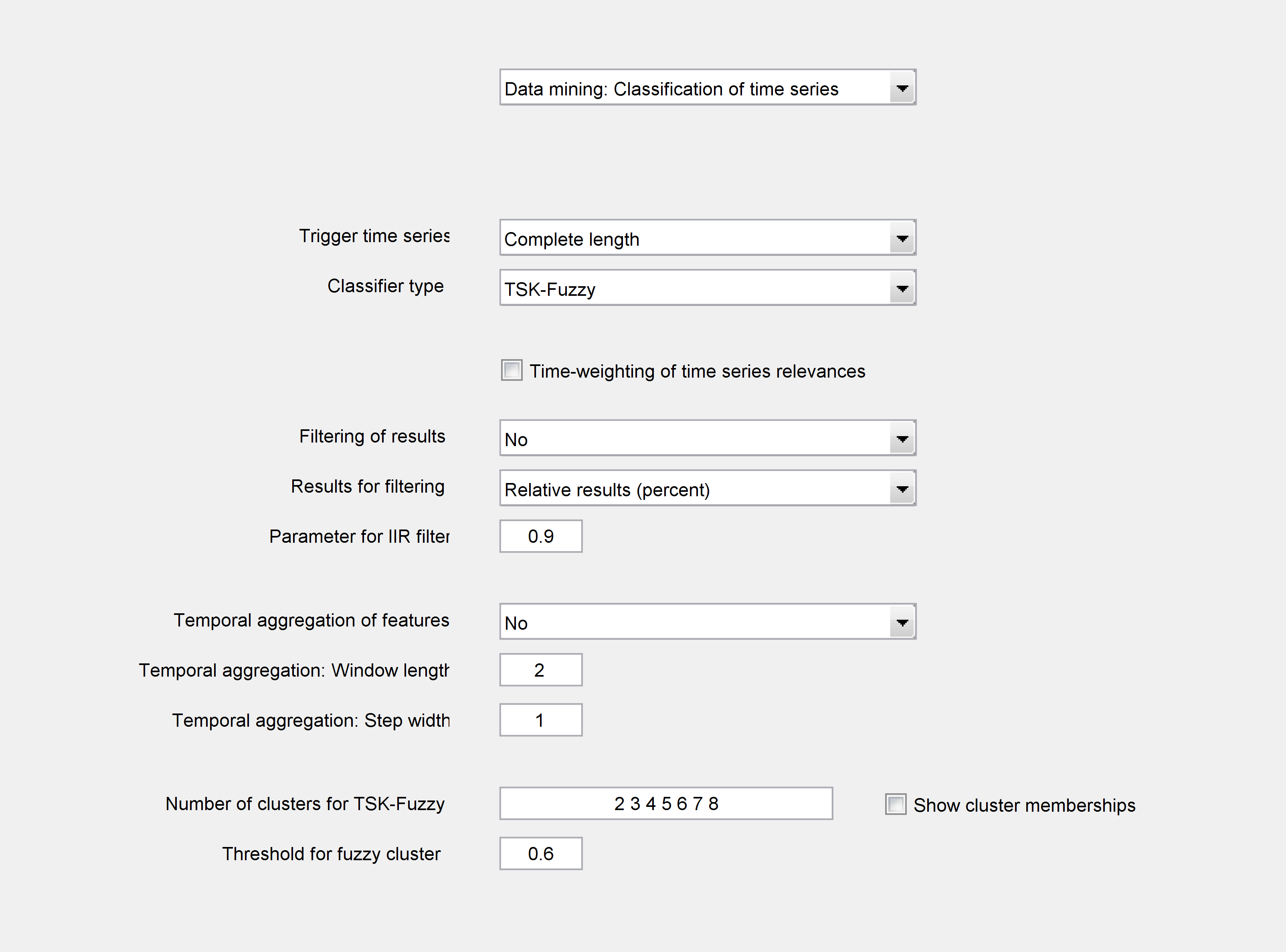}
 \caption{\label{fig:optionen12-15} Control elements for Data mining: Classification of time series - TSK-Fuzzy}
 \end{figure}
\begin{itemize}
\item {\bf Trigger time series}: \hypertarget{CE_ZRKlassi_Trigger}{} \newline
selects the \textindex{trigger time series} out of the existing time series. It is zero until the
\textindex{trigger} event, contains increasing values to measure the number of samples since the
trigger and zeros after the end of the recent event. If no trigger time series was selected, the
sample numbers of the complete time series are used: It starts with a value of one for the first
sample point and ends up with the length of the time series.
\item {\bf Classifier type}: \hypertarget{CE_ZRKlassi_Typ}{} \newline
selects a \textindex{classifier} type for the time series:

K1 contains only one classifier and ignores the trigger events. K5 is similar, but it uses the time
since the trigger event as additional input variable. K2 is a special kind of single feature
classifier for all sample points and aggregates results via a discriminant analysis. K3 and K4 have
separate sub-classifiers for each sample point since the trigger event. K3 has a fix selection of
time series, K4 a variable one for each sample point. TSK-Fuzzy is similar to K4, but it aggregates
similar sample points by means of fuzzy sets.

\cite{Burmeister06,Mikut08IEEE} give a detailed description of the classifiers.

The feature selection and the underlying classifiers (e.g. Bayes classifiers) are parameterized
using \hyperlink{CEO_DM_Grund}{\emph{Control elements:  Data mining: Classification of single features}} and \hyperlink{CE_Spezielle_Verfahren}{\emph{Control element: Data mining: Special methods - Method}}.
\item {\bf Time-weighting of time series relevances}: \hypertarget{CE_ZRKlassi_Rel_gewichten}{} \newline
gives the opportunity to prefer an earlier classification, e.g. for the feature selection of the
\textindex{K1} and \textindex{K3} classifier. The method modifies the feature relevances by
 $$R_k=\frac{R}{k-k_{trig}}, \text{ for } k > (k_{trig} + 0.15 \cdot k_{trig})$$
$k_{trig}$ is the sample point of the last trigger event. It increases feature relevances near the
\textindex{trigger event} in contrast to later ones.
\item {\bf Filtering of results}: \hypertarget{CE_ZRKlassi_Filter}{} \newline
select the method for filtering. The following options are available:

''No'' filter is used..

''\textindex{IIR-Filter}'' uses the filter parameters in \hyperlink{CE_ZRKlassi_IIR}{\emph{Control element: Data mining: Classification of time series - Parameter for IIR filter}}.

''Classification error for training data'' prefers samples with a better classification accuracy in
the training data set in contrast to samples with a lower accuracy. This results in time-variant
filter constants.

The ''Most frequent decision in a window'' selects the estimated class with the highest frequency
in a sliding window. The window length is defined in \hyperlink{CE_ZRKlassi_Aggregation}{\emph{Control element: Data mining: Classification of time series - Temporal aggregation of features}}. The first
sample points with a sample number up to the window length use all sample points up to this time
for decision.
\item {\bf Results for filtering}: \hypertarget{CE_ZRKlassi_Filtererg}{} \newline
switches between absolute and relative values as input values for the fusion by filtering. As an
example, absolute values of a \textindex{Bayes classifier} are the estimated  a-posteriori
probabilities, relative values the percentage of the estimated a-posteriori probabilities for the
given class relative to all classes.
\item {\bf Parameter for IIR filter}: \hypertarget{CE_ZRKlassi_IIR}{} \newline
set the parameter for the IIR filtering for the temporal fusion of classification results in time
series classification (see \hyperlink{CE_ZRKlassi_Filter}{\emph{Control element: Data mining: Classification of time series - Filtering of results}}).
\item {\bf Temporal aggregation of features}: \hypertarget{CE_ZRKlassi_Aggregation}{} \newline
tunes the temporal aggregation of sample points before the classification. The number and the names
of time series remain unchanged.

A selection of ''No'' does not use a temporal aggregation.

The ''Mean value'' is computed in a sliding window defined by
\hyperlink{CE_ZRKlassi_Aggregation_Groesse}{\emph{Control element: Data mining: Classification of time series - Temporal aggregation: Window length}} and \hyperlink{CE_ZRKlassi_Aggregation_Schritt}{\emph{Control element: Data mining: Classification of time series - Temporal aggregation: Step width}}.

''Minimum'', ''Maximum'' and ''ROM'' (Range of Motion) work with the same window, but with
different operators.

The aggregation is always causal, i.e. only past sample points are aggregated. The first
window does not start before the trigger event.
\item {\bf Temporal aggregation: Window length}: \hypertarget{CE_ZRKlassi_Aggregation_Groesse}{} \newline
defines the length of the sliding window for the temporal aggregation of features (see also
\hyperlink{CE_ZRKlassi_Aggregation}{\emph{Control element: Data mining: Classification of time series - Temporal aggregation of features}}). The aggregation is always causal, i.e. only past sample
points are aggregated. The first window does not start before the trigger event. Consequently, the
first window terminates at ''Trigger event + Window length - 1''.
\item {\bf Temporal aggregation: Step width}: \hypertarget{CE_ZRKlassi_Aggregation_Schritt}{} \newline
step width for the distance between two sliding windows. If all sample points should be used, a
value of~''1'' must be chosen.
\item {\bf Number of clusters for TSK-Fuzzy}: \hypertarget{CE_ZRKlassi_TSK_Anz}{} \newline
The \textindex{TSK fuzzy classifier} computes clusters to determine features with similar feature
relevances vs. time. The parameter sets the number of clusters to be found. An interval value (e.g.
2:5) computes a set of clusters for each candidate number of clusters. The number with the first
local minimum of the \textindex{separation} index will be selected.
\item {\bf Show cluster memberships}: \hypertarget{CE_ZRKlassi_TSK_ZGH}{} \newline
shows the result of the clustering by the search for time regions. It finds similar feature
relevances for the time series classifier.
\item {\bf Threshold for fuzzy cluster}: \hypertarget{CE_ZRKlassi_TSK_Schwellwert}{} \newline
contains a threshold used for deletion of clusters representing very short regions. It represents
the maximal value of the membership function. A large value simplifies the resulting classifier by
the reduction of the number of linguistic terms for the \textindex{TSK fuzzy classifier}. A too low
value might lead to an oversimplification.
\end{itemize}

\clearpage\clearpage

\section{Control elements for 'Data mining: Regression'}

 \begin{figure}[h] 
 \includegraphics[width=0.9\columnwidth]{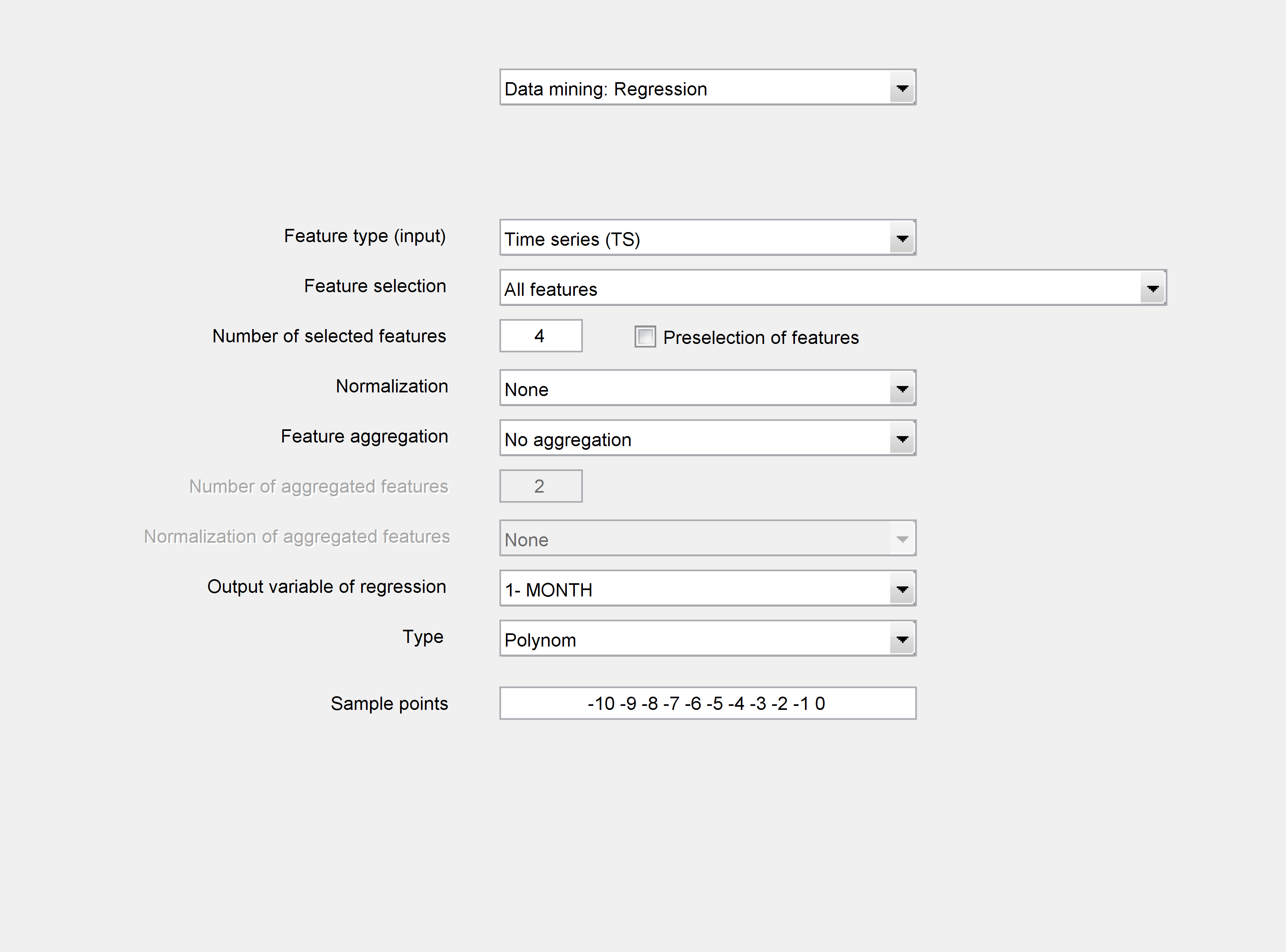}
 \caption{\label{fig:optionen13-16} Control elements for Data mining: Regression - Time series (TS)}
 \end{figure}
\begin{itemize}
\item {\bf Feature type (input)}: \hypertarget{CE_Regression_ZREM}{} \newline
switches between a \textindex{regression model} for single features with the model $y = f (x_1,
..., x_s)$ and for time series with the model $y[k] = f (x_1 [k-i], ..., x_1 [k-j], ..., x_s [k-i],
..., x_s [k-j]$). The selection of input variables $x_i$ controls
\hyperlink{CE_Regression_Merkmalsauswahl}{\emph{Control element: Data mining: Regression - Feature selection}}, the selection of output variables is done by
\hyperlink{CE_Regression_Output}{\emph{Control element: Data mining: Regression - Output variable of regression}}. For time series, the sample points  $i,...,j$ for the time shift are
selected by \hyperlink{CE_Regression_Abtastpunkte}{\emph{Control element: Data mining: Regression - Sample points}}.

The selection of the related model is only possible, if single features resp. time series exist in the
project.
\item {\bf Feature selection}: \hypertarget{CE_Regression_Merkmalsauswahl}{} \newline
selects the input variables of the regression model. Possible selections are ''All features'', ''Selected features'' (in \hyperlink{CE_Auswahl_EM}{\emph{Control element: Single features - Selection of single feature(s) (SF)}} or \hyperlink{CE_Auswahl_ZR}{\emph{Control element: Time series: General options - Selection of time series (TS)}}) and automatically by univariate resp. multivariate regression coefficients. To set the number of automatically selected features, use \hyperlink{CE_Regression_AnzahlMerkmale}{\emph{Control element: Data mining: Regression - Number of selected features}}.
\item {\bf Number of selected features}: \hypertarget{CE_Regression_AnzahlMerkmale}{} \newline
defines the number of selected features by an automatic feature selection. For time series, each combination
of time series and sample points is handled as separate feature. As an example, $x_1[k-1]$ and $x_1[k-2]$ are
two features.

The value will be ignored if ''All features'' or ''Selected features'' are chosen in
\hyperlink{CE_Regression_Merkmalsauswahl}{\emph{Control element: Data mining: Regression - Feature selection}}.
\item {\bf Preselection of features}: \hypertarget{CE_Regression_Preselection}{} \newline
preselects all manually selected features (by \hyperlink{CE_Auswahl_EM}{\emph{Control element: Single features - Selection of single feature(s) (SF)}}) for an automatic
\textindex{feature selection} for regression of single features.
\item {\bf Normalization}: \hypertarget{CE_Regression_Normierung_Merkmale}{} \newline
performs a normalization of features before a regression. It improves the results especially for Artificial
Neural Networks. Two approaches are implemented; to the interval $[0,1]$ or to a mean value of zero and a
standard deviation of one under the assumption of a normal distribution.
\item {\bf Feature aggregation}: \hypertarget{CE_Regression_Merkmalsaggregation}{} \newline
defines the method for the linear \textindex{feature aggregation} by means of a multiplication with a
weighting matrix. Possible options are ''No aggregation'',  a ''\textindex{Principal Component Analysis}'' or
the sum resp. mean value.
\item {\bf Number of aggregated features}: \hypertarget{CE_Regression_Anzahl_Aggregiert}{} \newline
defines the number of aggregated features ($s_d$) in a regression (see e.g. \cite{Reischl06}). The value is
ignored if ''No aggregation'' is chosen in \hyperlink{CE_Regression_Merkmalsaggregation}{\emph{Control element: Data mining: Regression - Feature aggregation}}. A value of $s_d=1$ is
always used for ''Sum'' and ''Mean value''.
\item {\bf Normalization of aggregated features}: \hypertarget{CE_Normierung_Aggregiert_Regr}{} \newline
performs a normalization of aggregated features before a regression. It improves the results especially for
Artificial Neural Networks. Two approaches are implemented; to the interval $[0,1]$ or to a mean value of
zero and a standard deviation of one under the assumption of a normal distribution.
\item {\bf Output variable of regression}: \hypertarget{CE_Regression_Output}{} \newline
selects a single feature or a time series at sample point $[k]$ as output of the
\textindex{regression model}. \hyperlink{CE_Regression_ZREM}{\emph{Control element: Data mining: Regression - Feature type (input)}} switches between single features and time
series.
\item {\bf Type}: \hypertarget{CE_Regression_Typ}{} \newline
selects the method for the \textindex{regression model}. The methods are parameterized in detail
using \hyperlink{CEO_DM_Klass}{\emph{Control elements:  Data mining: Special methods}}.
\item {\bf Sample points}: \hypertarget{CE_Regression_Abtastpunkte}{} \newline
defines the time shift for the regression of time series. The values must not be positive avoiding acausal
models. If a time series is input and output variable, the value $k=0$ is deleted for this time series.

Example:

The values  ''0 -1 -24'' with the time series  $x_1, x_2$ as input variables and the time series $x_1$ as
output variable are chosen. The resulting model is $x_1[k] = f (x_1 [k-1], x_1 [k-24], x_2 [k], x_2 [k-1],
x_2 [k-24])$.
\end{itemize}

\clearpage\clearpage

\section{Control elements for 'Data mining: Clustering'}

 \begin{figure}[h] 
 \includegraphics[width=0.9\columnwidth]{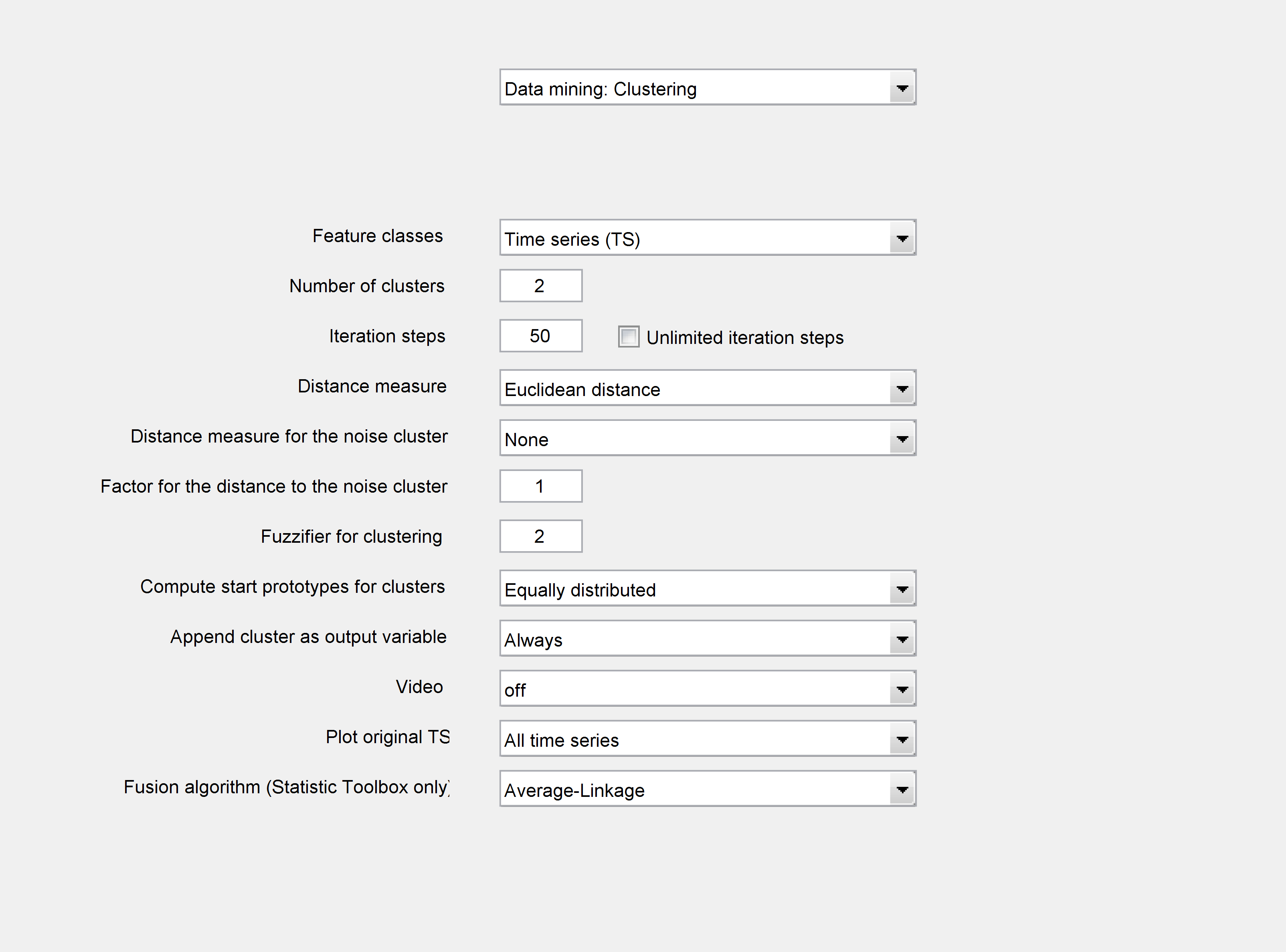}
 \caption{\label{fig:optionen14-17} Control elements for Data mining: Clustering }
 \end{figure}
\begin{itemize}
\item {\bf Feature classes}: \hypertarget{CE_Cluster_Merkmalsklassen}{} \newline
switches between a clustering of the selected time series (using the defined segment), the selected
single features, or a co-clustering of time series and single features.
\item {\bf Number of clusters}: \hypertarget{CE_Cluster_AnzCluster}{} \newline
defines the \textindex{number of clusters} to be computed by the \textindex{cluster algorithms}. A
multiple selection (e.g. 2:8) is possible. In this case, the number is defined according to the
first local minimum of the \textindex{separation index}.
\item {\bf Iteration steps}: \hypertarget{CE_Cluster_AnzSchritte}{} \newline
defines the maximal number of iteration steps of the cluster algorithm, if
\hyperlink{CE_Cluster_UnendlichSchritte}{\emph{Control element: Data mining: Clustering - Unlimited iteration steps}} is deactivated. If the algorithms converge before this step, the
algorithm is also terminated.
\item {\bf Unlimited iteration steps}: \hypertarget{CE_Cluster_UnendlichSchritte}{} \newline
executes the cluster algorithms until the internally defined convergence criterion was reached.
\item {\bf Distance measure}: \hypertarget{CE_Cluster_Abtandsmass}{} \newline
selects the metric for the distance between each data point and the cluster prototype.
\item {\bf Distance measure for the noise cluster}: \hypertarget{CE_Cluster_Method}{} \newline
defines the method for the definition of a noise cluster \cite{Dave02} for a \textindex{Fuzzy
C-Means method}. The option ''None'' does not generate a noise cluster. Alternatively, a new
cluster is defined to which all data points have the same distance. Outliers are mainly assigned to
this \textmainindex{noise cluster}. As a consequence, the influence of outliers to the other
clusters is reduced. Possible distance measures are the mean distance of all data points to all
other clusters \cite{Dave02} or the median of these distances. For the latter option, the distance
is scaled by \hyperlink{CE_Cluster_Factor}{\emph{Control element: Data mining: Clustering - Factor for the distance to the noise cluster}}. \index{outlier detection}
\item {\bf Factor for the distance to the noise cluster}: \hypertarget{CE_Cluster_Factor}{} \newline
defines a positive scaling factor for the distance to a noise cluster. A large value reduces the
assignment of data points to the noise cluster  (see also \hyperlink{CE_Cluster_Method}{\emph{Control element: Data mining: Clustering - Distance measure for the noise cluster}}).
\item {\bf Fuzzifier for clustering}: \hypertarget{CE_Fuzzifier}{} \newline
defines the \textindex{fuzzifier} in the fitness function of the \textindex{Fuzzy C-Means} method. The value
should be at least 1.1 for numerical reasons. A higher value generates increases the fuzziness of the cluster
membership values. For a value of 1, a crisp \textindex{k-means} method is used.
\item {\bf Compute start prototypes for clusters}: \hypertarget{CE_Cluster_Startzentren}{} \newline
selects the algorithms to compute \textindex{start prototypes} of the clusters. The possible
options are ''equally distributed'' clusters in the input space, ''random positions'' of start
prototypes, or ''random data points'' selected as start prototypes.
\item {\bf Append cluster as output variable}: \hypertarget{CE_Cluster_Anhaengen}{} \newline
adds the results of the cluster algorithms (the found crisp cluster memberships) to the project in
form of a new output variable.
\item {\bf Video}: \hypertarget{CE_Cluster_Video}{} \newline
shows the iterations of the cluster algorithms as video.
\item {\bf Plot original TS}: \hypertarget{CE_Cluster_OrigZR}{} \newline
shows the existing time series during the visualization of the cluster iterations. Possible options
are ''all time series'', ''mean values'', and ''none''.
\item {\bf Fusion algorithm (Statistic Toolbox only)}: \hypertarget{CE_Cluster_Fusionierung}{} \newline
select the fusion algorithm for the hierarchical cluster algorithm of the Statistic Toolbox of MATLAB (see
documentation of this toolbox).
\end{itemize}
\clearpage

\section{Control elements for 'Data mining: Special methods'}

 \begin{figure}[h] 
 \includegraphics[width=0.9\columnwidth]{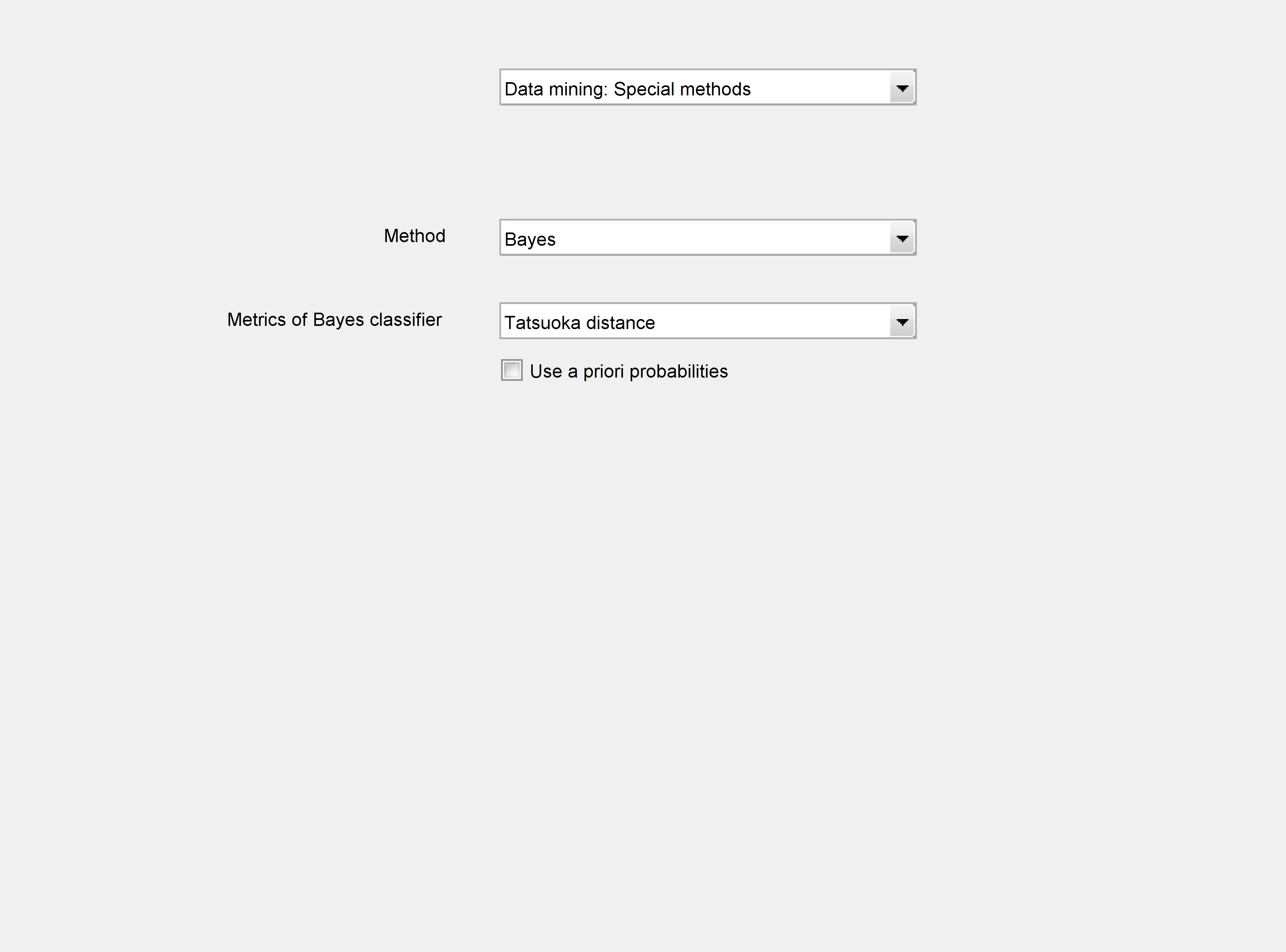}
 \caption{\label{fig:optionen15-18} Control elements for Data mining: Special methods - Bayes}
 \end{figure}
\begin{itemize}
\item {\bf Method}: \hypertarget{CE_Spezielle_Verfahren}{} \newline
parameterizes a method for the design of a classifier or \textindex{regression model}. The elements
are context-sensitive to the method selected by \hyperlink{CE_Spezielle_Verfahren}{\emph{Control element: Data mining: Special methods - Method}}.
\item {\bf Metrics of Bayes classifier}: \hypertarget{CE_Bayes_Metrik}{} \newline
defines the metric for the Bayes classifier.
\item {\bf Use a priori probabilities}: \hypertarget{CE_Bayes_Apriori}{} \newline
switches the use of estimated \textindex{a priori probabilities} of the different classes for the
Bayes classifier on resp. off. The estimation uses the training data set.
\end{itemize}

\clearpage
 \begin{figure}[h] 
 \includegraphics[width=0.9\columnwidth]{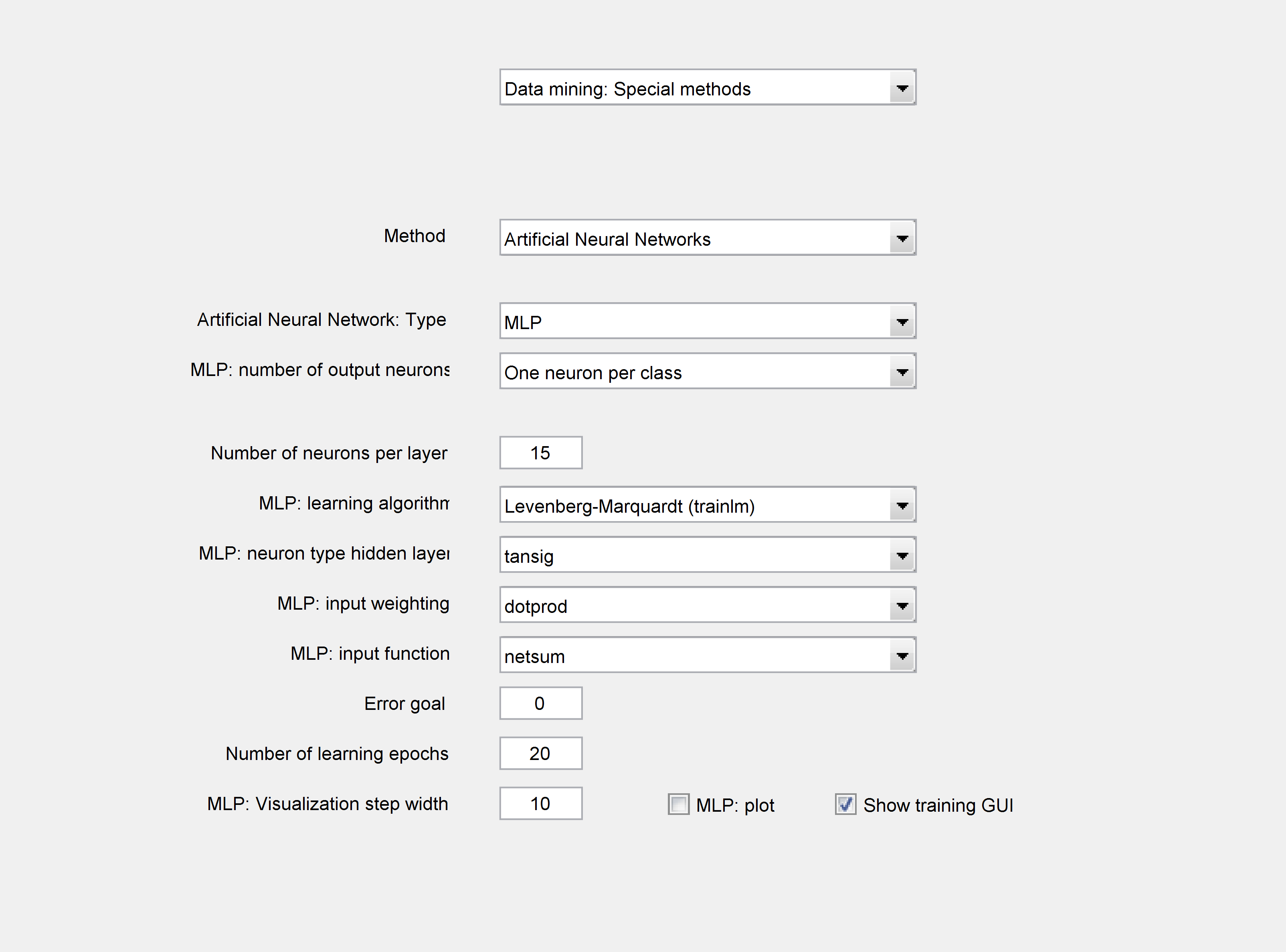}
 \caption{\label{fig:optionen15-19} Control elements for Data mining: Special methods - Artificial Neural Networks- MLP}
 \end{figure}
\begin{itemize}
\item {\bf Method}: \hypertarget{CE_Spezielle_Verfahren}{} \newline
parameterizes a method for the design of a classifier or \textindex{regression model}. The elements
are context-sensitive to the method selected by \hyperlink{CE_Spezielle_Verfahren}{\emph{Control element: Data mining: Special methods - Method}}.
\item {\bf Artificial Neural Network: Type}: \hypertarget{CE_NN_Typ}{} \newline
sets the type of the \textindex{Artificial Neural Network}. The available options include a
\textindex{Multi Layer Perceptron Net} (\textindex{MLP}) and a \textindex{Radial Base Function Net}
(\textindex{RBF}). The nets are used from the Neural Network Toolbox of Matlab.
\item {\bf MLP: number of output neurons}: \hypertarget{CE_MLP_Ausgang}{} \newline
defines the number of output neurons for \textindex{MLP} nets. Possible options are ''one neuron''
(values of the output variable: 1 to maximum class number) or ''one neuron per class'' (values of
the output variable: 0 ...1 for the related class membership). The latter case generates a higher
number of parameters in the net, but gives often better results. A main reason is a better fit for
more complex class topologies in the input space.

For regression models, only one output neuron will be used.
\item {\bf Number of neurons per layer}: \hypertarget{CE_MLP_Neuronen_Schicht}{} \newline
defines the number of neurons for the \textindex{hidden layer}.
\item {\bf MLP: learning algorithm}: \hypertarget{CE_MLP_Lern}{} \newline
defines the learning algorithms for the \textindex{MLP} net (see documentation of the Neural
Network Toolbox).
\item {\bf MLP: neuron type hidden layer}: \hypertarget{CE_MLP_Eingangsschicht}{} \newline
defines the type of neurons in the input layer of a \textindex{MLP} net (see documentation of the
Neural Network Toolbox).
\item {\bf MLP: input weighting}: \hypertarget{CE_MLP_Eingangswichtung}{} \newline
describes the method to combine the input of a neuron with the weights. The option ''dotprod'' uses
a multiplication.
\item {\bf MLP: input function}: \hypertarget{CE_MLP_Eingangsfunktion}{} \newline
describes the method to combine the weighted inputs of a neuron. The standard option ''netsum'' adds all weighted inputs.
\item {\bf Error goal}: \hypertarget{CE_MLP_Errorgoal}{} \newline
defines the error goal for the neural net. If the goal is reached, the design process will be terminated.
\item {\bf Number of learning epochs}: \hypertarget{CE_MLP_Lernepochen}{} \newline
defines the maximum number of learning epochs for \textindex{MLP} nets (see documentation of the
Neural Network Toolbox).
\item {\bf MLP: Visualization step width}: \hypertarget{CE_MLP_Internal_Stepwidth}{} \newline
defines the number of iterations between the visualizations of the MLP net. The value is only
relevant if \hyperlink{CE_MLP_Zeichnen}{\emph{Control element: Data mining: Special methods - MLP: plot}} is activated.

The design process of a neural net is restarted after each visualization with the last net as
starting parameter. However, some internal strategy parameter for the design might be lost for some
learning algorithms.  Larger values mostly result in lower errors and faster convergence.
\item {\bf MLP: plot}: \hypertarget{CE_MLP_Zeichnen}{} \newline
switches plots for the recent learning progress of \textindex{MLP} nets on or off (see
documentation of the Neural Network Toolbox).
\item {\bf Show training GUI}: \hypertarget{CE_MLP_ShowGUI}{} \newline
switches the use of the Learning GUI window of the Neural Network Toolbox on or off.
\end{itemize}

\clearpage
 \begin{figure}[h] 
 \includegraphics[width=0.9\columnwidth]{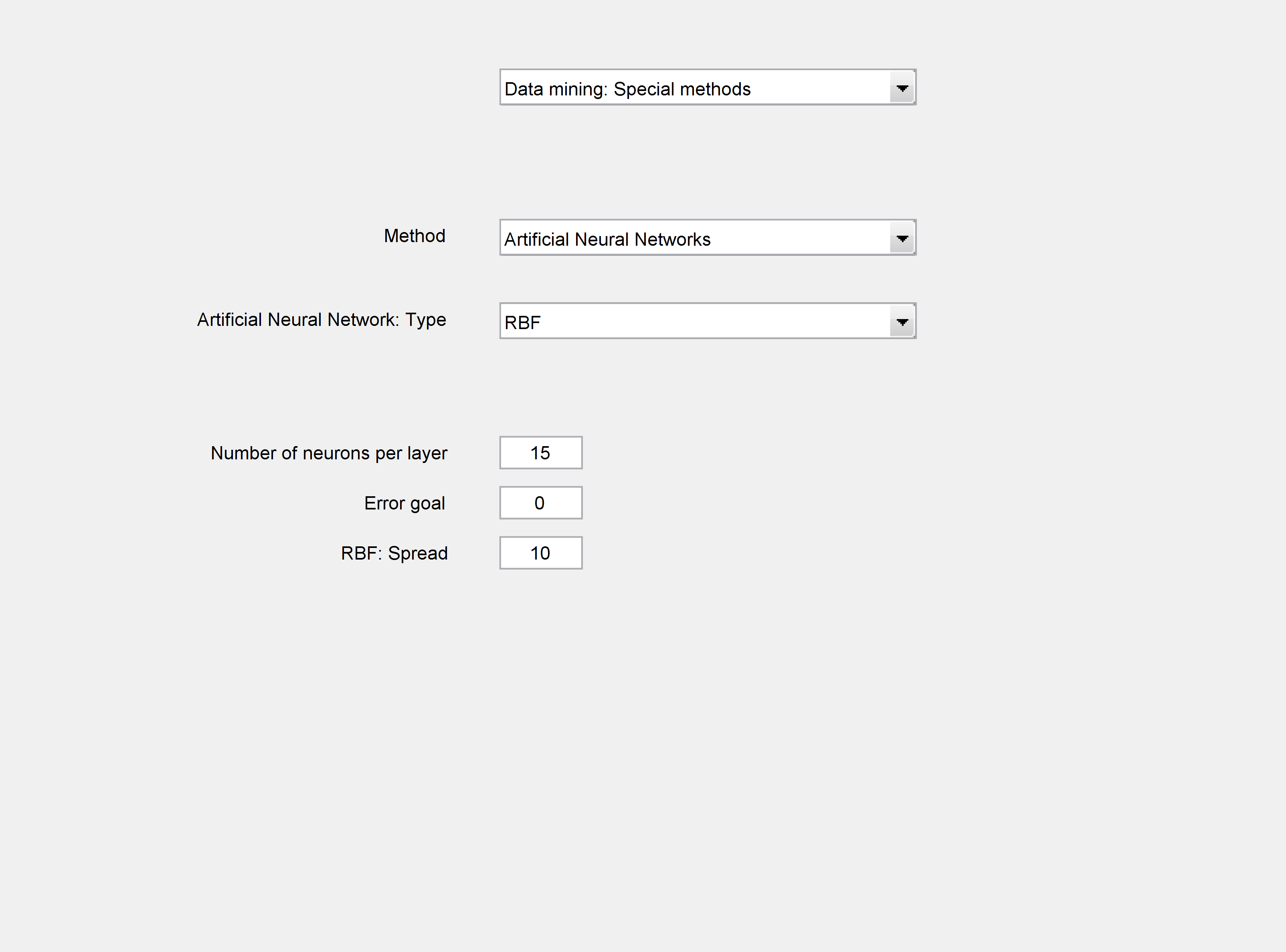}
 \caption{\label{fig:optionen15-20} Control elements for Data mining: Special methods - Artificial Neural Networks- RBF}
 \end{figure}
\begin{itemize}
\item {\bf Method}: \hypertarget{CE_Spezielle_Verfahren}{} \newline
parameterizes a method for the design of a classifier or \textindex{regression model}. The elements
are context-sensitive to the method selected by \hyperlink{CE_Spezielle_Verfahren}{\emph{Control element: Data mining: Special methods - Method}}.
\item {\bf Artificial Neural Network: Type}: \hypertarget{CE_NN_Typ}{} \newline
sets the type of the \textindex{Artificial Neural Network}. The available options include a
\textindex{Multi Layer Perceptron Net} (\textindex{MLP}) and a \textindex{Radial Base Function Net}
(\textindex{RBF}). The nets are used from the Neural Network Toolbox of Matlab.
\item {\bf Number of neurons per layer}: \hypertarget{CE_MLP_Neuronen_Schicht}{} \newline
defines the number of neurons for the \textindex{hidden layer}.
\item {\bf Error goal}: \hypertarget{CE_MLP_Errorgoal}{} \newline
defines the error goal for the neural net. If the goal is reached, the design process will be terminated.
\item {\bf RBF: Spread}: \hypertarget{CE_RBF_Spreizung}{} \newline
defines the spread of the Radial Base Function. A larger spread smoothes the function
approximation. Too large spreads can cause numerical problems.
\end{itemize}

\clearpage
 \begin{figure}[h] 
 \includegraphics[width=0.9\columnwidth]{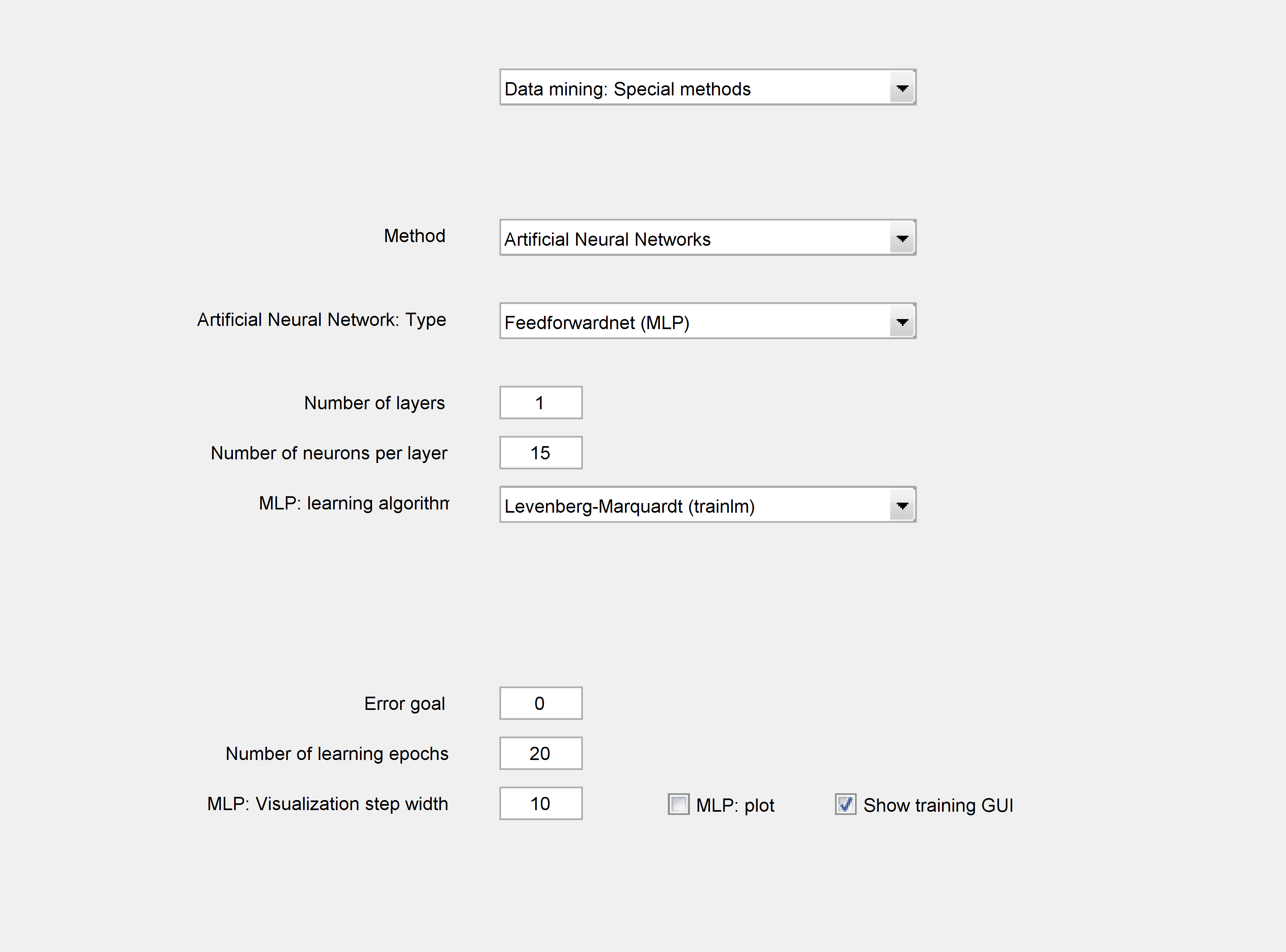}
 \caption{\label{fig:optionen15-21} Control elements for Data mining: Special methods - Artificial Neural Networks- Feedforwardnet (MLP)}
 \end{figure}
\begin{itemize}
\item {\bf Method}: \hypertarget{CE_Spezielle_Verfahren}{} \newline
parameterizes a method for the design of a classifier or \textindex{regression model}. The elements
are context-sensitive to the method selected by \hyperlink{CE_Spezielle_Verfahren}{\emph{Control element: Data mining: Special methods - Method}}.
\item {\bf Artificial Neural Network: Type}: \hypertarget{CE_NN_Typ}{} \newline
sets the type of the \textindex{Artificial Neural Network}. The available options include a
\textindex{Multi Layer Perceptron Net} (\textindex{MLP}) and a \textindex{Radial Base Function Net}
(\textindex{RBF}). The nets are used from the Neural Network Toolbox of Matlab.
\item {\bf Number of layers}: \hypertarget{CE_MLP_Neuronen_NumberofLayers}{} \newline
defines the number of hidden layers of feedforward nets.
\item {\bf Number of neurons per layer}: \hypertarget{CE_MLP_Neuronen_Schicht}{} \newline
defines the number of neurons for the \textindex{hidden layer}.
\item {\bf MLP: learning algorithm}: \hypertarget{CE_MLP_Lern}{} \newline
defines the learning algorithms for the \textindex{MLP} net (see documentation of the Neural
Network Toolbox).
\item {\bf Error goal}: \hypertarget{CE_MLP_Errorgoal}{} \newline
defines the error goal for the neural net. If the goal is reached, the design process will be terminated.
\item {\bf Number of learning epochs}: \hypertarget{CE_MLP_Lernepochen}{} \newline
defines the maximum number of learning epochs for \textindex{MLP} nets (see documentation of the
Neural Network Toolbox).
\item {\bf MLP: Visualization step width}: \hypertarget{CE_MLP_Internal_Stepwidth}{} \newline
defines the number of iterations between the visualizations of the MLP net. The value is only
relevant if \hyperlink{CE_MLP_Zeichnen}{\emph{Control element: Data mining: Special methods - MLP: plot}} is activated.

The design process of a neural net is restarted after each visualization with the last net as
starting parameter. However, some internal strategy parameter for the design might be lost for some
learning algorithms.  Larger values mostly result in lower errors and faster convergence.
\item {\bf MLP: plot}: \hypertarget{CE_MLP_Zeichnen}{} \newline
switches plots for the recent learning progress of \textindex{MLP} nets on or off (see
documentation of the Neural Network Toolbox).
\item {\bf Show training GUI}: \hypertarget{CE_MLP_ShowGUI}{} \newline
switches the use of the Learning GUI window of the Neural Network Toolbox on or off.
\end{itemize}

\clearpage
 \begin{figure}[h] 
 \includegraphics[width=0.9\columnwidth]{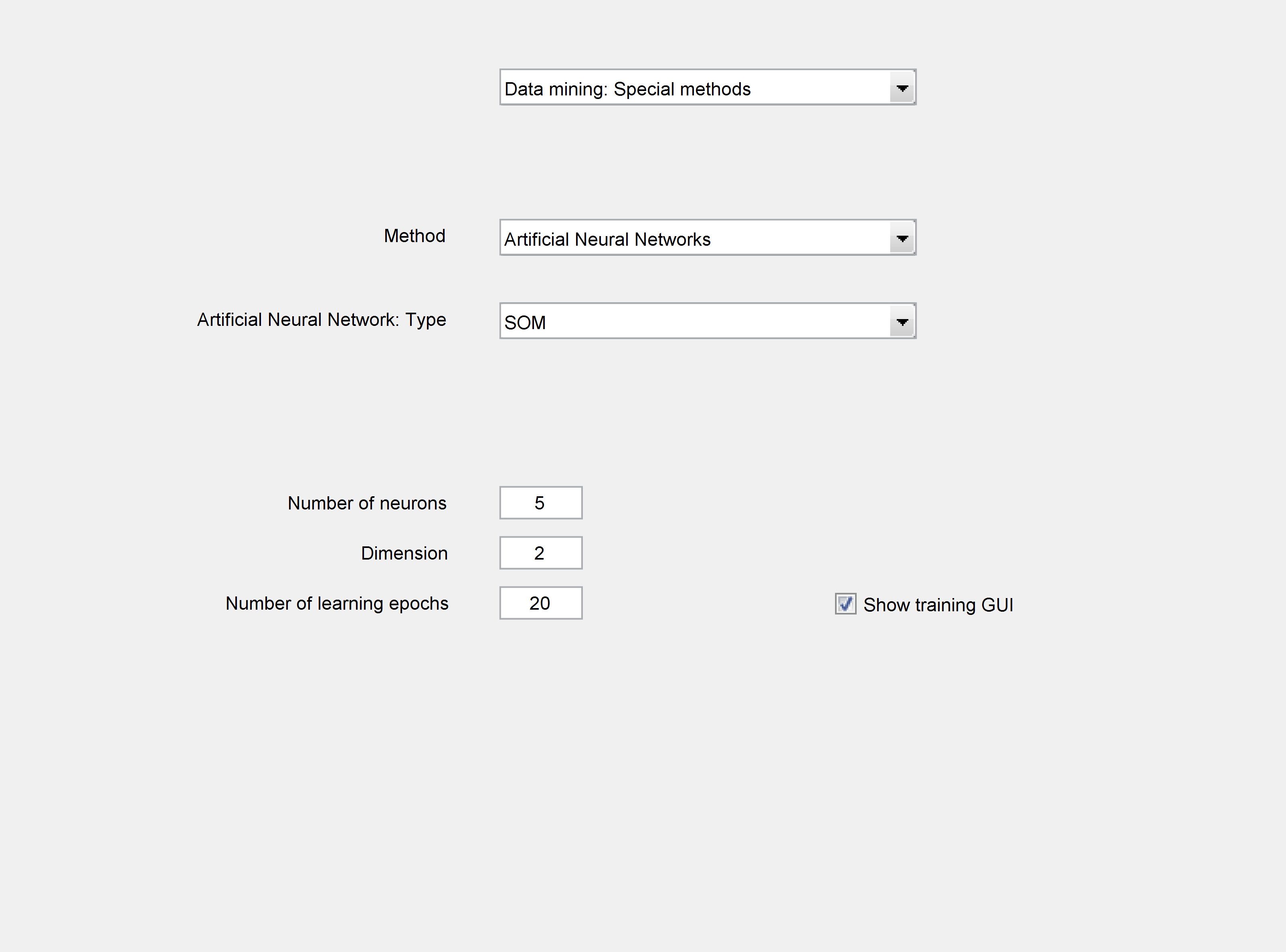}
 \caption{\label{fig:optionen15-22} Control elements for Data mining: Special methods - Artificial Neural Networks- SOM}
 \end{figure}
\begin{itemize}
\item {\bf Method}: \hypertarget{CE_Spezielle_Verfahren}{} \newline
parameterizes a method for the design of a classifier or \textindex{regression model}. The elements
are context-sensitive to the method selected by \hyperlink{CE_Spezielle_Verfahren}{\emph{Control element: Data mining: Special methods - Method}}.
\item {\bf Artificial Neural Network: Type}: \hypertarget{CE_NN_Typ}{} \newline
sets the type of the \textindex{Artificial Neural Network}. The available options include a
\textindex{Multi Layer Perceptron Net} (\textindex{MLP}) and a \textindex{Radial Base Function Net}
(\textindex{RBF}). The nets are used from the Neural Network Toolbox of Matlab.
\item {\bf Number of neurons}: \hypertarget{CE_SomNeurons}{} \newline
defines the number of neurons per dimension for the design of \textindex{Self Organizing Maps} with \hyperlink{MI_SomEn}{\emph{Data mining - Self Organizing Maps - Design}}. The number is used for all dimensions, e.g., a value of 8 with a dimensionality of 2 means the design of a net with 8x8 neurons.
\item {\bf Dimension}: \hypertarget{CE_SomDimension}{} \newline
defines the dimensionality for the design of \textindex{Self Organizing Maps} with \hyperlink{MI_SomEn}{\emph{Data mining - Self Organizing Maps - Design}}.
\item {\bf Number of learning epochs}: \hypertarget{CE_MLP_Lernepochen}{} \newline
defines the maximum number of learning epochs for \textindex{MLP} nets (see documentation of the
Neural Network Toolbox).
\item {\bf Show training GUI}: \hypertarget{CE_MLP_ShowGUI}{} \newline
switches the use of the Learning GUI window of the Neural Network Toolbox on or off.
\end{itemize}

\clearpage
 \begin{figure}[h] 
 \includegraphics[width=0.9\columnwidth]{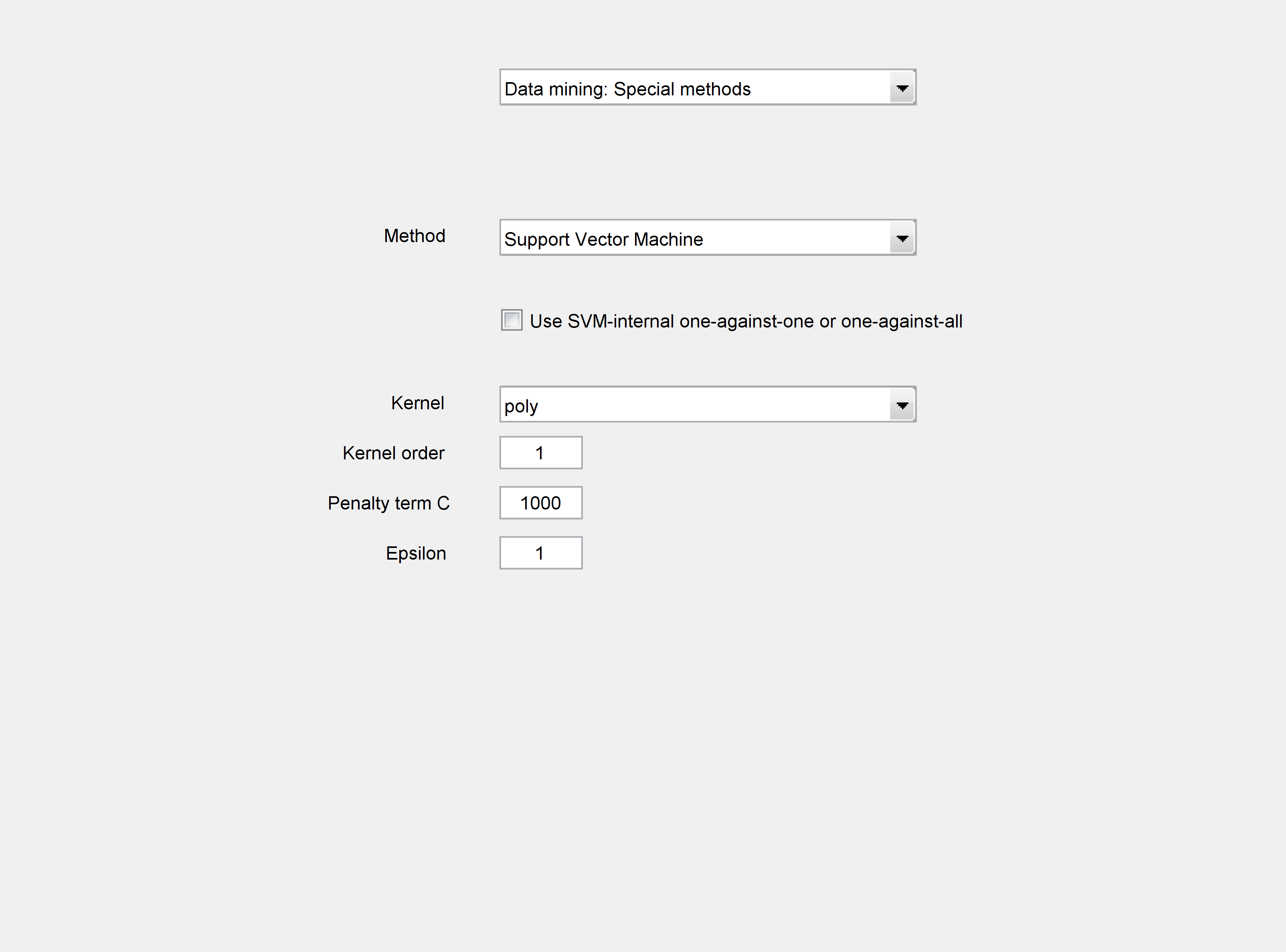}
 \caption{\label{fig:optionen15-23} Control elements for Data mining: Special methods - Support Vector Machine}
 \end{figure}
\begin{itemize}
\item {\bf Method}: \hypertarget{CE_Spezielle_Verfahren}{} \newline
parameterizes a method for the design of a classifier or \textindex{regression model}. The elements
are context-sensitive to the method selected by \hyperlink{CE_Spezielle_Verfahren}{\emph{Control element: Data mining: Special methods - Method}}.
\item {\bf Use SVM-internal one-against-one or one-against-all}: \hypertarget{CE_SVM_intern_oax}{} \newline
uses the internal decomposition of multi-class problems of the SVM toolbox. The function is faster
and uses less memory compared to the external decomposition of SciXMiner. This external
decomposition works with the one-against-one resp. the one-against-all option in
\hyperlink{CE_Klassifikation_Mehrklassen}{\emph{Control element: Data mining: Classification of single features - Multi-class problems}}.
\item {\bf Kernel}: \hypertarget{CE_SVM_Kernel}{} \newline
defines the used kernel function for \textindex{Support Vector Machines} \cite{Burges98}.
Homogenous polynomial kernels (without bias term), polynomial kernels (with bias term), and
Gaussian radial base functions are implemented. The order is tuned by  \hyperlink{CE_SVM_Ordnung}{\emph{Control element: Data preprocessing - Kernel order}}. A
first-order polynomial kernel means a linear hyperplane.
\item {\bf Kernel order}: \hypertarget{CE_SVM_Ordnung}{} \newline
set the order of the kernel of the SVM. It is used for the classification with the SVM and by
the SVM-based outlier detection.
\item {\bf Penalty term C}: \hypertarget{CE_SVM_Strafterm}{} \newline
tunes the compromise between a small number of classification errors and the simplification of the
hyperplanes to discriminate classes for the SVM. Larger values of $C$ lead to a higher number of
classification errors to simplify the hyperplanes.
\item {\bf Epsilon}: \hypertarget{CE_SVM_Epsilon}{} \newline
defines the width of the insensitive region of the loss function in Support Vector Regression.
\end{itemize}

\clearpage
 \begin{figure}[h] 
 \includegraphics[width=0.9\columnwidth]{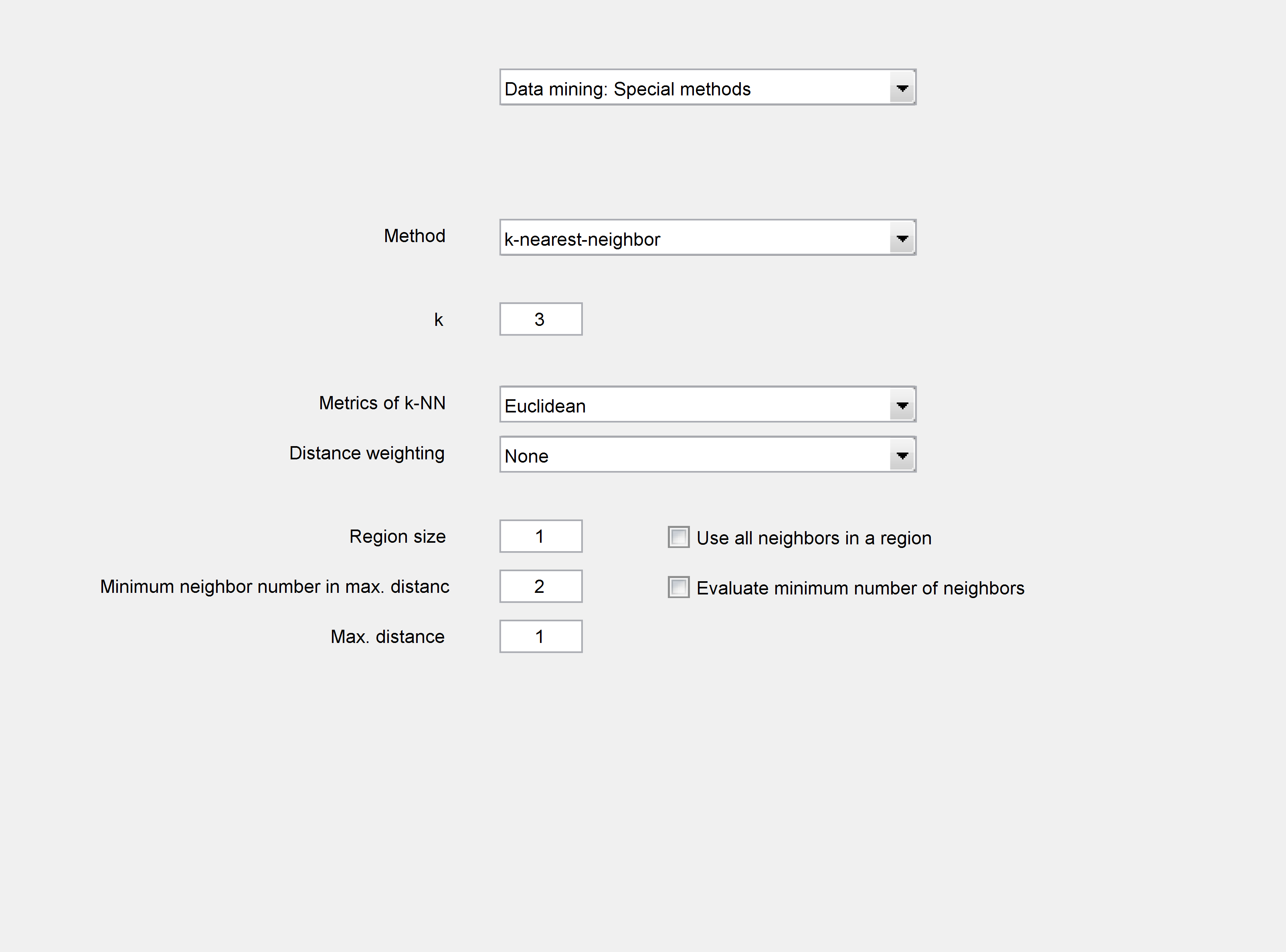}
 \caption{\label{fig:optionen15-24} Control elements for Data mining: Special methods - k-nearest-neighbor}
 \end{figure}
\begin{itemize}
\item {\bf Method}: \hypertarget{CE_Spezielle_Verfahren}{} \newline
parameterizes a method for the design of a classifier or \textindex{regression model}. The elements
are context-sensitive to the method selected by \hyperlink{CE_Spezielle_Verfahren}{\emph{Control element: Data mining: Special methods - Method}}.
\item {\bf k}: \hypertarget{CE_kNN_k}{} \newline
defines the number of the $k$ neighbors for the \textindex{$k$-nearest neighbor} classifier.
\item {\bf Metrics of k-NN}: \hypertarget{CE_kNN_Metrik}{} \newline
defines the metric to compute the nearest neighbors.
\item {\bf Distance weighting}: \hypertarget{CE_kNN_Wichtung}{} \newline
enables a weighting of different distances.

If ''None'' is selected, all neighbors contribute equally to the decision (standard option).

''Inverse linear'' prefers the nearest neighbor by a weighting with $1$. The farthest of the $k$
neighbors is weighted with $0$. All neighbors between have an equidistant weighting, e.g. $0.25,
0.5$ and $0.75$ for $k=4$.

''Inverse distance'' uses $1/Distance$ as weight to prefer neighbors with smaller distances. A
neighbor with distance 0 gets a maximum weight of  $1/1E-10$.

''Inverse exponential'' is similar but with more equal weights by using $e^{-d}$ with the distance
$d$.
\item {\bf Region size}: \hypertarget{CE_kNN_Region}{} \newline
specifies the type of environment around a data point. If the option \hyperlink{CE_kNN_AlleNachbarn}{\emph{Control element: Data mining: Special methods - Use all neighbors in a region}} was
activated, all neighbors in the defined environment are included as neighbors (and not necessary
only the k nearest).
\item {\bf Use all neighbors in a region}: \hypertarget{CE_kNN_AlleNachbarn}{} \newline
includes all neighbors in an environment defined by \hyperlink{CE_kNN_Region}{\emph{Control element: Data mining: Special methods - Region size}}. The classifier normalizes
all values to an interval $[0, \hdots, 1]$. Consequently, the value should be smaller than 1!

If less than k neighbors are localized in the environment, only the k nearest neighbors are used.
\item {\bf Minimum neighbor number in max. distance}: \hypertarget{CE_kNN_Mindestzahl_Nachbarn}{} \newline
defines the minimum number of neighbors in a maximum accepted distance defined by
\hyperlink{CE_kNN_MaxAbstand}{\emph{Control element: Data preprocessing - Max. distance}} to make a decision.
\item {\bf Evaluate minimum number of neighbors}: \hypertarget{CE_kNN_NachbarnPruefen}{} \newline
defines if a minimum number of neighbors in an environment (defined by a distance) should be used. If the number
is not reached and the option is activated, the decision is set to rejection. The distance is tuned
by \hyperlink{CE_kNN_MaxAbstand}{\emph{Control element: Data preprocessing - Max. distance}}, the number by \hyperlink{CE_kNN_Mindestzahl_Nachbarn}{\emph{Control element: Data mining: Special methods - Minimum neighbor number in max. distance}}.
\item {\bf Max. distance}: \hypertarget{CE_kNN_MaxAbstand}{} \newline
defines the accepted maximum distance for a neighborhood (see \hyperlink{CE_kNN_NachbarnPruefen}{\emph{Control element: Data mining: Special methods - Evaluate minimum number of neighbors}}).
\end{itemize}

\clearpage
 \begin{figure}[h] 
 \includegraphics[width=0.9\columnwidth]{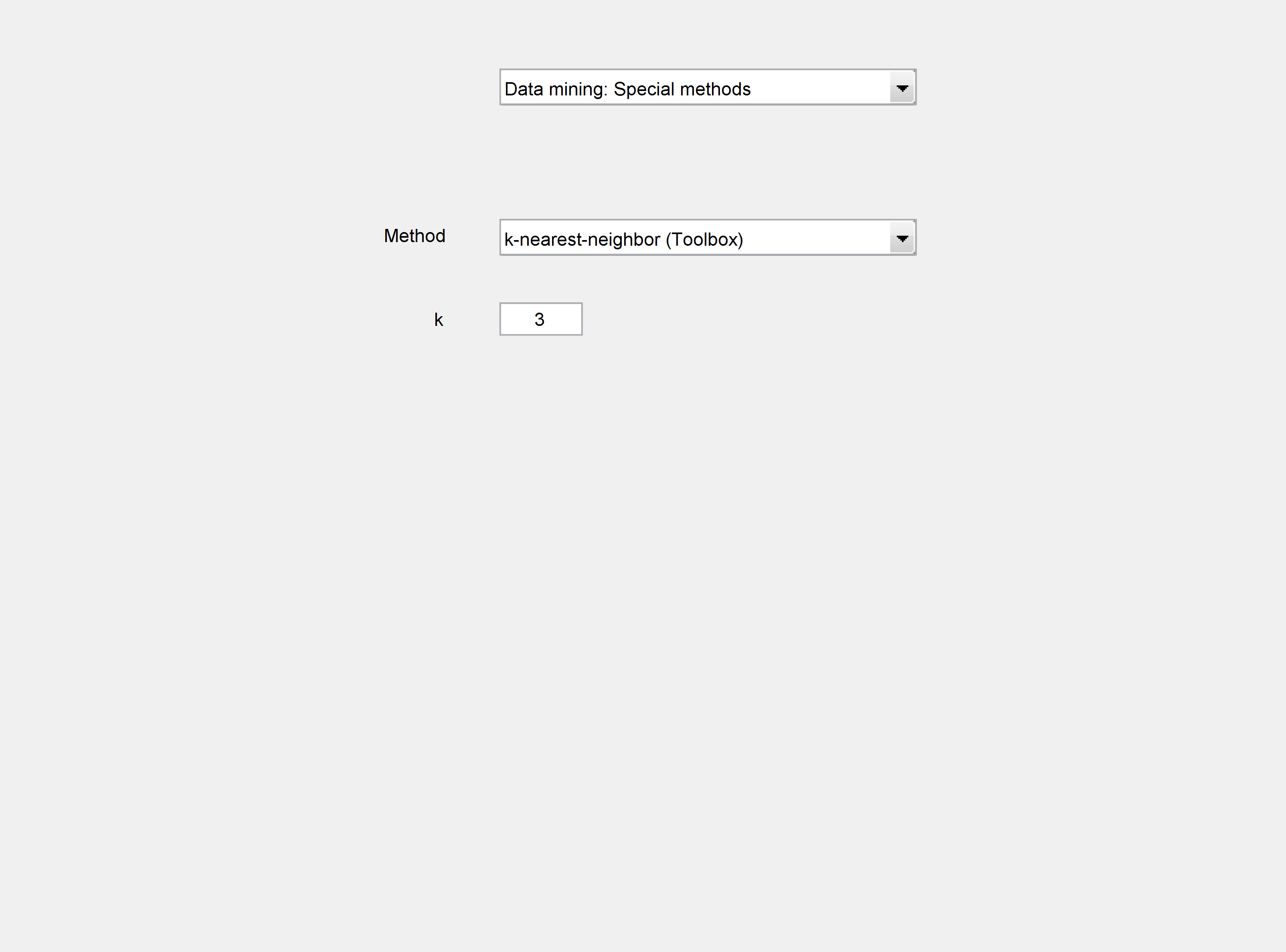}
 \caption{\label{fig:optionen15-25} Control elements for Data mining: Special methods - k-nearest-neighbor (Toolbox)}
 \end{figure}
\begin{itemize}
\item {\bf Method}: \hypertarget{CE_Spezielle_Verfahren}{} \newline
parameterizes a method for the design of a classifier or \textindex{regression model}. The elements
are context-sensitive to the method selected by \hyperlink{CE_Spezielle_Verfahren}{\emph{Control element: Data mining: Special methods - Method}}.
\item {\bf k}: \hypertarget{CE_kNN_k}{} \newline
defines the number of the $k$ neighbors for the \textindex{$k$-nearest neighbor} classifier.
\end{itemize}

\clearpage
 \begin{figure}[h] 
 \includegraphics[width=0.9\columnwidth]{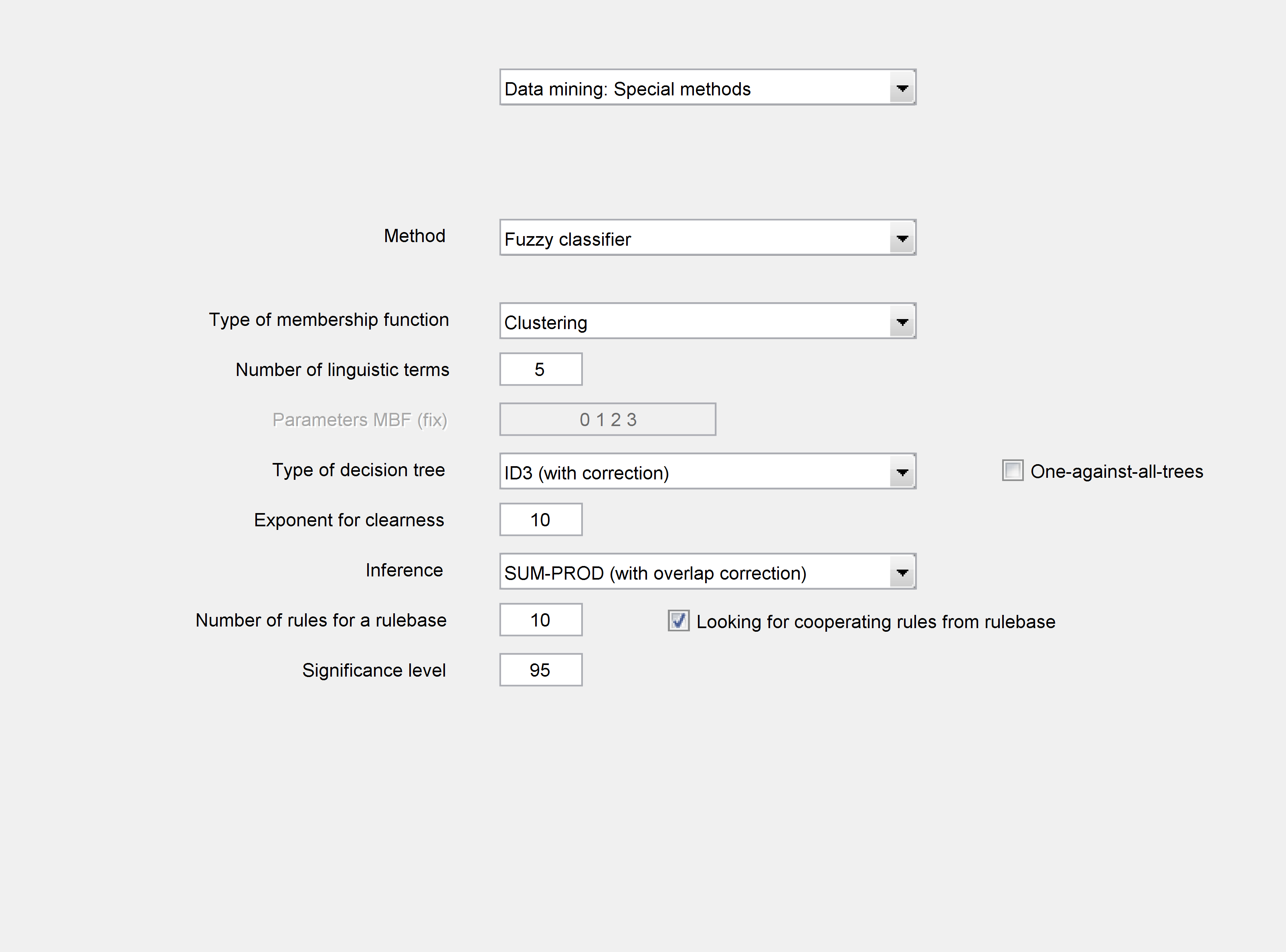}
 \caption{\label{fig:optionen15-26} Control elements for Data mining: Special methods - Fuzzy classifier}
 \end{figure}
\begin{itemize}
\item {\bf Method}: \hypertarget{CE_Spezielle_Verfahren}{} \newline
parameterizes a method for the design of a classifier or \textindex{regression model}. The elements
are context-sensitive to the method selected by \hyperlink{CE_Spezielle_Verfahren}{\emph{Control element: Data mining: Special methods - Method}}.
\item {\bf Type of membership function}: \hypertarget{CE_Fuzzy_TypeZGF}{} \newline
defines the algorithm for the design of a \textindex{membership function}  (see
\hyperlink{CE_Fuzzy_AnzLingTerme}{\emph{Control element: Data mining: Special methods - Number of linguistic terms}}).

''Median'' generates terms with approximately equal data point frequencies for all terms in the
training data set.

''Equal distribution'' generate terms with similar distances between the maximum values of the
membership functions.

''Clustering'' uses cluster prototypes as parameters.

''Fix'' set the parameters of the membership functions to the values defined by \hyperlink{CE_MBF_Fix}{\emph{Control element: Data mining: Special methods - Parameters MBF (fix)}}.

The choice ''with \textindex{interpretability}'' rounds the found parameters using a method
described in \cite{Mikut05} to improve the interpretability.

Remark:

All design methods set the random number generator of Matlab to a fix value. It improves the
reproducibility of found membership functions. Nevertheless, all methods with random elements might
be sensitive to the state of the random number generator (e.g. cluster algorithms with random start
clusters). It can also influence other methods using these membership functions, e.g. by different
selected features in a decision tree (e.g. \hyperlink{MI_EMAusw_Inform}{\emph{Data mining - Selection and evaluation of single features - Information theoretic measures}}). \index{information measures}
\index{start cluster}\index{cluster algorithms}
\item {\bf Number of linguistic terms}: \hypertarget{CE_Fuzzy_AnzLingTerme}{} \newline
defines the desired number of \textindex{linguistic terms} for the design of the membership
functions. The designed membership functions are \textindex{standard partitions}: The first and the
last term have trapezoidal membership functions, middle terms triangular membership functions. The
membership functions sum up to one for each value of the variable.

The term number might be automatically reduced by the designed method, e.g. if only two different
values exist and three terms should be designed. \index{membership function}
\item {\bf Parameters MBF (fix)}: \hypertarget{CE_MBF_Fix}{} \newline
defines the parameters of the membership functions if ''fix'' was chosen in \hyperlink{CE_Fuzzy_TypeZGF}{\emph{Control element: Data mining: Special methods - Type of membership function}}.
\item {\bf Type of decision tree}: \hypertarget{CE_Fuzzy_Entscheidungsbaum}{} \newline
defines the splitting criterion for the design of the \textindex{decision tree}. The type
\textindex{ID3} uses a maximized \textindex{mutual information}. The type \textindex{C4.5}
maximizes the mutual information divided by the \textindex{input entropy}. In addition, a
statistical correction of the entropy can be chosen. It reduces the influence of an overestimation
of feature relevances for small training data sets and avoids overfitted trees.

All trees use the design fuzzy membership functions instead of computed binary splits for each
node.
\item {\bf One-against-all-trees}: \hypertarget{CE_Fuzzy_Baum_klassenspezifisch}{} \newline
uses $C$ different decision trees for the rule generation. The trees are generated by a
\textindex{one-against-all} approach of a class against all other classes. \index{decision tree}
\item {\bf Exponent for clearness}: \hypertarget{CE_Fuzzy_Klarheit}{} \newline
defines the evaluation method for the pruning of fuzzy rules \cite{Mikut05}. An increasing
clearness prefers highly specialized error-free rules to more generalized rules with some errors.
\index{fuzzy system}
\item {\bf Inference}: \hypertarget{CE_Fuzzy_Inferenz}{} \newline
defines the inference method for the \textindex{fuzzy system}. The \textindex{MIN-MAX inference}
uses the minimum operator for all conjunctions (e.g. for the aggregation) and the maximum for all
disjunctions (e.g. for the accumulation of all rules with identical conclusions. The
\textindex{SUM-PROD inference} uses the bounded sum for all disjunctions and the product for all
conjunctions. An additional correction of overlapping rules \cite{MikutJG00:2} was
implemented for artifacts by rules with the same or concurrent conclusions in case of the SUM-PROD
inference.
\item {\bf Number of rules for a rulebase}: \hypertarget{CE_Fuzzy_NumberRulesRulebase}{} \newline
defines the maximal number of selected rules in a fuzzy rulebase.
\item {\bf Looking for cooperating rules from rulebase}: \hypertarget{CE_Fuzzy_RuleBaseSearch}{} \newline
switches the search for cooperating rule bases on or off. If the option is off, all found single rules are used for classification or regression.
\item {\bf Significance level}: \hypertarget{CE_Fuzzy_StatRele}{} \newline
influences the statistical evaluation of fuzzy rules. Values towards  100 \% tends to the generation of only few, but very confident rules.
\end{itemize}

\clearpage
 \begin{figure}[h] 
 \includegraphics[width=0.9\columnwidth]{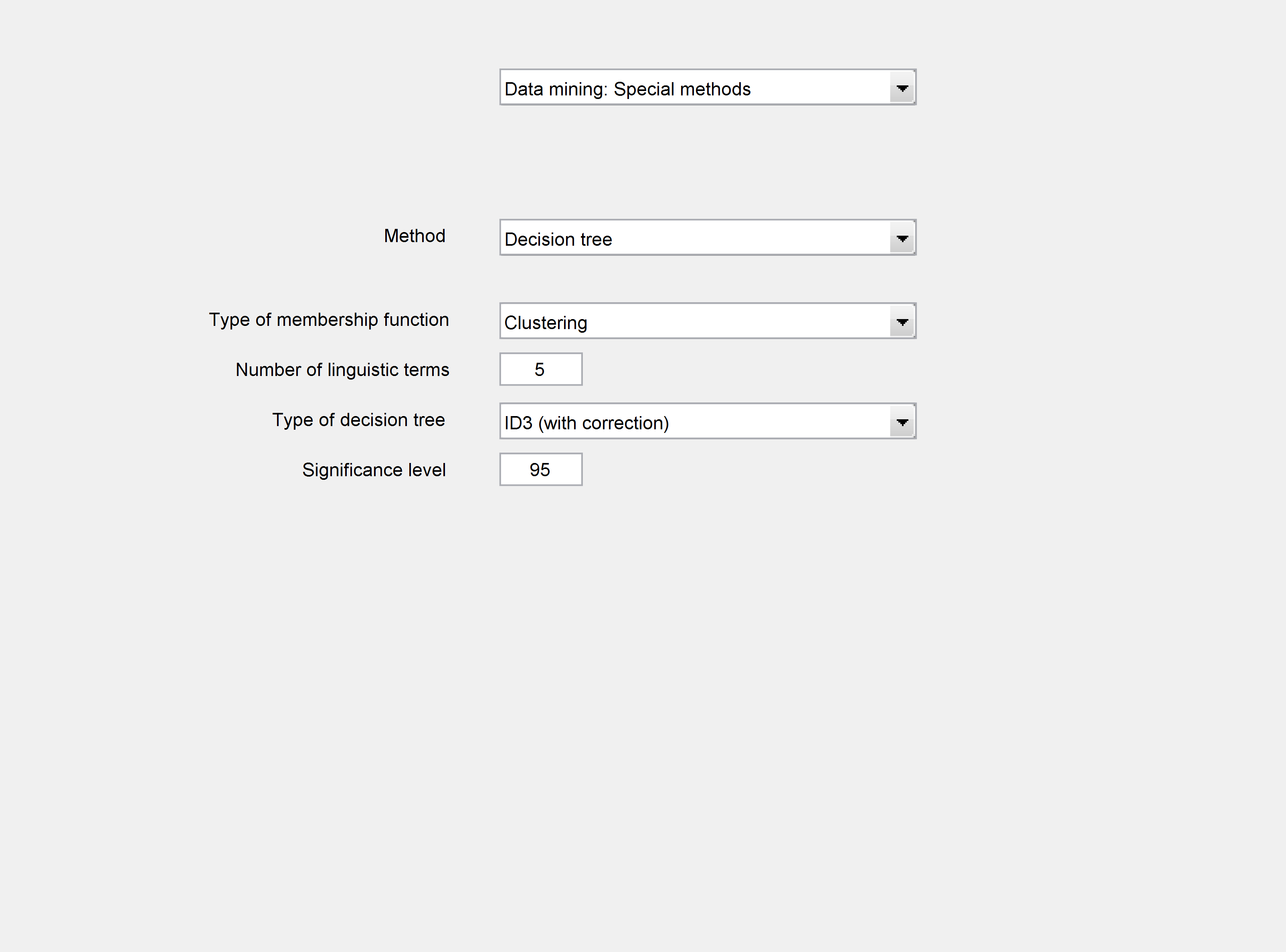}
 \caption{\label{fig:optionen15-27} Control elements for Data mining: Special methods - Decision tree}
 \end{figure}
\begin{itemize}
\item {\bf Method}: \hypertarget{CE_Spezielle_Verfahren}{} \newline
parameterizes a method for the design of a classifier or \textindex{regression model}. The elements
are context-sensitive to the method selected by \hyperlink{CE_Spezielle_Verfahren}{\emph{Control element: Data mining: Special methods - Method}}.
\item {\bf Type of membership function}: \hypertarget{CE_Fuzzy_TypeZGF}{} \newline
defines the algorithm for the design of a \textindex{membership function}  (see
\hyperlink{CE_Fuzzy_AnzLingTerme}{\emph{Control element: Data mining: Special methods - Number of linguistic terms}}).

''Median'' generates terms with approximately equal data point frequencies for all terms in the
training data set.

''Equal distribution'' generate terms with similar distances between the maximum values of the
membership functions.

''Clustering'' uses cluster prototypes as parameters.

''Fix'' set the parameters of the membership functions to the values defined by \hyperlink{CE_MBF_Fix}{\emph{Control element: Data mining: Special methods - Parameters MBF (fix)}}.

The choice ''with \textindex{interpretability}'' rounds the found parameters using a method
described in \cite{Mikut05} to improve the interpretability.

Remark:

All design methods set the random number generator of Matlab to a fix value. It improves the
reproducibility of found membership functions. Nevertheless, all methods with random elements might
be sensitive to the state of the random number generator (e.g. cluster algorithms with random start
clusters). It can also influence other methods using these membership functions, e.g. by different
selected features in a decision tree (e.g. \hyperlink{MI_EMAusw_Inform}{\emph{Data mining - Selection and evaluation of single features - Information theoretic measures}}). \index{information measures}
\index{start cluster}\index{cluster algorithms}
\item {\bf Number of linguistic terms}: \hypertarget{CE_Fuzzy_AnzLingTerme}{} \newline
defines the desired number of \textindex{linguistic terms} for the design of the membership
functions. The designed membership functions are \textindex{standard partitions}: The first and the
last term have trapezoidal membership functions, middle terms triangular membership functions. The
membership functions sum up to one for each value of the variable.

The term number might be automatically reduced by the designed method, e.g. if only two different
values exist and three terms should be designed. \index{membership function}
\item {\bf Type of decision tree}: \hypertarget{CE_Fuzzy_Entscheidungsbaum}{} \newline
defines the splitting criterion for the design of the \textindex{decision tree}. The type
\textindex{ID3} uses a maximized \textindex{mutual information}. The type \textindex{C4.5}
maximizes the mutual information divided by the \textindex{input entropy}. In addition, a
statistical correction of the entropy can be chosen. It reduces the influence of an overestimation
of feature relevances for small training data sets and avoids overfitted trees.

All trees use the design fuzzy membership functions instead of computed binary splits for each
node.
\item {\bf Significance level}: \hypertarget{CE_Fuzzy_StatRele}{} \newline
influences the statistical evaluation of fuzzy rules. Values towards  100 \% tends to the generation of only few, but very confident rules.
\end{itemize}

\clearpage
 \begin{figure}[h] 
 \includegraphics[width=0.9\columnwidth]{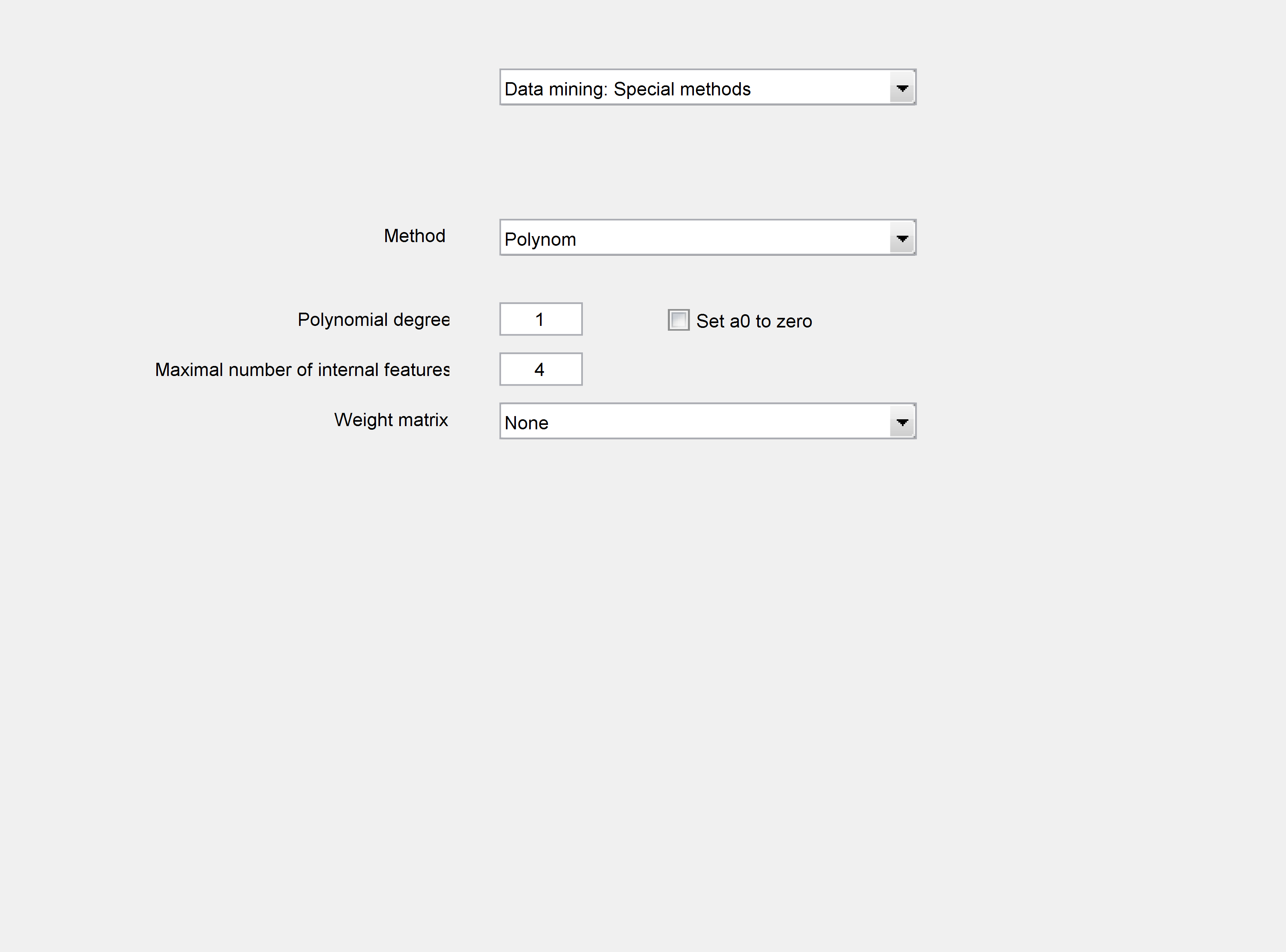}
 \caption{\label{fig:optionen15-28} Control elements for Data mining: Special methods - Polynom}
 \end{figure}
\begin{itemize}
\item {\bf Method}: \hypertarget{CE_Spezielle_Verfahren}{} \newline
parameterizes a method for the design of a classifier or \textindex{regression model}. The elements
are context-sensitive to the method selected by \hyperlink{CE_Spezielle_Verfahren}{\emph{Control element: Data mining: Special methods - Method}}.
\item {\bf Polynomial degree}: \hypertarget{CE_Regression_GradPolynom}{} \newline
defines the maximal degree $m$ of a polynomial function $y = a_0 +
a_1 x + ... + a_m x^m$.
\item {\bf Set a0 to zero}: \hypertarget{CE_Regression_NonAffineModel}{} \newline
forces a polynomial regression model to estimate a model without a constant term called a0.
\item {\bf Maximal number of internal features}: \hypertarget{CE_Regression_AnzahlPolyMerkmale}{} \newline
enables an internal feature reduction during the computation of the polynomial model. It is useful especially
for the selection of terms with the most relevant power coefficients.
\item {\bf Weight matrix}: \hypertarget{CE_Regression_PolynomWeighting}{} \newline
selects a weighting matrix for data points during a Least Square estimation. ''None'' uses equal weights (1), ''inverse proportional to class frequency'' prefers data points from rare linguistic terms of the selected output variable.
\end{itemize}

\clearpage
 \begin{figure}[h] 
 \includegraphics[width=0.9\columnwidth]{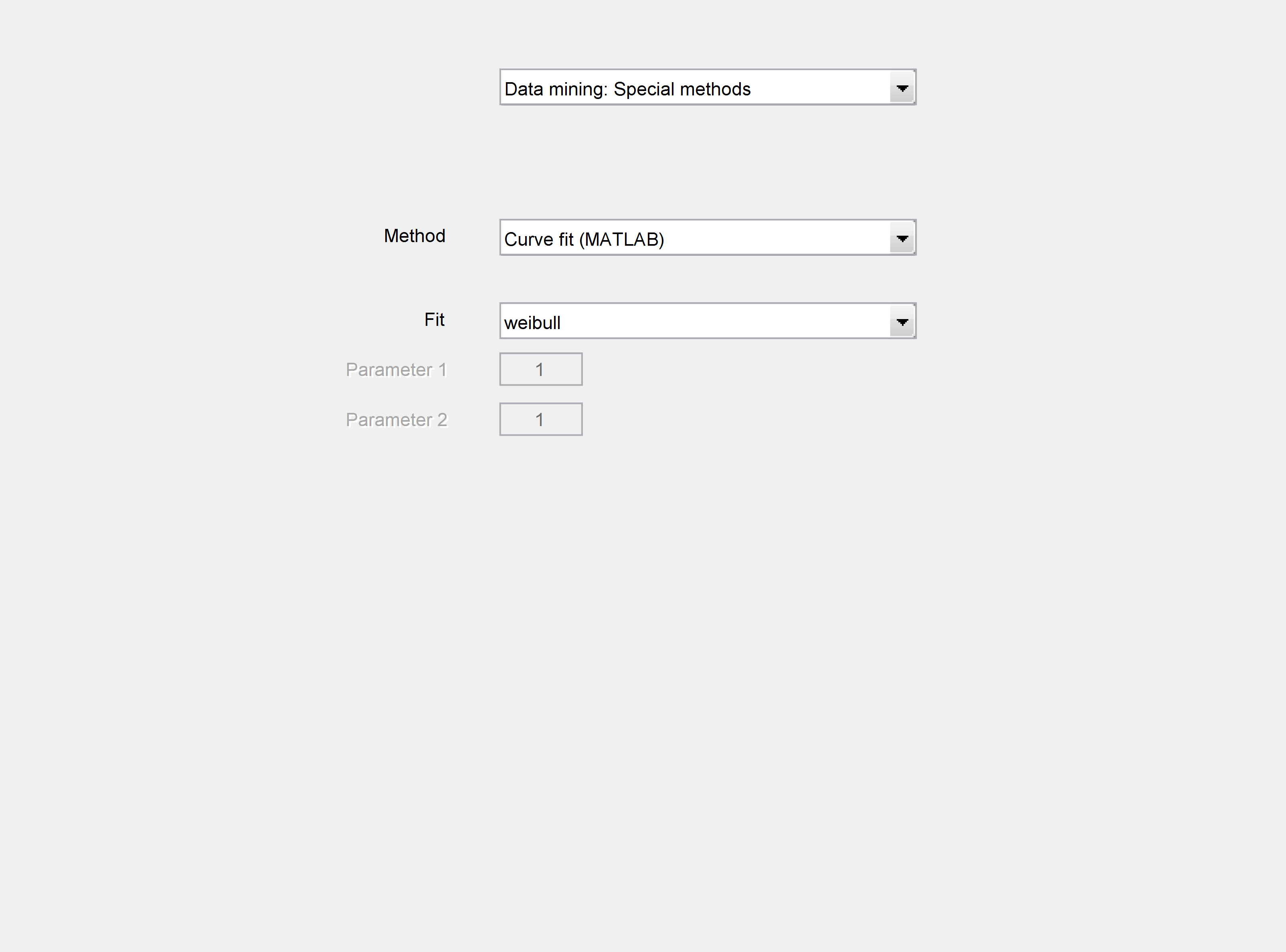}
 \caption{\label{fig:optionen15-29} Control elements for Data mining: Special methods - Curve fit (MATLAB)}
 \end{figure}
\begin{itemize}
\item {\bf Method}: \hypertarget{CE_Spezielle_Verfahren}{} \newline
parameterizes a method for the design of a classifier or \textindex{regression model}. The elements
are context-sensitive to the method selected by \hyperlink{CE_Spezielle_Verfahren}{\emph{Control element: Data mining: Special methods - Method}}.
\item {\bf Fit}: \hypertarget{CE_Regression_Fit}{} \newline
specifies the method for the interpolation with the MATLAB function fit (part of the Curve Fitting Toolbox). For more details type ''help fit'' in the MATLAB command window.
The following methods are implemented (in parentheses: number of input variables and specialties):
weibull (1D, no negative inputs),
exponential (1D),
fourier (1D),
gauss (1D),
cubicinterp (1D or 2D, no interpolation),
pchipinterp (1D, no interpolation),
poly curve (1D or 2D),
power (1D, no negative inputs),
rational (1D),
sinus (1D),
cubicspline (1D, no interpolation),
smoothingspline (1D),
biharmonicinterpolation (2D),
local linear regr (2D),
local quadratic regr (2D).
\item {\bf Parameter 1}: \hypertarget{CE_Regression_Fit_Par1}{} \newline
select context-sensitive the first parameter for the selected method in \hyperlink{CE_Regression_Fit}{\emph{Control element: Data mining: Special methods - Fit}}. This parameter is not used by every method.
\item {\bf Parameter 2}: \hypertarget{CE_Regression_Fit_Par2}{} \newline
select context-sensitive the second parameter for the selected method in \hyperlink{CE_Regression_Fit}{\emph{Control element: Data mining: Special methods - Fit}}. This parameter is not used by every method.
\end{itemize}

\clearpage
 \begin{figure}[h] 
 \includegraphics[width=0.9\columnwidth]{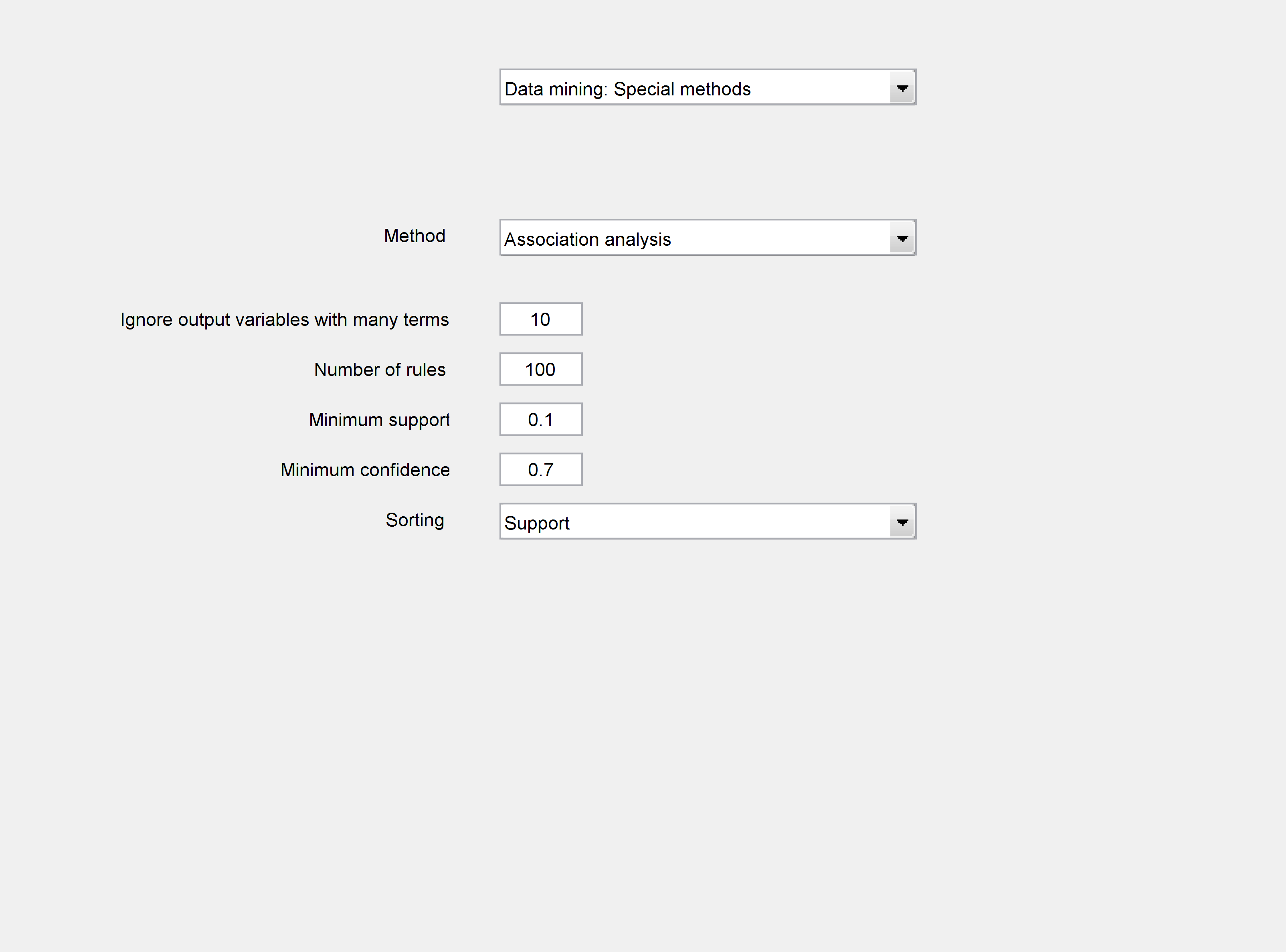}
 \caption{\label{fig:optionen15-30} Control elements for Data mining: Special methods - Association analysis}
 \end{figure}
\begin{itemize}
\item {\bf Method}: \hypertarget{CE_Spezielle_Verfahren}{} \newline
parameterizes a method for the design of a classifier or \textindex{regression model}. The elements
are context-sensitive to the method selected by \hyperlink{CE_Spezielle_Verfahren}{\emph{Control element: Data mining: Special methods - Method}}.
\item {\bf Ignore output variables with many terms}: \hypertarget{CE_AssociationIgnoreFrequent}{} \newline
defines the number of linguistic terms to ignore output variables with many terms (e.g., names or IDs) in an association analysis.
\item {\bf Number of rules}: \hypertarget{CE_AssociationNRules}{} \newline
defines the maximum number of rules for an association analysis.
\item {\bf Minimum support}: \hypertarget{CE_AssociationMinSupp}{} \newline
defines the minimal support P(A AND B) of a rule  IF A THEN B.
\item {\bf Minimum confidence}: \hypertarget{CE_AssociationMinKonf}{} \newline
defines the minimal confidence P(B|A) of a rule  IF A THEN B.
\item {\bf Sorting}: \hypertarget{CE_AssociationSorting}{} \newline
defines the sorting order for the found rules by support or by confidence.
\end{itemize}

\clearpage\clearpage

\section{Control elements for 'Data mining: Statistical options'}

 \begin{figure}[h] 
 \includegraphics[width=0.9\columnwidth]{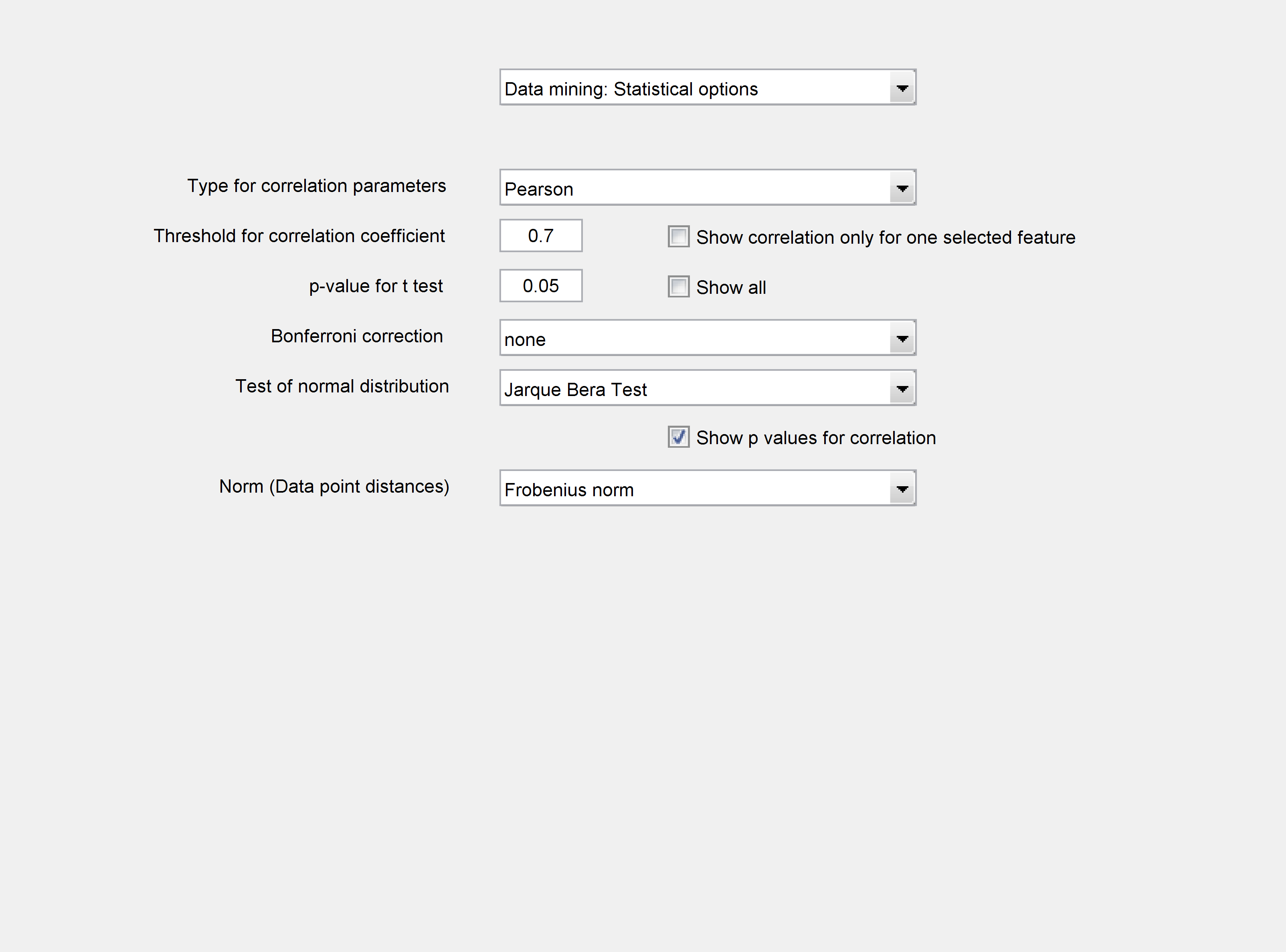}
 \caption{\label{fig:optionen16-31} Control elements for Data mining: Statistical options }
 \end{figure}
\begin{itemize}
\item {\bf Type for correlation parameters}: \hypertarget{CE_Corr_Type}{} \newline
defines the type of correlation (Linear: Pearson; Rank-oriented: Kendall, Spearman) to be used in visualizations or output files for correlations.
\item {\bf Threshold for correlation coefficient}: \hypertarget{CE_Krit_Koeff}{} \newline
defines a \textindex{threshold for the correlation coefficient}. A larger value is interpreted as a
relevant pair-wise correlation between two features. This value influences e.g. the listing
of relevant correlation coefficients or the downgrading of correlated features in the feature
selection.
\item {\bf Show correlation only for one selected feature}: \hypertarget{CE_Anzeige_CorrSelFeature}{} \newline
computes only the correlations of all selected single features to the feature defined by \hyperlink{CE_Regression_Output}{\emph{Control element: Data mining: Regression - Output variable of regression}}.
\item {\bf p-value for t test}: \hypertarget{CE_P_Wert}{} \newline
defines the p-value for the \textindex{t-test} (accepted error probability of a statistical test).
Usual values are 0.01...0.05.
\item {\bf Show all}: \hypertarget{CE_Krit_Koeff_Alle}{} \newline
defines whether all tested or only significant different single features (measured by the
\textindex{t-test}) for the output variable should be shown.
\item {\bf Bonferroni correction}: \hypertarget{CE_Corr_Bonferroni}{} \newline
defines if statistics tests and correlations are computed with or without (''no'') corrections for multiple testings. A \textindex{Bonferroni corrections} divides the $\alpha$ values in \hyperlink{CE_P_Wert}{\emph{Control element: Data mining: Statistical options - p-value for t test}} by the number of tests. The less conservative \textindex{Bonferroni-Holm correction} divides $\alpha$ by the number of test - the ranking of the tested values + 1. It accepts all successful tests before the first rejected test.
\item {\bf Test of normal distribution}: \hypertarget{CE_Normtest}{} \newline
selects a methods to test the normality of a distribution using \hyperlink{MI_NormTest}{\emph{Data mining - Selection and evaluation of single features - Test of normal distribution}}. Possible values are \textindex{Chi Square Test}, \textindex{Lillie Test}, \textindex{Anderson Darling Test} and \textindex{Jarque Bera Test} from the MATLAB Statistics Toolbox.
\item {\bf Show p values for correlation}: \hypertarget{CE_Anzeige_CorrPValues}{} \newline
defines the option for the reporting of p-values for correlation coefficients unequal zero in the correlation visualization (\hyperlink{MI_Anzeige_EM_Korr}{\emph{View - Single features - Correlation coefficients (Pearson)}} etc.).
\item {\bf Norm (Data point distances)}: \hypertarget{CE_DatDistNorm}{} \newline
select the used norm to compute data point distances.
\end{itemize}
\clearpage

\section{Control elements for 'Data mining: Validation'}

 \begin{figure}[h] 
 \includegraphics[width=0.9\columnwidth]{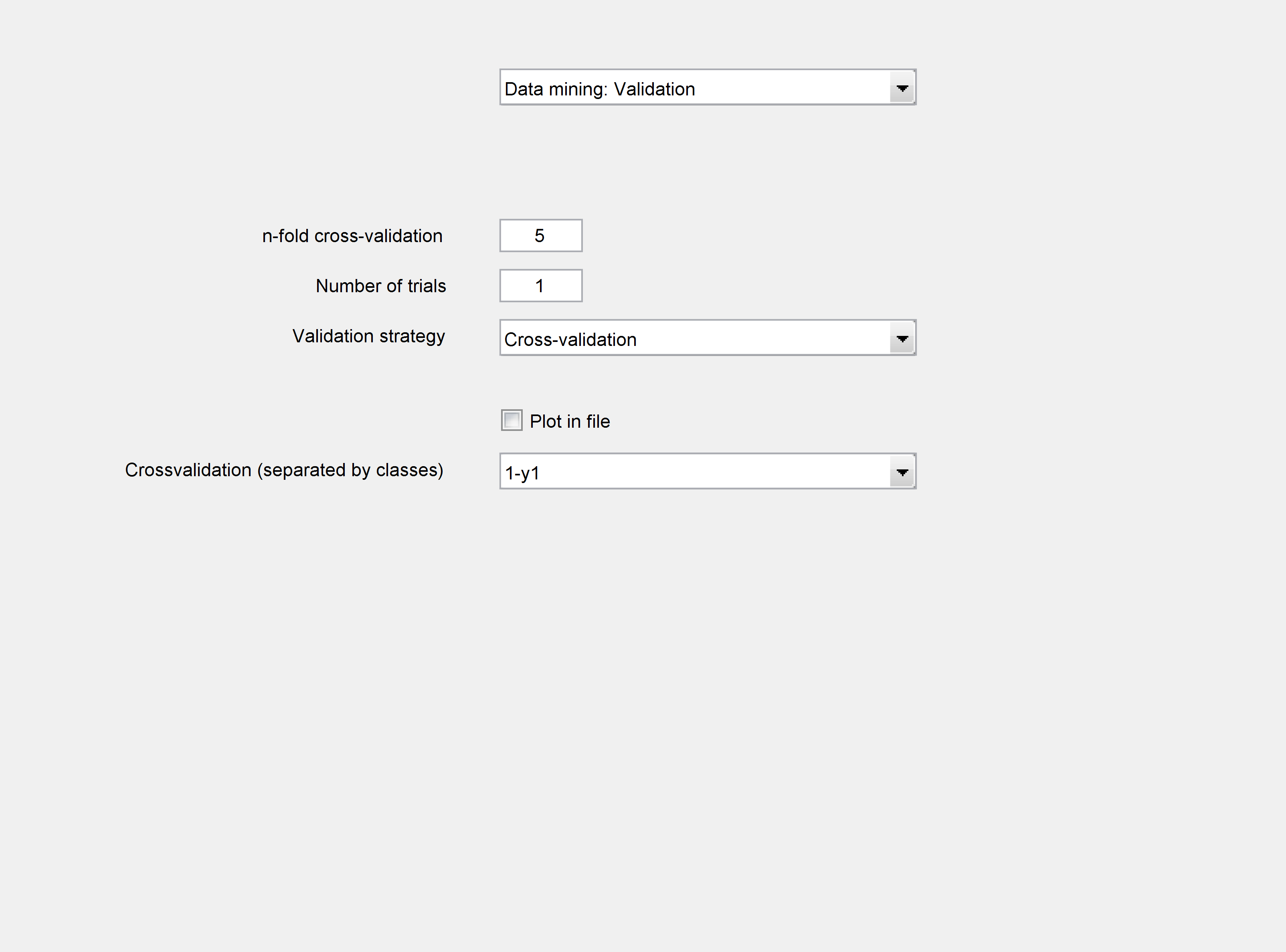}
 \caption{\label{fig:optionen17-32} Control elements for Data mining: Validation }
 \end{figure}
\begin{itemize}
\item {\bf n-fold cross-validation}: \hypertarget{CE_CV_n}{} \newline
defines the number $n$ of parts for a $n$-fold \textmainindex{cross-validation}. As an example, a
value of 10 means that the training data set is separated in 10 parts. For each run, 9 parts are
used as training data and one part as validation data set. Consequently, a complete trial of a
cross-validation requires 10 runs containing each selected data point exactly one times in the
validation data set. Typical values are  $n=2 \hdots 10$.

The value will be ignored for the selection of 'Leave-one-out' ($n$ is equal to the number of
selected data points) or 'Bootstrap' in \hyperlink{CE_CV_Typ}{\emph{Control element: Data mining: Validation - Validation strategy}}.
\item {\bf Number of trials}: \hypertarget{CE_CV_Versuche}{} \newline
defines the number of trials for a cross-validation. Internally, a random assignment of data points
to the different parts will be chosen. This assignment takes the output variables into account to
get an approximately equal distribution of linguistic output terms for the different parts. As a
consequence, the results of different trials will differ. A higher number of trials costs computing
time, but improves the estimation for the classification accuracy. Typical values are 1...50
depending on the necessary computing time.

The value will be ignored for the selection of 'Leave-one-out' in  \hyperlink{CE_CV_Typ}{\emph{Control element: Data mining: Validation - Validation strategy}} (always 1 due
to the deterministic selection of training data).
\item {\bf Validation strategy}: \hypertarget{CE_CV_Typ}{} \newline
switches the validation type between \textindex{cross-validation}, \textindex{leave-one-out}
($N$-fold cross-validation with the number $N$ of selected data points and exactly one data
point in each validation data set) and \textindex{bootstrap} method (training data set: random
selection of data points with replacement resulting usually in a multiple selection of some data
points, validation data set: all none selected data points) (see e.g. \cite{Efron95}).
\item {\bf Plot in file}: \hypertarget{CE_Auswertung_Datei}{} \newline
controls if the validation results are written into a protocol file or not.
\item {\bf Crossvalidation (separated by classes)}: \hypertarget{CE_Auswahl_CVOutputSelect}{} \newline
use the classes of the selected output variables to split the dataset. As a consequence, the data points belonging to a class are used as validation dataset, all other data points as training dataset. As usual for a crossvalidation, this split is rotated so that each selected data points is used exactly once as in the validation dataset.
\end{itemize}
\clearpage

\section{Control elements for 'General options'}

 \begin{figure}[h] 
 \includegraphics[width=0.9\columnwidth]{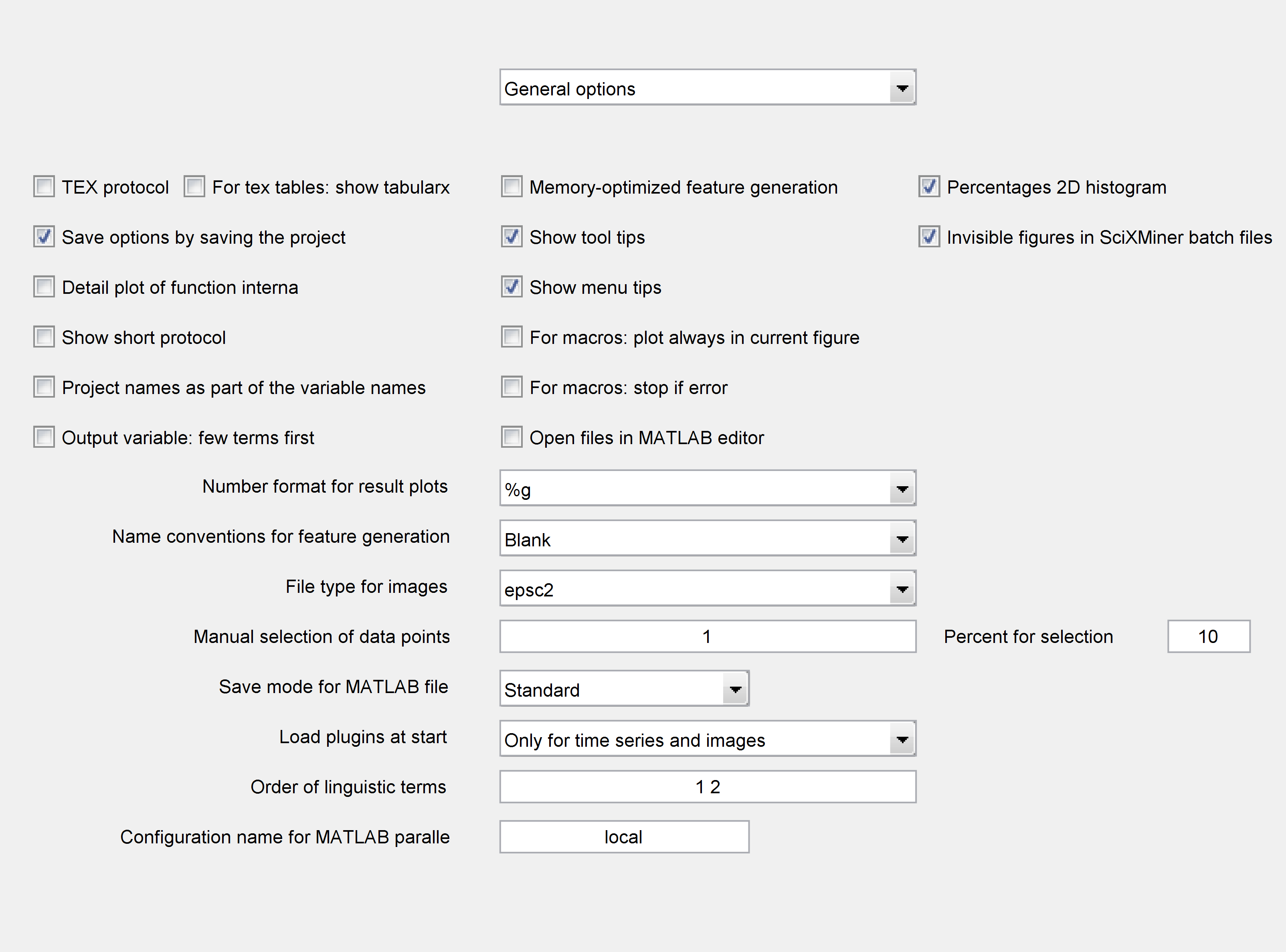}
 \caption{\label{fig:optionen18-33} Control elements for General options }
 \end{figure}
\begin{itemize}
\item {\bf TEX protocol}: \hypertarget{CE_Tex_Protokoll}{} \newline
generates most generated protocol files in a \LaTeX format. Otherwise, ASCII files are produced.
\item {\bf For tex tables: show tabularx}: \hypertarget{CE_Tex_Table_Options}{} \newline
switches the tabularx style in the generation of Latex tables on or off.
\item {\bf Memory-optimized feature generation}: \hypertarget{CE_Speicherschonen}{} \newline
modifies the algorithm for feature extraction from time series. If the option is chosen, the
computing time for extraction increases and the risk of \textindex{memory problems} decreases.
\item {\bf Percentages 2D histogram}: \hypertarget{CE_Anzeige_Percentage_2DOutput}{} \newline
switches the generation of percentage information (per row and per column) in the file containing quantitative information of 2D histograms on or off.
\item {\bf Save options by saving the project}: \hypertarget{CE_Optionen_Speichern}{} \newline
switches the saving of options (all GUI elements) on or off, if a project is saved.
\item {\bf Show tool tips}: \hypertarget{CE_Anzeige_Tooltips}{} \newline
switches the context-sensitive help for control elements on or off, if the mouse arrow is over an
element.
\item {\bf Invisible figures in SciXMiner batch files}: \hypertarget{CE_Anzeige_BatchInvisFigures}{} \newline
hides all figures during execution of a \textindex{SciXMiner batch file}. This option is useful for parallel working on the same computer without disturbances of opening figures. The SciXMiner main figure is also hidden. If a batch file stops by error, the hidden figures can be restored by the input of the command ''restore\_figures'' in the MATLAB command window.
\item {\bf Detail plot of function interna}: \hypertarget{CE_Anzeige_Interna}{} \newline
influences the level detail for some texts and figures with intermediate results  (marked = more
details).
\item {\bf Show menu tips}: \hypertarget{CE_Anzeige_Menutips}{} \newline
if activated, a short description of the clicked menu item is shown as part of the project overview (\hyperlink{CEO_Projekt}{\emph{Control elements:  Project overview}}).
\item {\bf Show short protocol}: \hypertarget{CE_Show_ShortProtocol}{} \newline
switches off the complete listing for all control elements in the protocol files. Here, only some project
characteristics are written into the files.
\item {\bf For macros: plot always in current figure}: \hypertarget{CE_Aktuelle_Figure}{} \newline
plots all visualizations in the recent figure. This option is useful for the generation of
user-defined subplots in macros. However, it requires a manual modification of the
\textindex{macro} with figure and sub-figure commands. Otherwise, figures are overwritten without
any warning.
\item {\bf Project names as part of the variable names}: \hypertarget{CE_ProjectNamesFusion}{} \newline
adds project names during the project fusion for new single features and time series to mark the source project of a single feature or time series.
\item {\bf For macros: stop if error}: \hypertarget{CE_Makro_Stop_Error}{} \newline
switches off the error handling with try/catch in macros. It allows a better backtracking of error messages.
\item {\bf Output variable: few terms first}: \hypertarget{CE_ShowFewTermsFirst}{} \newline
shows in \hyperlink{MI_Anzeige_Datentupel}{\emph{View - Classes for selected data points}} the output variables with only few terms first. This option is useful in projects with many linguistic terms and data points.
\item {\bf Open files in MATLAB editor}: \hypertarget{CE_OpenFilesInEditor}{} \newline
defines whether generated text files (e.g., for feature evaluation) will be immediately opened by the MATLAB editor or not.
\item {\bf Number format for result plots}: \hypertarget{CE_Anzeige_Zahlenformat}{} \newline
defines the format of numbers for most visualization and protocol functions, especially the number
of digits before and after the decimal point . ''\%g'' is the automatically chosen standard Matlab
format, ''\%f'' means floating points. The parameter does not influence export functions.
\item {\bf Name conventions for feature generation}: \hypertarget{CE_Namenskonventionen}{} \newline
defines the name conventions for extracted features. Possible options are a combination of name
fragments by blanks or underscores.
\item {\bf File type for images}: \hypertarget{CE_Allgemein_FileTypeImage}{} \newline
defines the file type for the hardcopy of all open figures into files. The abbreviations are the
same as for the MATLAB print command. The EPS format is subdivided into black-and-white images
(eps) and color images (epsc).
\item {\bf Manual selection of data points}: \hypertarget{CE_Select_DataPoints}{} \newline
sets the numbers of data points for the selection with \hyperlink{MI_Datenauswahl_GUI}{\emph{Edit - Select - Data point from GUI}}.
\item {\bf Percent for selection}: \hypertarget{CE_Select_PercDataPoints}{} \newline
defines the percentage of selected data points by using \hyperlink{MI_Datenauswahl_PercDataPoints}{\emph{Edit - Select - Random data points (defined percentage of selected data points)}}.
\item {\bf Save mode for MATLAB file}: \hypertarget{CE_Savemode}{} \newline
defines the compatibility mode for binary MATLAB files. '-V6' is compatible to the MATLAB versions
5 and 6, '-V7.0' to version 7 and '-V7.3' to version 7.3. For lower versions, only a reduced set of
modes is available.
\item {\bf Load plugins at start}: \hypertarget{CE_LoadPluginStart}{} \newline
defines the modus of loading and initializing of plugins for feature extraction during loading a project: never, always, only if the project contains time series or images.
\item {\bf Order of linguistic terms}: \hypertarget{CE_Anzeige_OrderofTerms}{} \newline
defines a new order for the linguistic terms of the selected output variable. The resorting is done by \hyperlink{MI_Trans_Term_Order}{\emph{Edit - Convert - Resort linguistic terms (order from GUI)}}.
\item {\bf Configuration name for MATLAB parallel}: \hypertarget{CE_AllgParConfig}{} \newline
defines the name of the configuration for the \textindex{MATLAB Parallel Computing Toolbox} that can be started by \hyperlink{MI_Extra_StartParallel}{\emph{Extras - Matlab Parallel - Start}} (see also Parallel - Manage Configurations in the menu of the MATLAB command window).
\end{itemize}
\clearpage\clearpage\clearpage\clearpage\clearpage\clearpage\clearpage

\chapter{Feature extraction from time series}\label{sec:plugins_und_einzuege}\thispagestyle{empty}
\section{Definition of feature types by plugins}\label{sec:plugins}
The \textmainindex{feature extraction} of time series is realized with \textmainindex{plugins}.
Please notice that SciXMiner was actually developed in German language, so some
variable names are German. Unfortunately, it was impossible to change this inconvenience.

Plugins are single Matlab functions named  \texttt{plugin\_*.m}, which are included in a special directory of the
SciXMiner installation (...$\backslash$scixminer$\backslash$plugins$\backslash$mgenerierung) or in the directory which
contains the loaded project. To work with SciXMiner properly, they have to abide by the following format:
\begin{verbatim}
 function [datenOut, ret, info] = pluginfct(paras, datenIn).
\end{verbatim}
The return value info contains the following elements:
\begin{itemize}
    \item{\textbf{info.beschreibung}: short description of the new feature with one or two words (e.g. minimum, maximum, velocity). Only one
    description is allowed, independent from the numbers of computed features. The description \textbf{must not} start
    with a hyphen (minus)!
    }
    \item{\textbf{info.bezeichner}: short identifier of the new feature (e.g. MIN, MAX, V). The matrix has to contain one
    identifier per feature or time series (every row is a single identifier). Use \verb=str2mat('String','OtherString')=
    to create a matrix with one identifier per row.
    }

    \item{\textbf{info.explanation}: explanation of the plugin (approx. 1-2 sentences)}
    \item{\textbf{info.explanation\_long}: optionally an additional longer text, used only for documentation}

    \item{\textbf{info.anz\_zr}: numbers of time series extracted by the plugin (has to be 0, if \texttt{info.anz\_em} $\not= 0$)}
    \item{\textbf{info.anz\_benoetigt\_zr}: numbers of time series needed for the computation of the new time series.
    If $>1$ the first anz\_benoetigt\_zr time series are always given to the plugin. The plugin have to check the
    correct amount of time series! If the plugin is able to deal with an arbitrary amount of time series (e.g. computation of minimum), use Inf for
    anz\_benoetigt\_zr.
    }

    \item{\textbf{info.anz\_em}: numbers of features extracted by the plugin (has to be 0, if \texttt{info.anz\_zr} $\not= 0$)}
     \item{\textbf{info.anz\_benoetigt\_em}: number of input single features of the plugin (will be used in future versions)}

    \item{\textbf{info.anz\_im}: number of images generated in the plugin (only used together with \textindex{Image Extension Package})}
    \item{\textbf{info.anz\_benoetigt\_im}: number of images needed for the generation of new images or single features (only used together with \textindex{Image Extension Package})}

  \item{\textbf{info.callback}: optional callback function to avoid the transfer of images into functions (only used together with \textindex{Image Extension Package})\footnote{Operations on images needs more time if images are transferred to functions. As an alternative, the callbacks are used in plugin sequences to allow a direct access to the variable myimage in the workspace.}}

    \item{\textbf{info.einzug\_OK}: If the plugin is able to deal with intervals other than the complete time series, set
    einzug\_OK to $1$, otherwise to $0$. For a description of intervals, see Section~\ref{sec:einzuege}).}
    \item{\textbf{info.richtung\_entfernen}: special parameter for plugins for the gait analysis. Removes identifier

    "left" or "right" if this variable is set to $1$. The default is $0$.
    }
    \item{\textbf{info.typ}: specifies the type of the result: 'TS': time series, 'SF': feature. The type 'Norm' requires additional \textindex{normative data}. For images, the additional types 'IM' (image) and 'IMSlice' (3D stack of images) are in preparation. A plugin must not extract both time series and features.
    }
\end{itemize}
Missing info elements are replaces by standard entries if it is possible.

Plugins can contain parameters. Here, the use of one or more \textmainindex{plugin parameters} with edit or popup elements is possible. For each parameter $p$, the following definitions exist:
\begin{itemize}
    \item{\textbf{info.commandline.description\{p\}}: name of the parameters}
    \item{\textbf{info.commandline.parameter\_commandline\{p\}}: default value (string, number or vector of numbers; scalar for popup elements means number of the selected string)}
    \item{\textbf{info.commandline.popup\_string\{p\}}: string with alternatives, e.g. 'yes|no' (only for popup elements)}
     \item{\textbf{info.commandline.tooltext\{p\}}: text for \textindex{tool tips}}
    \item{\textbf{info.commandline.wertebereich\{p\}}: possible range for numbers in edit elements, the two elements means the minimum and maximum value, e.g. \{1 Inf \}. The use of variables is possible, e.g.  \{1 'paras.par.laenge\_zeitreihe' \} for the length of the time series}
    \item{\textbf{info.commandline.ganzzahlig\{p\}}: optional restriction for the use of integer values in edit elements, if the value is 1}
    \item{\textbf{info.commandline.format\{p\}}: ignoring of the internal format supervision, if the value is ''any''}

\end{itemize}
The parameters are defined using \hyperlink{CE_Auswahl_PluginsCommandLine}{\emph{Control element: Plugin sequence - Plugin parameter}} for the recent selected plugin in \hyperlink{CE_Auswahl_Plugins}{\emph{Control element: Plugin sequence - Selection of plugins}} resp. \hyperlink{CE_Auswahl_PluginsList}{\emph{Control element: Plugin sequence - Selected plugin sequence}}. The change between different parameters of a plugin is done using \hyperlink{CE_Plugins_ParameterNumber}{\emph{Control element: Plugin sequence - No.}}. All parameters can be modified using a \textindex{macro}.

In a plugin, the access to the parameters is possible with  paras.parameter\_commandline\{2\} for the second parameter etc.

The inputs are as follows:
\begin{itemize}
    \item{\textbf{datenIn}
        \begin{itemize}
            \item{\textit{datenIn.dat}: matrix of the input data (size paras.par.anz\_dat $\times$ sample points in interval
            $\times$ info.anz\_benoetigt\_zr).
            The third dimension is missing, if info.anz\_benoetigt\_zr == 1!
            }
            \item{\textit{datenIn.ref}: data of a reference.\index{normative data})}
        \end{itemize}
    }
    \item{\textbf{paras}
        \begin{itemize}
            \item{\textit{paras.par}: par vector from SciXMiner}
            \item{\textit{paras.code}: currently selected output variable}
            \item{\textit{paras.code\_alle}: all output variables}
            \item{\textit{paras.var\_bez}: identifier of time series}
            \item{\textit{paras.merk\_red}: number of features to select}
            \item{\textit{paras.einzuege}: interval from which the data has been copied out of the complete time series
            (size $1 \times 2$)}
            \item{\textit{paras.ind\_zr\_merkmal}: indices of selected time series
             (size $1 \times$ anz\_benoetigt\_zr).}
            \item{\textit{paras.anz\_gew\_merk}: number of selected time series}
            \item{\textit{paras.parameter}: \textindex{SciXMiner parameter struct} (contains options from the GUI)}
                             \item{\textit{paras.parameter\_commandline}: cell array with values of the \textindex{plugin parameters}}

        \end{itemize}
        The parameters are set in the file "merkmalsgenerierung\_plugins.m". One can add new elements to the parameter
        struct. Use the Matlab command \texttt{isfield} to check, whether an element exists in the struct. The
        following elements has been added for some plugins:
        \begin{itemize}
           \item{iirfilter: parameter for an \textindex{IIR-Filter} (included due to plugin\_iirfilter)}
           \item{iirfilter\_aS\_aL\_aSigma: Some plugins (e.g. Trend and standard deviation estimation) use multiple IIR-Filters}
           \item{abtastfrequenz:  sampling rate of the time series [Hz]}
           \item{samplepunkt:  single sample for the computation of a new feature (e.g. extraction of a specific
           sample out of a given time series}
        \end{itemize}
    }
\end{itemize}

If the plugin is called with an empty matrix for \texttt{datenIn}, the \texttt{info}-Struct has to
be returned by the plugin. \texttt{datenOut} and \texttt{ret} remain empty. This proceeding is
necessary to query some information about the existing plugins. The \texttt{paras}-Struct is always
committed by the calling function. The first rows of the plugin could be:
\begin{small}
\begin{verbatim}
 info = struct('identifier', 'MIN', 'anz_zr', 1, ...
    'anz_em', 0, 'typ', 'ZR', ...);
 if (nargin < 2 | isempty(datenIn))
   datenOut = [];
   ret = [];
   return;
 end;
\end{verbatim}
\end{small}

The outputs have to contain the following elements:
\begin{itemize}
    \item{\textbf{datenOut}:
        \begin{itemize}
            \item{\textit{datenOut.dat\_zr}: new time series, if the plugin extracts time series (size
            paras.par.anz\_dat $\times$ paras.par.laenge\_zeitreihe $\times$ info.anz\_zr). Empty otherwise.}
            \item{\textit{datenOut.dat\_em}: new features, if the plugin extracts features (size paras.par.anz\_dat $\times$ info.anz\_em).
            Empty otherwise.
            }
        \end{itemize}
    }
    \item{\textbf{ret}:
        \begin{itemize}
            \item{\textit{ret.ungueltig}: set to $1$, if the extraction of the feature or time series has failed. The
            result of the plugin is not accepted by SciXMiner.}
            \item{\textit{ret.bezeichner}: contains the identifier of the time series or features if the amount of
            extracted time series or features depends on parameters in the GUI (e.g. Principal Component Analysis or
            IIR-Filter). If this element exists and is not empty it is used as identifier instead of the
            identifier from the info-Struct. The identifier from the info-Struct must not be changed.}
        \end{itemize}
    }
\end{itemize}

\section{Standard plugins in SciXMiner}
A list of include plugins in the standard installation of SciXMiner is shown in Section~\ref{sec:appendix_plugin}. In a project, the list of all available plugins can be shown by  \hyperlink{CE_Plugins_Anzeige}{\emph{Control element: Plugin sequence - Show plugins}}.

\section{Plugins for single features from single features}
Plugins for the extraction of new single features from existing single features are not yet implemented in a formal way. Two different alternatives exist:
\begin{itemize}
 \item implementation of individual plugins via macros (see two examples in the directory  standardmakros: feature\_plugin\_div2.makrog for the division of two selected single features and   feature\_plugin\_log.makrog to compute the logarithm of all selected features)
 \item computing aggregated features (see \hyperlink{MI_Extraktion_EMEMA}{\emph{Edit - Extract - Single features -> Single features (with the selected feature aggregation from Options-Data Mining: Classification of single features)}})
\end{itemize}

\section{Defining intervals via files}\label{sec:einzuege}
Intervals are used to restrict the feature extraction to specific sample points of the time series
(only available for plugins with \texttt{info.einzug\_OK} $=1$.

After loading a SciXMiner project, all \textit{*.einzug} files in the current directory and in the
subdirectory \textit{plugins/einzuggenerierung} of the SciXMiner installation are imported. An
\textmainindex{interval file} is an ASCII file including a chart of the intervals. The rows are
separated with line breaks, columns with tabs. To avoid an empty interval do not add a line break
after the last interval. The first row of the file has to include the following four column names (separated by tabs):
\begin{verbatim}
 Identifier  Short Identifier    Start   Stop
\end{verbatim}
The intervals can be defined in the following rows, e.g.:
\begin{verbatim}
 Standphase    ST_P  1  [FootOff]
 Schwungphase  SW_P  [FootOff]+1 -1
\end{verbatim}
Special features are:
\begin{itemize}
    \item{In the definition, features and mathematical operations can be used, e.g.
        \texttt{[OppositeFootOff]+([OppositeFootContact]-[OppositeFootOff])/2}.
        Pay attention, that used features exist in the project!}
    \item{The end of a time series can be labeled by \texttt{-1} or \texttt{maxtime}}.
\end{itemize}
Function calls or variables from the Matlab workspace can not be used.

\section{Definition of a feature ontology by categories}\label{sec:ontologie}
Categories are useful to define an ontology for single features and time series. Here, similar
features and time series can be grouped into classes. They defined one or more categories. Such
categories might be used in different analysis and visualization functions (see
\cite{Loose04,Wolf06}).

The assignment to categories can be done by plugins, segments, or name fragments.  It is defined in
one or more files with the extension \texttt{*.categories}. Here, all files in the current working
directory or in the subdirectory ''plugins/mgenerierung'' of SciXMiner or a user-defined extension
package are considered.

These functions are text files and have a pre-defined format:
\begin{verbatim}
  #CategoryName 'Temporal derivation'

  #DefaultTermName 'No'

  #TermName 'Velocity'
  #TermRelevance 0.8
  'plugin_geschwindigkeit'
  'plugin_geschwindigkeit_kausal'
  '_V'
  ' V '

  #TermName 'Acceleration'
  #TermRelevance 0.6
  'plugin_beschleunigung'
  '_A'
  ' A '
\end{verbatim}
The following key words are used:
\begin{itemize}
\item  \texttt{\#CategoryName}: name of the category.
\item  \texttt{\#TermName}: name of a term of the recent category followed by a string.
       In the next rows, identifiers for a mapping of time series or single features are defined.
       Such identifiers can be name fragments (e.g. \texttt{'\_V'}),
       names of used plugins (e.g. \texttt{'plugin\_geschwindigkeit'})
       or names of segments using the key word  \texttt{einzug\_short\_description} (e.g.
       \texttt{einzug\_ST\_P}). Consequently, each time series or single feature containing any of these identifiers
       belongs to the term 'Velocity' of the category 'Temporal derivation' if  \texttt{'\_V'} is part of the name,
       the plugin \texttt{'plugin\_geschwindigkeit'} was used for feature extraction by
       \hyperlink{MI_Extraktion_ZRZR}{\emph{Edit - Extract - Time series -> Time series, Time series -> Single features...}}.
\item  \texttt{\#DefaultTermName} (optional): defines the name for the default term. It used if as
       time series or a single feature does not belong to an other term of the category. Its
       standard name is 'unknown'.
\item  \texttt{\#TermRelevance} (optional): defines the \textindex{a priori relevances} for the recent term.
       It might be defined with one or two values between zero and one separated by a space.
       The first value is the interpretability weighted by the exponent alpha, the second the implementability weighted by the exponent beta.
       If the values are missing, the standard value is one.
\end{itemize}

\chapter{Conclusions and perspectives}
This manual describes the functionality of the open source Matlab toolbox SciXMiner. The aim of this
toolbox is to provide an interface to apply and compare data mining methods. The architecture of
the toolbox is chosen in a way that the developer crew from Karlsruhe, Germany and/or other
developer crews are easily able to enlarge the toolbox by further algorithms. Everyone is kindly
invited to support the further development of SciXMiner.

{\bf Thanks:}

Thanks to all the busy programmers, developers of algorithms, testers, especially to Rüdiger Alshut, Alessandro Angelin, Martin Ashby, Sebastian Beck, Sebastian Braun, Ole Burmeister, Joachim Dieterle, Patrick Gerland, Sebastian Gollmer, Andreas Gommlich, Lutz Gröll, Markus Grube, Eduard Hübner, Jens Jäkel, Sina Keller, Thilo Krüger, Jessica Legradi, Tobias Loose, Mihai Lipovei, Jörg Matthes, Dimitrios Patikas, Sebastian Pfeiffer, Tim Pychynski, Matthias Schablowski, Oliver Schill, Martin Wieland, Sebastian Wolf, Mohamed Zayani and Baifan Zhou. The support by the Deutsche Forschungsgemeinschaft (German research association) within the project 'Diagnosis support in gait analysis'' was a great help to build the gait analysis-specific part and
thus, to build a basis for the further development of the toolbox. The Collaborative Research Center for Humanoid Robots, also sponsored by DFG, has inspired us to many technical applications.
The present contribution was supported by the Helmholtz Association in the program ''BioInterfaces in Technology and Medicine'' and under the Joint Initiative “Energy System 2050–
A Contribution of the Research Field Energy”.

\clearpage

\begin{appendix}
\chapter{Important file structures}\label{sec:dateien}\thispagestyle{empty}
\begin{itemize}
\item A \textmainindex{SciXMiner batch file} (\emph{*.batch}) is an ASCII file. It contains directories, projects, options and macros for automated analyses.
\item A \textmainindex{classifier file} (\emph{*.class}) is a Matlab file. It contains data to apply a designed classifier as \emph{klass\_single}
struct array. It includes the selected single features, aggregation transformations and the
classifier itself as well as chosen standardizations. The mapping of single features and of the
output variables uses the feature names. One-against-x classifiers are described by a vector-
shaped \emph{klass\_single} struct.
\item A \textmainindex{frequency list file} (\emph{*.freq}) is a Matlab file. It contains a struct freqlist with two substructs:
freqlist.segments characterizes segments with elements freqlist.segments.f (vector with two elements for the lower and the upper frequency) and .name (segment name).
freqlist.multiples characterizes \textindex{overtones} with freqlist.segments.f (scalar for frequency), .ftol (tolerance for overtones) and .name (overtone name). An example is shown in sounds.freq in the directory  standardmakros.
\item A \textmainindex{fuzzy system file} (\emph{*.fuzzy}) is a Matlab file. It contains membership functions and a fuzzy rulebase.
The mapping to single features and to the output variable is done by variable names.
\item A \textmainindex{segment file} (\emph{*.einzug}) is an ASCII file. It defines segments for the feature
extraction from time series (see Section~\ref{sec:einzuege} for details).

\item A \textmainindex{macro file} (\emph{*.makrog}) is a text file containing a sequence of pushed menu items
and control elements. This file can be manually modified, e.g. by adding other Matlab commands.
\index{control elements}\index{menu items}
\item A \textmainindex{plugin sequence file} (\emph{*.plugseq}) is (binary) Matlab file. It contains a \textindex{plugin sequence} with plugin parameters.
\item A \textmainindex{SciXMiner project file} (\emph{*.prjz}) is a (binary) Matlab file. It contains at least the three-dimensional matrix
of time series (\emph{d\_orgs}) or the matrix with single features (\emph{d\_org}). In addition,
some additional data can be included as e.g. names of time series. This file can be loaded and
saved using the SciXMiner GUI or the Matlab command window (see
Section~\ref{ssec:manuelles_projekt}).
\item A \textmainindex{\LaTeX file} (\emph{*.tex}) is an ASCII file containing various results. It is prepared to be included using \LaTeX.
The user can switch between normal text files and \LaTeX files with the control element
\emph{Control element: General options - TEX protocol}. Some examples for result protocols are rule
bases (\emph{*\_cp*.tex}), feature relevances (\emph{*\_alle\_merkmale.tex},
\emph{*\_besten\_merkmale.tex}), mean values (\emph{*\_MWST.tex}) or \textindex{correlation}
coefficients (\emph{*\_CORR.tex}).
\item A \textmainindex{text file} (\emph{*.txt}) is an ASCII file containing various results. It can
be opened with any text editor. The user can switch between text files and \LaTeX files with the
control element \emph{Control element: General options - TEX protocol}. The results are similar to
\LaTeX files.
\item An \textmainindex{option file} (\emph{*.uihdg}) is a Matlab file containing all options defined by the control elements in Chapter~\ref{sec:optionen}.
\end{itemize}

\chapter{Important internal data structures}\label{sec:variablen}\thispagestyle{empty}
Please notice that SciXMiner was actually developed in German language. From this time, some
variable names are German. Unfortunately, it was impossible to change this inconvenience.

  \begin{center}
  \tablehead{\hline Name&Dimension&Remarks\\}
  \tabletail{\hline}

  \begin{supertabular}{|l|l|p{.6\textwidth}|}
   \hline
  \emph{bez\_code}
  & ($s_y,\text{any}$)
  & Vector (string array) with names of output variables\\ \hline
  \emph{cluster\_ergebnis}
  & (Struct)
  & Struct for results and parameters of cluster methods\\ \hline
  \emph{code}
  & ($N,1$)
  & Vector of the selected output variable, classes must be coded as serial integer values between 1 and $m_y$\\ \hline
  \emph{code\_alle}
  & ($N,s_y$)
  & Matrix of all existing output variables, classes must be coded as serial integer values between 1 and $m_y$
    \emph{code} contains a redundant copy for the selected output. \\ \hline
  \emph{d\_org}
  & ($N,s$)
  & Matrix of single features\\ \hline
  \emph{d\_orgs}
  & ($N,K,s_z$)
  & Matrix of time series (WARNING! Definition as global variable to avoid memory problems due to
  multiple copies in the called functions)\\ \hline
  \emph{dorgbez}
  & ($s,\text{any}$)
  & Matrix (string array). The rows contain names of single features. If the names are not defined in the project, standard names
  as x1 ... xs are chosen\\ \hline
  \emph{fuzzy\_system}
  & (Struct)
  & Struct for a designed fuzzy system \\ \hline
   \emph{gaitcad\_extern}
  & (Struct)
  & Struct for different variables (especially for macros and SciXMiner batch files), which are reloaded after the start of SciXMiner \\ \hline

  \emph{ind\_auswahl}
  & ($N_{red},1$)
  & Indices of selected data points\\ \hline
  \emph{interpret\_merk}
  & ($s,1$)
  & \textindex{a priori relevances} of single features (set to an empty variable by deactivation of \hyperlink{CE_A_Priori}{\emph{Control element: Single features - A priori feature relevances}}) \\ \hline
  \emph{interpret\_merk\_rett}
  & ($s,1$)
  & saved \textindex{a priori relevances} of single features (are not changed by deactivation of \hyperlink{CE_A_Priori}{\emph{Control element: Single features - A priori feature relevances}},
  but are necessary to restore \emph{interpret\_merk} after (re-activation)) \\ \hline
  \emph{klass\_single}
  & (Struct)
  & Struct with settings and parameters for feature classifiers \\ \hline
  \emph{klass\_zr}
  & (Struct)
  & Struct with settings and parameters for time-series classifiers\\ \hline
  \emph{L}
  & (Struct)
  & Struct with settings for decision theory. It can be initialized as empty. At the moment in use: L.ld: Matrix of \textindex{decision costs} for the recent output variable, L.ld\_alle (struct): Matrix of decision costs of all output variables
  \\ \hline
    \emph{makro\_lern}
  &($1,\text{any}$)
  & contains the macro as string, which is used for the design of classifier in a validation process\\ \hline
  \emph{makro\_test}
  &($1,\text{any}$)
  & contains the macro as string, which is used for the application of classifier in a validation process\\ \hline
  \emph{mu\_y}
  &($N,m_y$)
  & matrix, membership values of the estimated output variable as a result of the fuzzy analysis \\ \hline
  \emph{parameter}
  &(Struct)
  & Struct with parameters about the project. A detailed description is given later-on in this chapter\\ \hline
  \emph{pos}
  &$(N, 1)$
  & contains the estimated class membership for a chosen classifier. Data points which are not classified are marked with $0$.\\ \hline
  \emph{prz}
  &$(N, 1)$
  & contains the relative probability for a class estimation. Data points which are not classified are marked with $0$.\\ \hline
  \emph{var\_bez}
  & ($s_z,beliebig$)
  & matrix (string array), rows contain the denotations of the time series. They are set to standard terms, if they are not defined within the
  project file. After the last row of this variable the output variable is appended.\\ \hline
  \emph{zgf}
  & ($s+1,\max(m_l,m_y)$)
  & parameter of the membership functions (MBF) of the linguistic terms of the $s$ features and of the output variable in the element $s+1$,
    features: only trapezoid MBFs at the margins and triangular MBFs for all others permitted, output variable: singletons\\ \hline
  \emph{zgf\_bez} & ($s+1,\max(m_l,m_y)$) & array, contains the denotation of the linguistic terms of the
  $s$ features and of the output variable in the element \emph{name}. It is set to standard terms if it is not defined within the project file.\\ \hline
  \emph{zgf\_y\_bez} & ($s_y,\max(m_{y,i})$) & array,
  contains the denotations of the linguistic terms of the output variables in the element \emph{name}. It is set to standard terms if it is not defined
  within the project file.\\ \hline
  \end{supertabular}
  \end{center}

The \texttt{parameter}-Struct contains three elements, \texttt{allgemein, gui, projekt}. These elements again contain Structs.

\textit{parameter.allgemein} contains general information about paths, extensions, and status variables.

\textit{parameter.gui} encapsulates manages the \textindex{control elements} and contains variables, which correspond to the control elements.

By encapsulated functions it is ensured, that these variables always contain the current content.
Furthermore, the setting of these variables and the following actualization of the graphical user
interface is implemented, too.

\textit{parameter.projekt} contains values of the recent project, for instance file name, path of
the project and further status variables. In \textit{parameter.projekt.abtastfrequenz} the sampling
frequency of the time series within the project may be saved.

\chapter{Needed Standard Toolboxes}\label{sec:toolbox_Matlab} \thispagestyle{empty}
Some functions need standard toolboxes from Matlab:

the \textindex{Signal Processing Toolbox} is used
\begin{itemize}
 \item to filter time-series (functions \texttt{filter} and \texttt{butter}),
 \item to calculate spectrograms (function \texttt{specgram}),
 \item to calculate Cross and Auto Correlation Functions (function \texttt{xcorr}),\index{autocorrelation}
 \item to calculate Matlab wavelet decompositions (function \texttt{interp}) and
 \item to \textindex{import} time series with varying lengths (function \texttt{resample})
\end{itemize}

The use of \texttt{resample} is checked when time-series are imported\index{import}. If an error occurs, a warning will be displayed and the
import is proceeded with appended zeros.

The \textindex{Neural Network Toolbox} is necessary, to calculate and apply a \textindex{Artificial
Neural Network} as \textindex{classifier}.

The \textindex{Statistic Toolbox} is necessary,
\begin{itemize}
 \item to calculate paired or an unpaired \textindex{t-test},
 \item to use k-NN classifiers from this toolbox with other metrics than the Euclidean (function \texttt{pdist}) and
 \item to use the cluster-function (out of this toolbox).
\end{itemize}

The \textindex{Wavelet Toolbox} is needed, to do a Matlab \textindex{wavelet decomposition}. If
this toolbox is not available, a SciXMiner-integrated version may be used.

\chapter{Included External Toolboxes (GNU-License)}\label{sec:toolbox_gnu}\thispagestyle{empty}
\begin{lstlisting}

ICA - Toolbox
% FastICA for Matlab 5.x
% Version 2.1, January 15 2001
% Copyright (c)
%Hugo Gävert, Jarmo Hurri, Jaakko Särelä, and Aapo Hyvärinen (GNU-GPL).

fastica.m
fpica.m
pcamat.m
remmean.m
whiteenv.m

------------------------------------------------------------------------

% SVM and Kernel Methods Matlab Toolbox \cite{Canu03}
% http://asi.insa-rouen.fr/~arakotom/toolbox/index.html
% Copyright S CANU - scanu@insa-rouen.fr
 (GNU-GPL)

cout.m
monqp.m
svmclass.m
svmkernel.m
svmmulticlass.m
svmmulticlassoneagianstone.m
svmmultival.m
svmmultivaloneagianstone.m
svmreg.m
svmval.m

------------------------------------------------------------------------

SOM Toolbox
% Version 2.0beta, May 30 2002
% Copyright 1997-2000 by Esa Alhoniemi, Johan Himberg,
% Juha Parhankangas and Juha Vesanto (GNU-GPL)
% http://www.cis.hut.fi/projects/somtoolbox/
knn.m
som_eucdist2.m
vis_valuetype.m

------------------------------------------------------------------------
lp_solve 5.5.0.7

% http://lpsolve.sourceforge.net/5.5/
% Co-developers   : Michel Berkelaar, Kjell Eikland, Peter Notebaert
% Licence terms   : GNU LGPL (Lesser General Public Licence)
% Citation policy : General references as per LGPL
%                   Module specific references as specified therein

mxlpsolve.dll
lpsolve55.dll
mxlpsolve.m
lp_maker.m

The *.DLL have to be copied in <matlabroot>/bin/ !

------------------------------------------------------------------------
Kolmogorov-Smirnov-Test for distributions

% Armin Günther, Universität Greifswald, Germany

kstest.m

------------------------------------------------------------------------
XML read (BSD licencse)

%Jarek Tuszynski, SAIC
%http://www.mathworks.com/matlabcentral/fileexchange/12907-xmliotools

xmlread.m

------------------------------------------------------------------------
Fisherextest:Fisher's Exact Probability Test

%Trujillo-Ortiz, A., R. Hernandez-Walls, A. Castro-Perez, L.
%Rodriguez-Cardozo, N.A. Ramos-Delgado and R. Garcia-Sanchez.
%(2004). Fisherextest:Fisher's Exact Probability Test. A
%MATLAB file. [WWW document]. http://www.mathworks.com/
%matlabcentral/fileexchange/loadFile.do?objectId=5957

fisherextest.m

------------------------------------------------------------------------
VAT (visual assessment of tendency) - an algorithm to improve the visualization of data point distance matrices

% Copyright by Timothy Havens, Michigan Tech
% http://www.ece.mtu.edu/~thavens/code/VAT.m

vat.m

------------------------------------------------------------------------

Association Analysis (Apriori Algorithm):
Given a set of transactions, find rules that will predict the occurrence of an item based on the occurrences of other
items in the transaction

% author: Narine Manukyan 07/08/2013
% Copyright (c) 2013, Narine Manukyan

findRules.m

------------------------------------------------------------------------

Uipickfiles:
GUI program to select files and/or folders

% Version: 1.15, 2 March 2012
% Author:  Douglas M. Schwarz
% Email:   dmschwarz=ieee*org, dmschwarz=urgrad*rochester*edu
% Real_email = regexprep(Email,{'=','*'},{'@','.'})
% http://www.mathworks.com/matlabcentral/fileexchange/10867-uipickfiles--uigetfile-on-steroids

uipickfiles.m
uipickfiles_license.txt

------------------------------------------------------------------------
MATLAB2TIKZ    Save figure in native LaTeX (TikZ/Pgfplots).

%   Copyright (c) 2008--2014, Nico Schloemer <nico.schloemer@gmail.com>

m2tUpdater.m
cleanfigure.m
figure2dot.m
m2tInputParser.m
matlab2tikz.m

------------------------------------------------------------------------
LIBSVM

% http://www.csie.ntu.edu.tw/~cjlin/libsvm
% Version:	3.20
% Author:	Chih-Chung Chang and Chih-Jen Lin
% License:	modified BSD license

libsvm-3.20\*.*

------------------------------------------------------------------------
ARESLab

% http://www.cs.rtu.lv/jekabsons/
% Version:	1.13.0
% Author: 	Gints Jekabsons (gints.jekabsons@rtu.lv)
% License:	GNU GPL

------------------------------------------------------------------------
CSV read

% https://github.com/gideongfeller/matlab/blob/master/read_mixed_csv.m
% license:	none

read_mixed_csv.m
\end{lstlisting}

\chapter{Symbols and abbreviations}\label{sec:symbol}\thispagestyle{empty}
\begin{center}
\begin{tabular}{|l|p{.7\textwidth}|}
\hline
Symbol&Meaning \\
\hline
 ACF          & Auto correlation function \\
 ANOVA        & Analysis of Variances \\
 CCF          & Cross Correlation Function \\
 CCOEF        & Cross Correlation Function \\
 DA           & Discriminant analysis \\
 FFT          & Fast Fourier transformation \\
 ICA          & Independent Component Analysis \\
 k-NN         & $k$-nearest-neighbor \\
 MBF          & Membership function \\
 MEAN         & Mean value \\
 MLP          & Multi-layer perceptron \\
 MANOVA       & Multivariate Analysis of Variances \\
 PCA          & Principal Component Analysis \\
 SF           & Single feature \\
 STD          & Standard deviation \\
 SVM          & Support Vector Machines \\
 TS           & Time series \\
 $K$          & Number of \textindex{sample points}\\
 $m$          & Number of linguistic terms of all features\\
 $m_l$        & Number of linguistic terms of the $l$-th feature $x_l$ \\
 $m_y$        & Number of linguistic terms (classes) of the output variable  \\
 $N$          & Number of \textindex{data points} \\
 $s$          & Number of \textindex{features} \\
 $s_m$        & Number of selected features \\
 $s_d$        & Number of transformed (aggregated) features \\
 $s_z$        & Number of time-series \\

\hline
\end{tabular}
\end{center}

\chapter{Known errors and problems}\label{sec:anhang_fehler}\thispagestyle{empty}
\begin{itemize}
\item SciXMiner needs certain variable, which must not be overwritten or deleted. For instance, the variable \textit{parameter}
      encapsulates the  \textindex{control elements} and enables SciXMiner to internally access
      these  variables.  If this Variable is deleted or does not contain the needed elements, a further execution of SciXMiner is not
      possible. The only possible solution is a restart. Further necessary variables are listed in Appendix~\ref{sec:variablen}.
\item To execute a macro as well the needed features and/or time series have to exist in within the project as the output variables. There may
      be the possibility that a \textindex{macro} has to be recorded separately for each project to ensure a proper functionality.
\item If a macro is recorded, it saves the names of selected elements in a popup window for a reconstruction. If the name is not available
      during a replay, an error message ''Inconsistent selection fields (see MATLAB command window for further
      details)'' is shown. In this case, you should verify the available elements in this
      field. Possible reasons are new names for variables or changed names for program options.
\item Frequently opening and closing of the SciXMiner GUI leads to a slower processing of figures. This problem
      may be traced back to an incorrect deletion of control and menu items in Matlab.
\item At the moment, massive problems with lp\_solve occur due to version conflicts.
\item Using the novelty detection with the one-class method the following error message occurs:
      ''Error! lp\_solve library not found. Please copy *.dll from the toolbox path in ...$\backslash$bin'': The novelty detection with the
      one-class method uses an external toolbox, which needs a dynamic link libraries (DLL). Matlab is not able to find these libraries, if they are not contained in the general Matlab path for libraries. This directory is
      located in ''$<$matlabroot$>$$\backslash$bin''. The directory $<$matlabroot$>$ is given in Matlab by using the command \texttt{matlabroot}.
      The following files have to be copied from the SciXMiner subdirectory ''toolbox'' to the above mentioned directory: ''mxlpsolve.dll''
      and ''lpsolve55.dll''.
\item Applying the Matlab \textindex{wavelet decomposition} the following error message occurs: ''Error using ==$>$ interp Length of data
      sequence must be at least 9 You either need more data or a shorter filter (L).'':
      The integrated Matlab function calculates time-series which are shorter than the original time series. A plugin interpolates them to the correct
      length. The used command for interpolation (\texttt{interp}) interrupts if time series are too short. Indeed, the used number of sample points
      may be reduced for the interpolation. However, this error correlates with another error in the \textindex{wavelet decomposition}, which
      originates from a too high number of levels. To solve the problem reduce the number of levels or deactivate the Matlab wavelet decomposition.
      Then, an alternative implementation will be used.
\item Errors occur for the \textindex{macro} recording if popup elements have digits as first characters.
\item Missing MATLAB functions can cause SciXMiner errors in older MATLAB versions:
\begin{itemize}
\item The MATLAB function ''iscolumn'' is available from MATLAB version 2011a.
\end{itemize}
\end{itemize}
\chapter{Version history}\label{sec:versionen}

\section{Versions}
\begin{itemize}
   \item earlier Gait-CAD-Versions, see  version history in Gait-CAD documentation
   \item Version 2016b (13.03.2017) 
   \item Version 2017a (12.04.2017) 
\end{itemize}

 \section{Selected changes between Gait-CAD version 2014b and SciXMiner version 2016b}
\begin{itemize}
 \item integration of support vector regression
 \item option to generate polynomial models without absolute value ($a_0$ term)
 \item handling of timestamp data (conversion from strings in linguistic terms of output variables to timestamps resp. from timestamps to date and time)
 \item visualization of multidimensional single features
 \item optional hiding of figures in batch files
 \item extended conversion options from data points into time series
 \item export of the error of the recent regression model into a single feature
 \item deletion of selected or unselected sample points in time series
 \item deletion of unselected single features
 \item additional style options for protocol files of classes and single features
 \item switchable percentage statistics in 2D histograms
 \item various internal changes related to the transfer from Gait-CAD to SciXMiner (naming, internal code optimizations etc.)
 \item large variety of bug fixes
 \item compatibility to MATLAB 2016b
 \end{itemize}

 \section{Selected changes between SciXMiner versions 2016b and 2017a}
\begin{itemize}
 \item compatibility to MATLAB 2017a
\end{itemize}

\chapter{Plugins}\label{sec:appendix_plugin}
\begin{itemize}

 \item {\bf COG SF (COG)}: computes the center of gravity of a time series or a time series segment.
 \begin{itemize}
\item Function name:    plugin\_cog\_em.m
\item Type:             SF
\item Time series: 1 inputs, 0 outputs, Segments possible:   yes
\item Single features: 0 inputs, 1 outputs
\item Images: 0 inputs, 0 outputs
\item Direct callback:  none
\item Number of parameters: 0
\end{itemize}
 \item {\bf Maximum (MAX)}: computes the maximum of a time series or a time series segment as single feature.
 \begin{itemize}
\item Function name:    plugin\_max.m
\item Type:             SF
\item Time series: 1 inputs, 0 outputs, Segments possible:   yes
\item Single features: 0 inputs, 1 outputs
\item Images: 0 inputs, 0 outputs
\item Direct callback:  none
\item Number of parameters: 0
\end{itemize}
 \item {\bf Maximum position (MAPO)}: computes the position of the maximum (number of the sample point) for a time series or a time series segment as single feature.
 \begin{itemize}
\item Function name:    plugin\_mapo.m
\item Type:             SF
\item Time series: 1 inputs, 0 outputs, Segments possible:   yes
\item Single features: 0 inputs, 1 outputs
\item Images: 0 inputs, 0 outputs
\item Direct callback:  none
\item Number of parameters: 0
\end{itemize}
 \item {\bf Mean value SF (MEAN)}: computes the mean of a time series or a time series segment as single feature.
 \begin{itemize}
\item Function name:    plugin\_mean\_em.m
\item Type:             SF
\item Time series: 1 inputs, 0 outputs, Segments possible:   yes
\item Single features: 0 inputs, 1 outputs
\item Images: 0 inputs, 0 outputs
\item Direct callback:  none
\item Number of parameters: 0
\end{itemize}
 \item {\bf Mean value SF NaN (MEAN)}: computes the mean of a time series or a time series segment as single feature. NaN values in a time series will be ignored.
 \begin{itemize}
\item Function name:    plugin\_mean\_em\_nan.m
\item Type:             SF
\item Time series: 1 inputs, 0 outputs, Segments possible:   yes
\item Single features: 0 inputs, 1 outputs
\item Images: 0 inputs, 0 outputs
\item Direct callback:  none
\item Number of parameters: 0
\end{itemize}
 \item {\bf Median SF (MEDIAN)}: computes the median of a time series or a time series segment as single feature.
 \begin{itemize}
\item Function name:    plugin\_median\_em.m
\item Type:             SF
\item Time series: 1 inputs, 0 outputs, Segments possible:   yes
\item Single features: 0 inputs, 1 outputs
\item Images: 0 inputs, 0 outputs
\item Direct callback:  none
\item Number of parameters: 0
\end{itemize}
 \item {\bf Median SF NaN (MEDIAN)}: computes the median of a time series or a time series segment as single feature. NaN values in a time series will be ignored.
 \begin{itemize}
\item Function name:    plugin\_median\_em\_nan.m
\item Type:             SF
\item Time series: 1 inputs, 0 outputs, Segments possible:   yes
\item Single features: 0 inputs, 1 outputs
\item Images: 0 inputs, 0 outputs
\item Direct callback:  none
\item Number of parameters: 0
\end{itemize}
 \item {\bf Minimum (MIN)}: computes the minimum of a time series or a time series segment as single feature.
 \begin{itemize}
\item Function name:    plugin\_min.m
\item Type:             SF
\item Time series: 1 inputs, 0 outputs, Segments possible:   yes
\item Single features: 0 inputs, 1 outputs
\item Images: 0 inputs, 0 outputs
\item Direct callback:  none
\item Number of parameters: 0
\end{itemize}
 \item {\bf Minimum position (MIPO)}: computes the position of the minimum (number of the sample point) for a time series or a time series segment as single feature.
 \begin{itemize}
\item Function name:    plugin\_mipo.m
\item Type:             SF
\item Time series: 1 inputs, 0 outputs, Segments possible:   yes
\item Single features: 0 inputs, 1 outputs
\item Images: 0 inputs, 0 outputs
\item Direct callback:  none
\item Number of parameters: 0
\end{itemize}
 \item {\bf Norm deviation (absolute value) (ND Abs)}: computes the mean absolute norm deviation of a time series in a segment as single feature (mean value of  Eq. (3) in \cite{Wolf06} for a segment)
 \begin{itemize}
\item Function name:    plugin\_normzahl\_betrag.m
\item Type:             SF
\item Time series: 1 inputs, 0 outputs, Segments possible:   yes
\item Single features: 0 inputs, 1 outputs
\item Images: 0 inputs, 0 outputs
\item Direct callback:  none
\item Number of parameters: 0
\end{itemize}
 \item {\bf Norm deviation (direction) (ND Dir)}: computes the mean value of the directed deviation from a time series to a norm time series in a segment as a single feature. (mean value of the directed norm deviation for a segment,  Eq. (3) in \cite{Wolf06} without the absolute value in the nominator)
 \begin{itemize}
\item Function name:    plugin\_normzahl\_richtung.m
\item Type:             SF
\item Time series: 1 inputs, 0 outputs, Segments possible:   yes
\item Single features: 0 inputs, 1 outputs
\item Images: 0 inputs, 0 outputs
\item Direct callback:  none
\item Number of parameters: 0
\end{itemize}
 \item {\bf Range of Motion (ROM)}: computes the range (of motion) of a time series or a time series segment as single feature.
 \begin{itemize}
\item Function name:    plugin\_rom.m
\item Type:             SF
\item Time series: 1 inputs, 0 outputs, Segments possible:   yes
\item Single features: 0 inputs, 1 outputs
\item Images: 0 inputs, 0 outputs
\item Direct callback:  none
\item Number of parameters: 0
\end{itemize}
 \item {\bf STD SF (STD SF)}: computes the standard deviation of all values of a time series or a time series segment. Normalization to K instead of K-1
 \begin{itemize}
\item Function name:    plugin\_std\_em.m
\item Type:             SF
\item Time series: 1 inputs, 0 outputs, Segments possible:   yes
\item Single features: 0 inputs, 1 outputs
\item Images: 0 inputs, 0 outputs
\item Direct callback:  none
\item Number of parameters: 0
\end{itemize}
 \item {\bf Sum SF (SUM)}: computes the sum if a time series or of a time series segment as single feature.
 \begin{itemize}
\item Function name:    plugin\_sum\_em.m
\item Type:             SF
\item Time series: 1 inputs, 0 outputs, Segments possible:   yes
\item Single features: 0 inputs, 1 outputs
\item Images: 0 inputs, 0 outputs
\item Direct callback:  none
\item Number of parameters: 0
\end{itemize}
 \item {\bf TS->DISCR SF MEAN (TERMQ1TERMQ2TERMQ3TERMQ4TERMQ5)}: computes the frequencies for  (crisp) discretizations in time series.
 \begin{itemize}
\item Function name:    plugin\_discr\_em.m
\item Type:             SF
\item Time series: 1 inputs, 0 outputs, Segments possible:   yes
\item Single features: 0 inputs, 5 outputs
\item Images: 0 inputs, 0 outputs
\item Direct callback:  none
\item Number of parameters: 2

 \begin{itemize}
\item Number of terms (Number of terms for discretization)
\item Type of membership function (defines the algorithm for the design of a membership function  (see
''Control element: Data mining: Special methods - Number of linguistic terms'').

''Median'' generates terms with approximately equal data point frequencies for all terms in the
training data set.

''Equal distribution'' generate terms with similar distances between the maximum values of the
membership functions.

''Clustering'' uses cluster prototypes as parameters.

''Fix'' set the parameters of the membership functions to the values defined by ''Control element: Data mining: Special methods - Parameters MBF (fix)''.

The choice ''with interpretability'' rounds the found parameters using a method
described in Literature: Mikut05 to improve the interpretability.

Remark:

All design methods set the random number generator of Matlab to a fix value. It improves the
reproducibility of found membership functions. Nevertheless, all methods with random elements might
be sensitive to the state of the random number generator (e.g. cluster algorithms with random start
clusters). It can also influence other methods using these membership functions, e.g. by different
selected features in a decision tree (e.g. ''Data mining - Selection and evaluation of single features - Information theoretic measures''). )
\end{itemize}\end{itemize}
 \item {\bf TS->FUZZY SF MEAN (TERM1TERM2TERM3TERM4TERM5)}: computes the frequency of fuzzy terms in a fuzzified time series as single features.
 \begin{itemize}
\item Function name:    plugin\_fuzzy\_em.m
\item Type:             SF
\item Time series: 1 inputs, 0 outputs, Segments possible:   yes
\item Single features: 0 inputs, 5 outputs
\item Images: 0 inputs, 0 outputs
\item Direct callback:  none
\item Number of parameters: 2

 \begin{itemize}
\item Number of terms (Number of terms for discretization)
\item Type of membership function (defines the algorithm for the design of a membership function  (see
''Control element: Data mining: Special methods - Number of linguistic terms'').

''Median'' generates terms with approximately equal data point frequencies for all terms in the
training data set.

''Equal distribution'' generate terms with similar distances between the maximum values of the
membership functions.

''Clustering'' uses cluster prototypes as parameters.

''Fix'' set the parameters of the membership functions to the values defined by ''Control element: Data mining: Special methods - Parameters MBF (fix)''.

The choice ''with interpretability'' rounds the found parameters using a method
described in Literature: Mikut05 to improve the interpretability.

Remark:

All design methods set the random number generator of Matlab to a fix value. It improves the
reproducibility of found membership functions. Nevertheless, all methods with random elements might
be sensitive to the state of the random number generator (e.g. cluster algorithms with random start
clusters). It can also influence other methods using these membership functions, e.g. by different
selected features in a decision tree (e.g. ''Data mining - Selection and evaluation of single features - Information theoretic measures''). )
\end{itemize}\end{itemize}
 \item {\bf TS->PC SF (PC1PC2)}: computes new single features using a Principal Component Analysis of a time series or a time series segment (K -> $s\_d$).
 \begin{itemize}
\item Function name:    plugin\_zrhkem.m
\item Type:             SF
\item Time series: 1 inputs, 0 outputs, Segments possible:   yes
\item Single features: 0 inputs, 2 outputs
\item Images: 0 inputs, 0 outputs
\item Direct callback:  none
\item Number of parameters: 2

 \begin{itemize}
\item Number of aggregated PCA features (Number s\_d of principal components from a time series (Transformation number of sample points K -> s\_d))
\item Normalize standard deviations (defines the normalization for the variances of the sample points)
\end{itemize}\end{itemize}
 \item {\bf TS->SF (TSSP)}: extracts the value of the time series for one sample point as single feature.
 \begin{itemize}
\item Function name:    plugin\_zr\_em.m
\item Type:             SF
\item Time series: 1 inputs, 0 outputs, Segments possible:   none
\item Single features: 0 inputs, 1 outputs
\item Images: 0 inputs, 0 outputs
\item Direct callback:  none
\item Number of parameters: 1

 \begin{itemize}
\item Number of the sample point (Sample point for Time series -> Single feature)
\end{itemize}\end{itemize}
 \item {\bf below/above norm mean value (ND Sign)}: computes a single feature indicating values above or below the norm time series
 \begin{itemize}
\item Function name:    plugin\_normzahl\_mittelwert.m
\item Type:             SF
\item Time series: 1 inputs, 0 outputs, Segments possible:   yes
\item Single features: 0 inputs, 1 outputs
\item Images: 0 inputs, 0 outputs
\item Direct callback:  none
\item Number of parameters: 0
\end{itemize}
 \item {\bf Absolute value (ABS)}: rectifies a time series by computing the absolute values
 \begin{itemize}
\item Function name:    plugin\_abs.m
\item Type:             TS
\item Time series: 1 inputs, 1 outputs, Segments possible:   none
\item Single features: 0 inputs, 0 outputs
\item Images: 0 inputs, 0 outputs
\item Direct callback:  none
\item Number of parameters: 0
\end{itemize}
 \item {\bf Acausal median of a window (FE-MED-AC)}: sets all values of the time series in the window to the median value of the window. Here, an acausal filter is used.
 \begin{itemize}
\item Function name:    plugin\_zr\_fenster\_medac.m
\item Type:             TS
\item Time series: 1 inputs, 1 outputs, Segments possible:   none
\item Single features: 0 inputs, 0 outputs
\item Images: 0 inputs, 0 outputs
\item Direct callback:  none
\item Number of parameters: 1

 \begin{itemize}
\item Window length (determines the window length in sample points for the filtering with a sliding window.)
\end{itemize}\end{itemize}
 \item {\bf Acceleration (A)}: computes the second time derivation of a time series. see Eq. (3.3) in \cite{Loose04}).
 \begin{itemize}
\item Function name:    plugin\_beschleunigung.m
\item Type:             TS
\item Time series: 1 inputs, 1 outputs, Segments possible:   none
\item Single features: 0 inputs, 0 outputs
\item Images: 0 inputs, 0 outputs
\item Direct callback:  none
\item Number of parameters: 1

 \begin{itemize}
\item Time (defines the scaling (per sample point resp. per time using the sampling frequency from the GUI))
\end{itemize}\end{itemize}
 \item {\bf Addition of time series (ADDTS)}: adds all selected time series.
 \begin{itemize}
\item Function name:    plugin\_add\_zr.m
\item Type:             TS
\item Time series: Inf inputs, 1 outputs, Segments possible:   none
\item Single features: 0 inputs, 0 outputs
\item Images: 0 inputs, 0 outputs
\item Direct callback:  none
\item Number of parameters: 0
\end{itemize}
 \item {\bf Change sign (SIGNIN)}: multiplies a time series with $-1$
 \begin{itemize}
\item Function name:    plugin\_vorzeichen\_umkehr.m
\item Type:             TS
\item Time series: 1 inputs, 1 outputs, Segments possible:   none
\item Single features: 0 inputs, 0 outputs
\item Images: 0 inputs, 0 outputs
\item Direct callback:  none
\item Number of parameters: 0
\end{itemize}
 \item {\bf Compute square root (ROOT)}: Compute root of the amplitudes of a time series.
 \begin{itemize}
\item Function name:    plugin\_root.m
\item Type:             TS
\item Time series: 1 inputs, 1 outputs, Segments possible:   none
\item Single features: 0 inputs, 0 outputs
\item Images: 0 inputs, 0 outputs
\item Direct callback:  none
\item Number of parameters: 0
\end{itemize}
 \item {\bf Compute trend (Trend)}:  Here, two first order IIR filters with the parameter aF (fast) and aS (slow) are used. The computation is explained in \cite{Reischl02}.
 \begin{itemize}
\item Function name:    plugin\_iirfilter\_trend.m
\item Type:             TS
\item Time series: 1 inputs, 1 outputs, Segments possible:   none
\item Single features: 0 inputs, 0 outputs
\item Images: 0 inputs, 0 outputs
\item Direct callback:  none
\item Number of parameters: 1

 \begin{itemize}
\item Parameter aF aS (to 0: no smoothing, to 1: strong smoothing. aSlow must be greater than aFast.)
\end{itemize}\end{itemize}
 \item {\bf Difference of two time series (DIFF)}: computes the difference between the first and second time series.
 \begin{itemize}
\item Function name:    plugin\_diffzr.m
\item Type:             TS
\item Time series: 2 inputs, 1 outputs, Segments possible:   none
\item Single features: 0 inputs, 0 outputs
\item Images: 0 inputs, 0 outputs
\item Direct callback:  none
\item Number of parameters: 0
\end{itemize}
 \item {\bf Estimate standard deviation (StdTS)}: applies different digital lowpass filters $x\_f [k+1] = a * x\_f [k]  + (1-a) * x [k]$ with parameters aFast, aSlow and aSigma to compute the standard deviation. see Eq. (2-6) in \cite{Reischl02} with aF = aSigma, aL = aSlow, aS = aFast.
 \begin{itemize}
\item Function name:    plugin\_iirfilter\_stdschaetzer.m
\item Type:             TS
\item Time series: 1 inputs, 1 outputs, Segments possible:   none
\item Single features: 0 inputs, 0 outputs
\item Images: 0 inputs, 0 outputs
\item Direct callback:  none
\item Number of parameters: 1

 \begin{itemize}
\item Parameters aFast aSlow aSigma (to 0: no smoothing, to 1: strong smoothing. aSlow must be greater than aFast.)
\end{itemize}\end{itemize}
 \item {\bf Filtered maximum (Fil-MAX)}: sliding maximum value with exponential forgetting. Here, a causal filter is used.
 \begin{itemize}
\item Function name:    plugin\_zr\_gefiltert\_max.m
\item Type:             TS
\item Time series: 1 inputs, 1 outputs, Segments possible:   none
\item Single features: 0 inputs, 0 outputs
\item Images: 0 inputs, 0 outputs
\item Direct callback:  none
\item Number of parameters: 0
\end{itemize}
 \item {\bf Filtered minimum (Fil-MIN)}: sliding minimum value with exponential forgetting. Here, a causal filter is used.
 \begin{itemize}
\item Function name:    plugin\_zr\_gefiltert\_min.m
\item Type:             TS
\item Time series: 1 inputs, 1 outputs, Segments possible:   none
\item Single features: 0 inputs, 0 outputs
\item Images: 0 inputs, 0 outputs
\item Direct callback:  none
\item Number of parameters: 0
\end{itemize}
 \item {\bf Filtering (FIL)}: filters a time series.  A \textindex{Butterworth filter} is used.This plugin needs the Signal Processing Toolbox of Matlab. \index{Highpass filter}\index{Lowpass filter}The Bode plot can be shown under Show - Time series.
 \begin{itemize}
\item Function name:    plugin\_filter.m
\item Type:             TS
\item Time series: 1 inputs, 1 outputs, Segments possible:   none
\item Single features: 0 inputs, 0 outputs
\item Images: 0 inputs, 0 outputs
\item Direct callback:  none
\item Number of parameters: 4

 \begin{itemize}
\item Filter type (determines the filter characteristics in the frequency domain.)
\item Frequencies (defines the critical frequencies for the filter using the unit of the sample frequency (e.g. in Hz). The second parameter is only used for bandpass filters.)
\item Filter order (FIL) (determines the order of a filter.)
\item Initial values (computes initial values of the filter, Filtic+Static assumes a steady state filter at k=1, computed with MATLAB function filtic.m)
\end{itemize}\end{itemize}
 \item {\bf Filtering with Morlet wavelet (Morl)}: 
 \begin{itemize}
\item Function name:    plugin\_morletfilter.m
\item Type:             TS
\item Time series: 1 inputs, 1 outputs, Segments possible:   none
\item Single features: 0 inputs, 0 outputs
\item Images: 0 inputs, 0 outputs
\item Direct callback:  none
\item Number of parameters: 3

 \begin{itemize}
\item Morlet wavelet: frequency (This frequency describes the center of the region which is not damped by the Morlet-Wavelet.)
\item Morlet wavelet: frequency (The width of the region is defined by the eigenfrequency of the Morlet wavelet)
\item Causal Morlet wavelet (Use the same transformation matrix (e.g. PCA) for time series reduction of all data points)
\end{itemize}\end{itemize}
 \item {\bf IIR filter (IIR)}: applies a digital lowpass filter $x\_f [k+1] = a * x\_f [k]  + (1-a) * x [k]$.
 \begin{itemize}
\item Function name:    plugin\_iirfilter.m
\item Type:             TS
\item Time series: 1 inputs, 1 outputs, Segments possible:   none
\item Single features: 0 inputs, 0 outputs
\item Images: 0 inputs, 0 outputs
\item Direct callback:  none
\item Number of parameters: 1

 \begin{itemize}
\item Parameter a (to 0: no smoothing, to 1: strong smoothing)
\end{itemize}\end{itemize}
 \item {\bf Individual norm deviation (NABW)}:  Here, the plugins IIR (Parameter aF) and StdTS are used for the estimation
 \begin{itemize}
\item Function name:    plugin\_normabweichung.m
\item Type:             TS
\item Time series: 1 inputs, 1 outputs, Segments possible:   none
\item Single features: 0 inputs, 0 outputs
\item Images: 0 inputs, 0 outputs
\item Direct callback:  none
\item Number of parameters: 1

 \begin{itemize}
\item Parameters aFast aSlow aSigma (to 0: no smoothing, to 1: strong smoothing. aSlow must be greater than aFast.)
\end{itemize}\end{itemize}
 \item {\bf Jerk (J)}: computes the jerk (3rd time derivation) of a time series (acausal).
 \begin{itemize}
\item Function name:    plugin\_ruck.m
\item Type:             TS
\item Time series: 1 inputs, 1 outputs, Segments possible:   none
\item Single features: 0 inputs, 0 outputs
\item Images: 0 inputs, 0 outputs
\item Direct callback:  none
\item Number of parameters: 1

 \begin{itemize}
\item Time (defines the scaling (per sample point resp. per time using the sampling frequency from the GUI))
\end{itemize}\end{itemize}
 \item {\bf Jump to zero (ZEROJump)}: set value  = 1 for a jump to zeros, and value = 0 otherwise.
 \begin{itemize}
\item Function name:    plugin\_nsprung.m
\item Type:             TS
\item Time series: 1 inputs, 1 outputs, Segments possible:   none
\item Single features: 0 inputs, 0 outputs
\item Images: 0 inputs, 0 outputs
\item Direct callback:  none
\item Number of parameters: 0
\end{itemize}
 \item {\bf Logarithm 10 TS (LOG10)}: computes the logarithm to the base 10 for all values of a time series.
 \begin{itemize}
\item Function name:    plugin\_log10\_zr.m
\item Type:             TS
\item Time series: 1 inputs, 1 outputs, Segments possible:   none
\item Single features: 0 inputs, 0 outputs
\item Images: 0 inputs, 0 outputs
\item Direct callback:  none
\item Number of parameters: 0
\end{itemize}
 \item {\bf Maximum of a window (FE-MAX)}: sets all values of the time series in the window to the maximum value of the window. Here, a causal filter is used.
 \begin{itemize}
\item Function name:    plugin\_zr\_fenster\_max.m
\item Type:             TS
\item Time series: 1 inputs, 1 outputs, Segments possible:   none
\item Single features: 0 inputs, 0 outputs
\item Images: 0 inputs, 0 outputs
\item Direct callback:  none
\item Number of parameters: 1

 \begin{itemize}
\item Window length (determines the window length in sample points for the filtering with a sliding window.)
\end{itemize}\end{itemize}
 \item {\bf Maximum of multiple time series (MAXTS)}: computes the maximum for all time series in the configuration window.
 \begin{itemize}
\item Function name:    plugin\_max\_zr.m
\item Type:             TS
\item Time series: Inf inputs, 1 outputs, Segments possible:   none
\item Single features: 0 inputs, 0 outputs
\item Images: 0 inputs, 0 outputs
\item Direct callback:  none
\item Number of parameters: 0
\end{itemize}
 \item {\bf Mean of a window (FE-MEAN)}: sets all values of the time series in the window to the mean value of the window. Here, a causal filter is used.
 \begin{itemize}
\item Function name:    plugin\_zr\_fenster\_mean.m
\item Type:             TS
\item Time series: 1 inputs, 1 outputs, Segments possible:   none
\item Single features: 0 inputs, 0 outputs
\item Images: 0 inputs, 0 outputs
\item Direct callback:  none
\item Number of parameters: 1

 \begin{itemize}
\item Window length (determines the window length in sample points for the filtering with a sliding window.)
\end{itemize}\end{itemize}
 \item {\bf Mean value of multiple time series (MEANTS)}: computes the mean for all time series in the configuration window.
 \begin{itemize}
\item Function name:    plugin\_mean\_zr.m
\item Type:             TS
\item Time series: Inf inputs, 1 outputs, Segments possible:   none
\item Single features: 0 inputs, 0 outputs
\item Images: 0 inputs, 0 outputs
\item Direct callback:  none
\item Number of parameters: 0
\end{itemize}
 \item {\bf Median of a window (FE-MED)}: sets all values of the time series in the window to the median value of the window. Here, a causal filter is used.
 \begin{itemize}
\item Function name:    plugin\_zr\_fenster\_median.m
\item Type:             TS
\item Time series: 1 inputs, 1 outputs, Segments possible:   none
\item Single features: 0 inputs, 0 outputs
\item Images: 0 inputs, 0 outputs
\item Direct callback:  none
\item Number of parameters: 1

 \begin{itemize}
\item Window length (determines the window length in sample points for the filtering with a sliding window.)
\end{itemize}\end{itemize}
 \item {\bf Minimum of a window (FE-MIN)}: sets all values of the time series in the window to the minimum value of the window. Here, a causal filter is used.
 \begin{itemize}
\item Function name:    plugin\_zr\_fenster\_min.m
\item Type:             TS
\item Time series: 1 inputs, 1 outputs, Segments possible:   none
\item Single features: 0 inputs, 0 outputs
\item Images: 0 inputs, 0 outputs
\item Direct callback:  none
\item Number of parameters: 1

 \begin{itemize}
\item Window length (determines the window length in sample points for the filtering with a sliding window.)
\end{itemize}\end{itemize}
 \item {\bf Minimum of multiple time series (MINTS)}: computes the minimum for all time series in the configuration window.
 \begin{itemize}
\item Function name:    plugin\_min\_zr.m
\item Type:             TS
\item Time series: Inf inputs, 1 outputs, Segments possible:   none
\item Single features: 0 inputs, 0 outputs
\item Images: 0 inputs, 0 outputs
\item Direct callback:  none
\item Number of parameters: 0
\end{itemize}
 \item {\bf Multiplication of time series (MULTTS)}: multiplies all time series that are selected in the configuration window.
 \begin{itemize}
\item Function name:    plugin\_mult\_zr.m
\item Type:             TS
\item Time series: Inf inputs, 1 outputs, Segments possible:   none
\item Single features: 0 inputs, 0 outputs
\item Images: 0 inputs, 0 outputs
\item Direct callback:  none
\item Number of parameters: 0
\end{itemize}
 \item {\bf Multiplication with a constant (CONST)}: Multiplication with a constant.
 \begin{itemize}
\item Function name:    plugin\_mult\_const.m
\item Type:             TS
\item Time series: 1 inputs, 1 outputs, Segments possible:   none
\item Single features: 0 inputs, 0 outputs
\item Images: 0 inputs, 0 outputs
\item Direct callback:  none
\item Number of parameters: 1

 \begin{itemize}
\item Constant gain
\end{itemize}\end{itemize}
 \item {\bf Norm deviation time series absolute value (NDTS ABS)}: computes the absolute value of the norm deviation as time series. (Eq. (3) in \cite{Wolf06})
 \begin{itemize}
\item Function name:    plugin\_normzeitreihe\_abs.m
\item Type:             TS
\item Time series: 1 inputs, 1 outputs, Segments possible:   none
\item Single features: 0 inputs, 0 outputs
\item Images: 0 inputs, 0 outputs
\item Direct callback:  none
\item Number of parameters: 0
\end{itemize}
 \item {\bf Norm time series (NDTS)}: computes the deviation from a time series to a norm time series as a new time series. (Eq. (3.9) in \cite{Loose04})
 \begin{itemize}
\item Function name:    plugin\_normzeitreihe.m
\item Type:             TS
\item Time series: 1 inputs, 1 outputs, Segments possible:   none
\item Single features: 0 inputs, 0 outputs
\item Images: 0 inputs, 0 outputs
\item Direct callback:  none
\item Number of parameters: 0
\end{itemize}
 \item {\bf Normalized time series (NORM)}: normalizes a time series or a time series segment.
 \begin{itemize}
\item Function name:    plugin\_normalized\_ts.m
\item Type:             TS
\item Time series: 1 inputs, 1 outputs, Segments possible:   none
\item Single features: 0 inputs, 0 outputs
\item Images: 0 inputs, 0 outputs
\item Direct callback:  none
\item Number of parameters: 1

 \begin{itemize}
\item Type of normalization (defines the type of normalization for a time series)
\end{itemize}\end{itemize}
 \item {\bf Normalized to mean value (NORMMEAN)}: normalizes the time series to the mean value normalizes the time series to the mean value
 \begin{itemize}
\item Function name:    plugin\_mean\_norm\_ts.m
\item Type:             TS
\item Time series: 1 inputs, 1 outputs, Segments possible:   none
\item Single features: 0 inputs, 0 outputs
\item Images: 0 inputs, 0 outputs
\item Direct callback:  none
\item Number of parameters: 0
\end{itemize}
 \item {\bf ROM of a window (FE-ROM)}: sets all values of the time series in the window to the range of the window. Here, a causal filter is used.
 \begin{itemize}
\item Function name:    plugin\_zr\_fenster\_rom.m
\item Type:             TS
\item Time series: 1 inputs, 1 outputs, Segments possible:   none
\item Single features: 0 inputs, 0 outputs
\item Images: 0 inputs, 0 outputs
\item Direct callback:  none
\item Number of parameters: 1

 \begin{itemize}
\item Window length (determines the window length in sample points for the filtering with a sliding window.)
\end{itemize}\end{itemize}
 \item {\bf Relative ratio of two time series (RELRAT)}: computes the relative ratio of the first time series to the sum of both time series selected in the configuration window.
 \begin{itemize}
\item Function name:    plugin\_verhzr.m
\item Type:             TS
\item Time series: 2 inputs, 1 outputs, Segments possible:   none
\item Single features: 0 inputs, 0 outputs
\item Images: 0 inputs, 0 outputs
\item Direct callback:  none
\item Number of parameters: 0
\end{itemize}
 \item {\bf Remove trend (DETREND)}: corrects the linear trend of a time series (MATLAB function detrend) corrects the linear trend of a time series (MATLAB function detrend)
 \begin{itemize}
\item Function name:    plugin\_detrend.m
\item Type:             TS
\item Time series: 1 inputs, 1 outputs, Segments possible:   none
\item Single features: 0 inputs, 0 outputs
\item Images: 0 inputs, 0 outputs
\item Direct callback:  none
\item Number of parameters: 1

 \begin{itemize}
\item Detrend method (<html>Direction of model:<br>linear   - removes a continuous, piecewise linear trend;<br>constant - removes just the mean value<html>)
\end{itemize}\end{itemize}
 \item {\bf Sorted time series (SORT TS)}: sorts all values of a time series in ascending order.
 \begin{itemize}
\item Function name:    plugin\_sort\_zr.m
\item Type:             TS
\item Time series: 1 inputs, 1 outputs, Segments possible:   none
\item Single features: 0 inputs, 0 outputs
\item Images: 0 inputs, 0 outputs
\item Direct callback:  none
\item Number of parameters: 0
\end{itemize}
 \item {\bf Square (SQR)}: computes the square of the amplitudes of a time series.
 \begin{itemize}
\item Function name:    plugin\_square.m
\item Type:             TS
\item Time series: 1 inputs, 1 outputs, Segments possible:   none
\item Single features: 0 inputs, 0 outputs
\item Images: 0 inputs, 0 outputs
\item Direct callback:  none
\item Number of parameters: 0
\end{itemize}
 \item {\bf TS->DISCR TS MEAN (TERMQ)}: computes a discretized time series with a tunable number of terms for discretization.
 \begin{itemize}
\item Function name:    plugin\_discr\_zr.m
\item Type:             TS
\item Time series: 1 inputs, 1 outputs, Segments possible:   yes
\item Single features: 0 inputs, 0 outputs
\item Images: 0 inputs, 0 outputs
\item Direct callback:  none
\item Number of parameters: 2

 \begin{itemize}
\item Number of terms (Number of terms for discretization)
\item Type of membership function (defines the algorithm for the design of a membership function  (see
''Control element: Data mining: Special methods - Number of linguistic terms'').

''Median'' generates terms with approximately equal data point frequencies for all terms in the
training data set.

''Equal distribution'' generate terms with similar distances between the maximum values of the
membership functions.

''Clustering'' uses cluster prototypes as parameters.

''Fix'' set the parameters of the membership functions to the values defined by ''Control element: Data mining: Special methods - Parameters MBF (fix)''.

The choice ''with interpretability'' rounds the found parameters using a method
described in Literature: Mikut05 to improve the interpretability.

Remark:

All design methods set the random number generator of Matlab to a fix value. It improves the
reproducibility of found membership functions. Nevertheless, all methods with random elements might
be sensitive to the state of the random number generator (e.g. cluster algorithms with random start
clusters). It can also influence other methods using these membership functions, e.g. by different
selected features in a decision tree (e.g. ''Data mining - Selection and evaluation of single features - Information theoretic measures''). )
\end{itemize}\end{itemize}
 \item {\bf TS->PC TS (PCTS1PCTS2)}: computes new time series by means of a Principal Component Analysis of time series (Transformation number of time series $s\_z$ -> $s\_d$).
 \begin{itemize}
\item Function name:    plugin\_zrhkzr.m
\item Type:             TS
\item Time series: Inf inputs, 2 outputs, Segments possible:   none
\item Single features: 0 inputs, 0 outputs
\item Images: 0 inputs, 0 outputs
\item Direct callback:  none
\item Number of parameters: 3

 \begin{itemize}
\item Number of aggregated PCA features (Number s\_d of principal components computed of a time series)
\item Normalize standard deviations (defines the normalization for the variances of the sample points)
\item Transformation matrix (Use the same transformation matrix (e.g. PCA) for time series reduction of all data points)
\end{itemize}\end{itemize}
 \item {\bf Time shift (SHIFT)}: shift the time series K samples (positive values shift to the future).
 \begin{itemize}
\item Function name:    plugin\_shift\_ts.m
\item Type:             TS
\item Time series: 1 inputs, 1 outputs, Segments possible:   none
\item Single features: 0 inputs, 0 outputs
\item Images: 0 inputs, 0 outputs
\item Direct callback:  none
\item Number of parameters: 1

 \begin{itemize}
\item Shift of a time series (positive values shift to the future, negative values to the past)
\end{itemize}\end{itemize}
 \item {\bf Value in region (REGION)}: set all values in a region of a time series to 1 and all other to 0.
 \begin{itemize}
\item Function name:    plugin\_timeseries\_region.m
\item Type:             TS
\item Time series: 1 inputs, 1 outputs, Segments possible:   none
\item Single features: 0 inputs, 0 outputs
\item Images: 0 inputs, 0 outputs
\item Direct callback:  none
\item Number of parameters: 2

 \begin{itemize}
\item Lower threshold (Lower threshold for the binarization of the time series)
\item Upper threshold (Upper threshold for the binarization of the time series)
\end{itemize}\end{itemize}
 \item {\bf Value larger than threshold (THRES)}: set all values of a time series above a threshold to 1, all other values to  0.
 \begin{itemize}
\item Function name:    plugin\_timeseries\_threshold.m
\item Type:             TS
\item Time series: 1 inputs, 1 outputs, Segments possible:   none
\item Single features: 0 inputs, 0 outputs
\item Images: 0 inputs, 0 outputs
\item Direct callback:  none
\item Number of parameters: 1

 \begin{itemize}
\item Threshold (Threshold for a binarization of time series)
\end{itemize}\end{itemize}
 \item {\bf Velocity (V)}: computes the first time derivation of a time series. The result is a single time series. Here, an acausal filter is used. see Eq. (3.2) in \cite{Loose04})
 \begin{itemize}
\item Function name:    plugin\_geschwindigkeit.m
\item Type:             TS
\item Time series: 1 inputs, 1 outputs, Segments possible:   none
\item Single features: 0 inputs, 0 outputs
\item Images: 0 inputs, 0 outputs
\item Direct callback:  none
\item Number of parameters: 1

 \begin{itemize}
\item Time (defines the scaling (per sample point resp. per time using the sampling frequency from the GUI))
\end{itemize}\end{itemize}
 \item {\bf Velocity (causal) (V\_kausal)}: computes the first time derivation of a time series. The result is a single time series. Here, a causal filter is used.
 \begin{itemize}
\item Function name:    plugin\_geschwindigkeit\_kausal.m
\item Type:             TS
\item Time series: 1 inputs, 1 outputs, Segments possible:   none
\item Single features: 0 inputs, 0 outputs
\item Images: 0 inputs, 0 outputs
\item Direct callback:  none
\item Number of parameters: 1

 \begin{itemize}
\item Time (defines the scaling (per sample point resp. per time using the sampling frequency from the GUI))
\end{itemize}\end{itemize}
 \item {\bf Wavedec (db1, Low 4 db1, High 1db1, High 2db1, High 3db1, High 4)}: 
 \begin{itemize}
\item Function name:    plugin\_wavedec.m
\item Type:             TS
\item Time series: 1 inputs, 5 outputs, Segments possible:   none
\item Single features: 0 inputs, 0 outputs
\item Images: 0 inputs, 0 outputs
\item Direct callback:  none
\item Number of parameters: 3

 \begin{itemize}
\item Wavelet (defines the wavelet type.)
\item Wavelets: number of levels (defines the number of levels.)
\item Matlab wavelet decomposition (defines the used implementation.)
\end{itemize}\end{itemize}\end{itemize}
 
\end{appendix}

\printindex

\end{document}